\documentclass[preprint]{elsarticle}

\makeatletter
\def\ps@pprintTitle{%
	\let\@oddhead\@empty
	\let\@evenhead\@empty
	\let\@oddfoot\@empty
	\let\@evenfoot\@oddfoot
}
\makeatother
\usepackage[12pt]{extsizes}
\usepackage{amsmath,amssymb,amsbsy,amsfonts,latexsym,amsopn,amstext,amsxtra,euscript,amscd,hyperref,lipsum}
\usepackage{mathptmx} 
\usepackage{hyperref}
\usepackage{tabu}
\usepackage{pifont}
\usepackage{enumitem}
\usepackage{bm}
\usepackage{soul} 
\usepackage{listings}
\usepackage[dvipsnames]{xcolor}
\usepackage{setspace}
\usepackage{algorithm,algpseudocode}
\usepackage{longtable}
\usepackage{caption}
\usepackage{lscape}
\usepackage{multirow}
\usepackage{footnote}
\usepackage{subcaption}
\usepackage{float}
\graphicspath{{images/}}
\usepackage[utf8]{inputenc}
\usepackage{outlines}
\usepackage{natbib}
\usepackage{graphics,graphicx}
\usepackage{epsfig,epstopdf}
\usepackage{array,tikz}
\usepackage{caption}
\usepackage{subcaption}
\usepackage{wrapfig}
\usepackage{setspace}
\usepackage{titlesec}
\usepackage{lscape}
\usepackage{booktabs}
\usepackage[autostyle]{csquotes}

\newcolumntype{C}{@{}c@{}}
\usepackage{float}
\usepackage[top=1.5cm, left=1.5cm,right=1.6cm,bottom=1.28cm]{geometry}
\newcounter{myeqno}
\usepackage{chngcntr}

\usepackage{tikz}
\usetikzlibrary{shadows,shadows.blur}
\usetikzlibrary{shapes.geometric,arrows}
\usetikzlibrary{calc}
\usetikzlibrary{shapes.geometric,arrows,fit,matrix,positioning,shapes.multipart}
\usetikzlibrary{arrows.meta}
\tikzstyle{arrow} = [thick,-Stealth]
\tikzstyle{arrow} =[draw, -latex']
\tikzstyle{arrow} = [thick,->,>=stealth]
\tikzset{block/.style={rectangle, rounded corners,  text width=6cm, 
 minimum height=1cm, text centered, draw, fill=blue!20},
cloud/.style={rounded corners, inner sep=-4pt,  minimum 
 height=1cm, text centered, draw, ellipse,fill=blue!20, text width=2.8cm},
line/.style={draw, -latex'}}
\usepackage{newtxtext}   
\usepackage{newtxmath}
\usepackage{longtable,lscape}
\usepackage{rotating}
\usepackage{multirow,multicol}
\usepackage{natbib}
\usepackage{hyperref}
\usepackage{makecell}

\usepackage{stfloats}

\usepackage{rotating}
\usepackage{pdflscape}
\usepackage{xcoffins}
\NewCoffin\tablecoffin

\begin{document}
	\begin{frontmatter}
		\title{\textbf{Explainable Information Processing in Particle Swarm Optimization through Landscape and Search Behavior Analysis}}
         \author[NITJ]{Nitin Gupta}
 \author[ujaen]{Bapi Dutta}
\author[NITJ]{Anupam Yadav\corref{cor1}}
\cortext[cor1]{Corresponding author}
\ead{anupam@nitj.ac.in}
\address[NITJ]{Department of Mathematics and Computing\\
Dr. B. R. Ambedkar National Institute of Technology Jalandhar, 
Jalandhar - 144008, INDIA}

\address[ujaen]{Department of Computer Science\\ University of Jaén, Jaén, SPAIN}

\begin{abstract}
Swarm-based optimization algorithms have demonstrated remarkable success in solving complex problems, yet their widespread adoption remains limited due to poor transparency in how algorithmic components influence performance. This work presents a multi-faceted explainability framework for Particle Swarm Optimization (PSO) through two complementary perspectives: landscape-based and algorithmic explainability. From the landscape-based perspective, we develop a comprehensive characterization framework using Exploratory Landscape Analysis (ELA) to quantify problem difficulty, multimodality, and ruggedness, extracting ELA meta-features, dispersion measures, and information content statistics, while a machine learning approach employing Decision Tree and Random Forest classifiers enables prediction of optimal topology-specific hyperparameter configurations for unseen problems. From the algorithmic explainability perspective, we integrate IOHxplainer for temporal convergence profiling and Search Trajectory Networks (STN) for spatial navigation mapping, proposing three novel STN metrics-Connectivity Density, Fragmentation Score, and Search Efficiency-that enhance visual explainability by quantifying topology-specific search organization and transition effectiveness. Through systematic experimentation across 24 benchmark functions in multiple dimensions with Star, Ring, and Von Neumann topologies, we establish practical guidelines for topology selection and parameter configuration. Our findings uncover the black-box nature of PSO, providing greater transparency and interpretability to swarm intelligence systems. The source code is available at \textcolor{blue}{\url{https://github.com/GitNitin02/ioh_pso}}.

\end{abstract}

\end{frontmatter}

\section{Introduction} \label{sec:Intro}
Swarm Intelligence (SI) represents a fascinating and powerful field of study that draws inspiration from the collective behaviour observed in decentralized, self-organized natural systems~\cite{bonabeau1999swarm, gonzalez2021addressing}. The complex collective intelligence observed in nature, from the synchronized movements of bird flocks to the optimized foraging of ant colonies, provides the foundational inspiration for sophisticated optimization algorithms. At its core, Swarm Intelligence (SI) is characterized by a population of simple, autonomous agents that follow basic rules and interact locally with their environment and neighbors. Without any centralized guidance, these limited interactions collectively produce sophisticated global patterns and solve complex problems. The approach is fundamentally defined by principles such as decentralization, emergence \cite{bonabeau1999swarm} , adaptation, and positive feedback.
Any swarm based optimization algorithm includes three fundamental components: (i) a population that evolves iteratively, (ii) a reproduction process where current population of candidate solutions generates new set of candidate solutions in the next generation, and (iii) a curated information sharing along with an efficient learning mechanism that determines next generation of candidate solutions. These algorithms are inherently scalable, adaptable, and robust, characterized by the simplicity of their individual candidate solutions, making them highly effective for solving a wide array of complex optimization problems \cite{bonyadi2017particle,bianchi2009survey, munoz2015algorithm}. At the core of the success of SI algorithms lie a critical balance between exploration – the search for novel solutions to the search problem in previously unexplored parts of search space – and exploitation – the search for better and better solutions in already explored regions of the search space. Exploitation, on the other hand, is the process of synthesizing around good ideas that are already known, in order to close in quickly and efficiently on a potentially successful solution. The ability to balance these two dynamics is very important if the goal is to efficiently converge towards reasonable solutions, while at the same time not getting trapped in less than ideal local optimum solutions. The method's key strengths – robustness and adaptability – stem directly from this decentralized architecture \cite{bonabeau1999swarm}. The solutions that evolve are the results of bottom-up interactions of the swarm elements, as contrasted with top-down optimization. Such systems are typically adaptive in nature, and are intrinsically flexible and robust, even in the face of a dynamic, high-dimensional problem space, where the loss of one agent will not disrupt the whole process. The intelligence is spread out, and not at a single point, so if one solution fails, or a local condition changes, the swarm of solutions can still adapt and find new solutions. The main advantage of SI algorithms is that they are suited to uncertain, dynamic, and/or too complex environments to incorporate a pre-programmed control mechanism \cite{engelbrecht2005fundamentals}. Several SI algorithms have been successful and have emerged from the many that exist. One of the most popular algorithms in this group is Particle Swarm Optimization (PSO) ~\cite{kennedy1995new, yu2025particle} which is inspired by the social behaviour of flocks of birds and schooling fish. The particles (candidate solutions) travel through a search space and always change their own position (pbest) according to their own best solution (pbest) and the Position of the Global best known (gbest) by the whole swarm. One of the major advantages of PSO is its suitability for continuous optimization problems ~\cite{li2021particle} that include important cross-functional optimization tasks like parameter tuning ~\cite{hossain2021machine} in machine learning modelling. There are several other swarm based optimization algorithms such as artificial bee colony (ABC) \cite{karaboga2007powerful, coleto2020artificial}, ant colony optimization (ACO) \cite{dorigo2018introduction, liu2023improved} and many more but our focus in this article is particle swarm optimization (PSO) which is one of the earliest swarm based optimization algorithms. In the application of swarm based optimization algorithms, parameters are crucial in making the effectiveness of a particular meta-heuristic to solve a particular decision problem, which has a profound impact on the algorithm's efficiency \cite{banks2007review}. The most important role of these is to optimally balance the convergence and diversity of the search process, which directly translates to the balance between exploration, or search in new areas of the solution space and exploitation, or search refinement around the currently known promising solution areas \cite{shi1998modified}. If such control parameters are not optimized, they can cause problems, including the premature convergence in which the swarm converge to a sub-optimal solution quickly, and the computational time increased. Parameter settings have a direct effect on the quality of solutions obtained, computational power required, the diversity and levels of exploration achieved and the overall leadership structure of the swarm \cite{bonyadi2017particle}.

The focus is on PSO as the performance of PSO is very sensitive to its control parameters, specifically the inertia weight $(w)$ and the acceleration parameters $(c_{1}, c_{2})$  \cite{shi1998modified}.Throughout the search process it is necessary to dynamically balance exploration and exploitation and therefore develop adaptive approaches for these parameters. The above points improve our understanding of PSO parameters:

\begin{enumerate}[label=\roman*.]
\item {Inertia Weight $(w)$}: This parameter determines the momentum of particles and thus the effect of their earlier velocities on the motion of particles. High $w$ will drive the particles into the region that they haven't visited before, which helps to do a global search; and when $w$ is low, the particles will stay in the region that they think is good to search, which helps to carry out a local search. The inertia weight has significant influence on the PSO performance.
\item {Acceleration Coefficients $(c_{1}, c_{2})$}: These values are often referred to as ``trust parameters,'' these coefficients, along with random numbers $(r_{1}, r_{2})$, control the stochastic influence of the cognitive (personal best, $c_{1}$) and social (global best, $c_{2}$) components on a particle's velocity.
\begin{enumerate}[label=\alph*.]
    \item $c_{1}$ (\textit{cognitive component}) reflects a particle's tendency to learn from its own experience and move towards its personal best position.
    \item $c_{2}$ (\textit{social component}) represents a particle's tendency to follow the collective knowledge of the swarm and move towards the global best position.
    \item A proper balance between $c_{1}$ and $c_{2}$ is essential for effective search~\cite{985692}. If $c_{1}$ is significantly greater than $c_{2}$, particles may wander excessively. Conversely, if $c_{2}$ is significantly greater than $c_{1}$, particles might rush prematurely to local optima.
\end{enumerate}
\item {Swarm Size}: The total number of particles within the swarm also impacts PSO's performance.
\item {Number of Iterations}: While the number of iterations is often just a stopping criterion, studies suggest it has a less substantial impact on performance compared to key parameters such as inertia weight and swarm size~\cite{blackwell2007particle}.
\end{enumerate}

Many times the PSO algorithm's stochastic nature and non-linear dynamics lead to a complexity of the parameter setting space, where the convergence speed is connected to the risk of being stuck in the local optimum mode of operation \cite{shi1998modified}. This is a tricky task if performed manually. SI algorithms (including PSO) are stochastic and the algorithms behaviour when optimizing differs significantly as a result of small variations in the initial algorithm parameters, from diverging to prematurely converging to a solution or even reaching the optimal solution. For a given problem, there are very few direct methods of parameter optimization for PSO presented in the literature, and for some problems, the parameters are found to be interdependent and non-linearity. Therefore, researchers are forced to take a trial and error solution, a method which is more of an intuition and long experimentation rather than computation \cite{eiben2011parameter}. This fundamental challenge of parameter sensitivity is the primary driver for the development of more sophisticated, adaptive parameter strategies, as static and empirically derived parameters are rarely optimal across diverse problem landscapes in the entire optimization process \cite{pedersen2010tuning}.

A big problem with these parameters, however, is that they're not particularly easy to interpret. The impact of these can sometimes be complex and case-specific and is not easy to formulate rules for. The next paragraph throws some lights on the need of interpretability and explainability of PSO framework.

\subsection{Need of Explainabilty in PSO and Current Status of Research} Although used effectively, PSO is often used as a “black box” that hides the decisionmaking processes within it, making it hard to instill trust – especially in certain areas, such as healthcare and autonomous systems~\cite{yang2025meta}. There are many different angles in which these systems are understood that are found in the current literature; in particular, the post-hoc surrogate modeling approach has been used~\cite{liang2024survey}, social network analysis approach has been used ~\cite{kennedy2002population} and fitness landscape approach has been used ~\cite{malan2013survey} and the visual analytics approach has been used~\cite{zhang2008sequential}. But it is difficult to make evaluation metrics, and real-time human-in-the-loop designs, standard. It is not the complexity of PSO, but emergent, non-linear interactions among these: among particles, between the particles' movements, and among the information particles that make it opaque. This simplicity/complexity paradox demands collective behavior and parameter sensitivity based methods for XAI, instead of architectural decomposing methods. Consequently, EAI has come to the fore to make models explainable and foster confidence among users \cite{doshi2017towards}.

The success of the Metaheuristic optimization algorithms like PSO, in the past, was mainly judged in terms of efficiency, convergence rate and the quality of the solutions generated. \textit{However, recent work~\cite{gupta2025enhancing} on analysing the impact of swarm topologies provides an early work in the direction of ``enhancing explainability and trustworthiness" in PSO. This work is an extension of the work carried out in~\cite{gupta2025enhancing}}. This indicates an important evolution in the area of computational intelligence. It implies that if we really seek optimal solutions, then this is no longer what we should be aiming for as the criteria of successful algorithms. On the other hand, an increased awareness that for the use and deployment of AI systems to be legitimate and widely adopted, it is essential that these systems are transparent and understandable, especially in critical use cases of AI \cite{BARREDOARRIETA202082}. This is a testament to the broader ethical and practical requirement of trusted AI. This paradigm shift indicates that the assessment should be done in a new way, not only optimized with respect to the quality of the solution and/or computational costs, but also quantitatively with respect to explaining. It also suggests that, for new algorithms, routine research on XAI needs to be conducted to include XAI-aware annotations as early as in the initial design phase of the algorithm, and not only post-hoc explanations \cite{slack2021reliable}. The follwoing subsections further throws light on some key factors which have the highest impact in the interpretability and explainability of PSO.


\subsubsection{Impact of Communication Topologies in PSO}
The structure with which the particles of the PSO swarm communicate with each other, or the topology, has a significant impact on the effectiveness and the search behaviour. The common topologies are Ring, \cite{liu2016topology} Star  \cite{miranda2008stochastic, ni2013new} and Von Neumann \cite{von1935complete, lynn2018population}. Research examines how the different topologies affect critical PSO operations like information sharing, diversity preservation, convergence rates and the balance between exploration (exploring new regions) and exploitation (improving good regions) \cite{zhang2019enhancing}. Such influences can be investigated systematically and recommendations drawn for the decision-making processes of PSO, not only in terms of their intrinsic interpretation but also with regard to the recommendations for selecting the best topology for specific optimization problems. The description of the impact of different communication topologies on the convergence, diversity and exploration/exploitation balance of PSO is clear and has high intrinsic interpretability. The advantage of this approach is that the black box PSO approach was not included. Rather it is dedicated to understanding what its core elements are, and designing them to have transparency by themselves. It implies that changes in the internal structure and rules of interaction of the algorithm directly influence the interpretability, and, consequently, the more transparent and trustworthy the algorithm becomes, the more promising to develop more transparent and trustworthy metaheuristics becomes, in the future.

\subsection{Summary of Contributions}
\begin{figure*}[h]
    \centering
    \includegraphics[width=8cm]{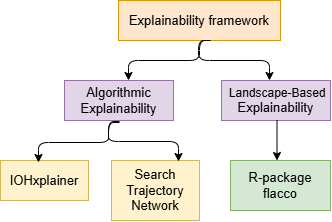}

      \caption{Overview of the explainability framework Toolkit, comprising algorithmic explainability and landscape-based explainability.} \label{fig:XAI}
\end{figure*}

Although many developments have been made in the understanding of Particle Swarm Optimization (PSO) and its communication topologies, there are still many research issues in making PSO explainable, generalizable and trustworthy. Current research generally adopts a strategy of empirical performance benchmarking, which, however, does not provide clear explanations of the causal mechanisms for PSO's actions in different landscapes. There is currently no common framework for explaining the way that hyperparameters, topological structures and features from the problem space interact in PSO, which restricts our ability to build interpretable insights. IOHxplainer, Search Trajectory Network (STNs), and Exploratory Landscape Analysis (ELA) are highly complementary technologies that provide a major step forward to close these research gaps in explainable and trustworthy PSO systems. All three of these frameworks reinforce each other from different angles, both algorithmically and problematically, in regards to optimization behaviour. The explainability framework Toolkit is used to combine algorithmic-level analysis, as provided by IOHxplainer and Search Trajectory Networks (STNs), with landscape-level analysis as provided by the flacco package as shown in Fig. \ref{fig:XAI}.

The key contributions of the manuscript can be summarized as follows:
 
\begin{enumerate}[label=\roman*.]
\item Introduces a dual-perspective algorithmic explainability framework combining IOHxplainer and Search Trajectory Networks (STNs) to reveal how Star, Ring, and Von Neumann topologies distinctly influence PSO's exploration-exploitation balance, stagnation patterns, and search space transitions over time.

\item Proposed three novel STN-based metrics, Connectivity Density, Fragmentation Score, and Search Efficiency, that enhance visual explainability by transparently quantifying topology-specific search organization, network cohesion, and transition effectiveness during PSO optimization.

\item Leveraged Pflacco library to extract ELA features (ruggedness, modality, neutrality) from benchmark functions, enabling a landscape-aware explainability framework that reveals how problem structure influences topology-specific search behavior and performance.

\item Utilized IOHxplainer with SHAP-based analysis for explainable benchmarking and hyperparameter impact quantification.

\item Developed a landscape-aware predictive automated methodology framework using ELA meta-features and ML models (Decision Tree, Random Forest) to predict optimal PSO hyperparameters for each topology on unseen problems.

\end{enumerate}


This builds on our previous work that explains the communication topology of PSOs and enhances the explainability of its communication topology using ELA-based and IOHxplainer-based analysis~\cite{gupta2025enhancing} to further improve interpretability. The rest of this paper is structured as follows: Section~\ref{sec:2} introduces the ELA framework for visualization used to benchmark. In Section~\ref{sec:3} the proposed Algorithmic explainability framework for PSO is introduced while this is followed by Section~\ref{sec:4} detailing the experimental setup and corresponding results in Section~\ref{sec:4}. The effect of hyperparameters on algorithmic performance is studied in Section~\ref{sec:5} and the data-driven, interpretable configuration learning methodology is explained in Section~\ref{sec:6}. Finally, in Section~\ref{sec:7} some conclusions are drawn and ideas for future research are suggested.

\section{ Landscape-Based Explainability }\label{sec:2}
Now, as the performance of algorithms depends on the landscape of a problem, the selection of an effective optimizer depends on those prior knowledge. There have been many high level characterizations proposed so far, based on categorical description of properties such as modality, separability, and variable scaling ~\cite{mersmann2011exploratory} which serve as a starting point but have severe drawbacks. They lack quantitative precision, they need domain expertise to understand them, they tend to miss some nuanced features, and they have a much more detailed understanding of a problem than is likely to be available in practice.

This lack of knowledge is compounded by the typical tools of the trade. Although platforms such as the BBOB benchmark suites contained in the COCO platform \cite{hansen2021coco} or the CEC function suites \cite{wu2017problem}  and the various sets available in the Nevergrad platform \cite{rapin2018nevergrad} offer the necessary platforms for the fair and reproducible comparison of algorithms, they tend to strengthen a ``black-box” evaluation paradigm. One of the main problems is that there isn't too much generalizability of their results—algorithms that work well for these particular test functions may not work well for other real-world problems. This is often a result of inadequate diversity of the set of benchmarks, the use of non-representative baselines, and the fact that the internal properties of the problem that determine the problem difficulty are often not known. Opaque reasons for algorithms' success or failure can cause comparisons to be unfair and may stifle progress.

In order to mitigate the shortcomings of high-level classification and of black-box benchmarking, in this work we explore quantitative, data-driven approaches, such as low-level features \cite{bischl2012algorithm}. Exploratory Landscape Analysis (ELA) will solve these problems by systematically analyzing the problem with sampled data. This method involves characterizing optimization problems by identifying some meta-features from the candidate solutions, generated by sampling methods like Latin hypercube sampling \cite{shields2016generalization, burhenne2011sampling} or random sampling. Based on these sampled solutions, the landscape features that represent important aspects of the problem are computed. ELA is a set of quantitative meta-features, extracted from a set of candidate solutions created through sampling, including convexity, distribution of the objective values, local search structure, information content, and clustering. ELA helps uncover the reasons behind an algorithm’s performance. For instance, if a landscape is very rugged, then extensive global exploration might be required, while if the problem is separable it can be accommodated best by coordinate-wise methods. These are important to give the jump from trial-and-error experimentation to principled algorithm design and selection. In essence, ELA features have potential to bridge the gap between ``black-box” benchmarking and meaningful interpretation. This provides deeper, more objective insights into the properties that influence algorithm performance.

One way to simplify this is with tools such as the R-package \texttt{flacco} and its Python counterpart \texttt{Pflacco} \cite{prager2024pflacco} that handle the process: Given a single initial design, they process more than 300 numerical features across 17 feature sets. The package handles the computational overhead by maintaining counts of the number of calls to the function and time used for each set of features. The system provides key mathematical function definitions to feature sets that require further assessments, and also enables special processing techniques like cell mapping, which require splitting up blocks by dimension.

In addition to feature calculation, \texttt{flacco} includes visualization techniques to provide greater understanding of the landscape. This type of visualization is provided, for example, by the function \texttt{plotCellMapping} \cite{kerschke2014cell} which breaks the continuous decision space into cells and displays the problem landscape. This approach involves identifying the following cells: attractor cells (black cells), uncertain cells that are influenced by more than one attractor (grey cells), and certain cells that are the basins of attraction (white cells). The arrows show the direction of the attractor from the other cells, and the thickness of the arrow shows the probability of attraction to the attractor. The interactions and distributions of these cells create other features and represent a strong visual summary of landscape structure.


These cell mapping bases can be used to make more advanced constructs, like the barrier trees as introduced in \cite{kerschke2019comprehensive}. The local optima (valleys) are represented as leaves (circles) and ridges are represented as nodes (diamonds, no filler). A hierarchical model of the structure of fitness landscape is shown. Visualization of the cell mapping plots and barrier trees plot are shown in Tables \ref{Tab ELA-1}-\ref{Tab ELA-3}. Likewise, as described in Table \ref{Tab ELA-4} Information Content of Fitness Sequences (ICoFiS) is based on the analysis of entropy of fitness sequences obtained from random walks with different step sizes which are related to properties like smoothness, ruggedness and neutrality. This approach effectively adapts discrete information-theoretic measures for continuous optimization domains, with the metrics in Table \ref{Tab ELA-4} serving as a direct output of this analysis. The comprehensive nature of this approach is summarized in Tables \ref{Tab E01}-\ref{Tab E02}, which details feature sets such as those computed by \texttt{calculate\_ela\_meta}, \texttt{calculate\_ela\_distribution}, \texttt{calculate\_nbc}, \texttt{calculate\_dispersion}, and \texttt{calculate\_information\_content} used to capture the global structure, value distribution, clustering behaviour, dispersion, and informational complexity of a selected BBOB function's landscape. Section \ref{sec:6} details a classification methodology where an expanded suite of ELA features (building on Tables \ref{Tab E01} and \ref{Tab E02}) is used to predict categorical performance of the PSO algorithm, as measured by its AOCC. The remaining set of ELA features, used for the classification task, has been made publicly available on GitHub.

\begin{landscape}
\begin{table*}[htp]
\centering
 \caption{Analysis of ELA features BBOB functions on d=2.}
\label{Tab ELA-1}
    \centering
   \scriptsize
   \setlength{\tabcolsep}{3pt}
   \begin{tabular}{ccccccc}
    \hline
      \textbf{Function}  & \textbf{Contour Plot} &  \textbf{Surface Plot}  &  \textbf{Cell- Mapping} &  \textbf{Barrier Tree-2D} &  \textbf{Barrier Tree-3D}  &  \textbf{Information Content}
      \\ \hline
      \multirow{1}{*}{$f_1$}  & \includegraphics[width=3.5cm]{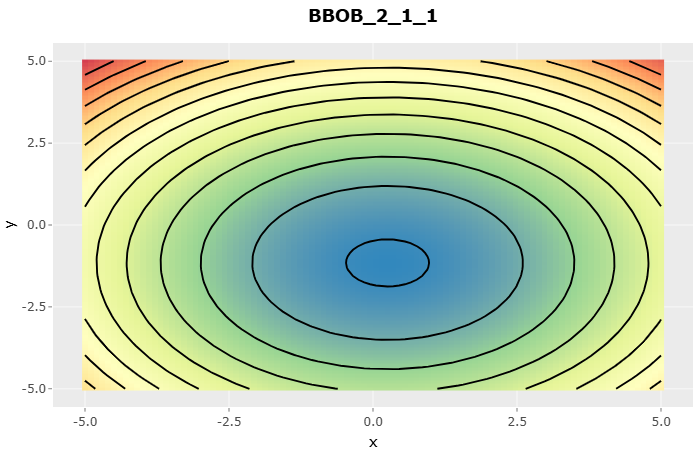}
        & \includegraphics[width=3.5cm]{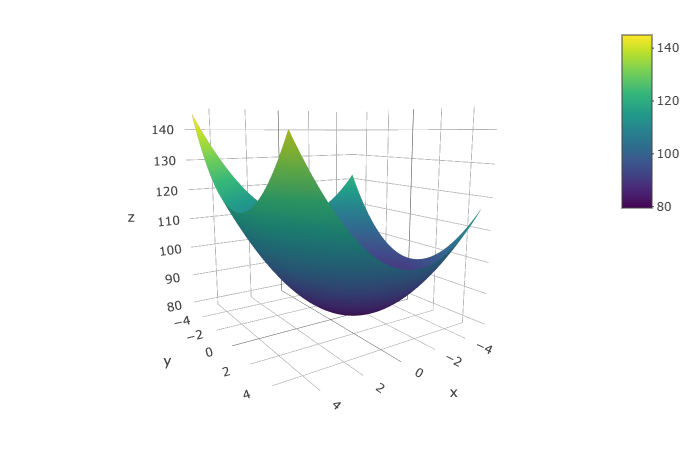} 
        & \includegraphics[width=3.5cm]{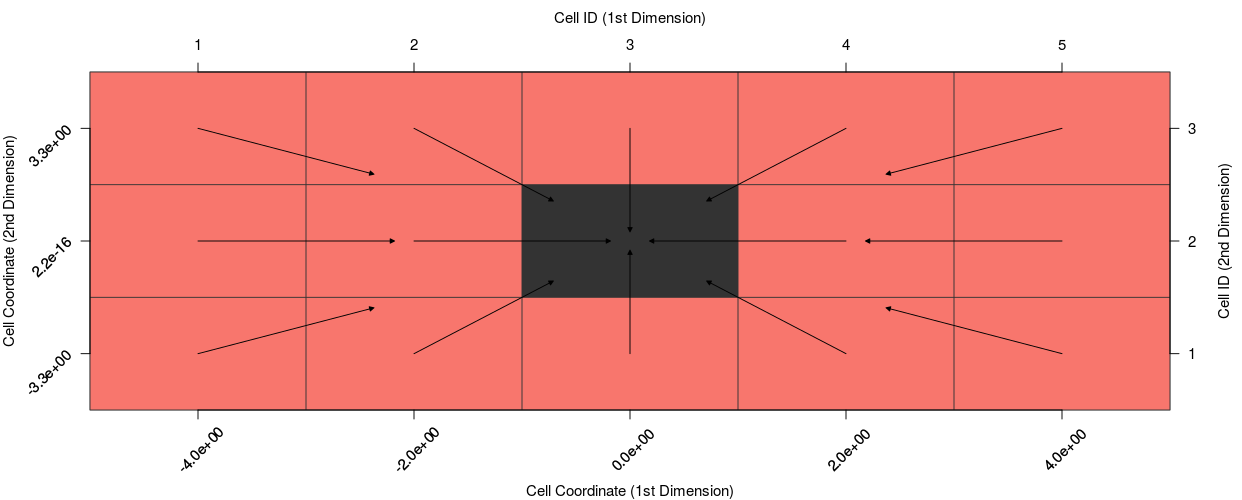}
        & \includegraphics[width=3.5cm]{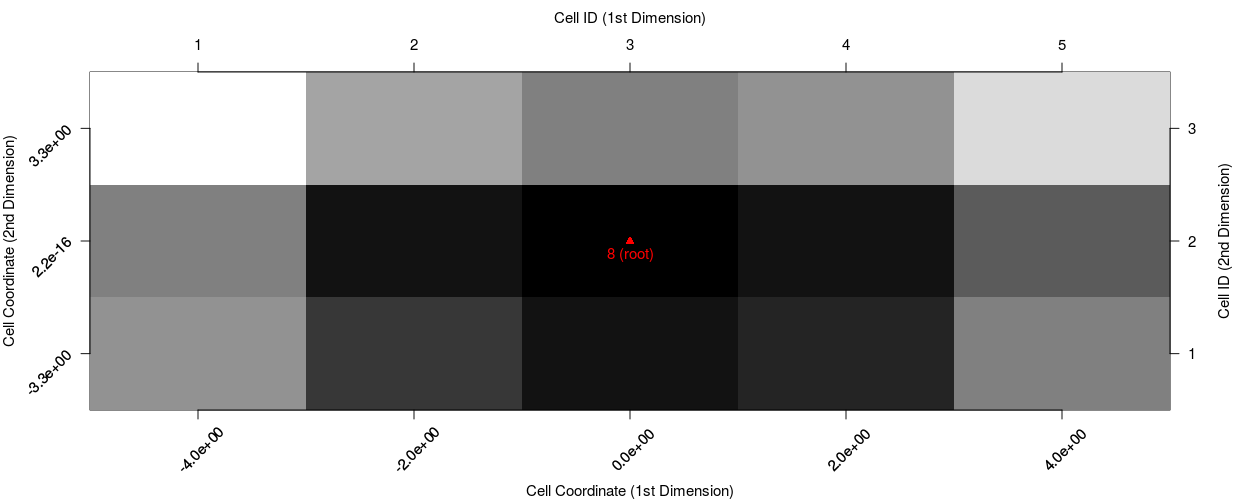}
        & \includegraphics[width=4.5cm]{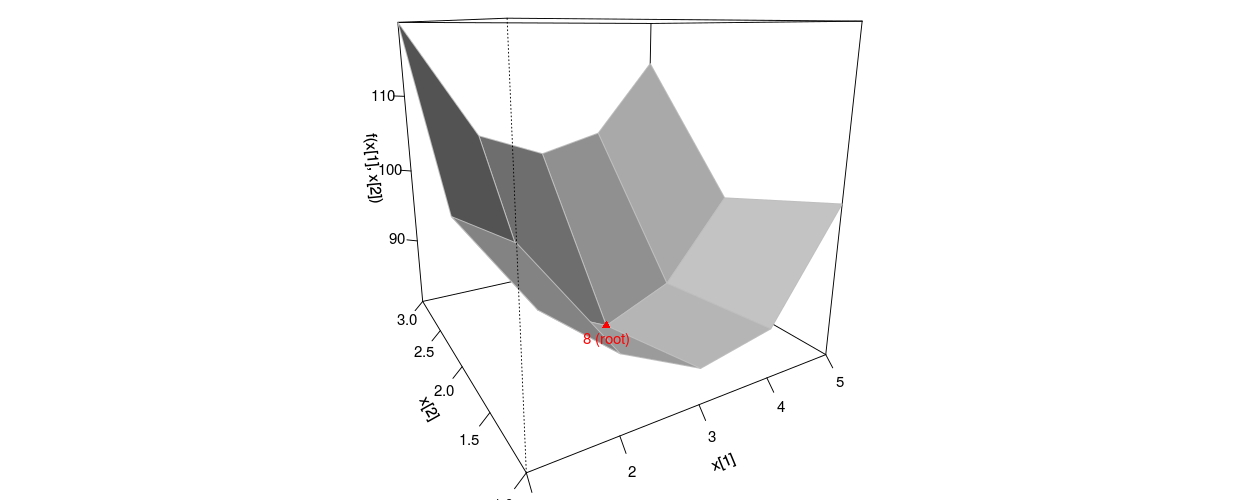}
        & \includegraphics[width=3.5cm]{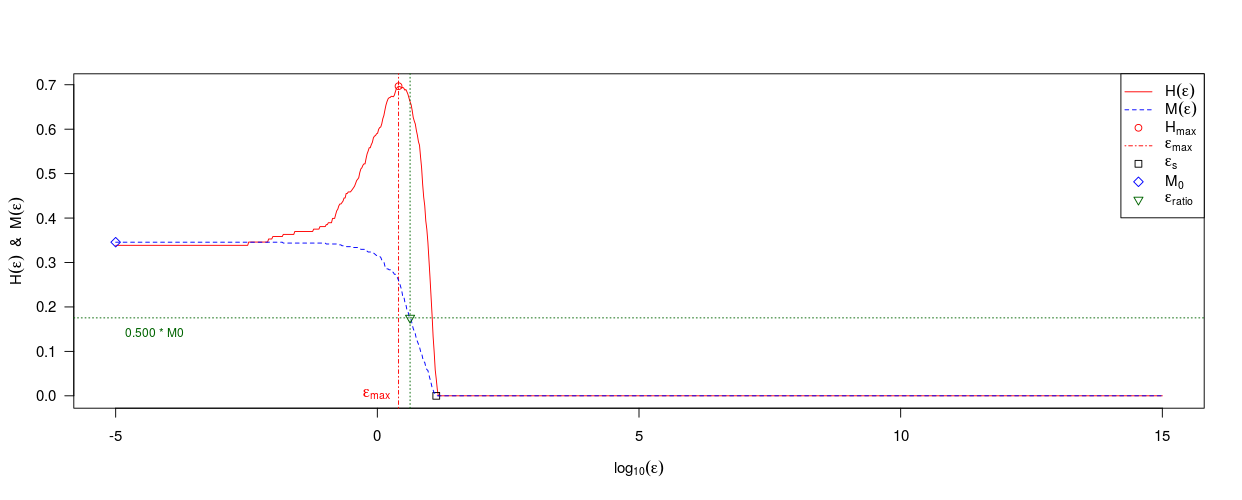}
        \\
 \hline
 \multirow{1}{*}{$f_2$}  & \includegraphics[width=3.5cm]{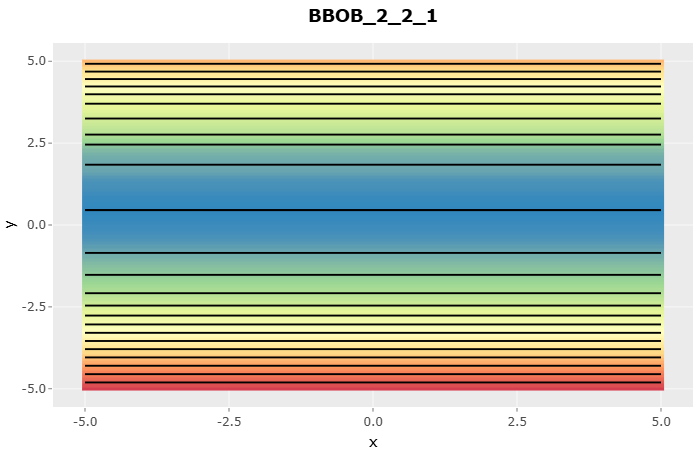}
        & \includegraphics[width=3.5cm]{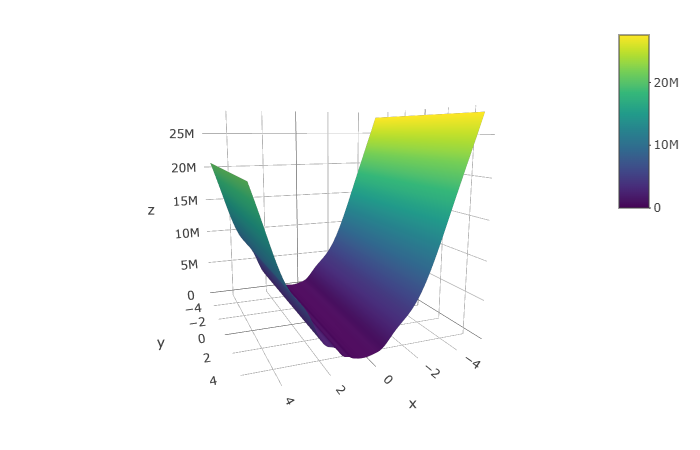} 
        & \includegraphics[width=3.5cm]{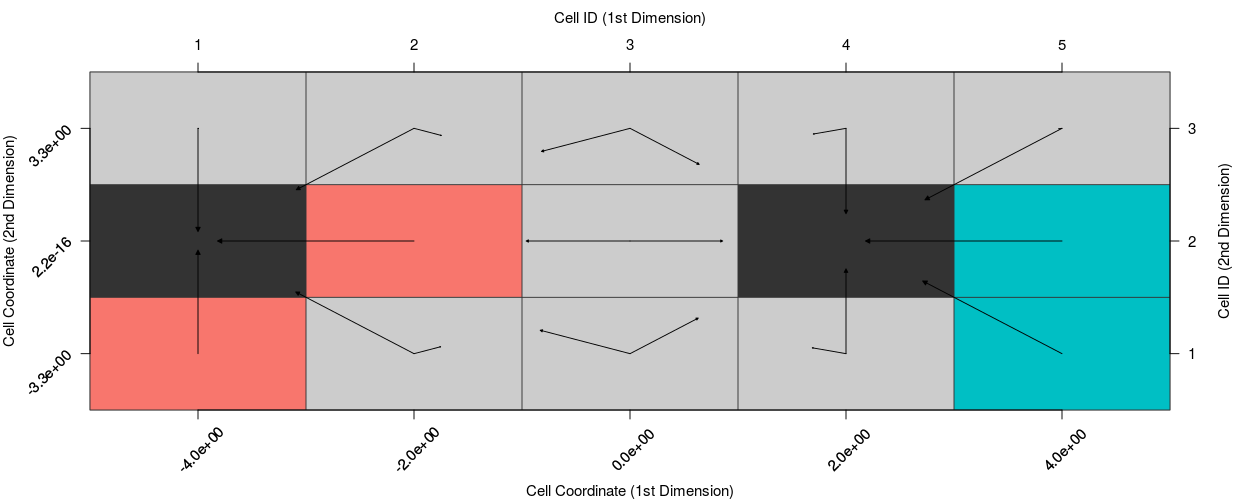}
        & \includegraphics[width=3.5cm]{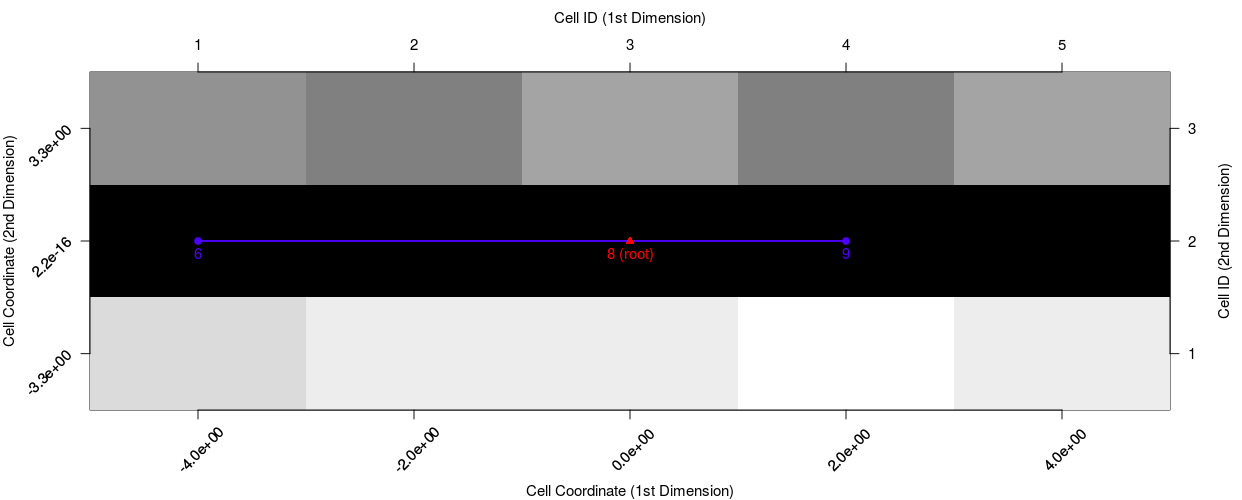}
        & \includegraphics[width=4.5cm]{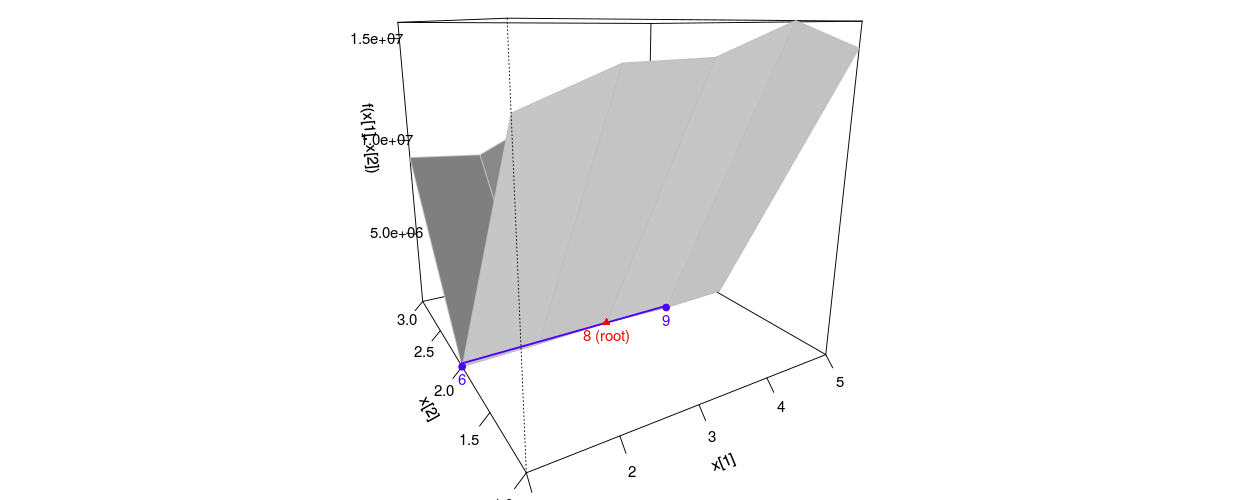}
        & \includegraphics[width=3.5cm]{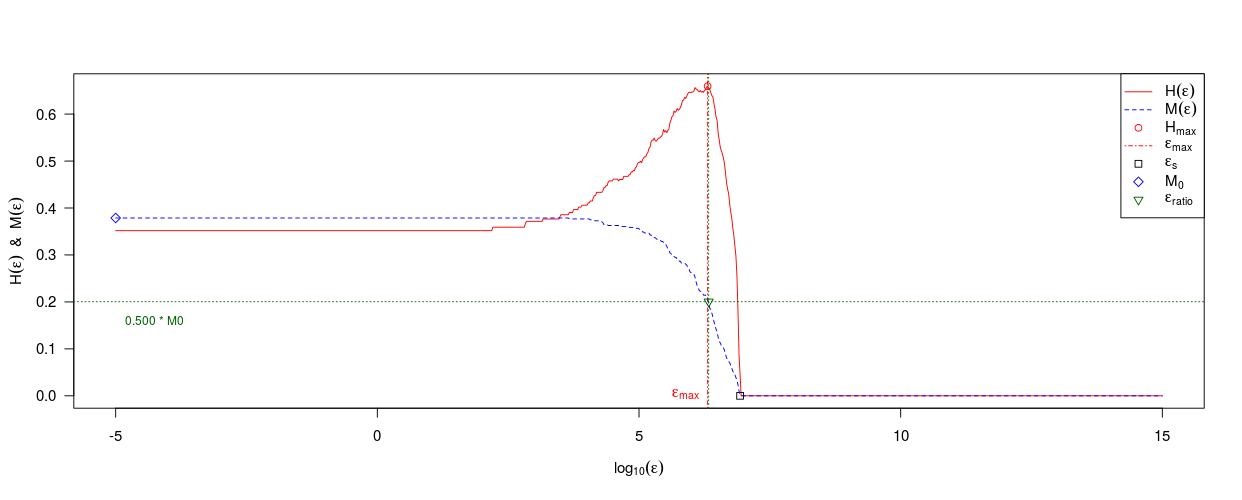}
        \\
 \hline
 \multirow{1}{*}{$f_3$}  & \includegraphics[width=3.5cm]{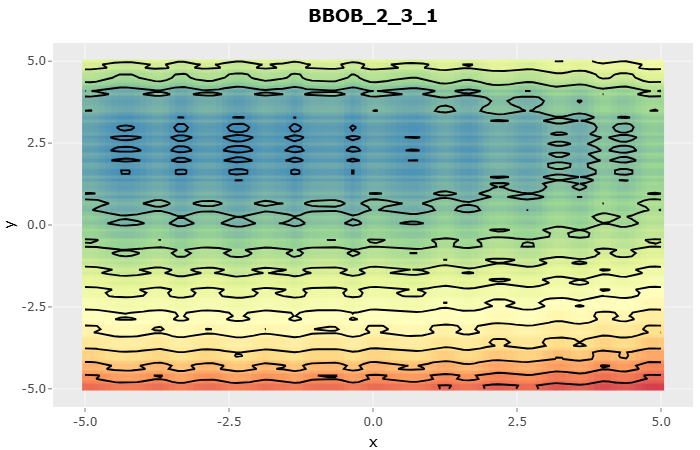}
        & \includegraphics[width=3.5cm]{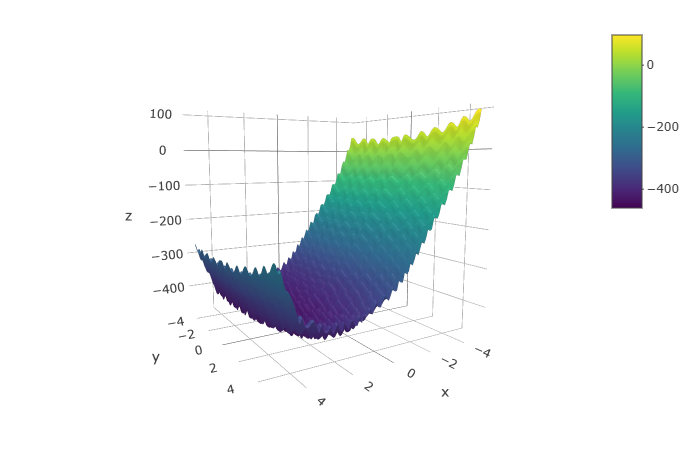} 
        & \includegraphics[width=3.5cm]{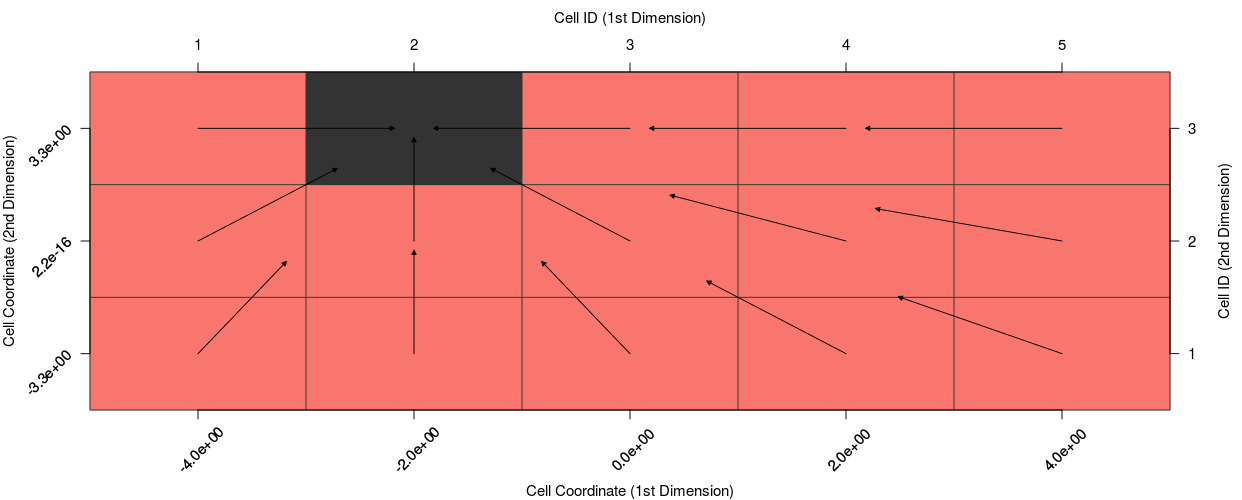 }
        & \includegraphics[width=3.5cm]{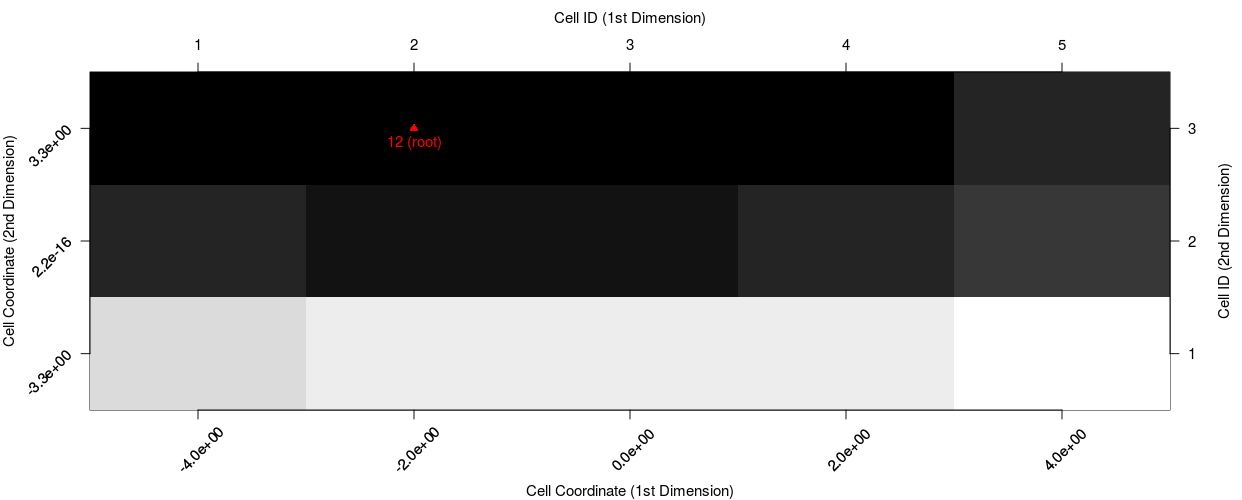 }
        & \includegraphics[width=4.5cm]{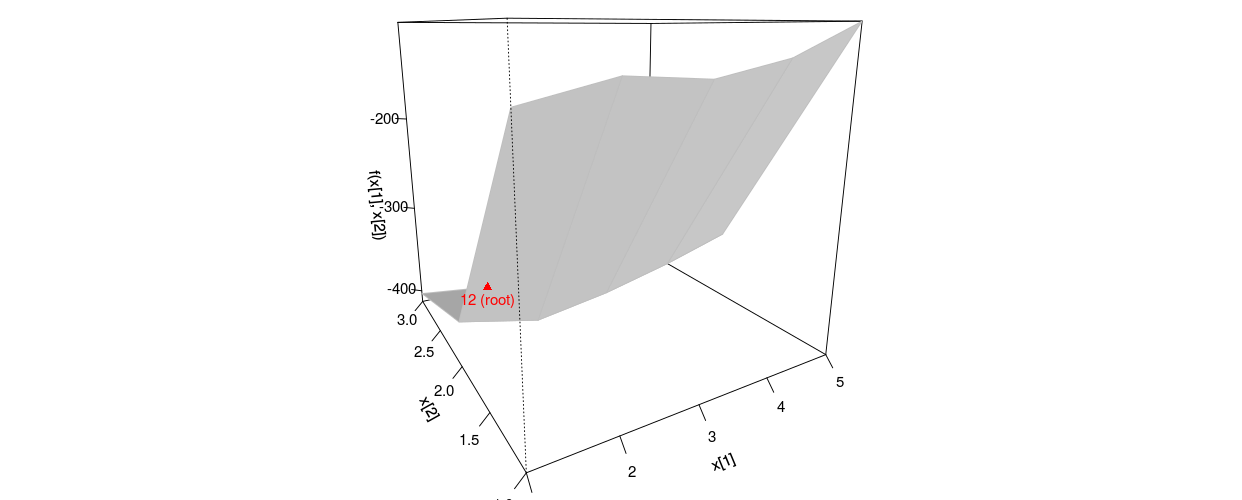 }
        & \includegraphics[width=3.5cm]{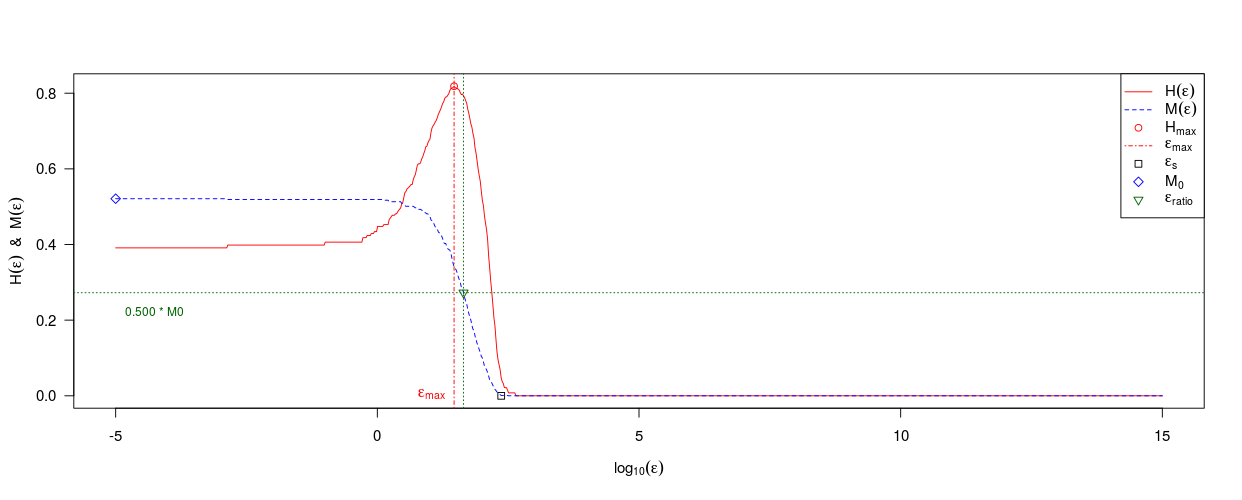}
        \\
 \hline
 \multirow{1}{*}{$f_4$}  & \includegraphics[width=3.5cm]{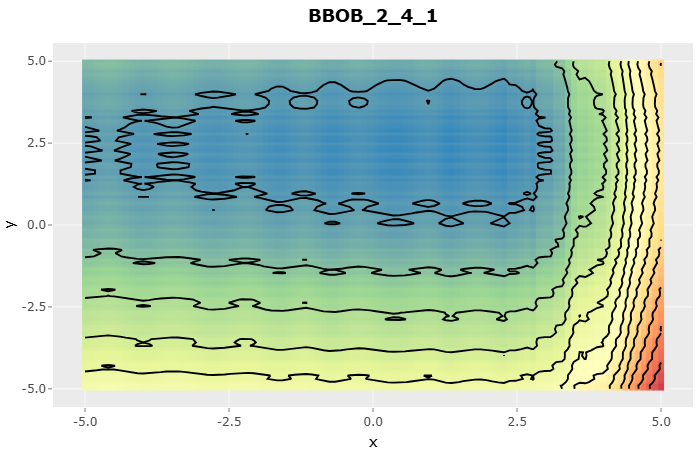}
        & \includegraphics[width=3.5cm]{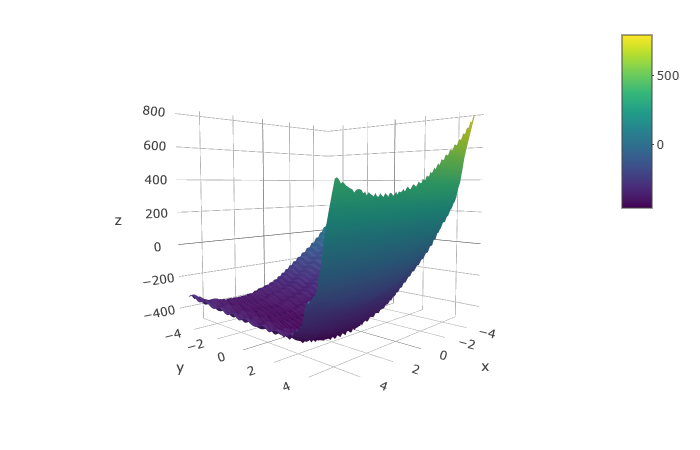} 
        & \includegraphics[width=3.5cm]{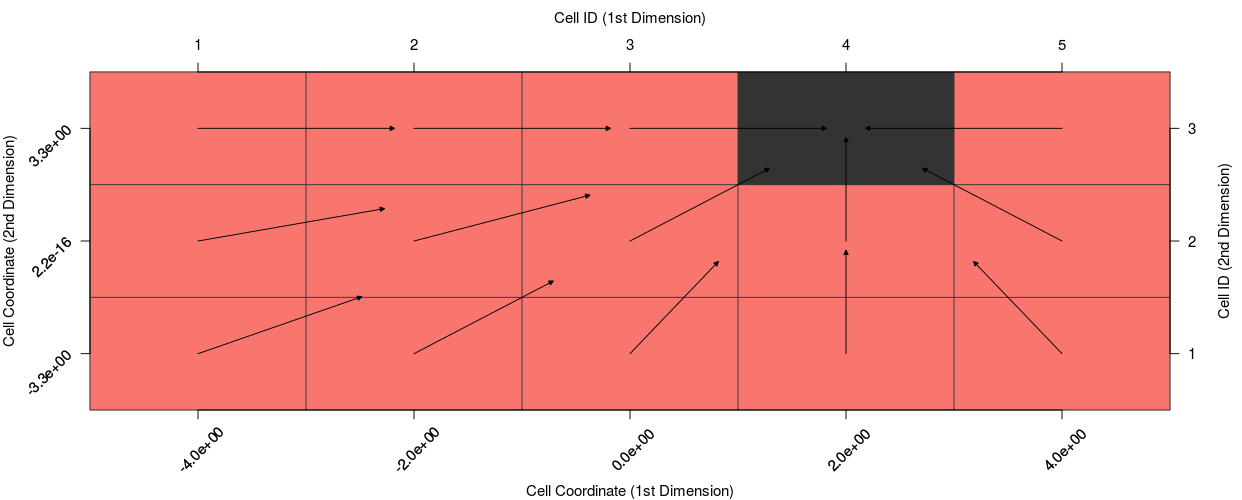}
        & \includegraphics[width=3.5cm]{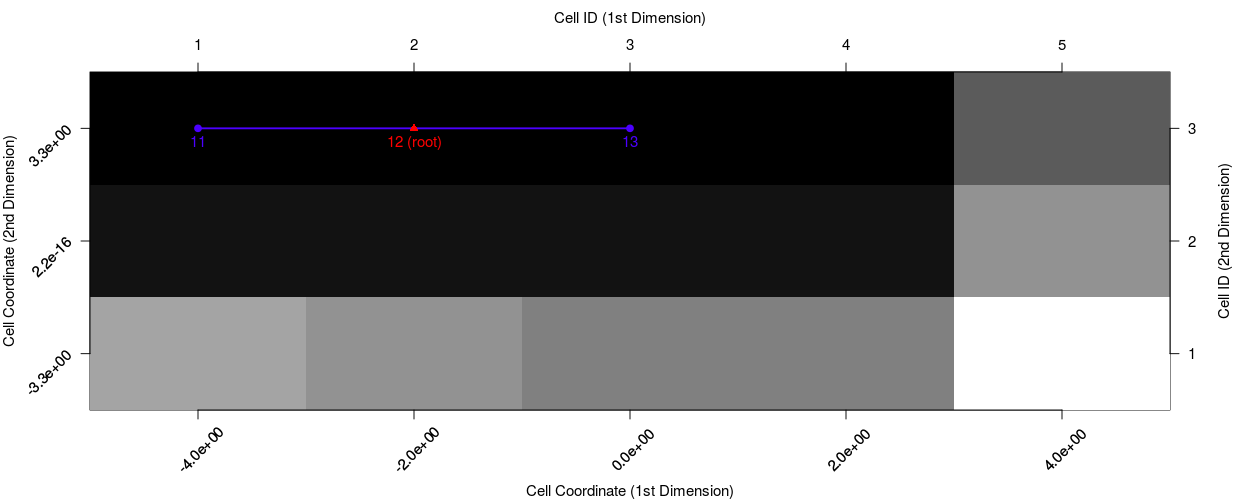}
        & \includegraphics[width=4.5cm]{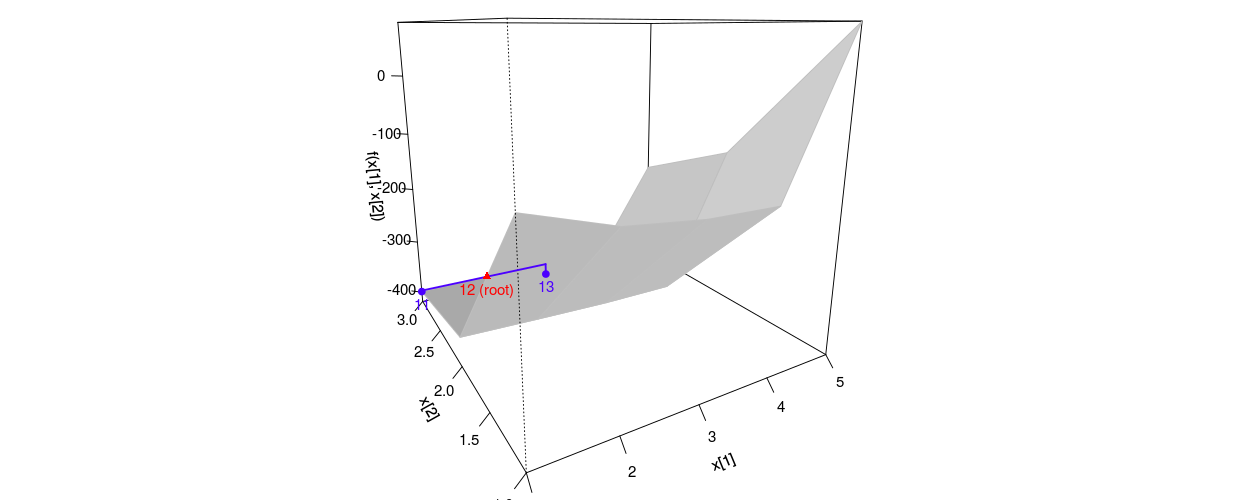 }
        & \includegraphics[width=3.5cm]{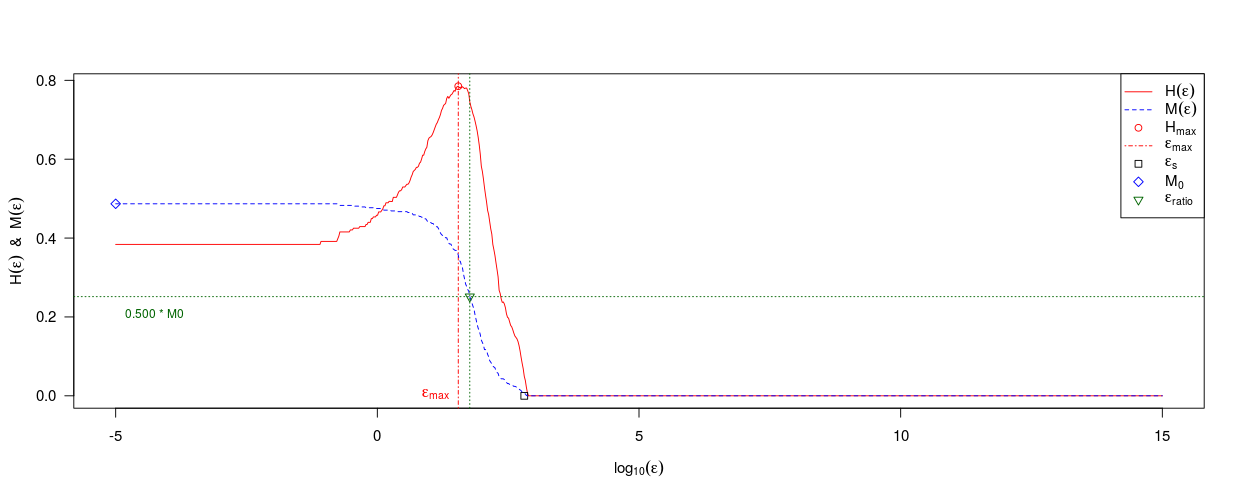 }
        \\
 \hline
\multirow{1}{*}{$f_5$}  & \includegraphics[width=3.5cm]{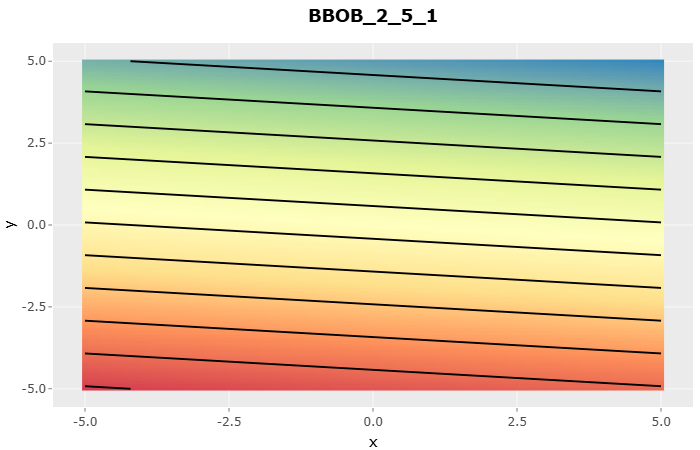}
        & \includegraphics[width=3.5cm]{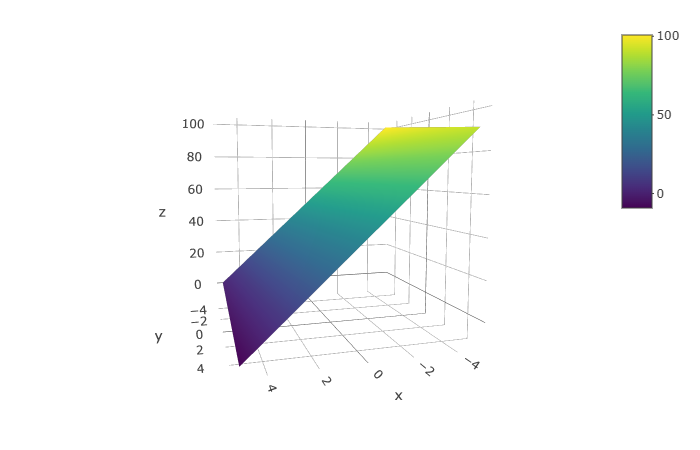} 
        & \includegraphics[width=3.5cm]{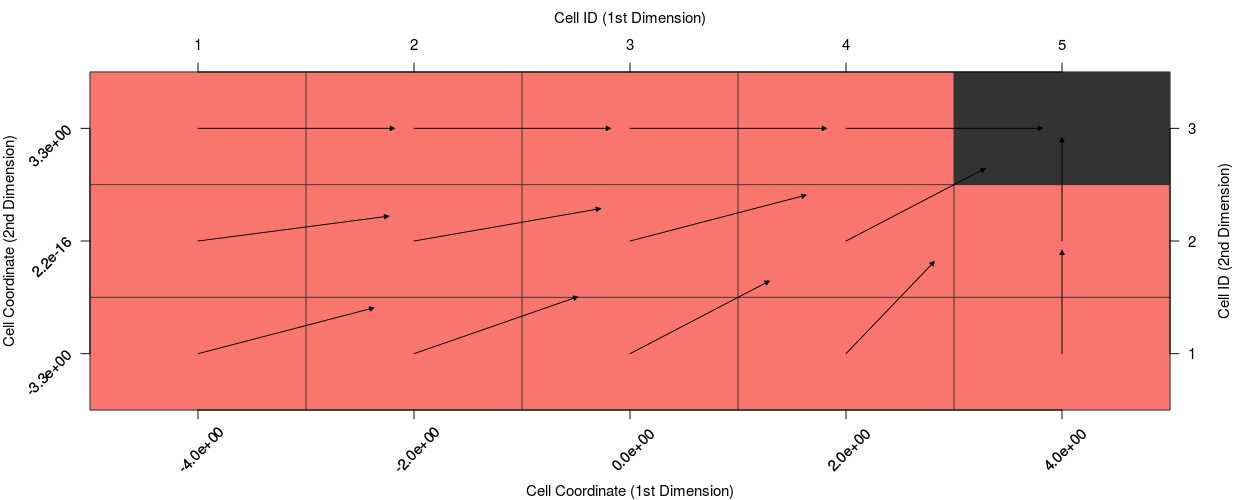}
        & \includegraphics[width=3.5cm]{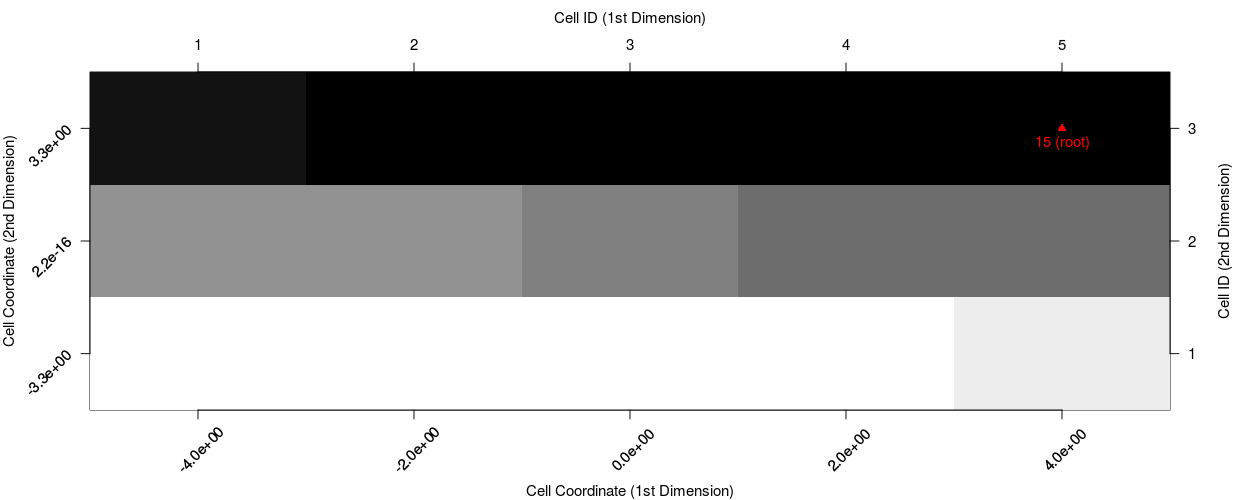}
        & \includegraphics[width=4.5cm]{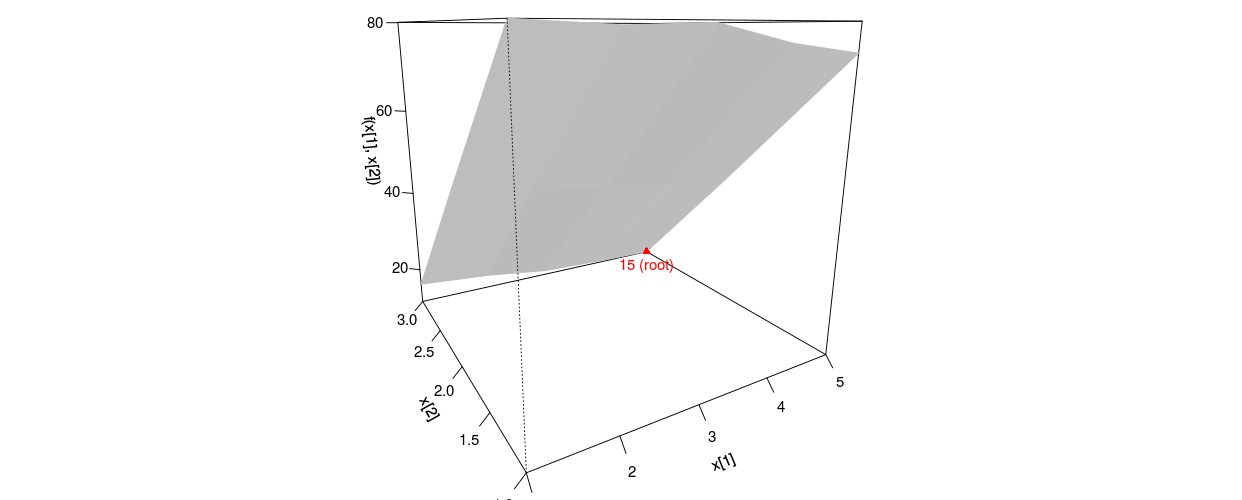}
        & \includegraphics[width=3.5cm]{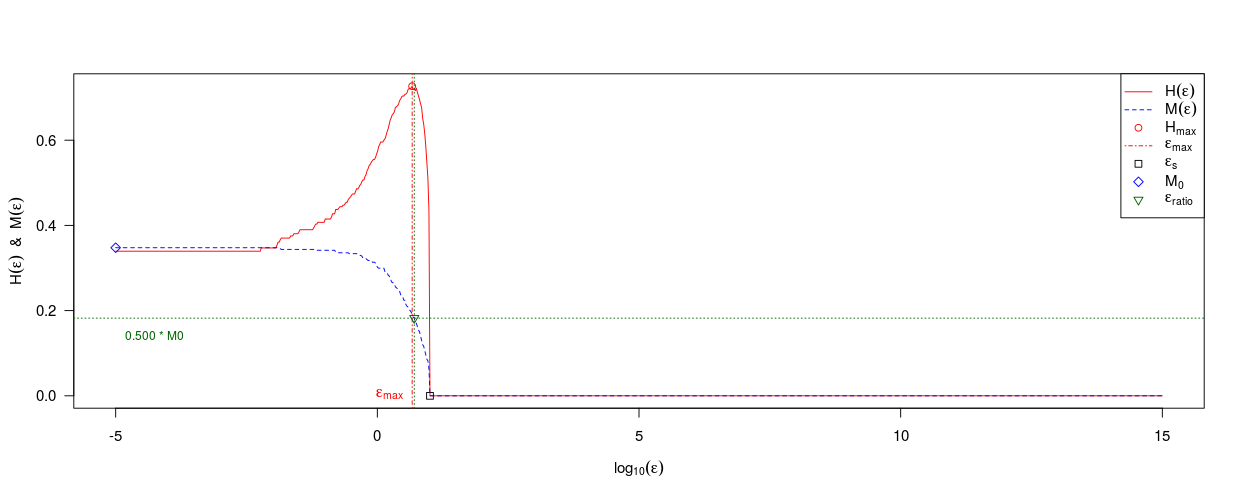}
        \\
        \hline
\multirow{1}{*}{$f_6$}  & \includegraphics[width=3.5cm]{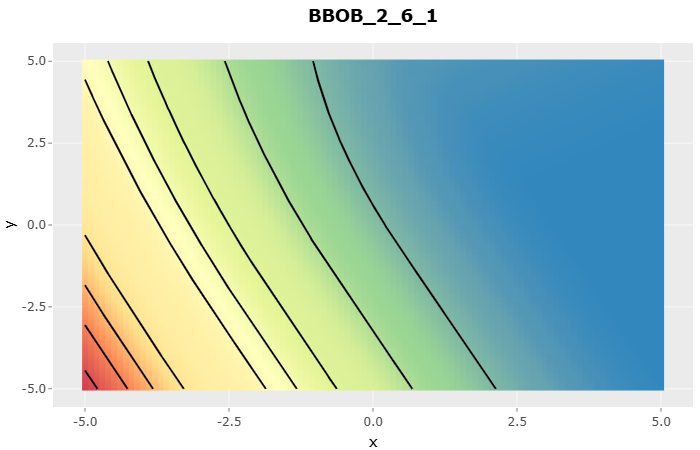}
        & \includegraphics[width=3.5cm]{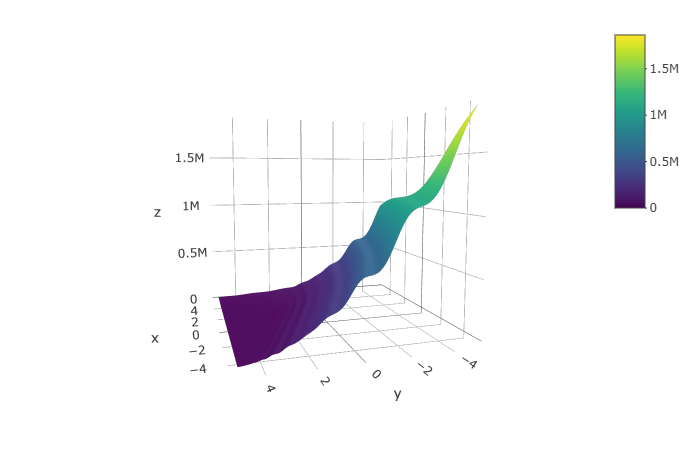} 
        & \includegraphics[width=3.5cm]{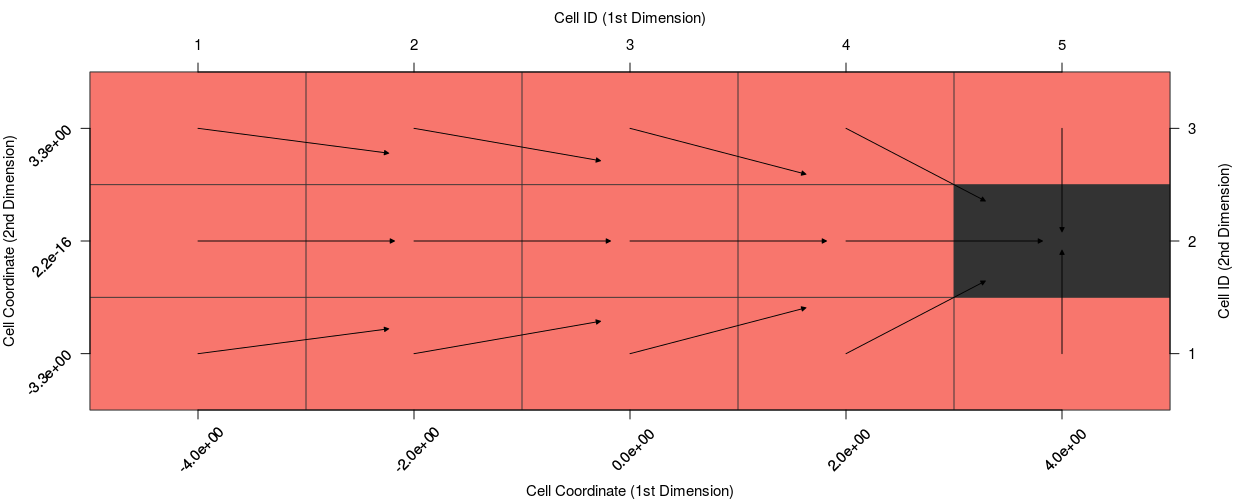}
        & \includegraphics[width=3.5cm]{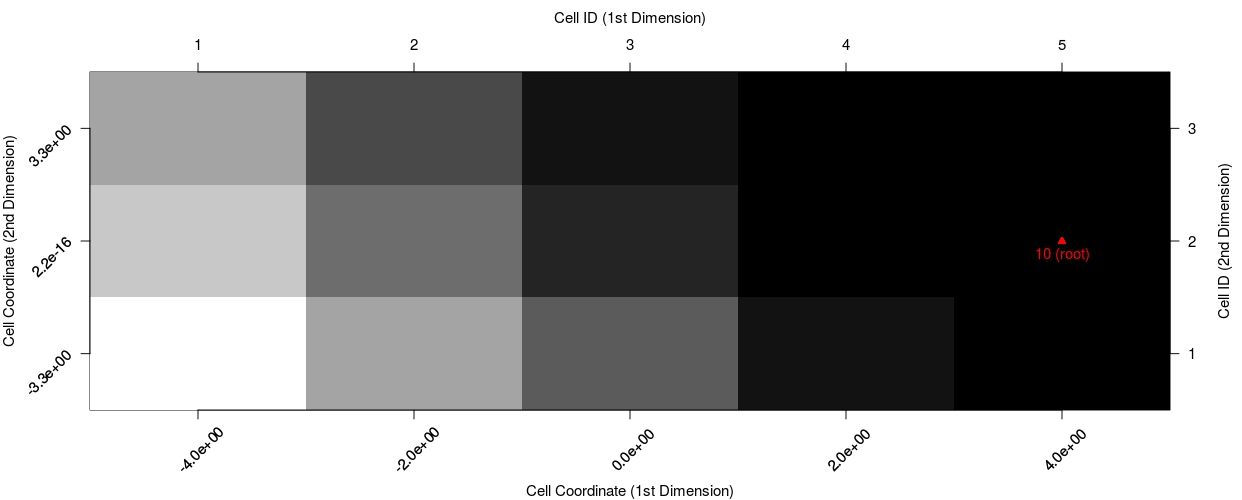}
        & \includegraphics[width=4.5cm]{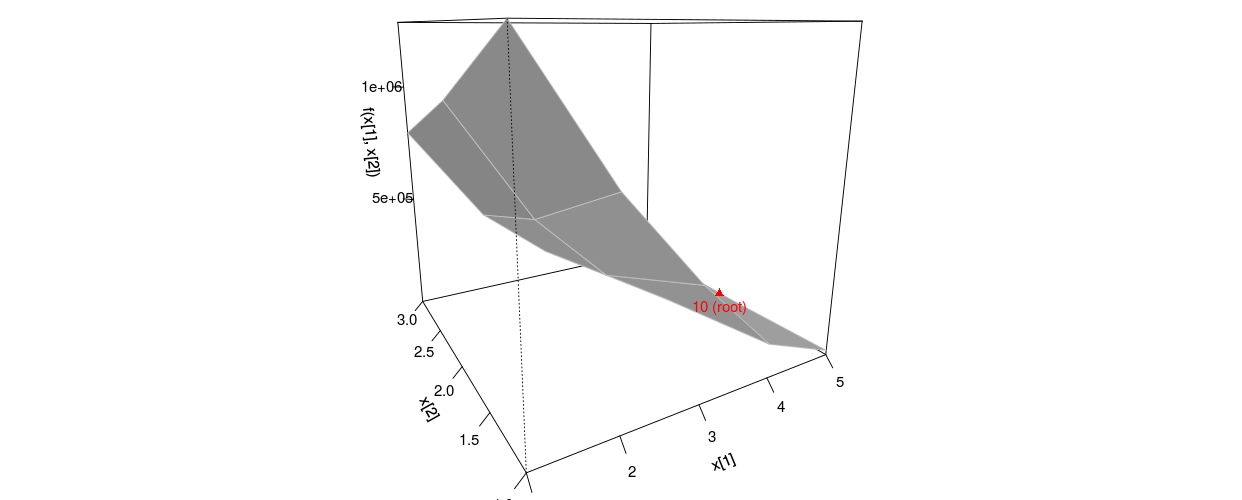}
        & \includegraphics[width=3.5cm]{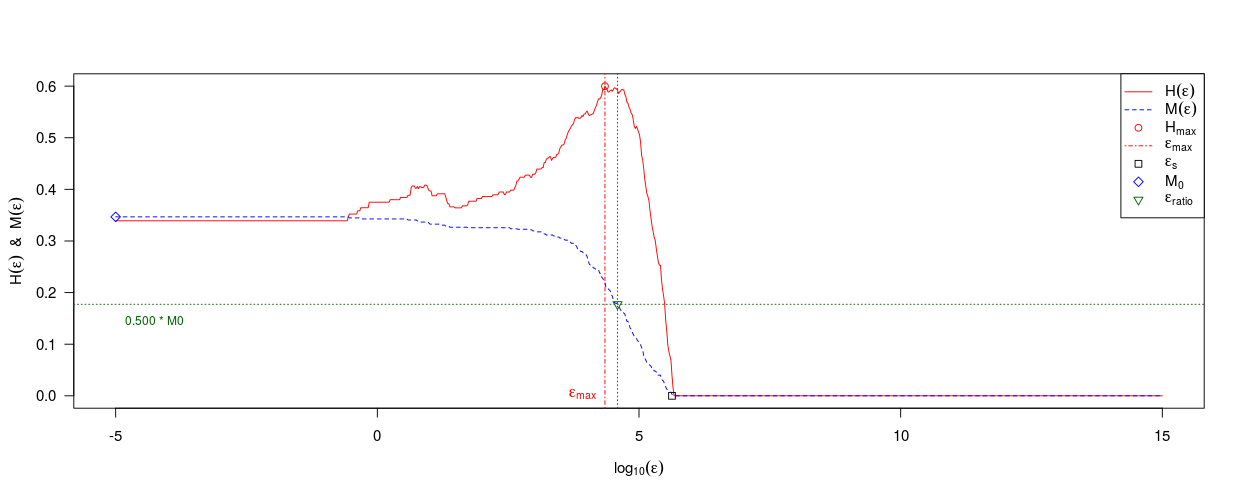}
        \\
 \hline
     \multirow{1}{*}{$f_7$}  & \includegraphics[width=3.5cm]{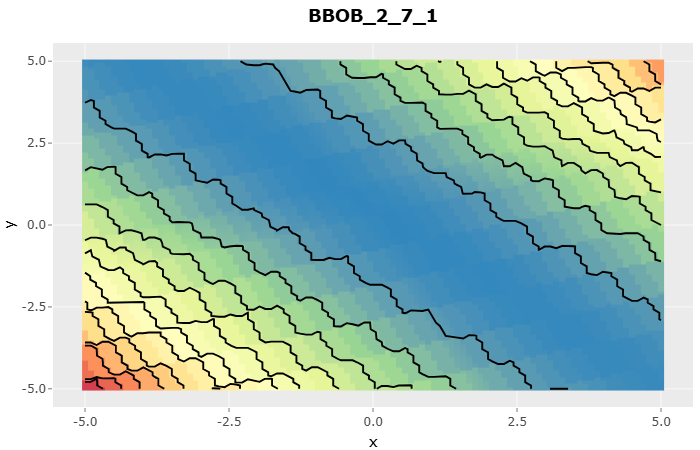}
        & \includegraphics[width=3.5cm]{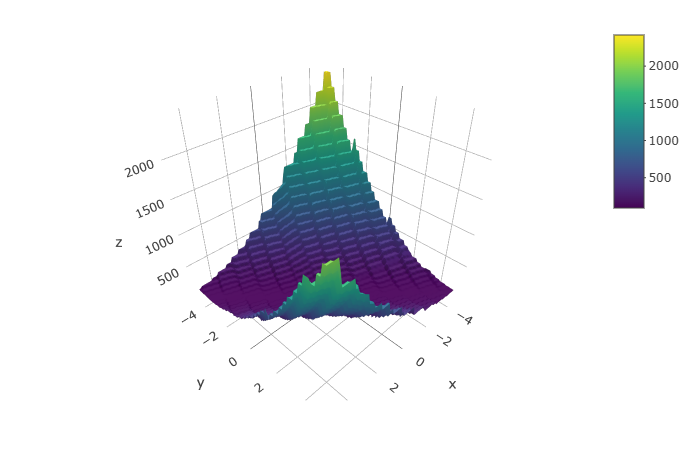} 
        & \includegraphics[width=3.5cm]{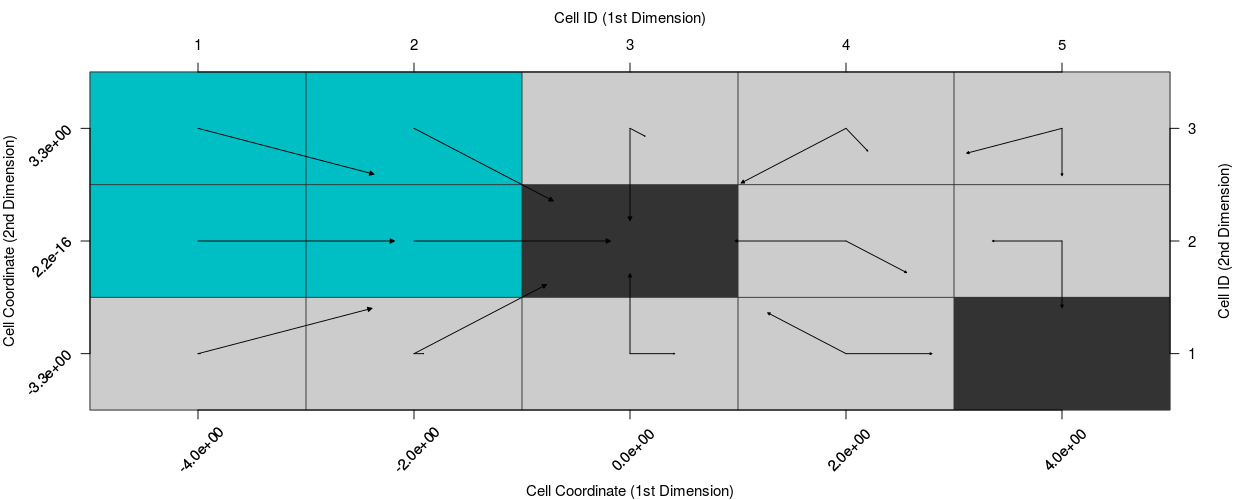 }
        & \includegraphics[width=3.5cm]{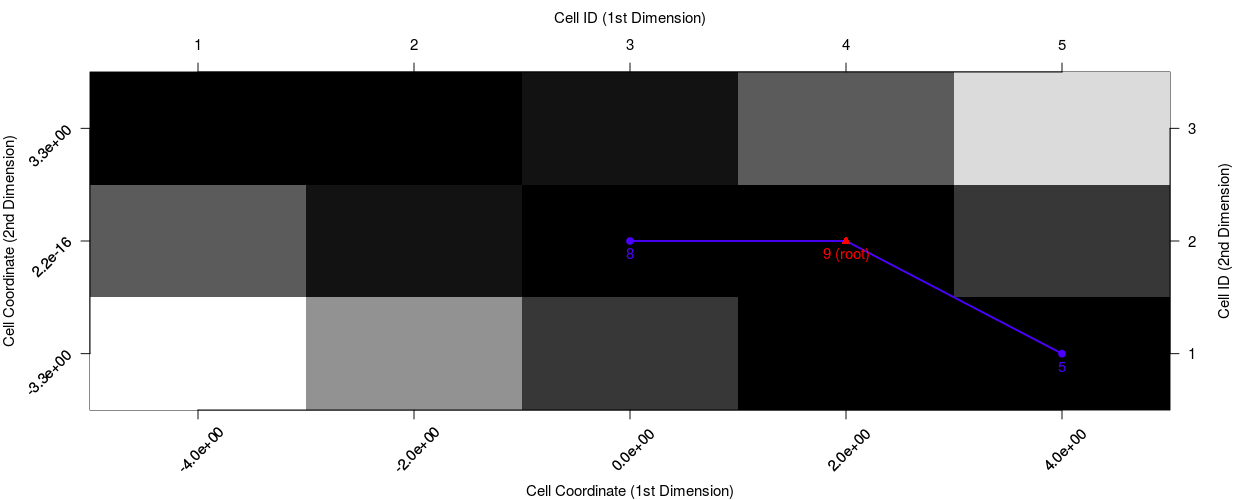 }
        & \includegraphics[width=4.5cm]{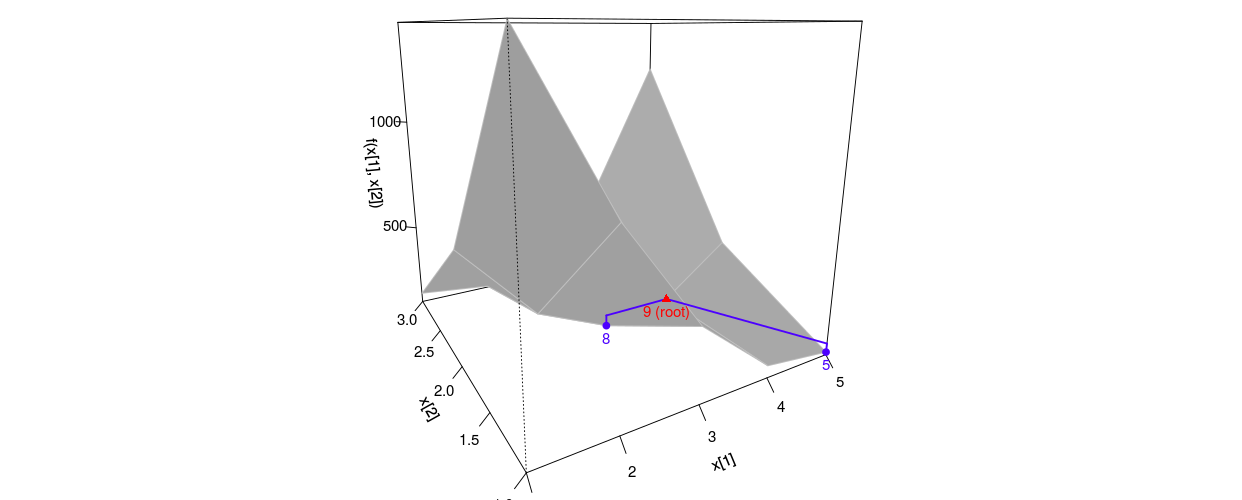 }
        & \includegraphics[width=3.5cm]{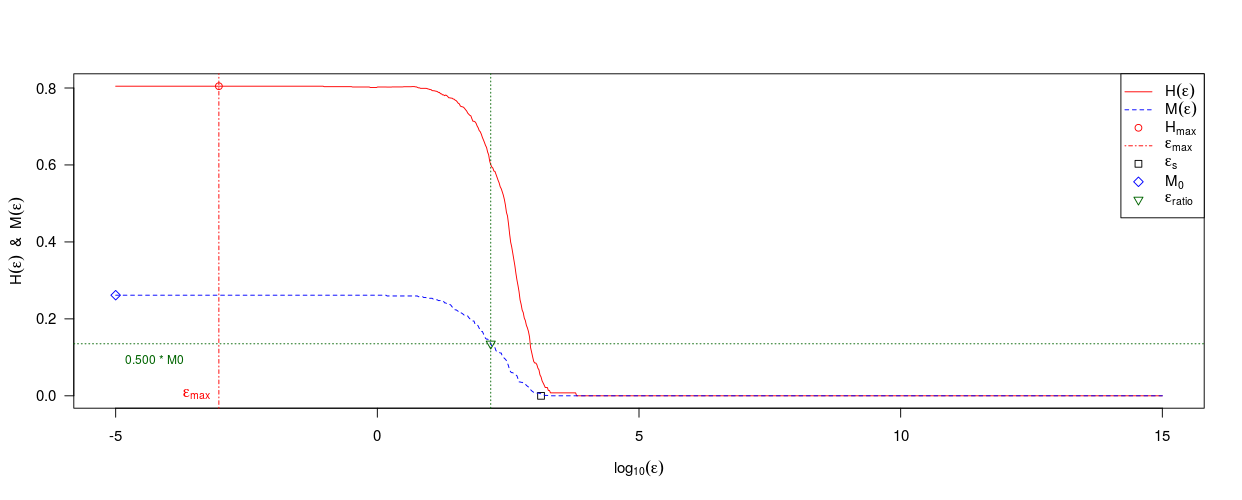}
        \\
 \hline
    \multirow{1}{*}{$f_8$}  & \includegraphics[width=3.5cm]{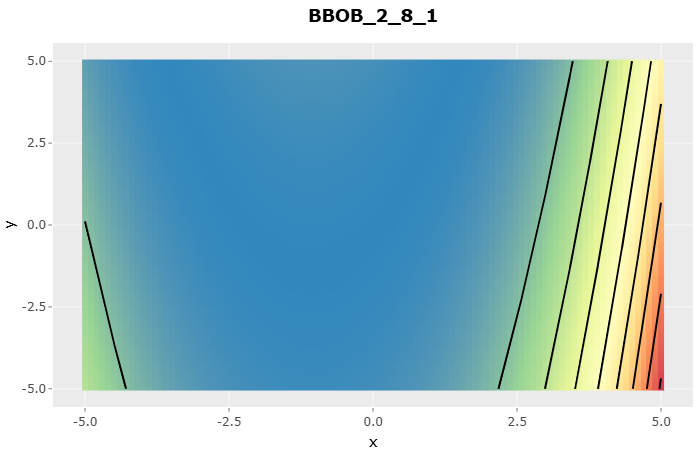}
        & \includegraphics[width=3.5cm]{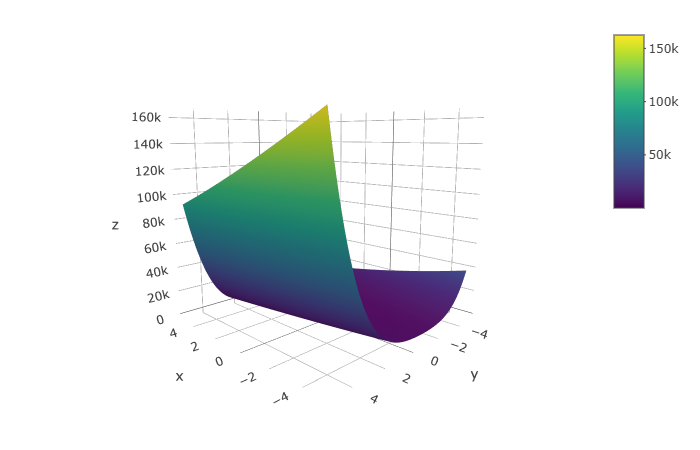} 
        & \includegraphics[width=3.5cm]{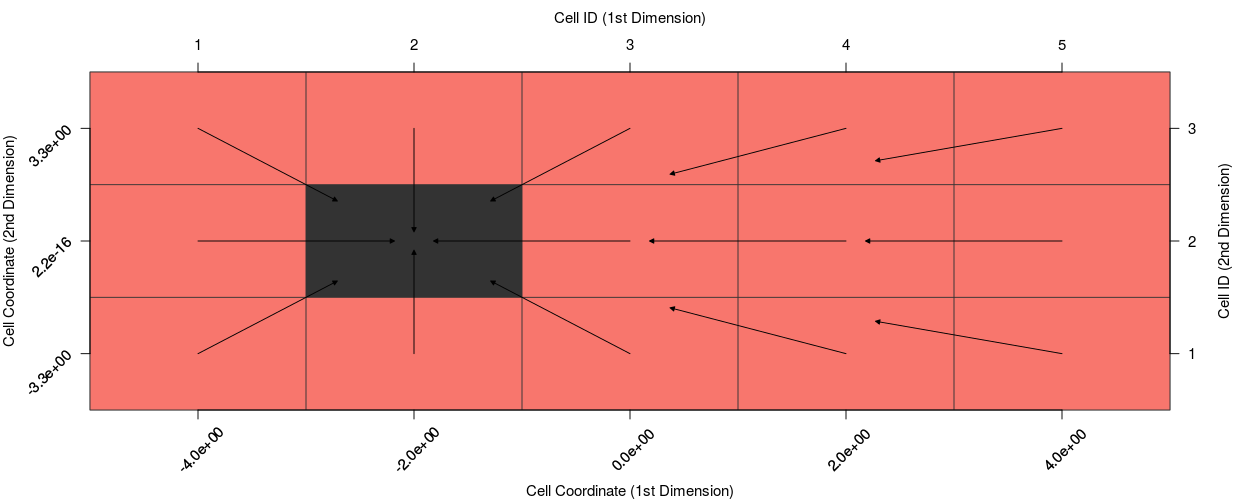 }
        & \includegraphics[width=3.5cm]{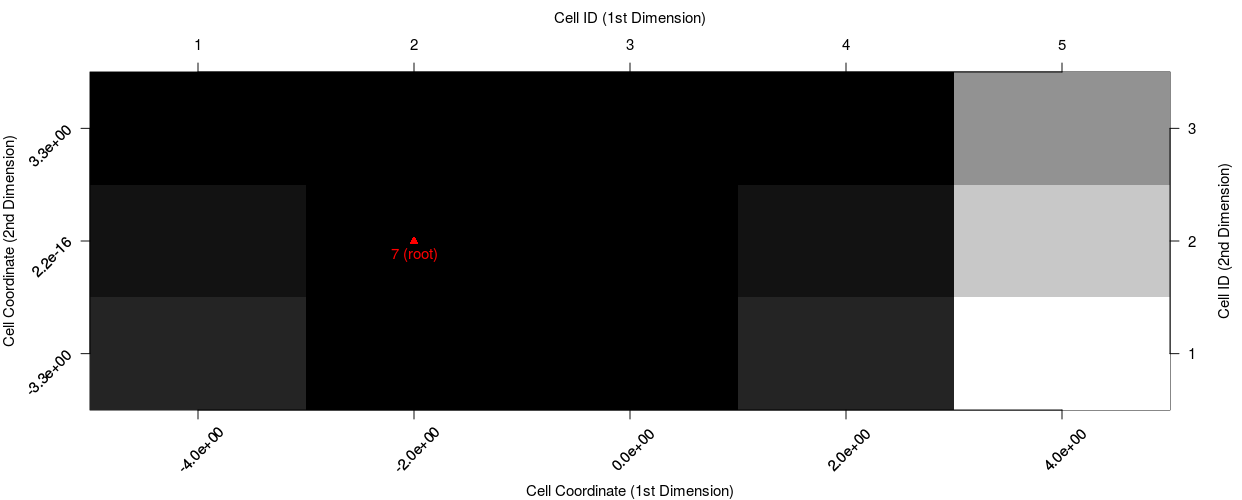}
        & \includegraphics[width=4.5cm]{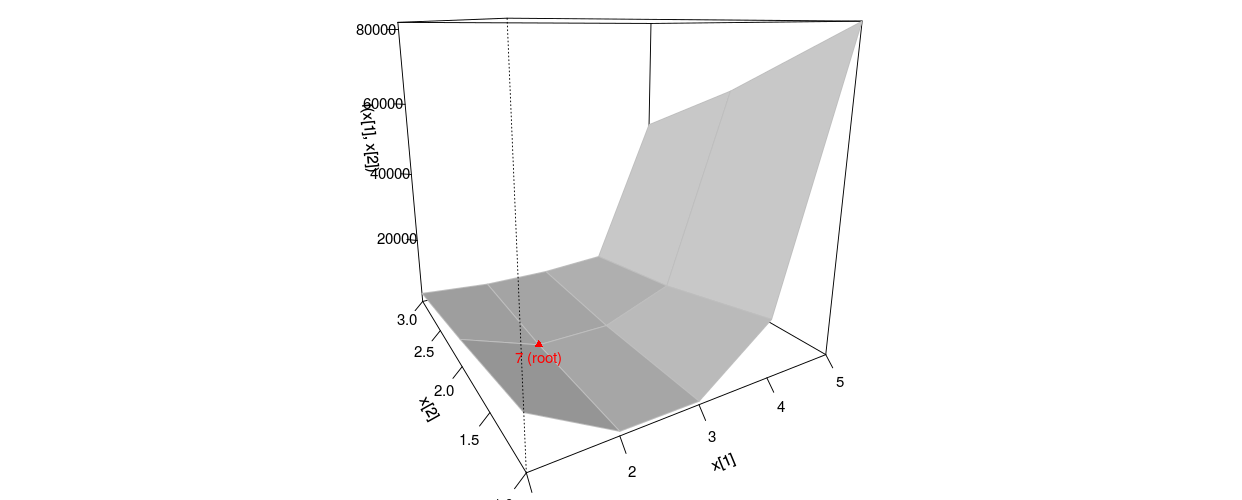}
        & \includegraphics[width=3.5cm]{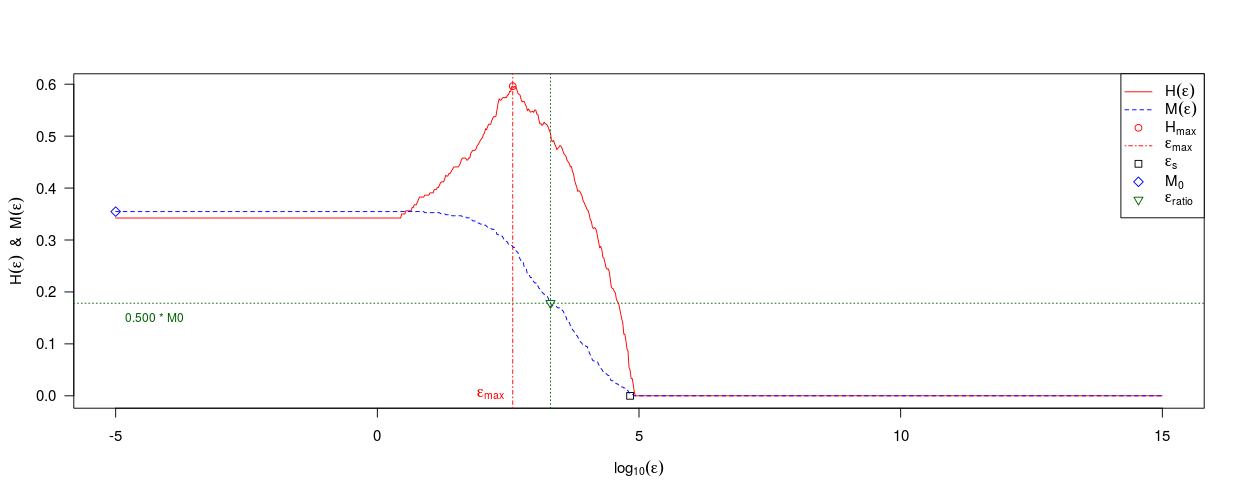 }
        \\
 \hline
 \end{tabular}
\end{table*}
\end{landscape}

\begin{landscape}
\begin{table*}[htp]
\centering
 \caption{Analysis of ELA features BBOB functions on d=2.}
\label{Tab ELA-2}
    \centering
   \scriptsize
   \setlength{\tabcolsep}{3pt}
   \begin{tabular}{ccccccc}
    \hline
      \textbf{Function}  & \textbf{Contour Plot} &  \textbf{Surface Plot}  &  \textbf{Cell- Mapping} &  \textbf{Barrier Tree-2D} &  \textbf{Barrier Tree-3D}  &  \textbf{Information Content}\\ \hline

          \multirow{1}{*}{$f_9$}  & \includegraphics[width=3.5cm]{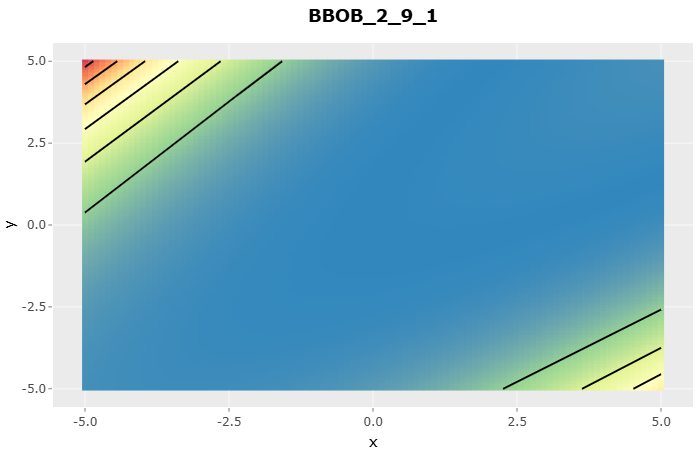}
        & \includegraphics[width=3.5cm]{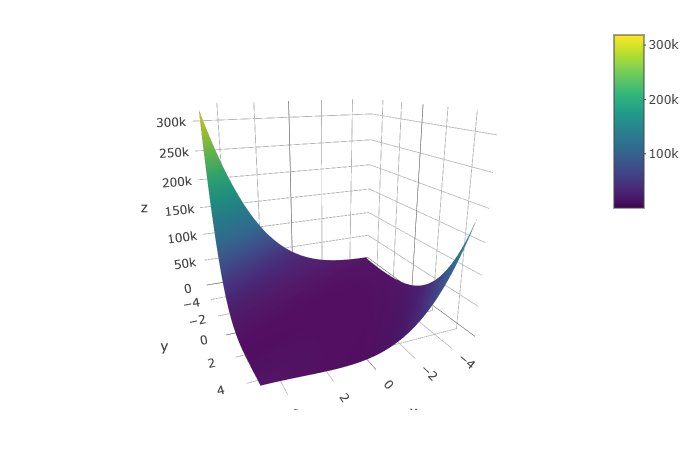} 
        & \includegraphics[width=3.5cm]{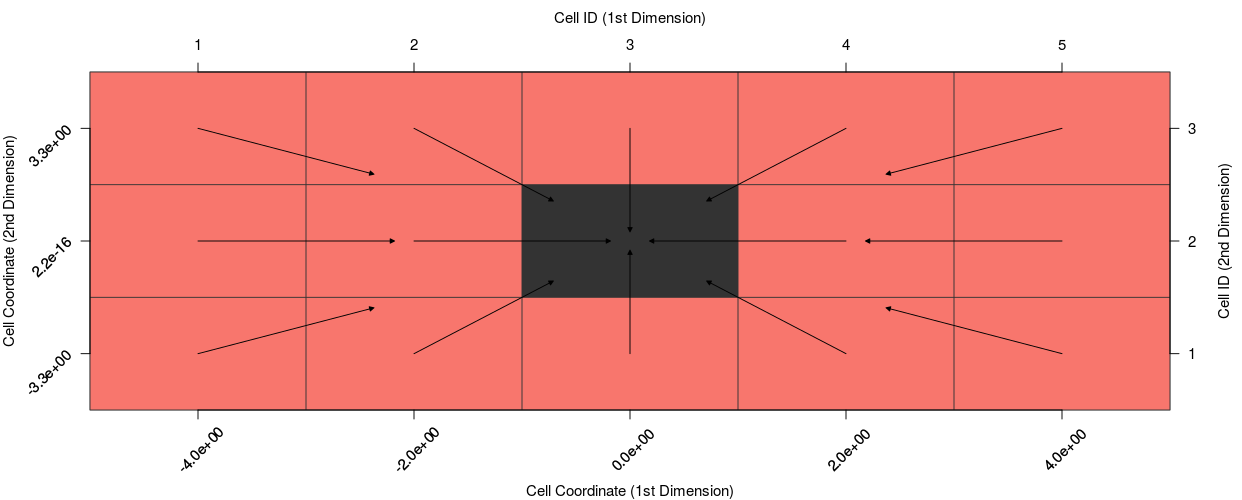 }
        & \includegraphics[width=3.5cm]{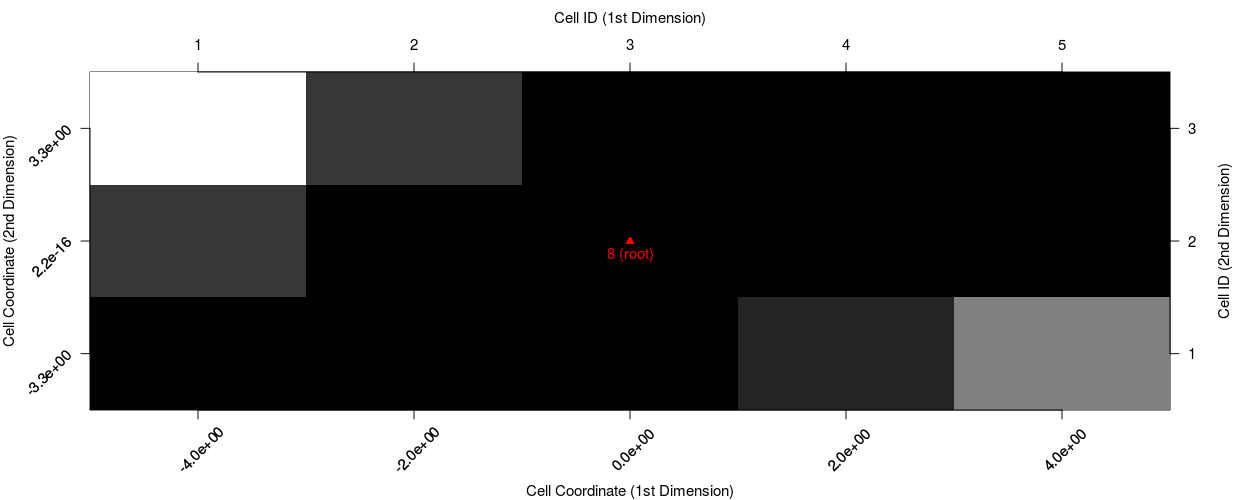}
        & \includegraphics[width=4.5cm]{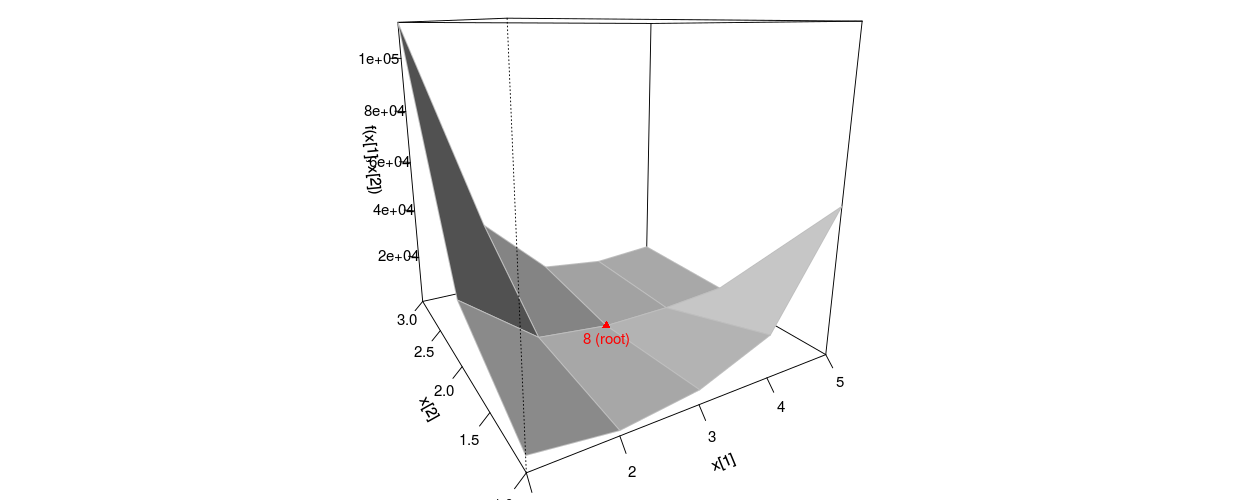}
        & \includegraphics[width=3.5cm]{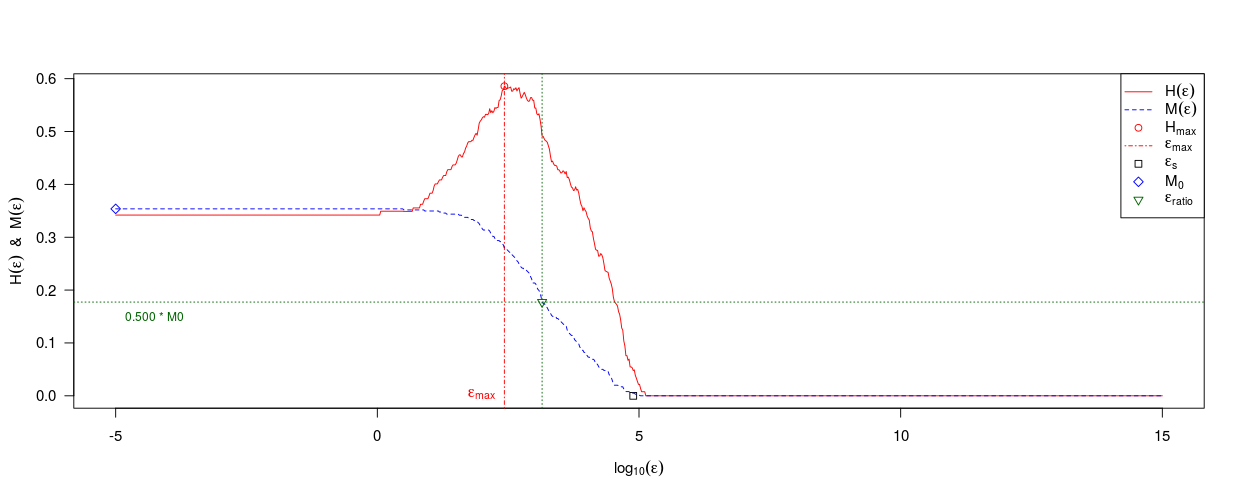}
        \\
 \hline
        \multirow{1}{*}{$f_{10}$}  & \includegraphics[width=3.5cm]{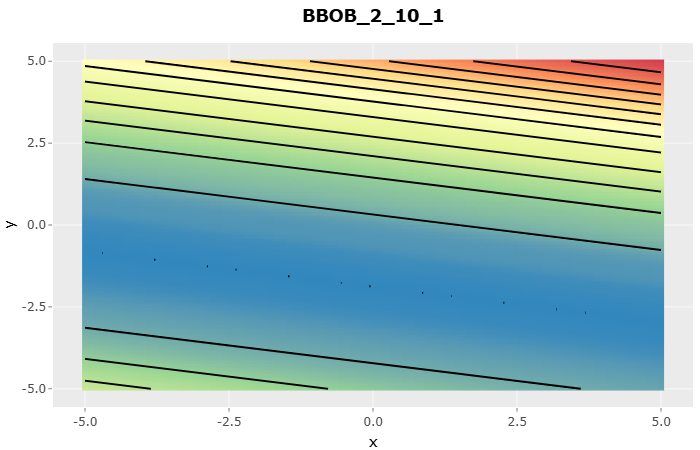}
        & \includegraphics[width=3.5cm]{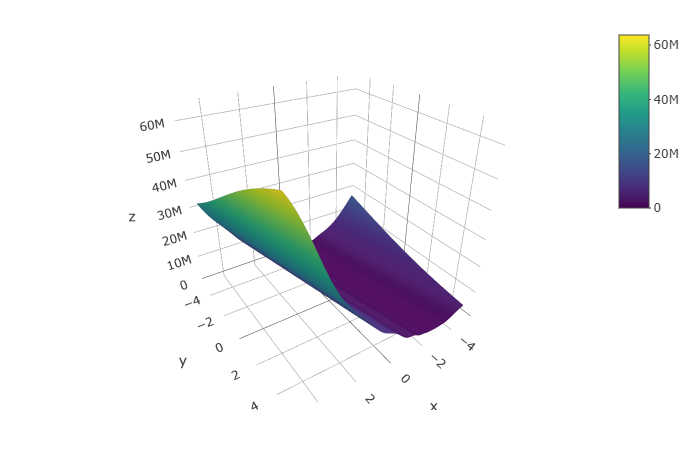} 
        & \includegraphics[width=3.5cm]{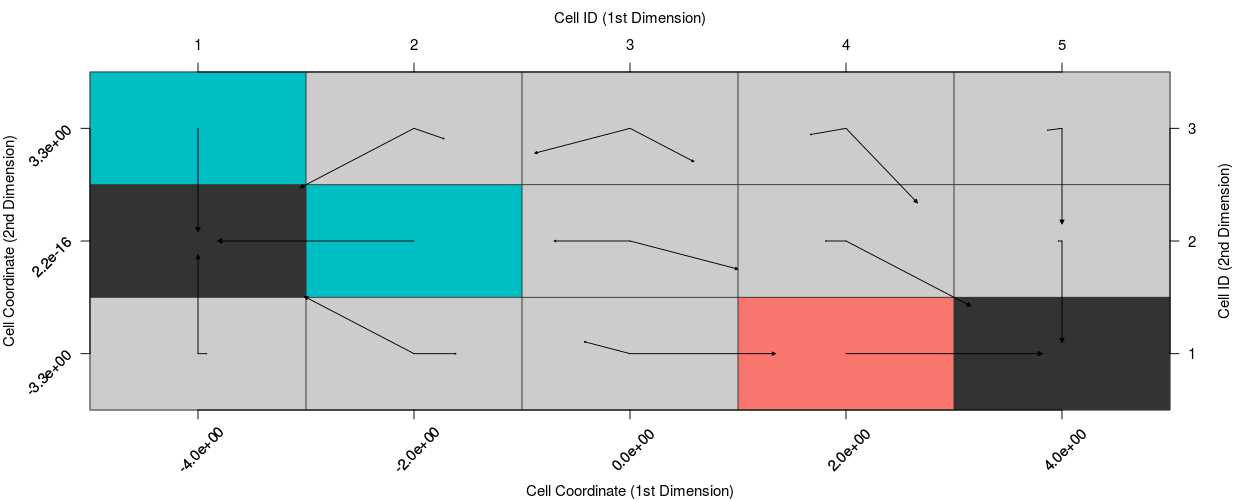 }
        & \includegraphics[width=3.5cm]{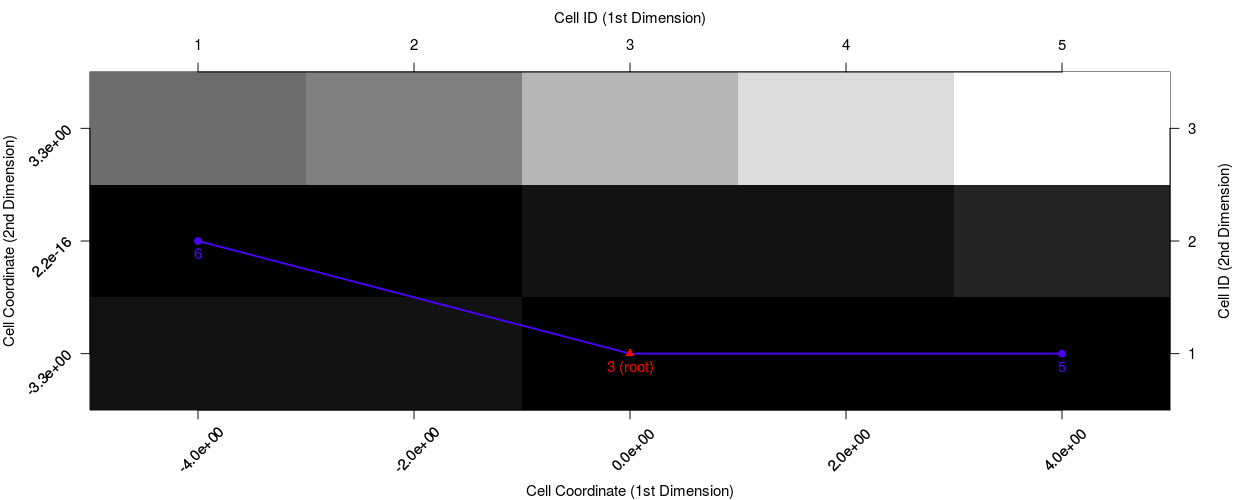 }
        & \includegraphics[width=4.5cm]{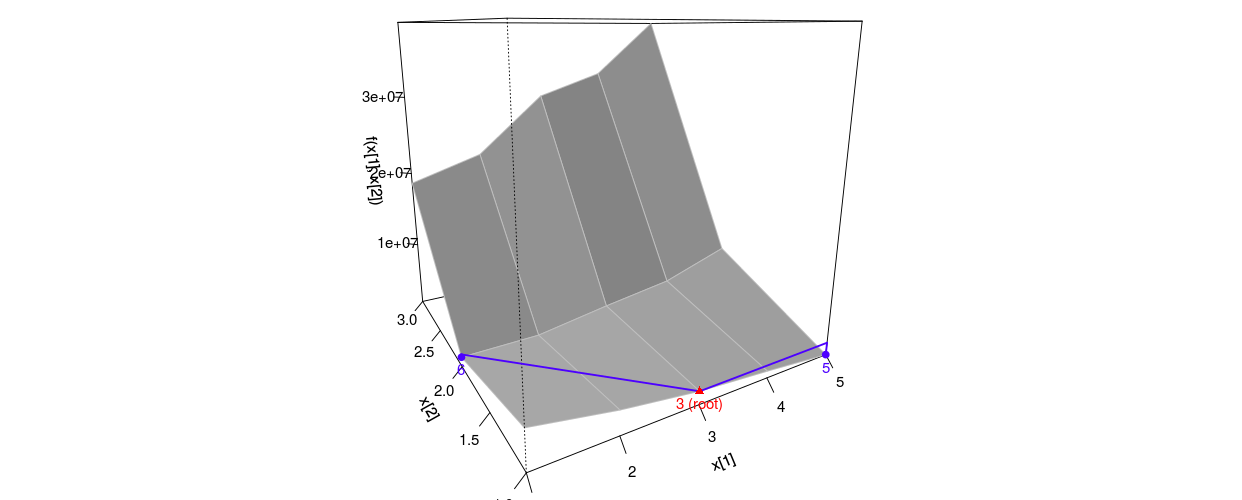 }
        & \includegraphics[width=3.5cm]{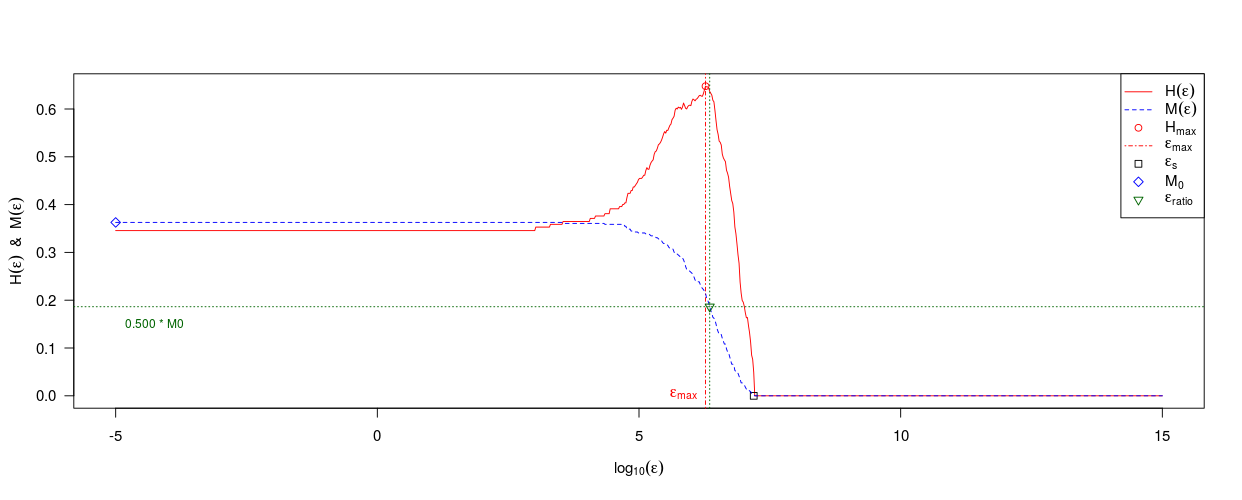}
        \\
 \hline
 
        \multirow{1}{*}{$f_{11}$}  & \includegraphics[width=3.5cm]{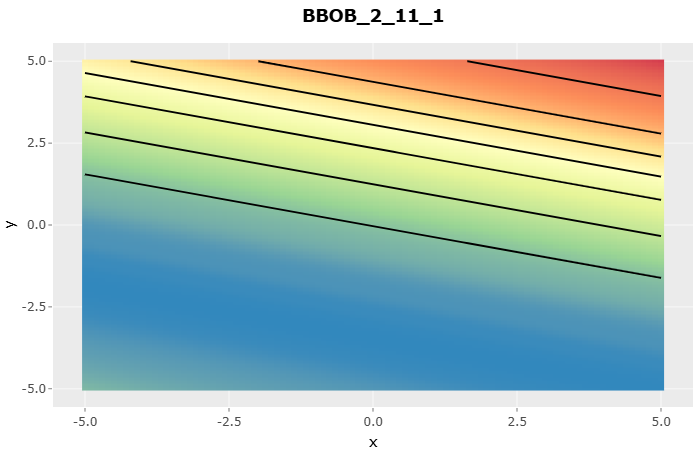}
        & \includegraphics[width=3.5cm]{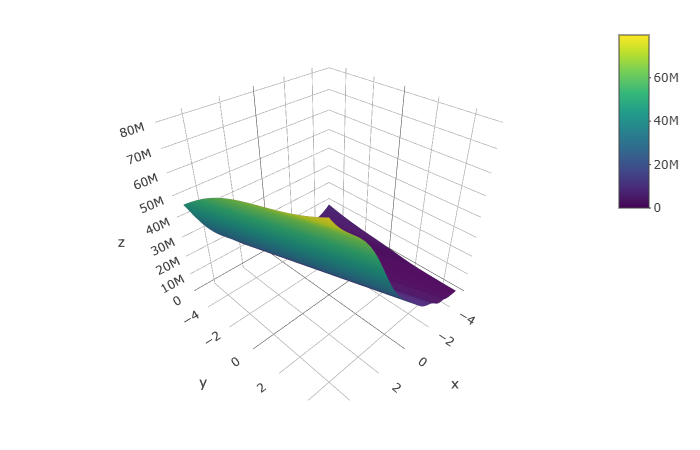} 
        & \includegraphics[width=3.5cm]{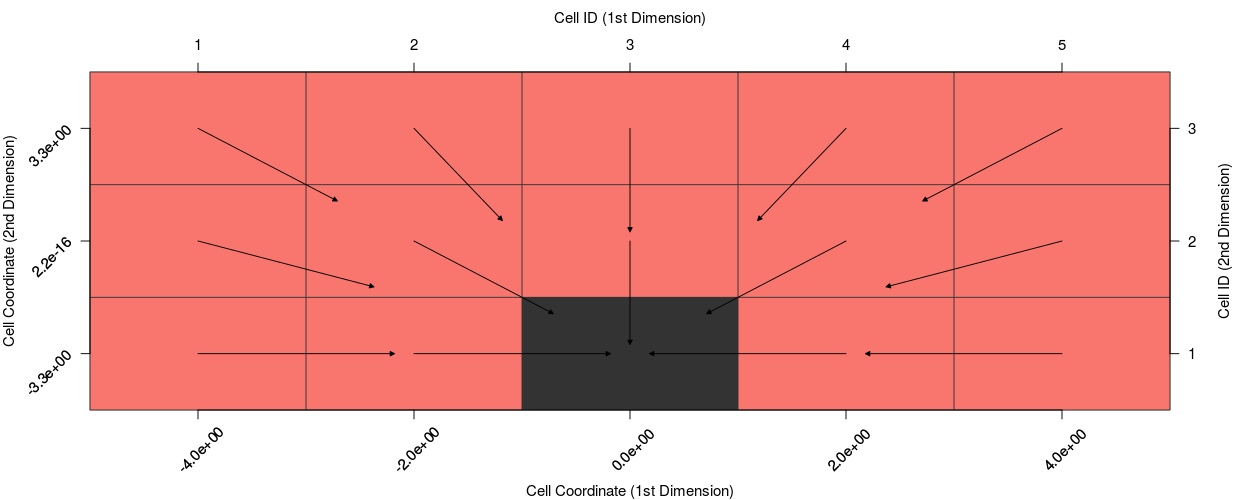  }
        & \includegraphics[width=3.5cm]{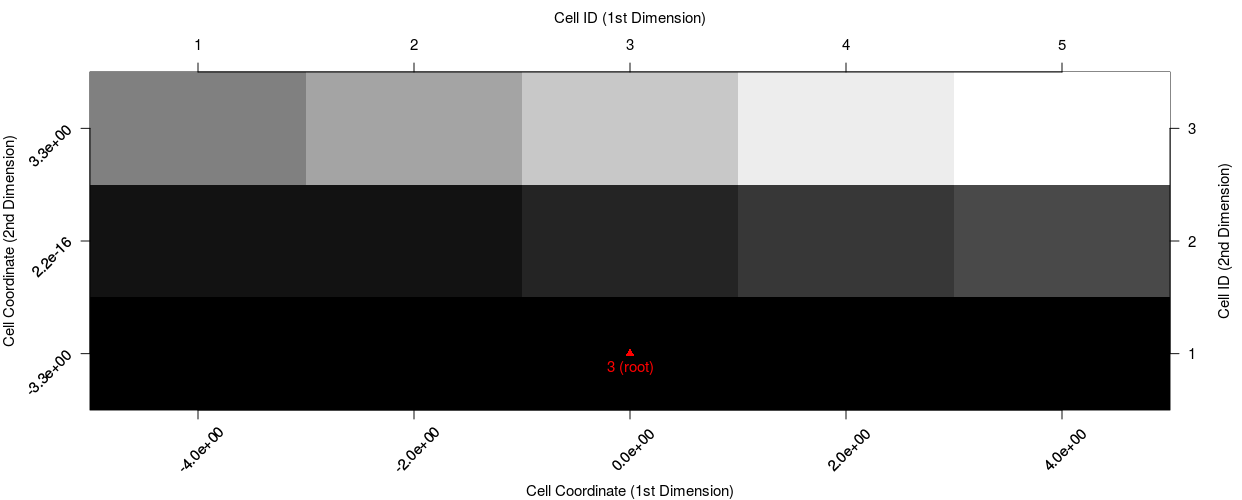}
        & \includegraphics[width=4.5cm]{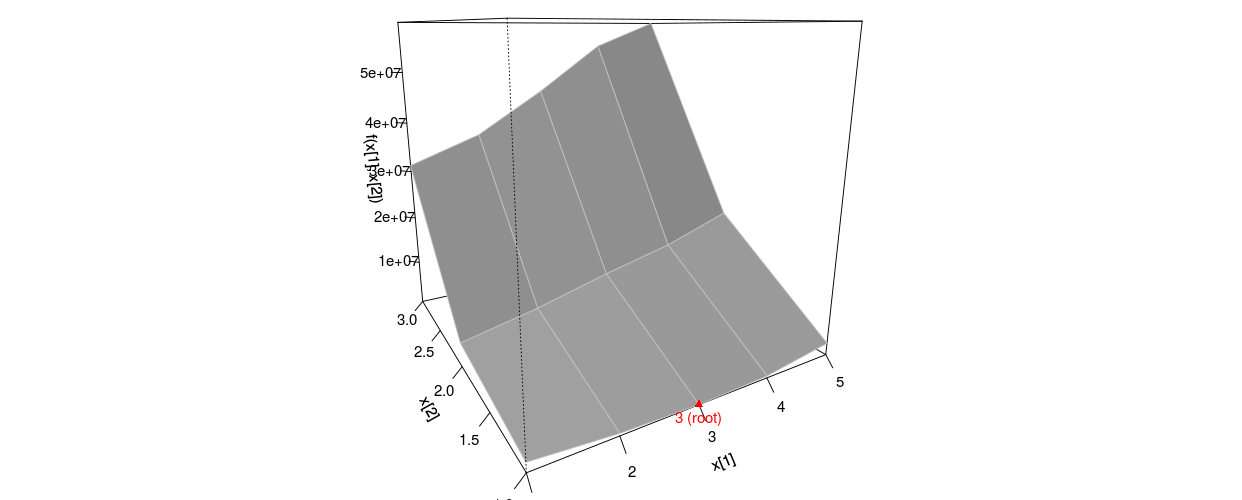}
        & \includegraphics[width=3.5cm]{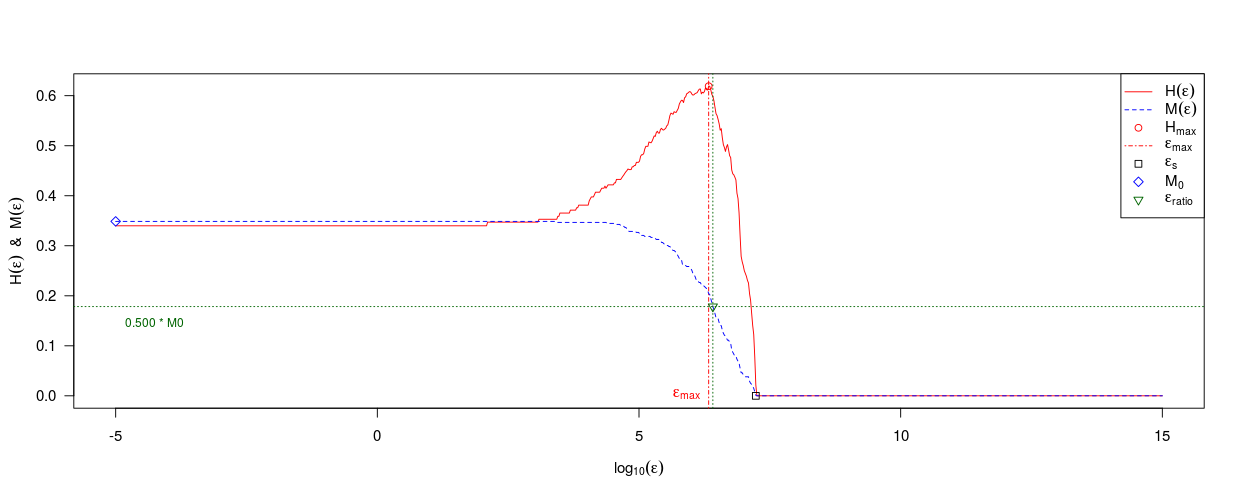 }
        \\
 \hline

          \multirow{1}{*}{$f_{12}$}  & \includegraphics[width=3.5cm]{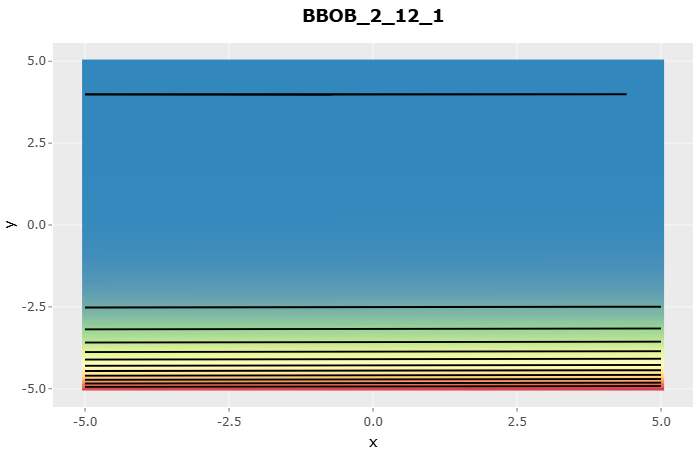}
        & \includegraphics[width=3.5cm]{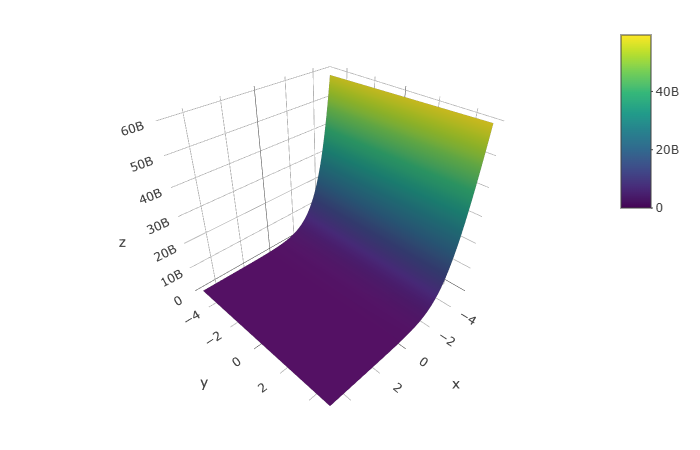} 
        & \includegraphics[width=3.5cm]{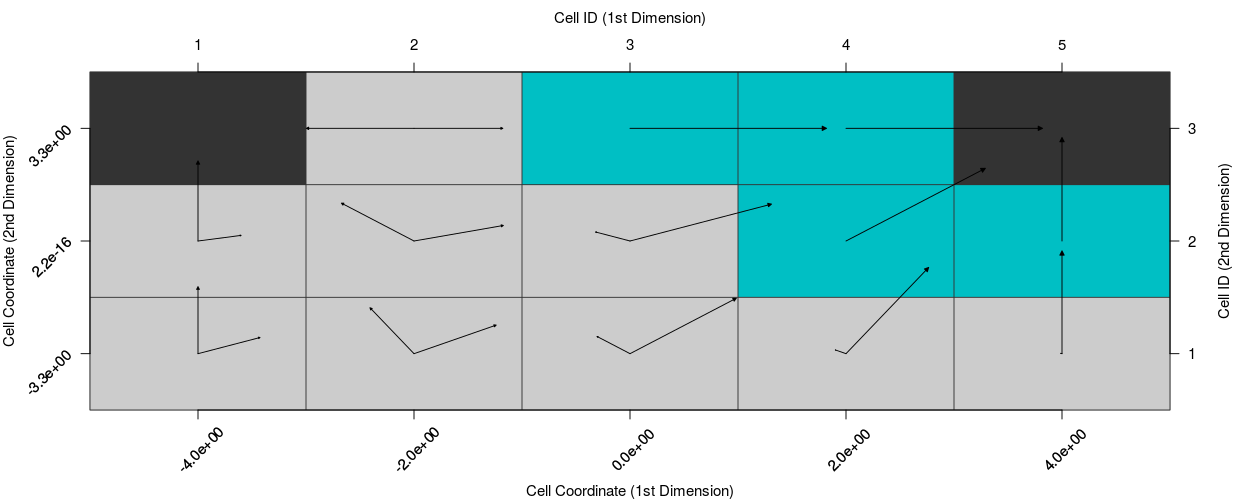  }
        & \includegraphics[width=3.5cm]{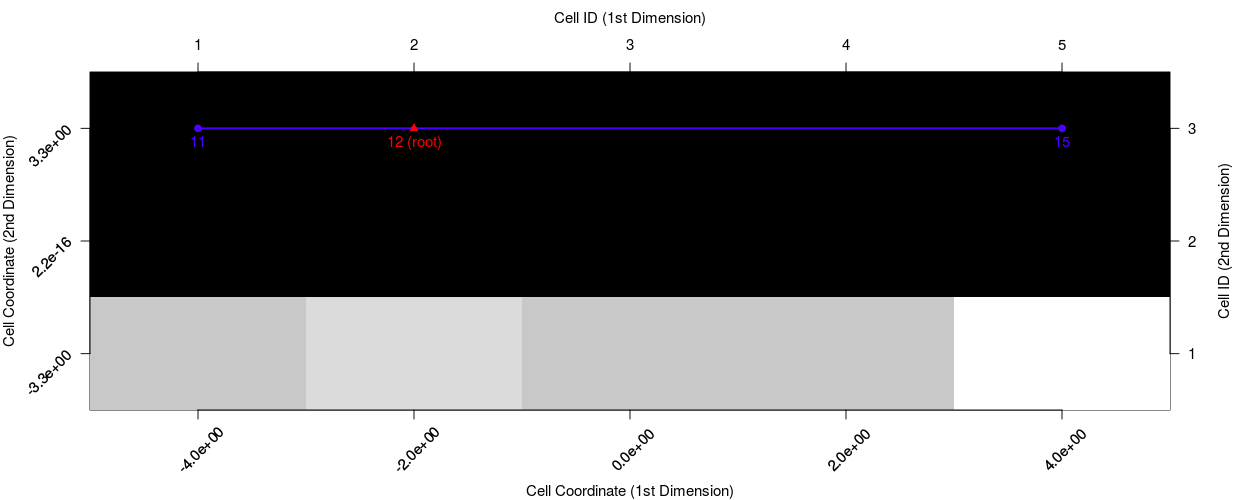 }
        & \includegraphics[width=4.5cm]{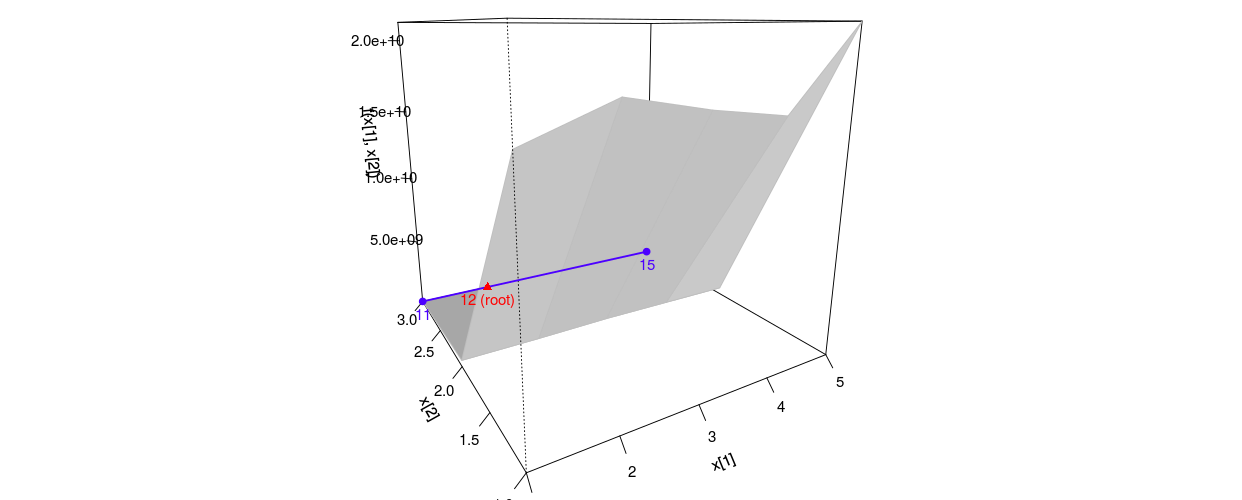 }
        & \includegraphics[width=3.5cm]{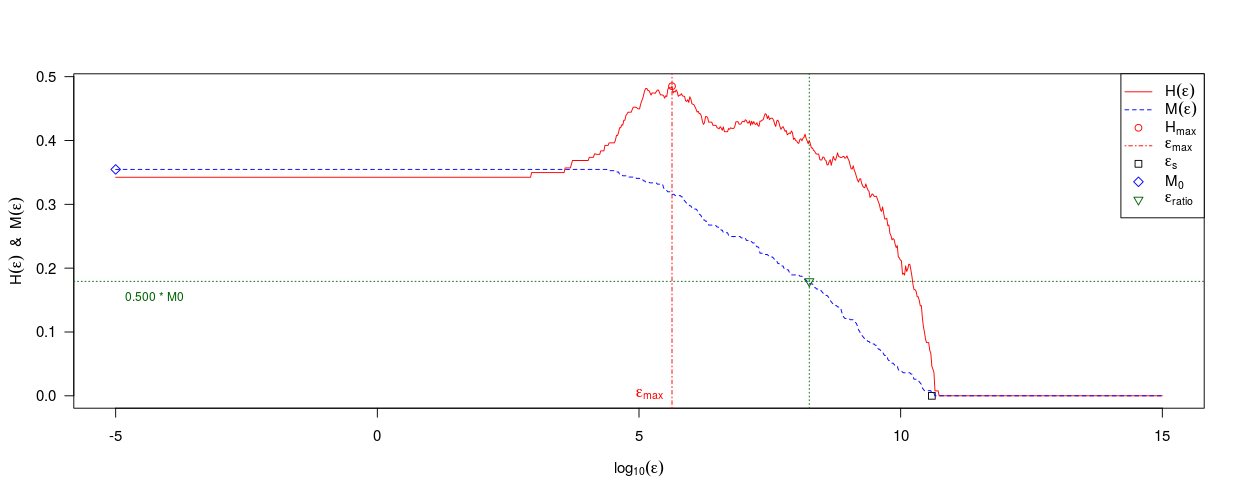}
        \\
 \hline
         \multirow{1}{*}{$f_{13}$}  & \includegraphics[width=3.5cm]{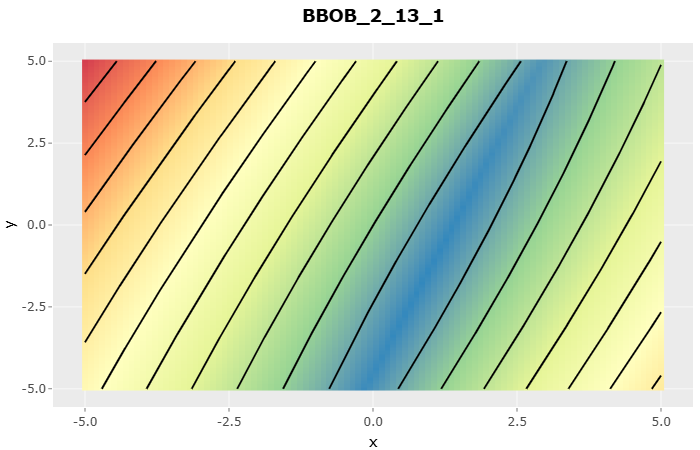}
        & \includegraphics[width=3.5cm]{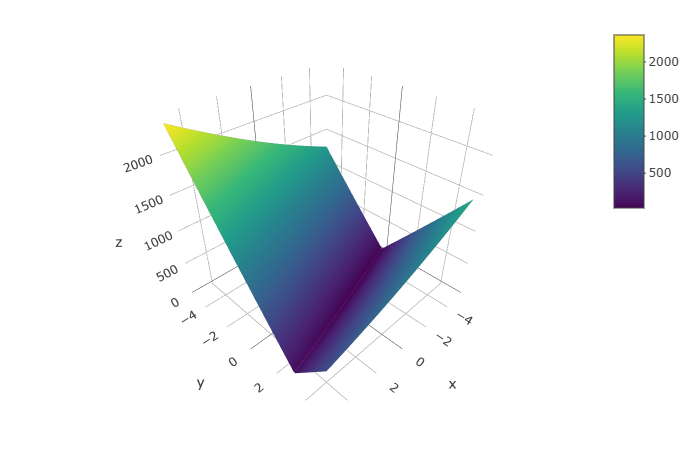} 
        & \includegraphics[width=3.5cm]{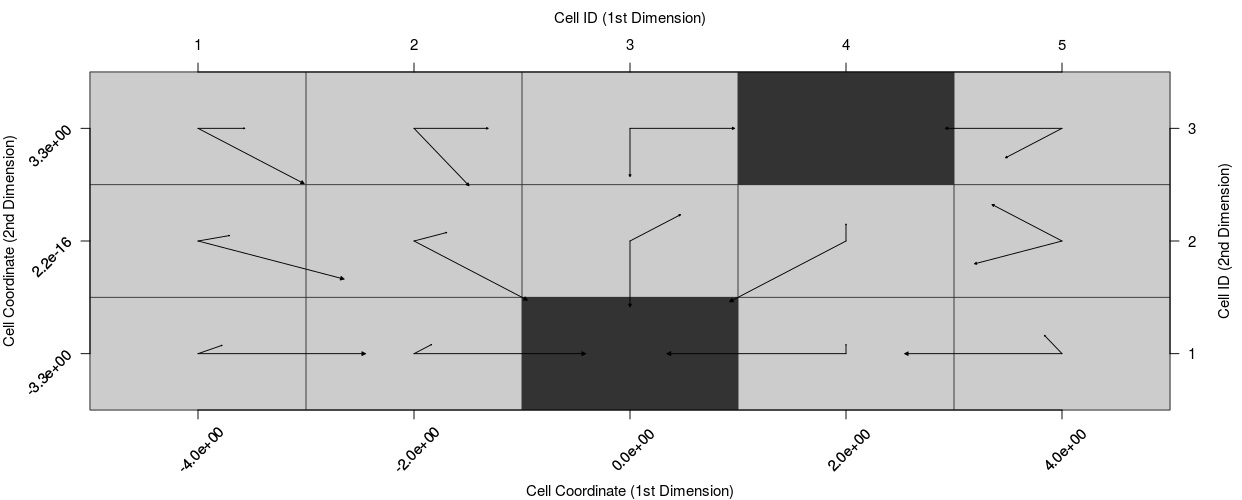  }
        & \includegraphics[width=3.5cm]{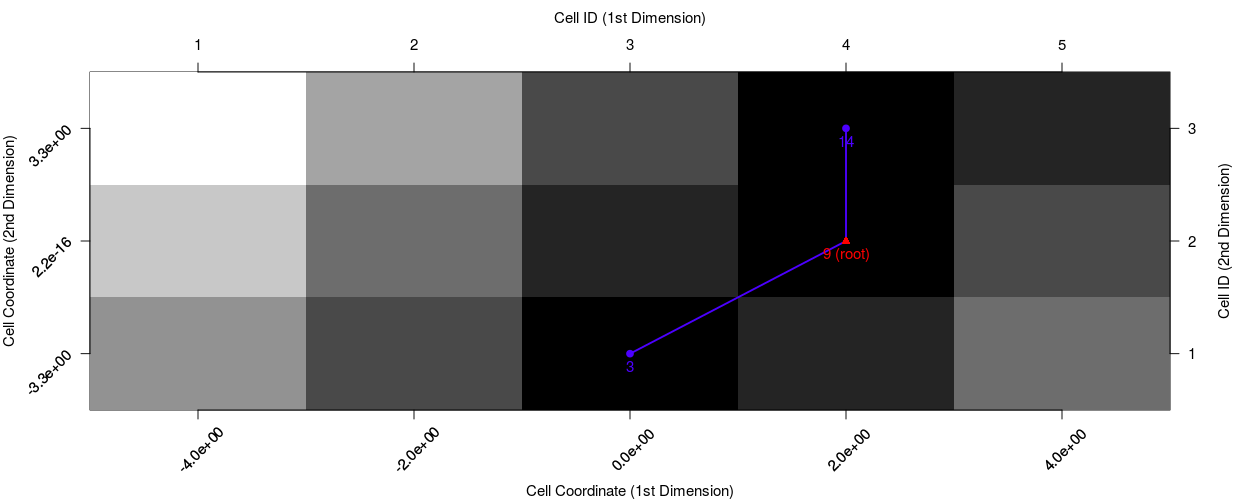  }
        & \includegraphics[width=4.5cm]{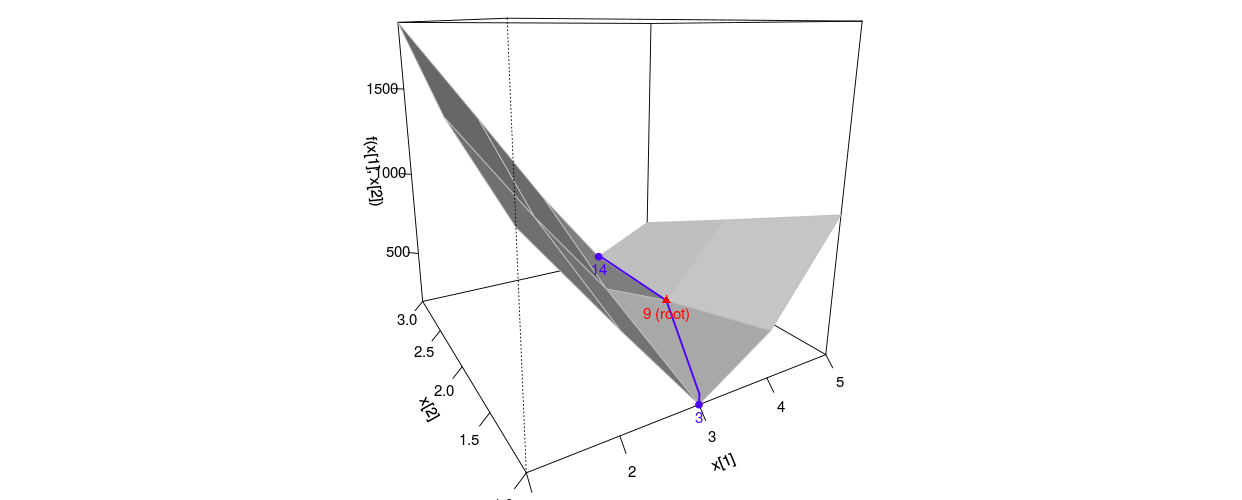  }
        & \includegraphics[width=3.5cm]{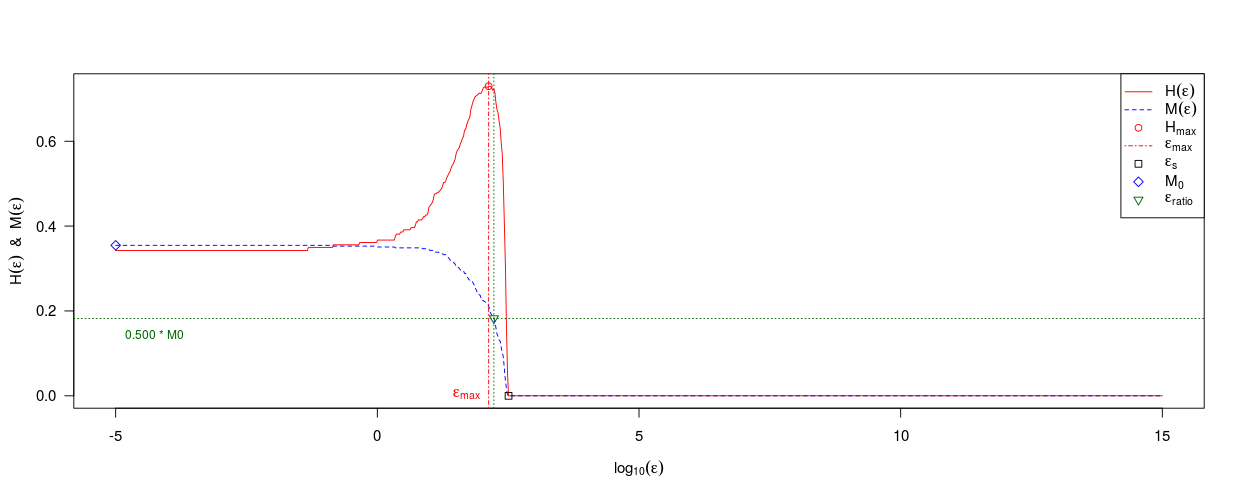 }
        \\
 \hline
         \multirow{1}{*}{$f_{14}$}  & \includegraphics[width=3.5cm]{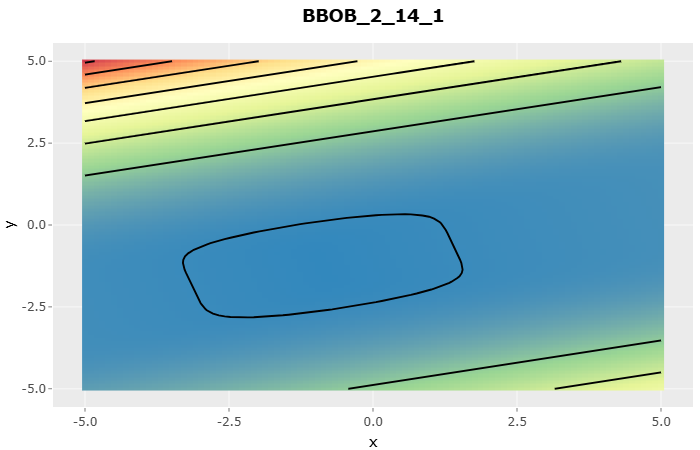}
        & \includegraphics[width=3.5cm]{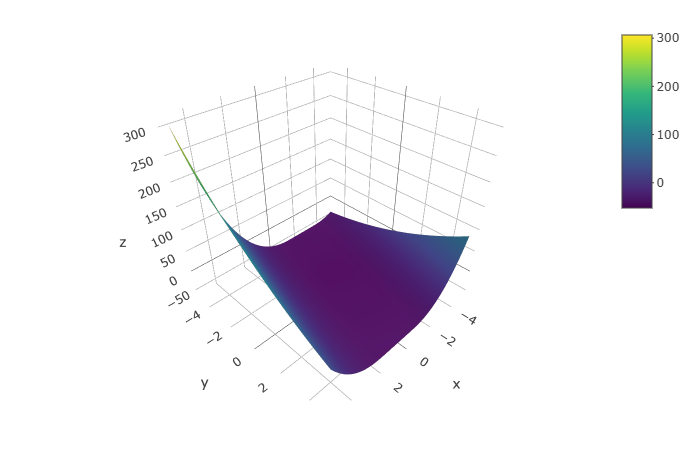} 
        & \includegraphics[width=3.5cm]{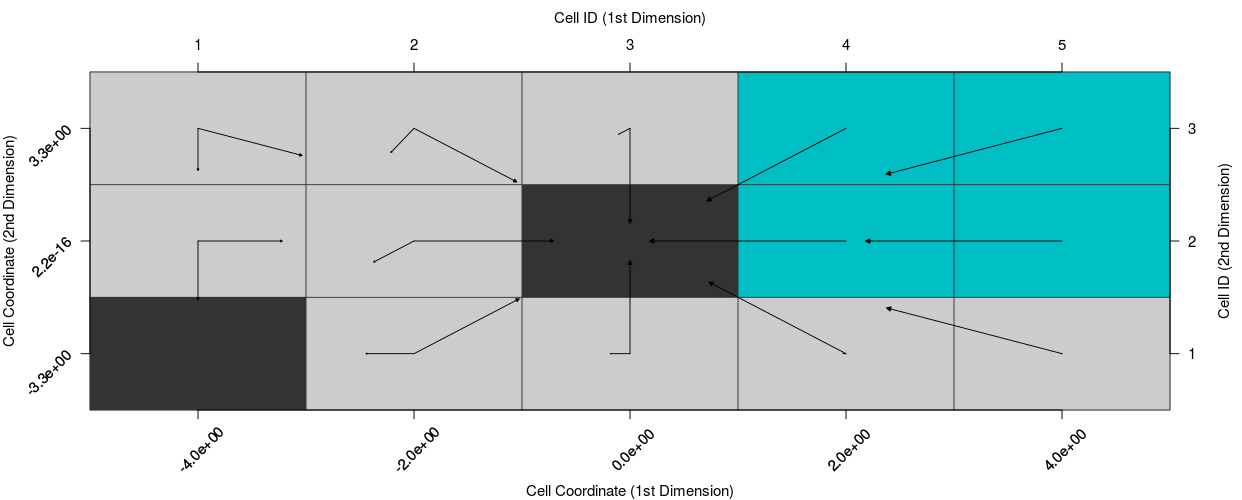   }
        & \includegraphics[width=3.5cm]{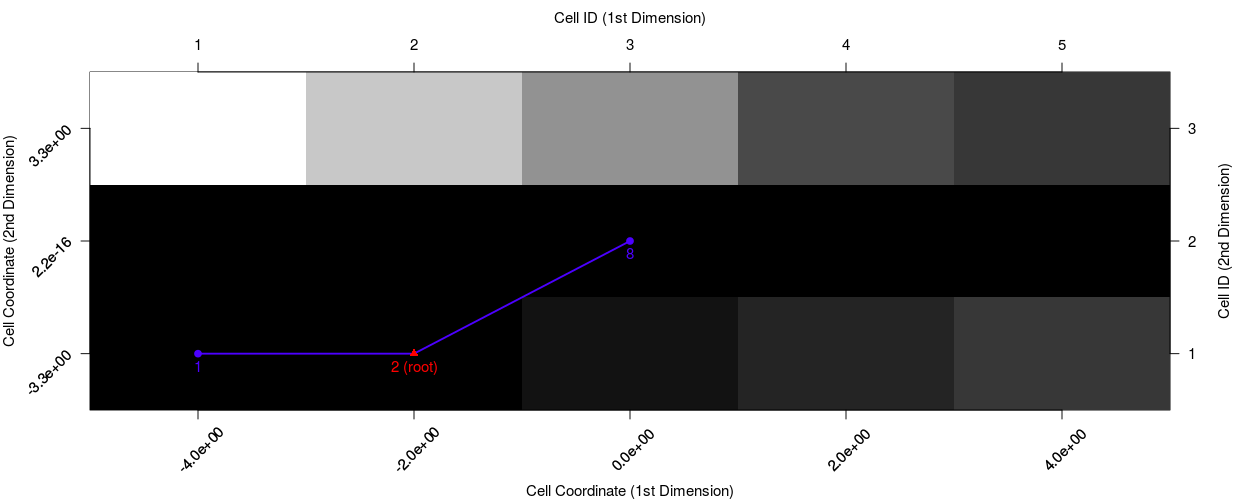   }
        & \includegraphics[width=4.5cm]{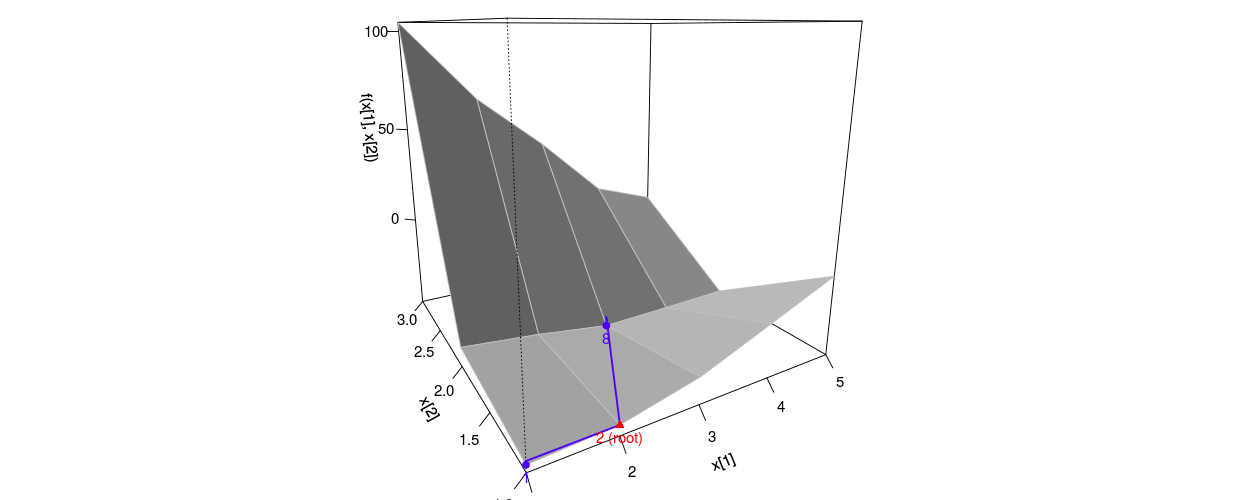   }
        & \includegraphics[width=3.5cm]{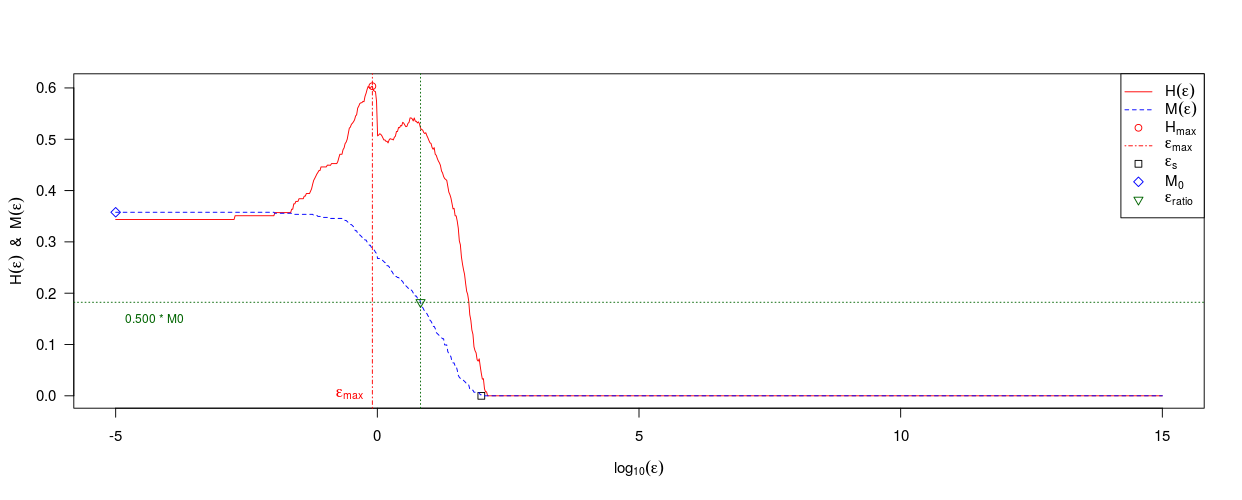  }
        \\
 \hline

        \multirow{1}{*}{$f_{15}$}  & \includegraphics[width=3.5cm]{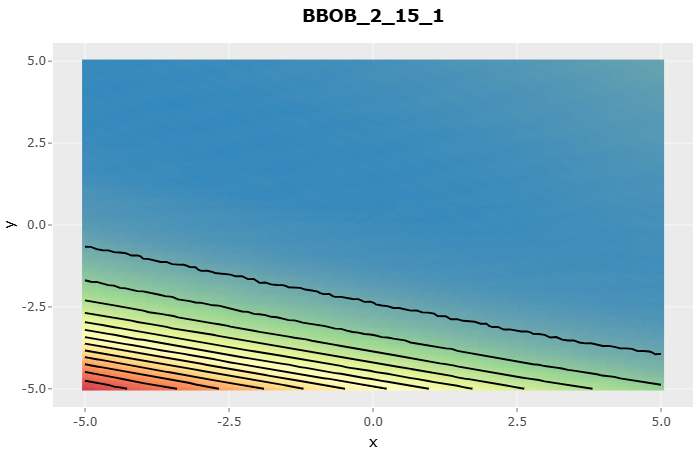}
        & \includegraphics[width=3.5cm]{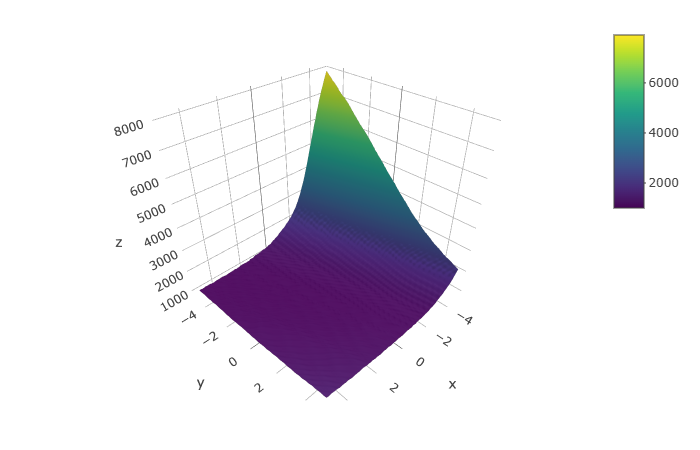} 
        & \includegraphics[width=3.5cm]{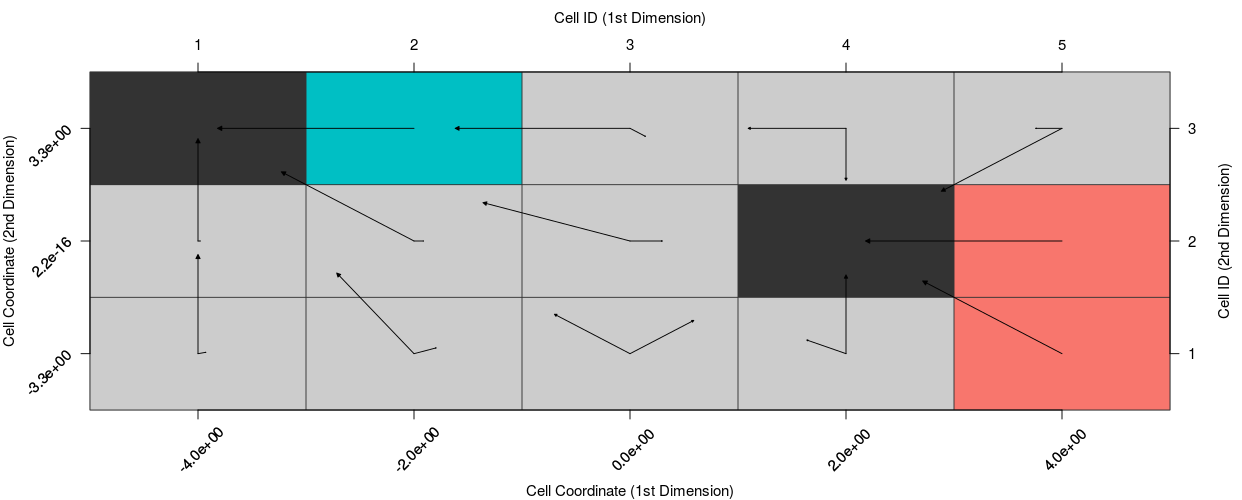}
        & \includegraphics[width=3.5cm]{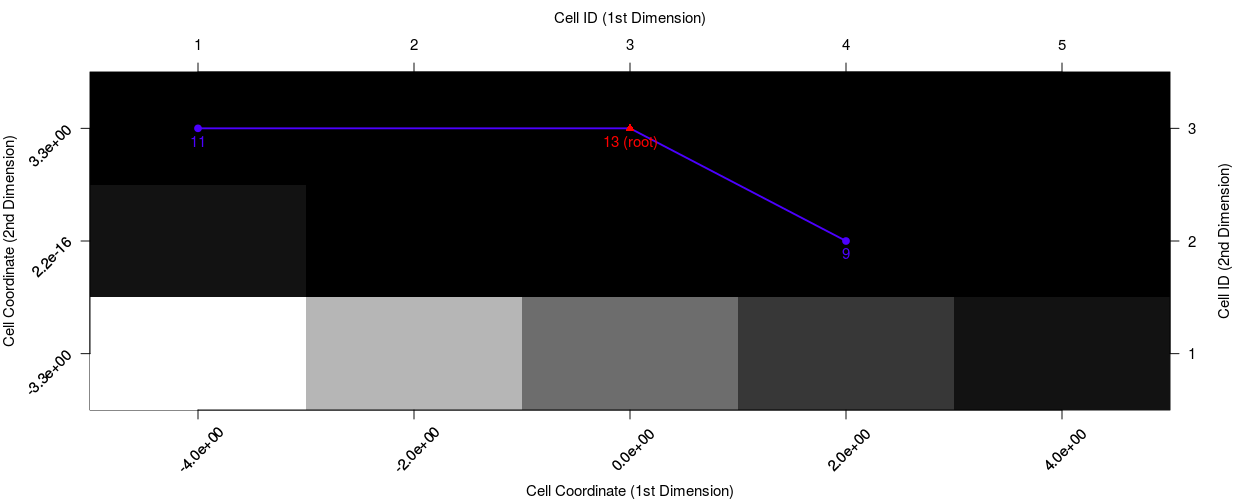}
        & \includegraphics[width=4.5cm]{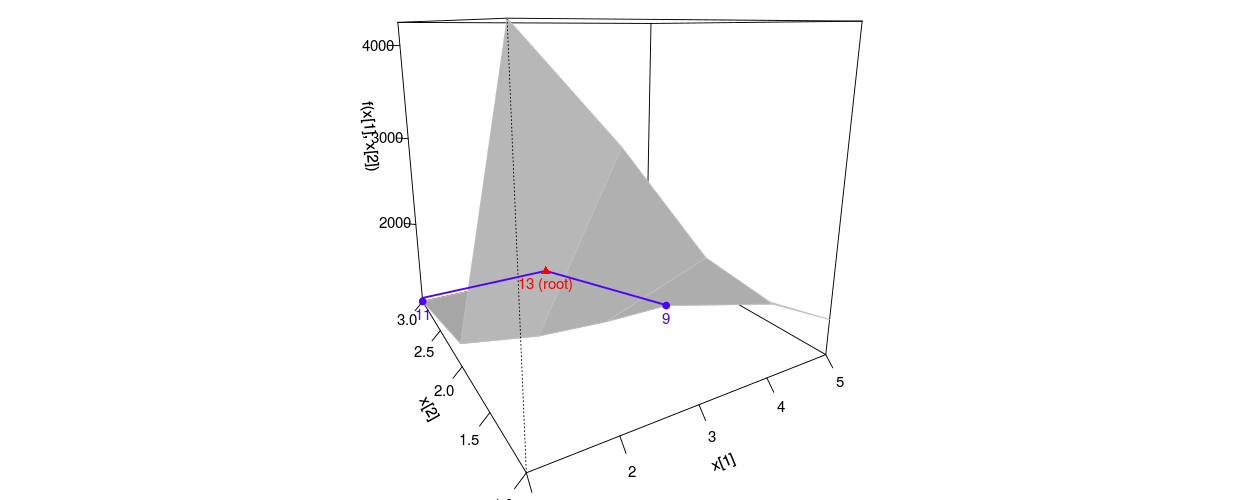}
        & \includegraphics[width=3.5cm]{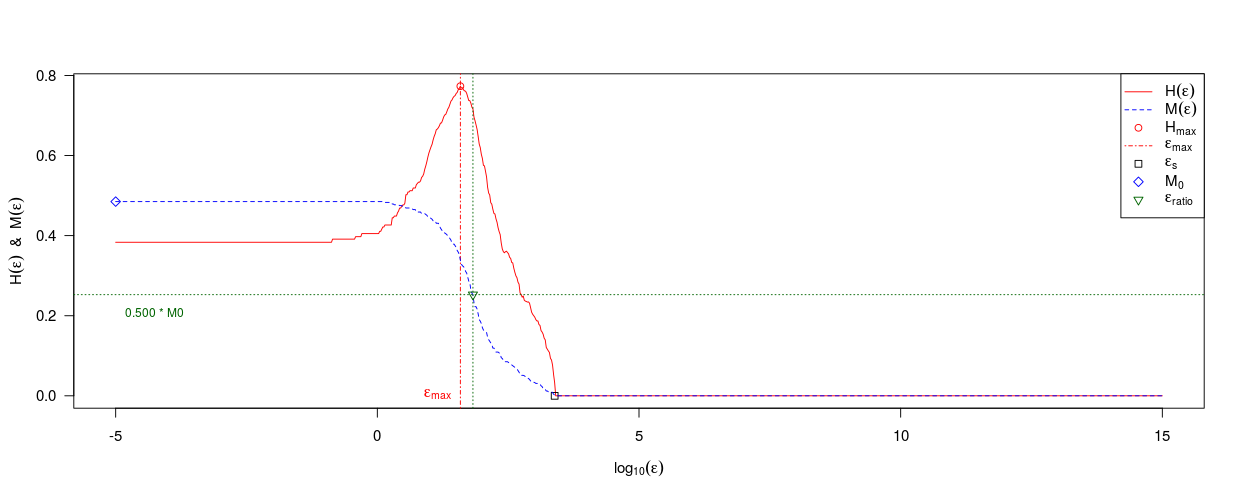}
        \\
 \hline

          \multirow{1}{*}{$f_{16}$}  & \includegraphics[width=3.5cm]{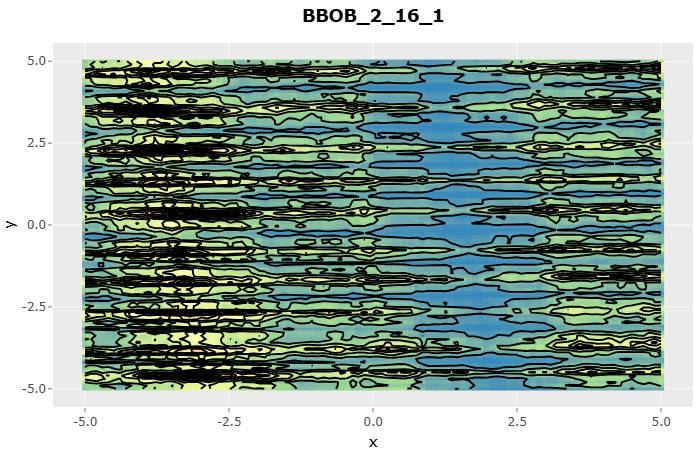}
        & \includegraphics[width=3.5cm]{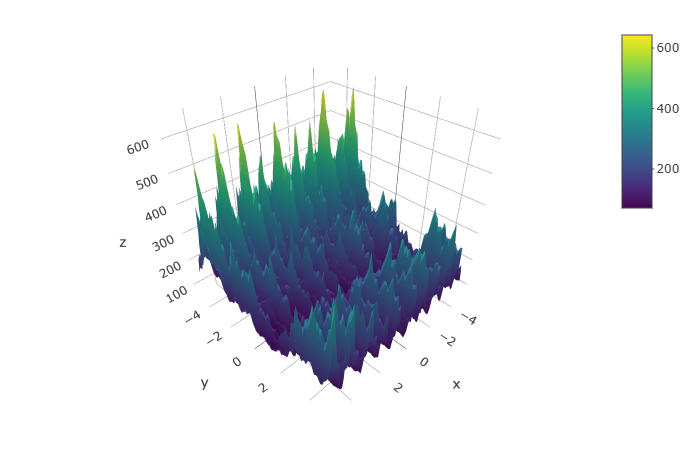} 
        & \includegraphics[width=3.5cm]{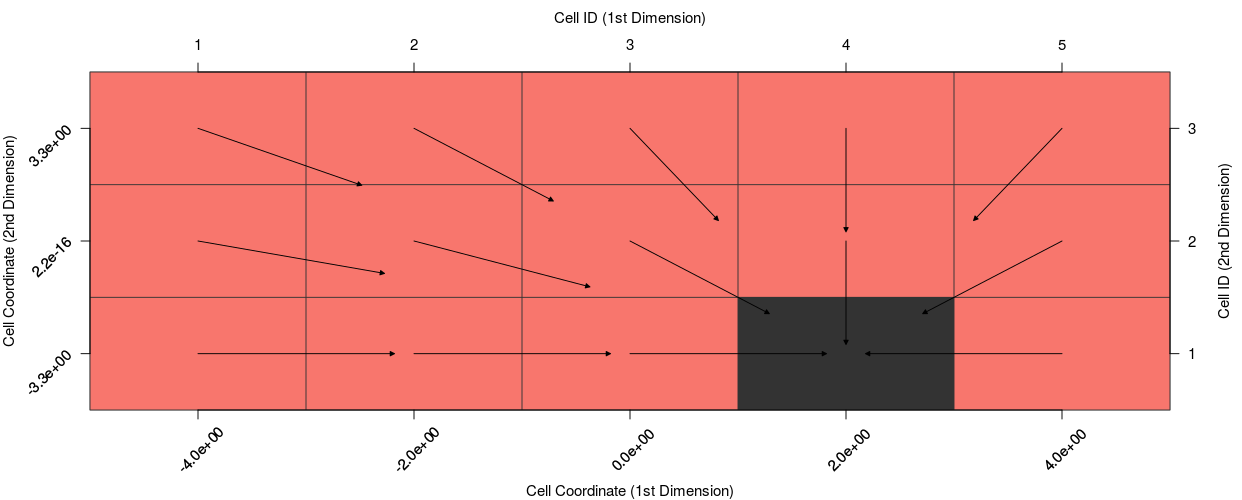 }
        & \includegraphics[width=3.5cm]{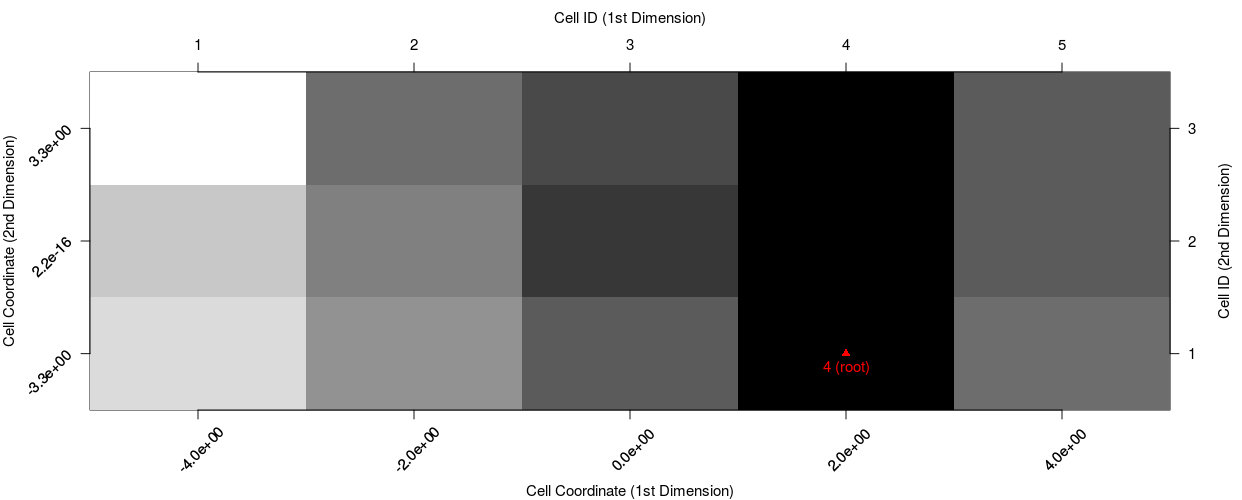 }
        & \includegraphics[width=4.5cm]{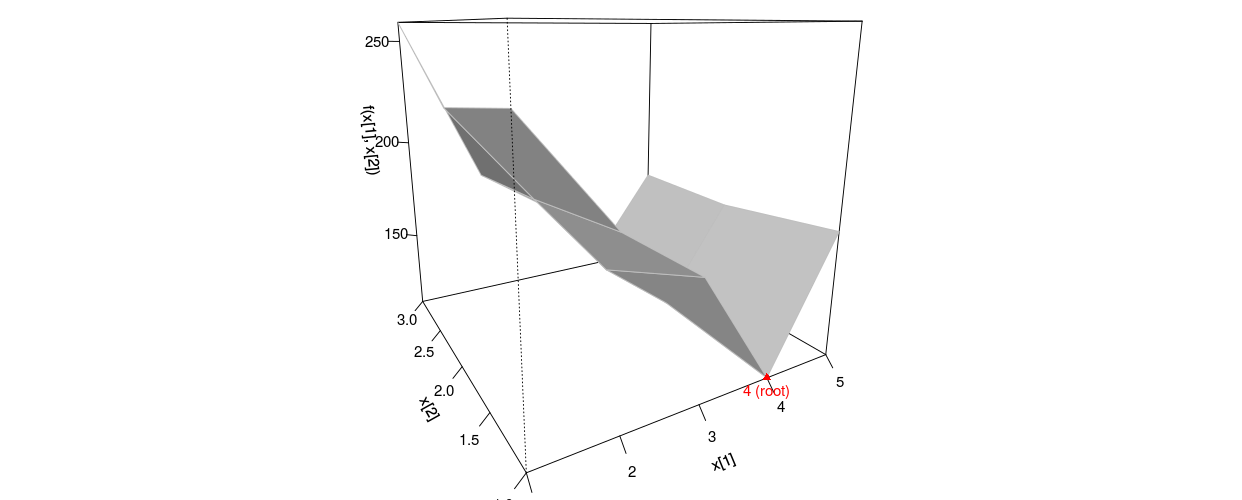 }
        & \includegraphics[width=3.5cm]{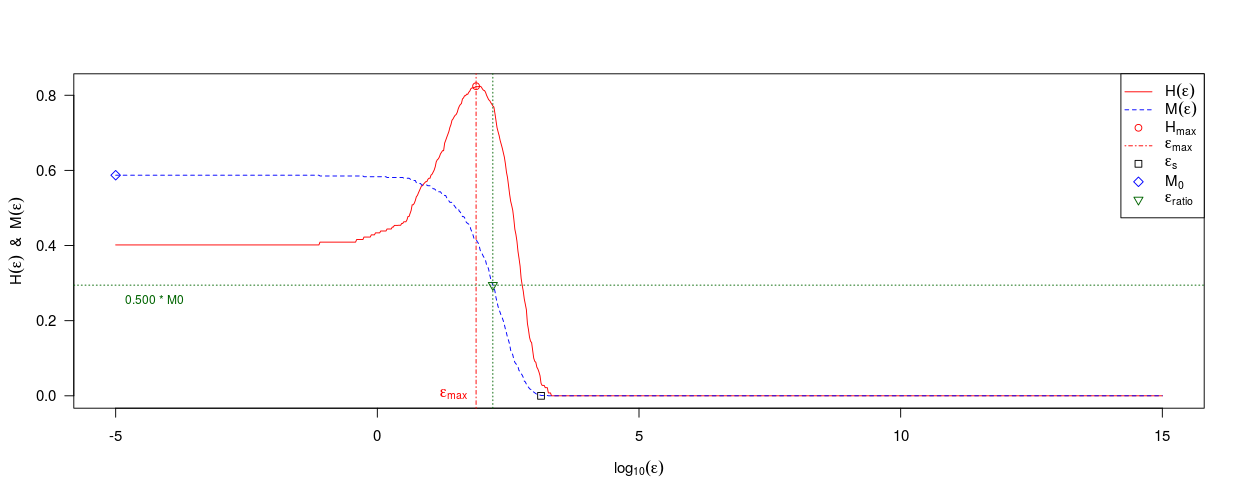 }
        \\
 \hline

 \end{tabular}
\end{table*}
\end{landscape}

\begin{landscape}
\begin{table*}[htp]
\centering
 \caption{Analysis of ELA features BBOB functions on d=2.}
\label{Tab ELA-3}
    \centering
   \scriptsize \begin{tabular}{c c c c c c c}
    \hline
      \textbf{Function}  & \textbf{Contour Plot} &  \textbf{Surface Plot}  &  \textbf{Cell- Mapping} &  \textbf{Barrier Tree-2D} &  \textbf{Barrier Tree-3D}  &  \textbf{Information Content}\\ \hline

        \multirow{1}{*}{$f_{17}$}  & \includegraphics[width=3.5cm]{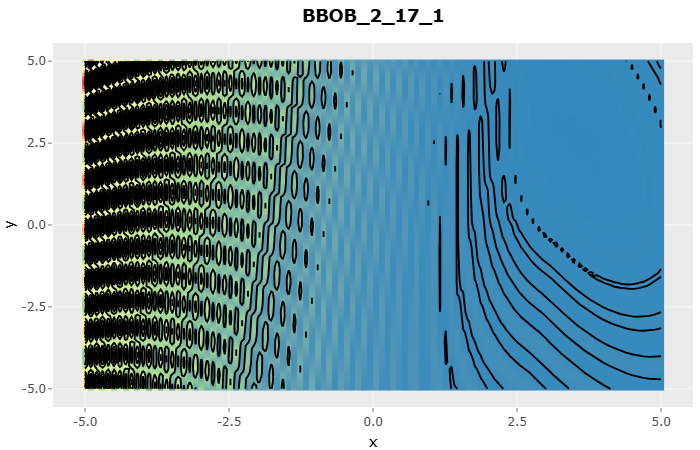}
        & \includegraphics[width=3.5cm]{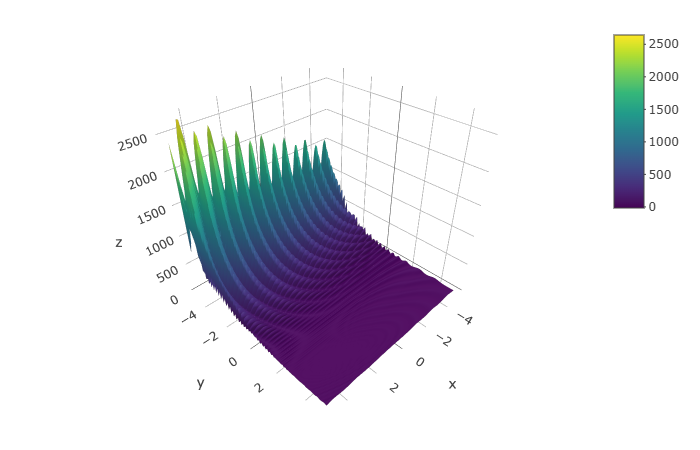} 
        & \includegraphics[width=3.5cm]{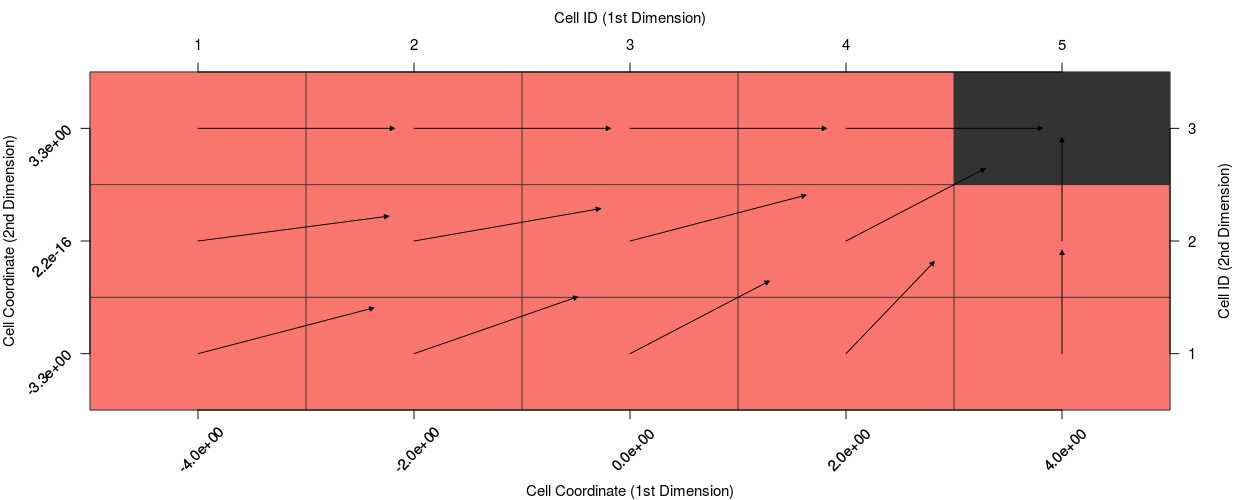  }
        & \includegraphics[width=3.5cm]{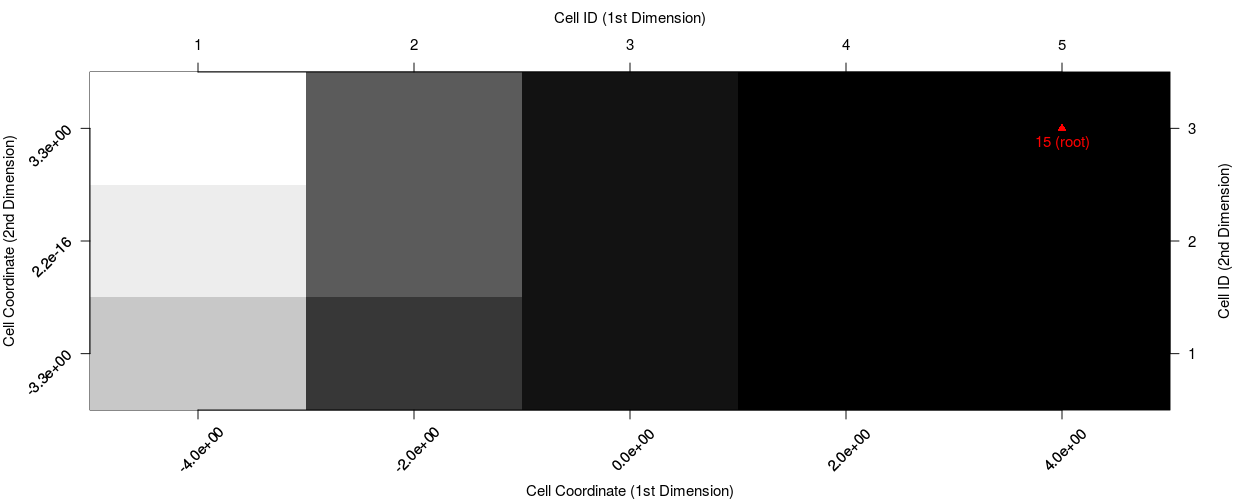  }
        & \includegraphics[width=4.5cm]{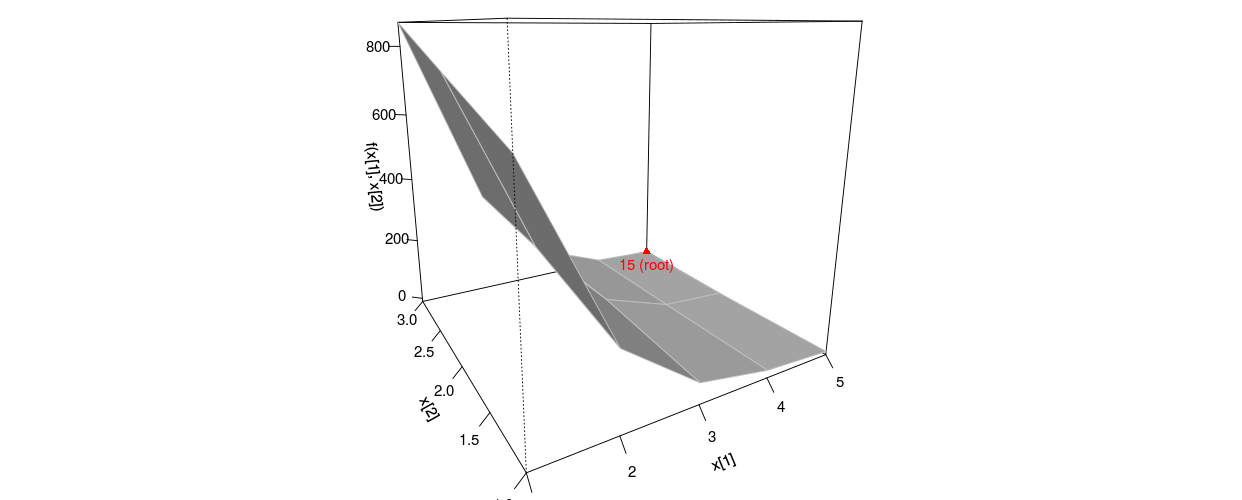  }
        & \includegraphics[width=3.5cm]{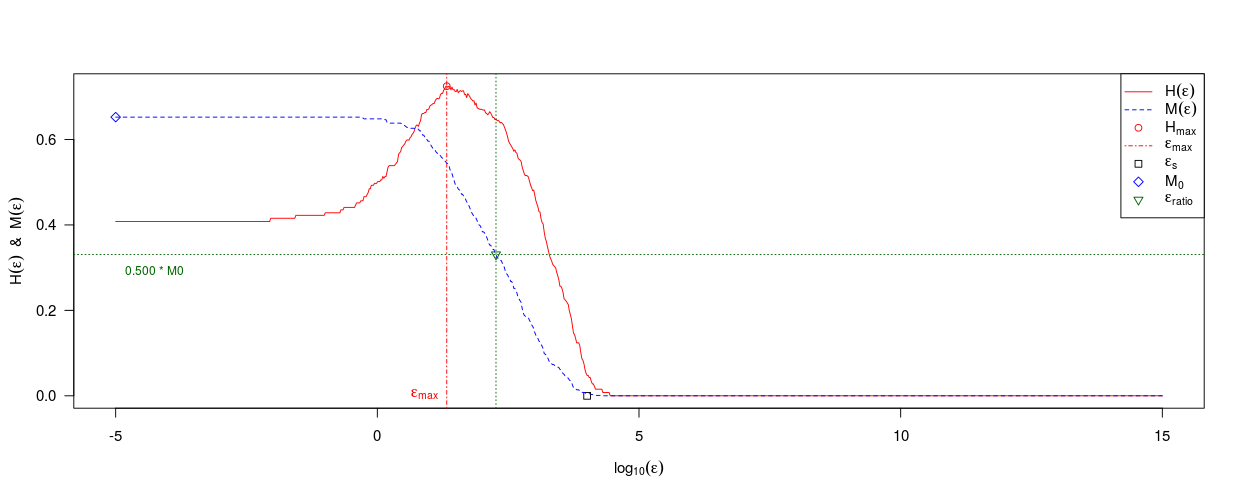 }
        \\
 \hline
 
        \multirow{1}{*}{$f_{18}$}  & \includegraphics[width=3.5cm]{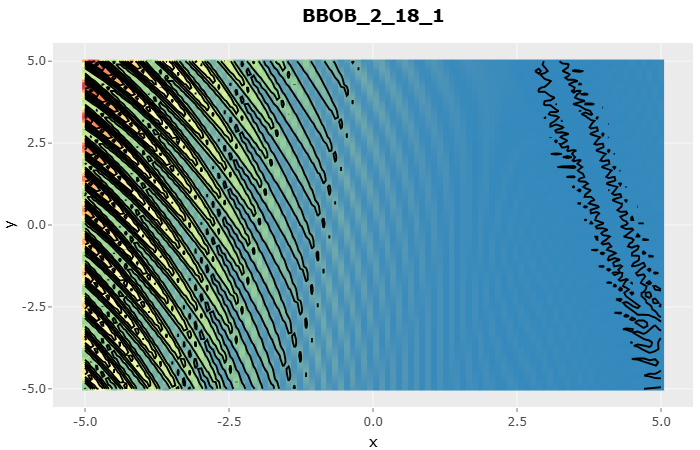}
        & \includegraphics[width=3.5cm]{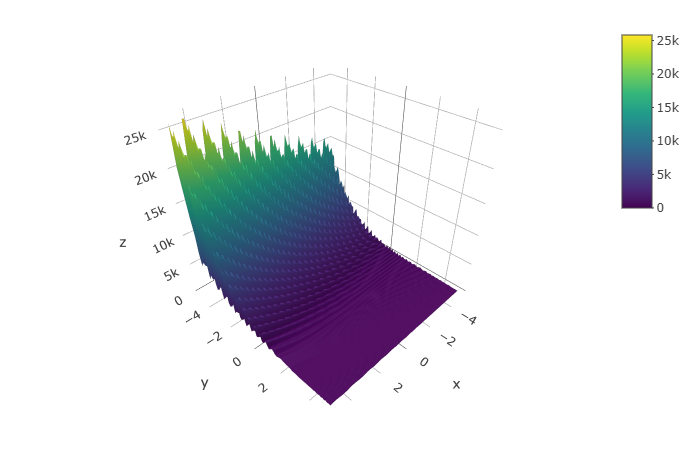} 
        & \includegraphics[width=3.5cm]{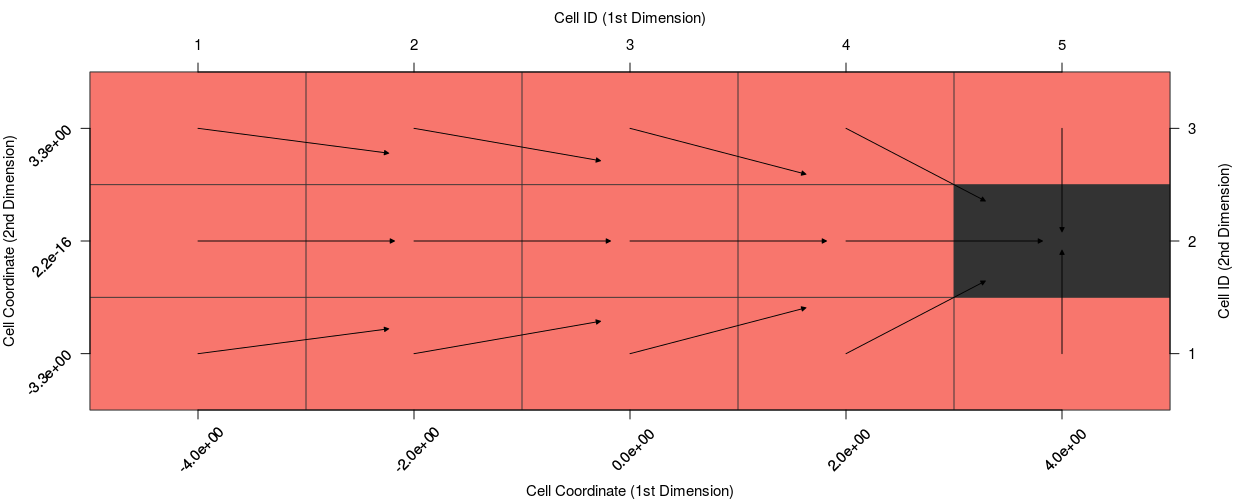}
        & \includegraphics[width=3.5cm]{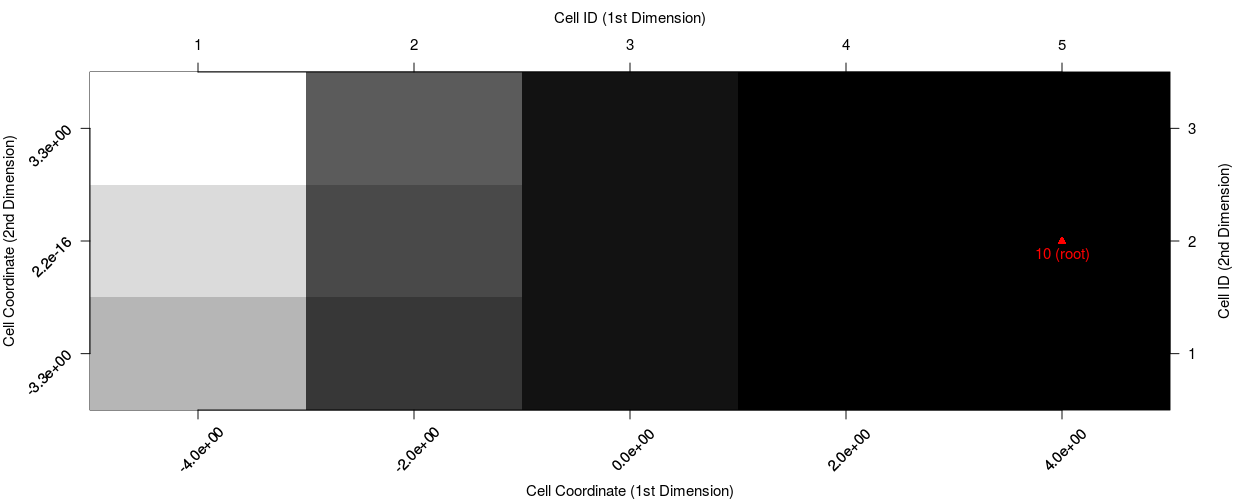}
        & \includegraphics[width=4.5cm]{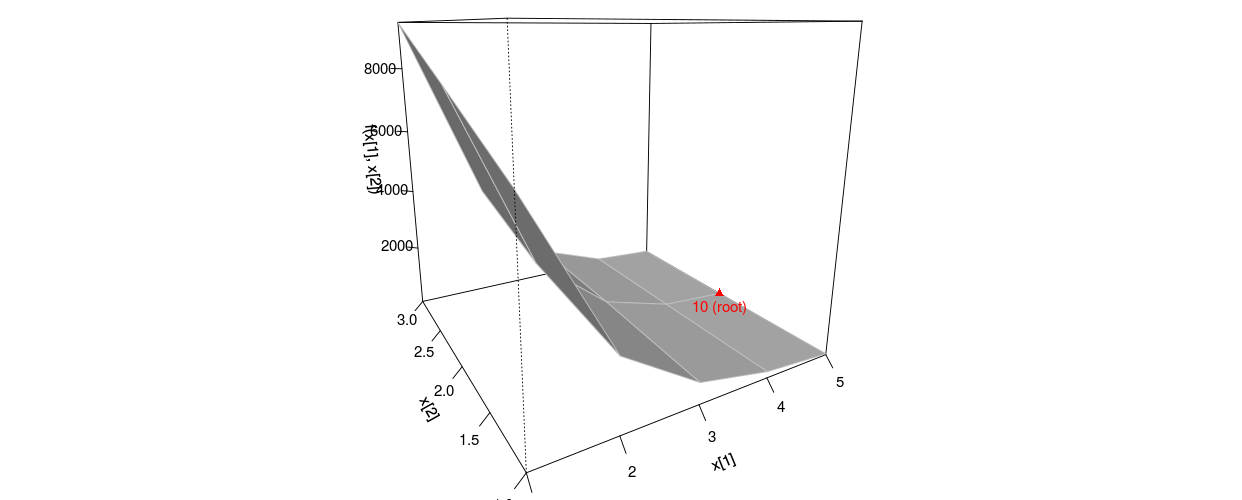 }
        & \includegraphics[width=3.5cm]{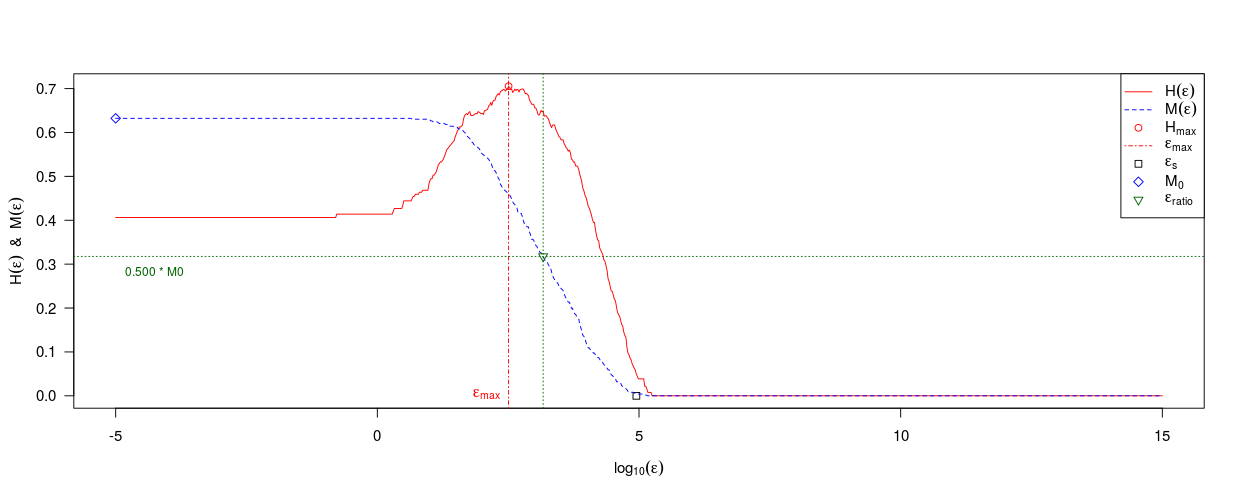  }
        \\
 \hline

          \multirow{1}{*}{$f_{19}$}  & \includegraphics[width=3.5cm]{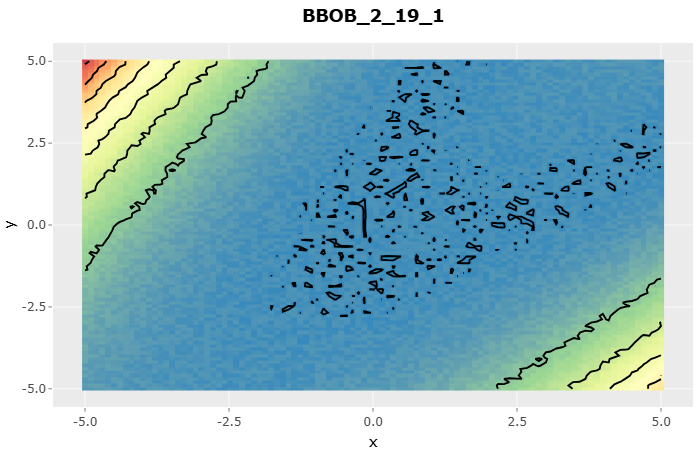}
        & \includegraphics[width=3.5cm]{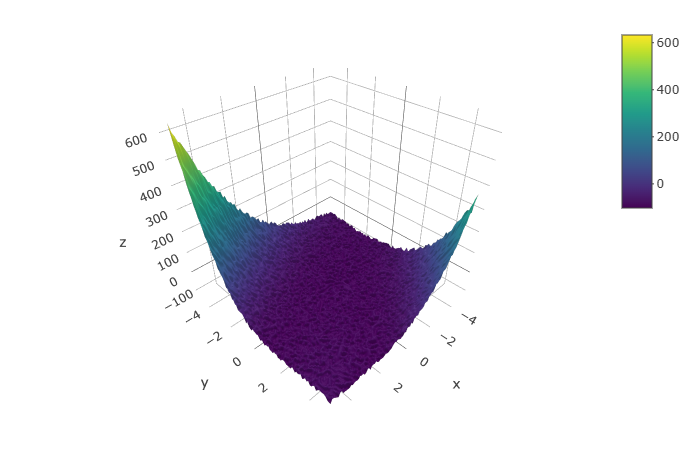} 
        & \includegraphics[width=3.5cm]{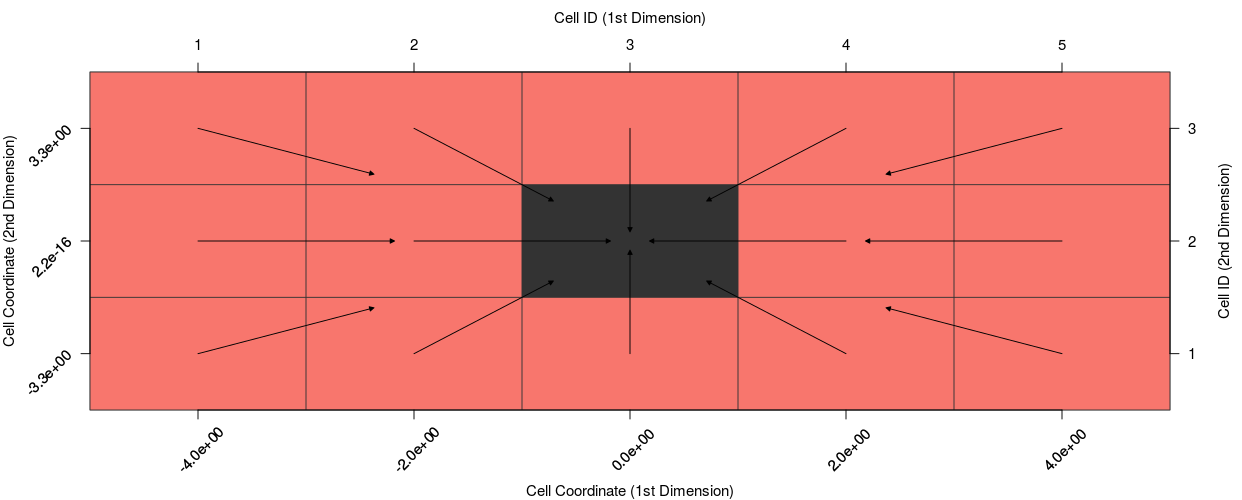}
        & \includegraphics[width=3.5cm]{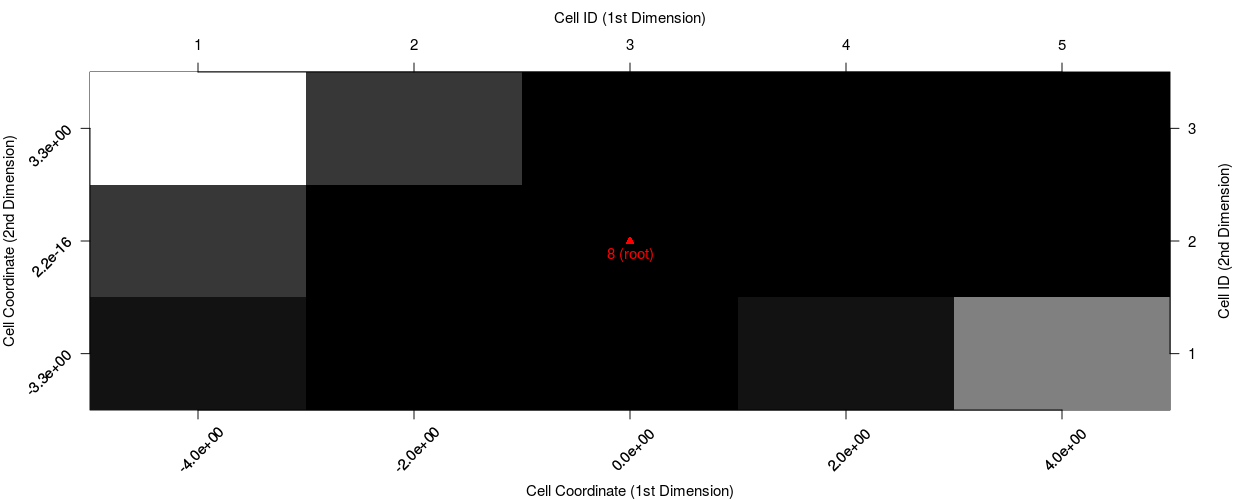}
        & \includegraphics[width=4.5cm]{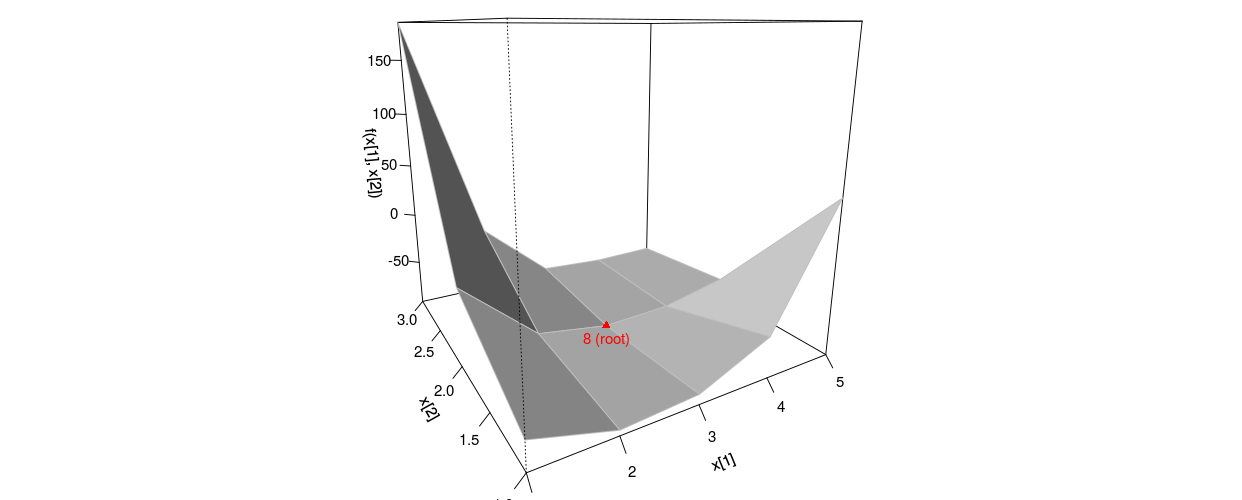}
        & \includegraphics[width=3.5cm]{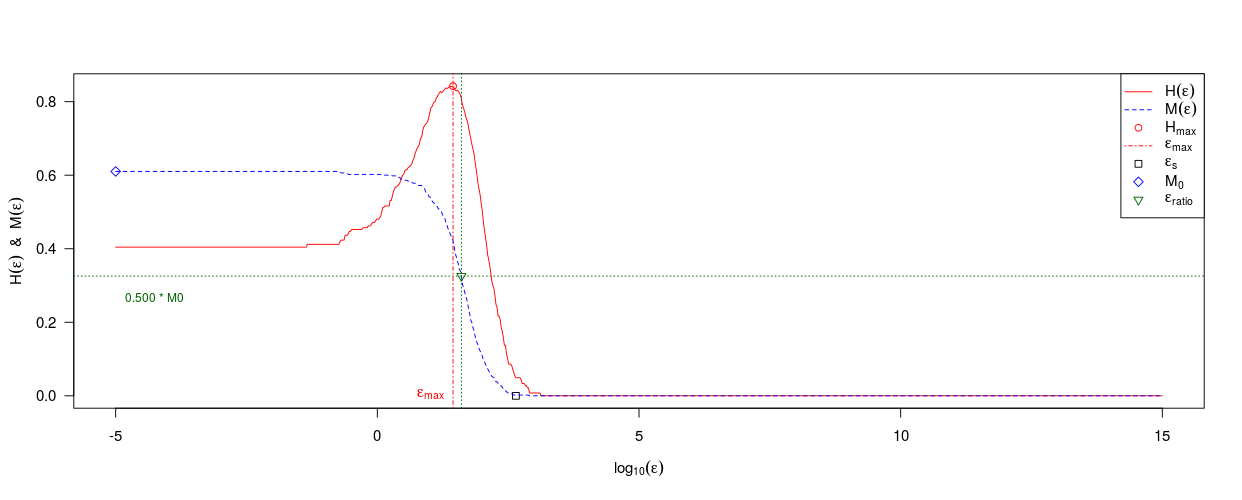}
        \\
 \hline
         \multirow{1}{*}{$f_{20}$}  & \includegraphics[width=3.5cm]{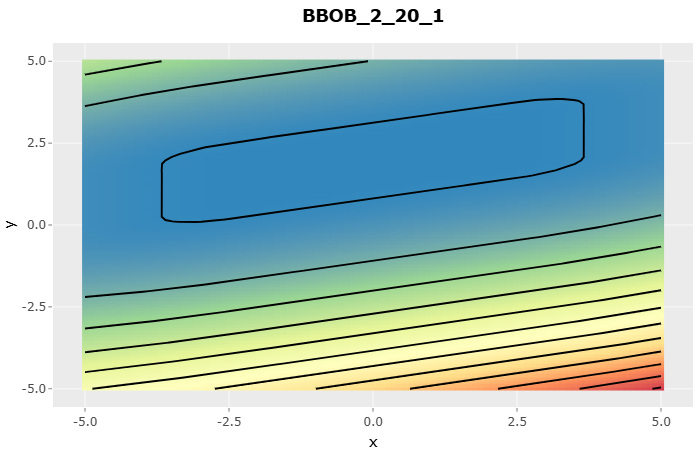}
        & \includegraphics[width=3.5cm]{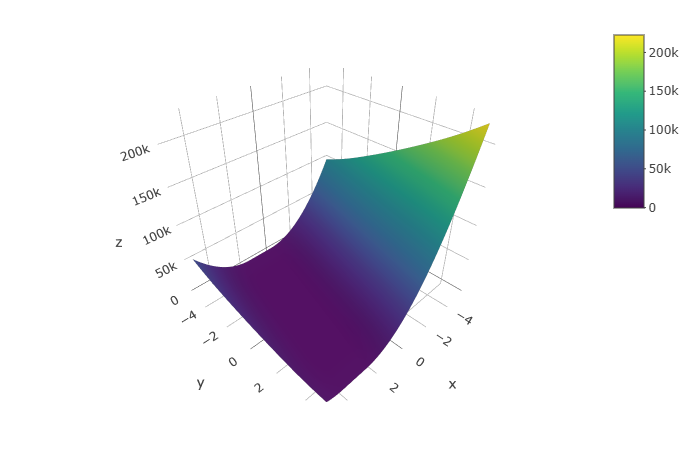} 
        & \includegraphics[width=3.5cm]{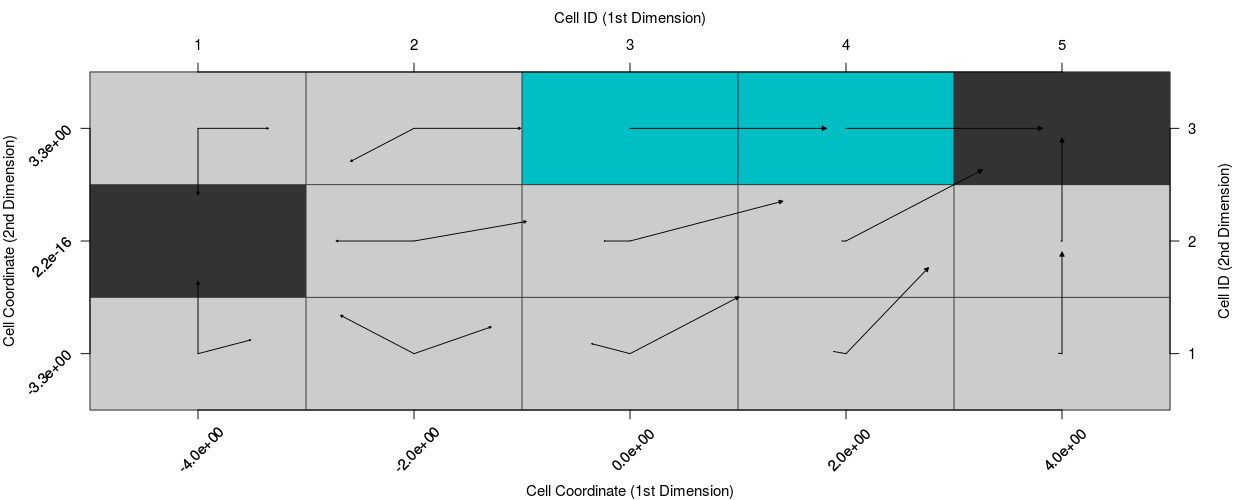 }
        & \includegraphics[width=3.5cm]{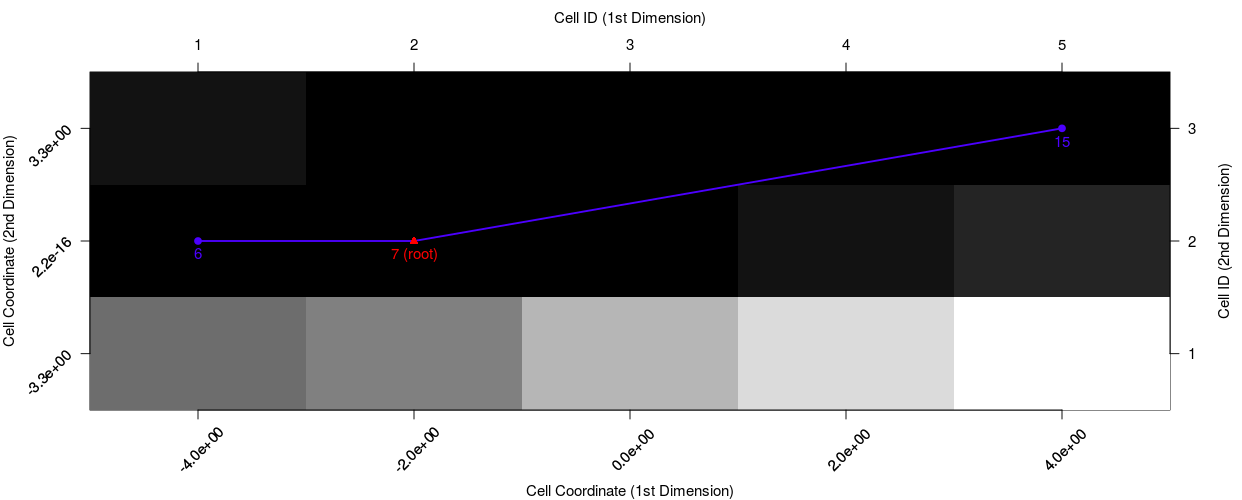 }
        & \includegraphics[width=4.5cm]{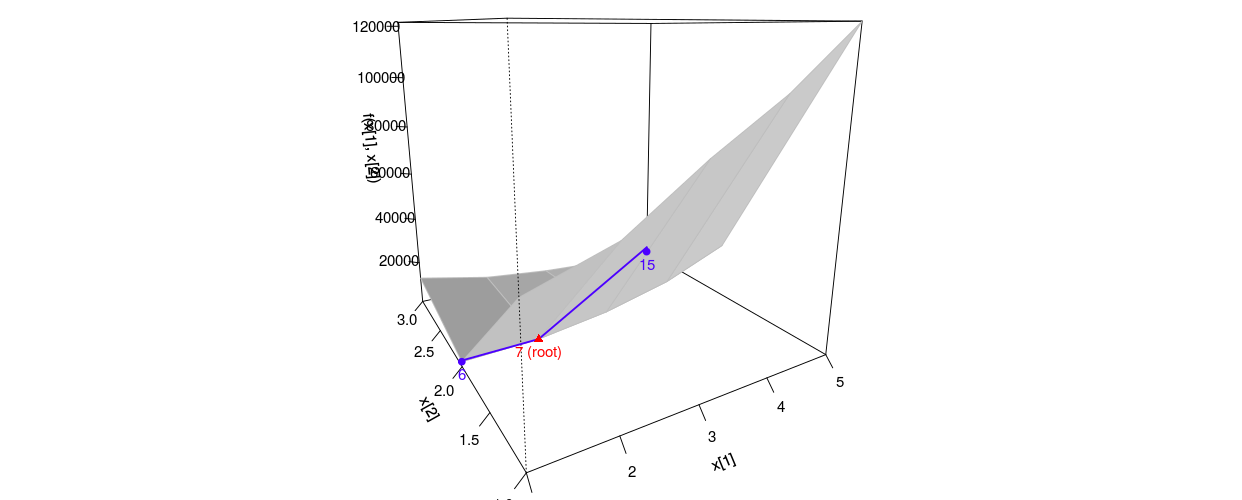  }
        & \includegraphics[width=3.5cm]{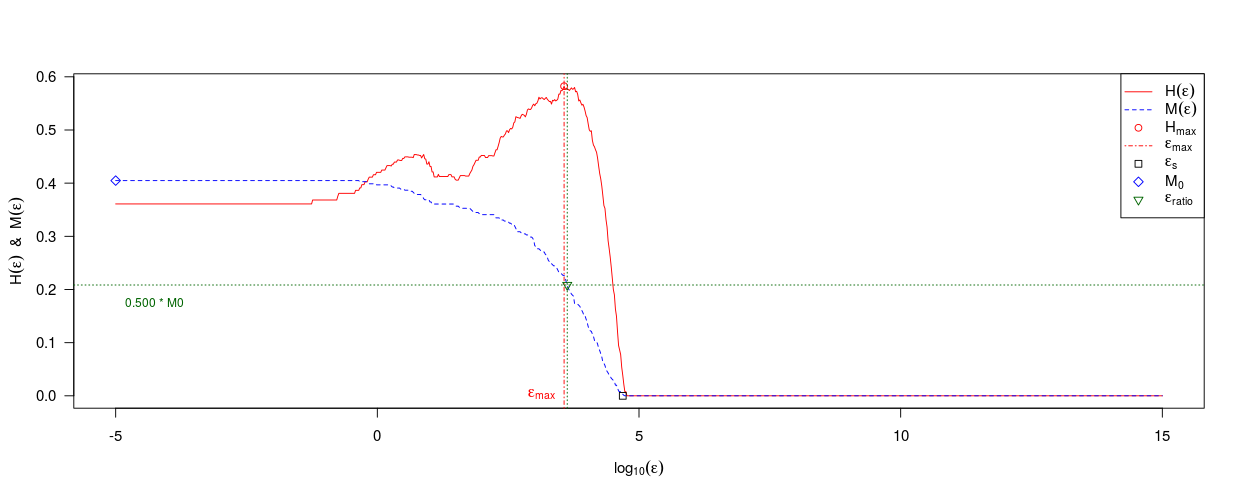 }
        \\
 \hline
         \multirow{1}{*}{$f_{21}$}  & \includegraphics[width=3.5cm]{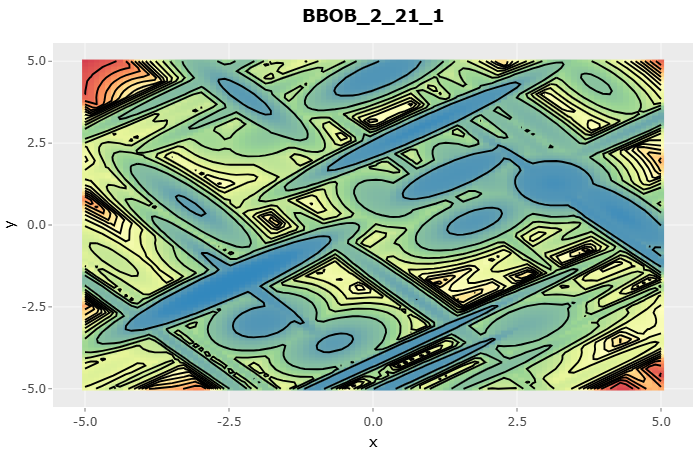}
        & \includegraphics[width=3.5cm]{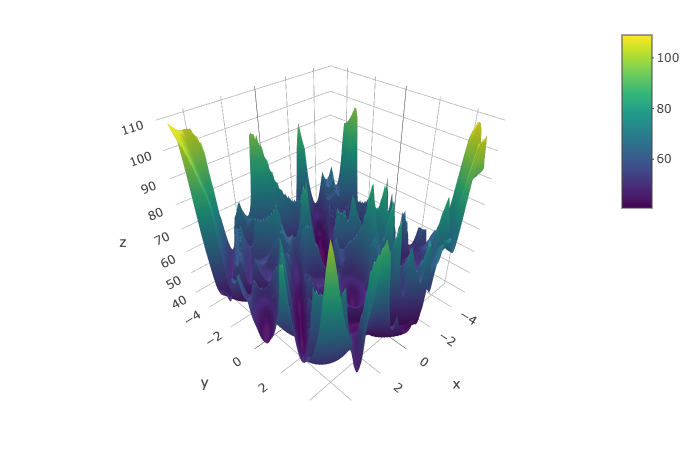} 
        & \includegraphics[width=3.5cm]{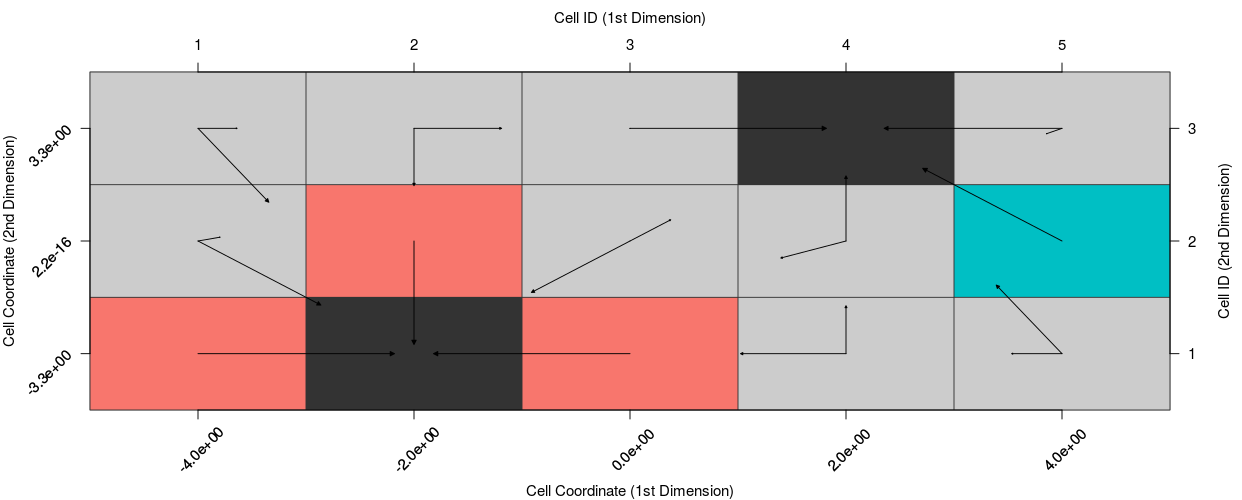    }
        & \includegraphics[width=3.5cm]{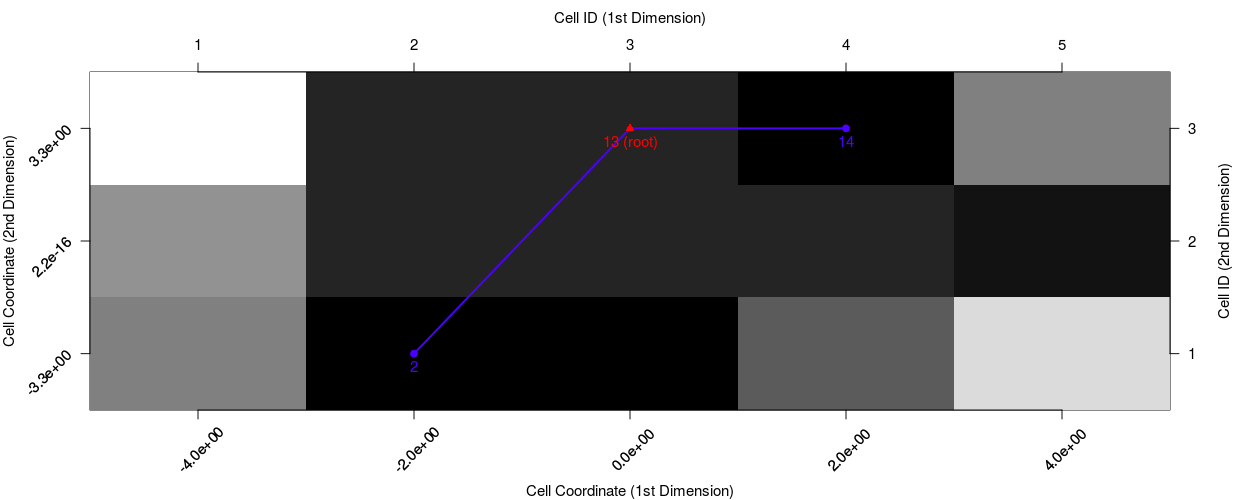 }
        & \includegraphics[width=4.5cm]{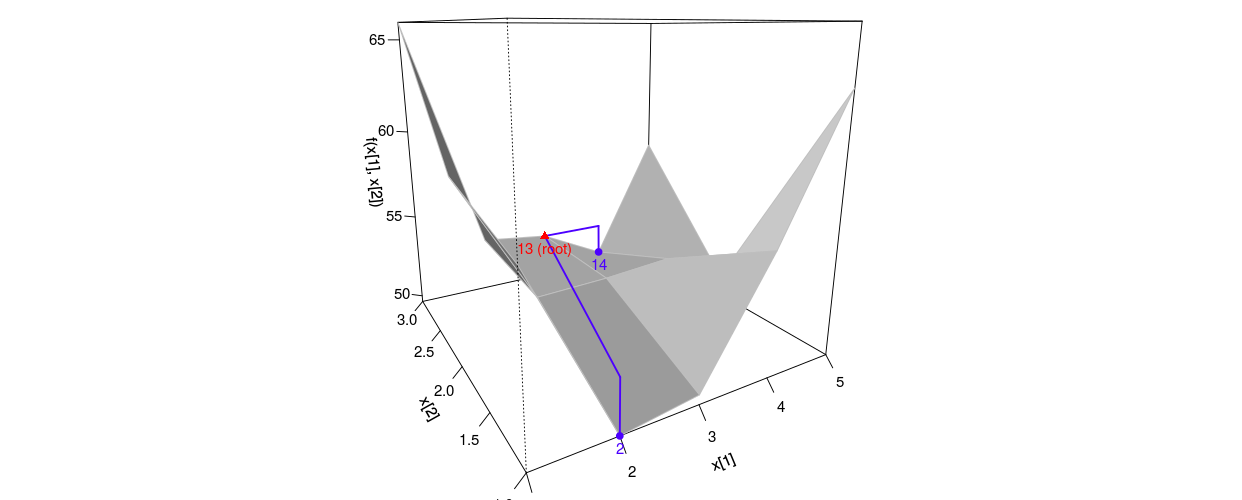   }
        & \includegraphics[width=3.5cm]{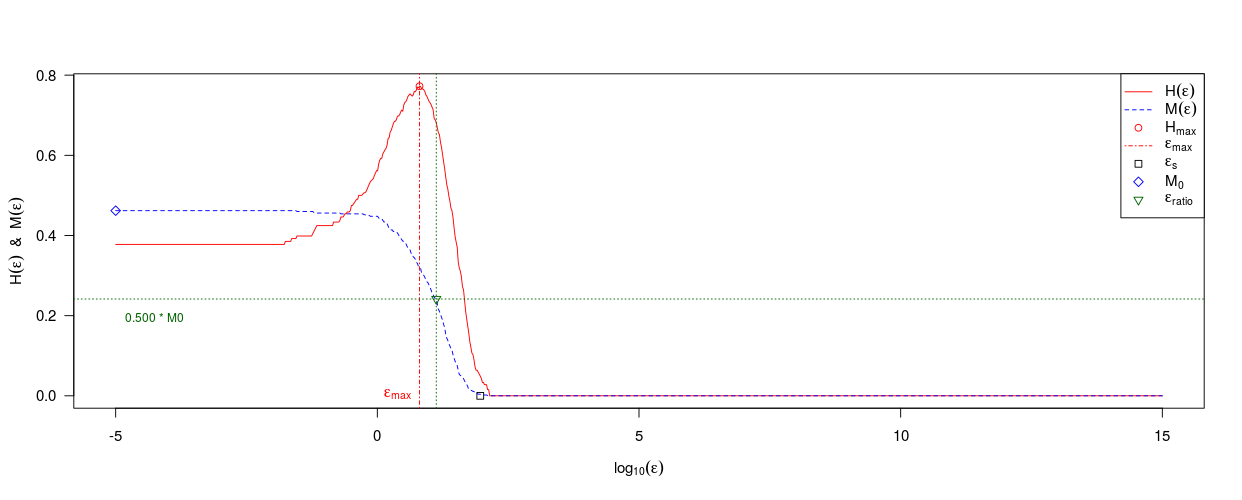  }
        \\
 \hline

        \multirow{1}{*}{$f_{22}$}  & \includegraphics[width=3.5cm]{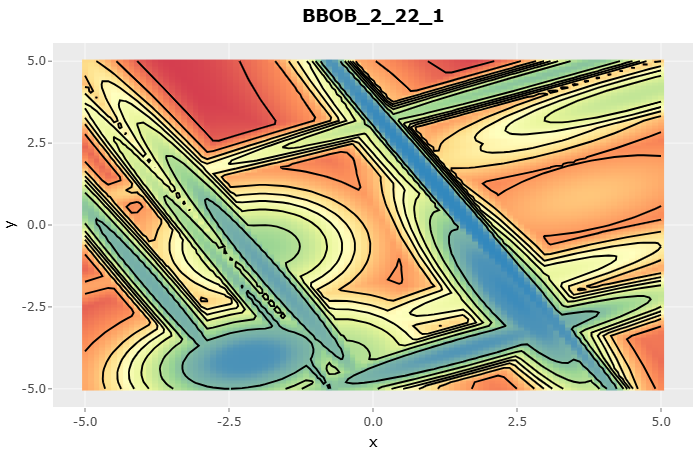}
        & \includegraphics[width=3.5cm]{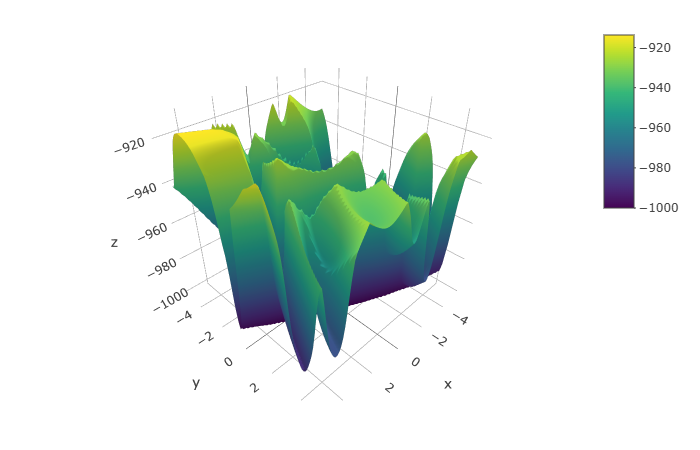} 
        & \includegraphics[width=3.5cm]{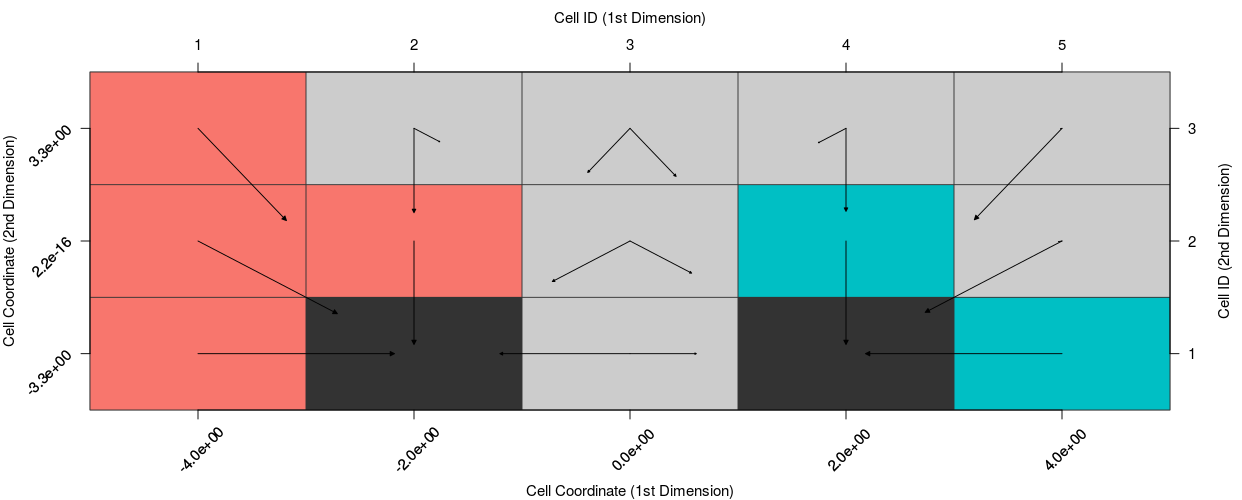}
        & \includegraphics[width=3.5cm]{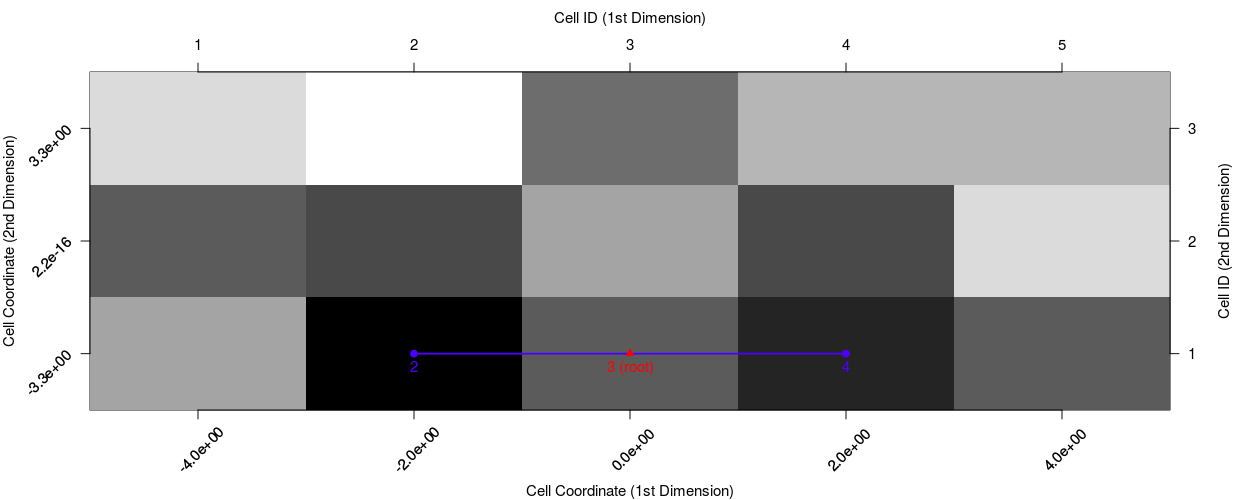}
        & \includegraphics[width=4.5cm]{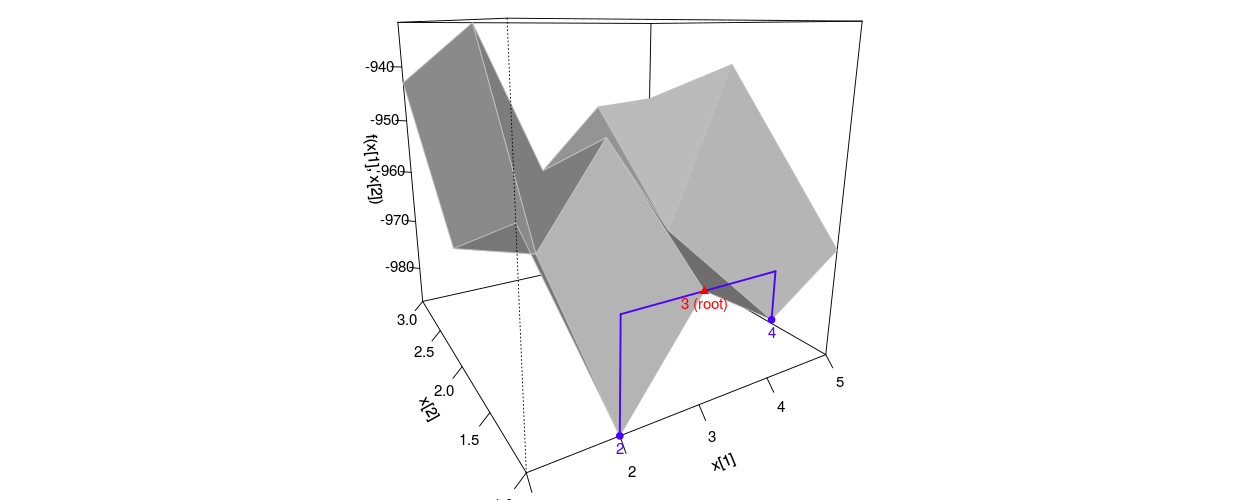}
        & \includegraphics[width=3.5cm]{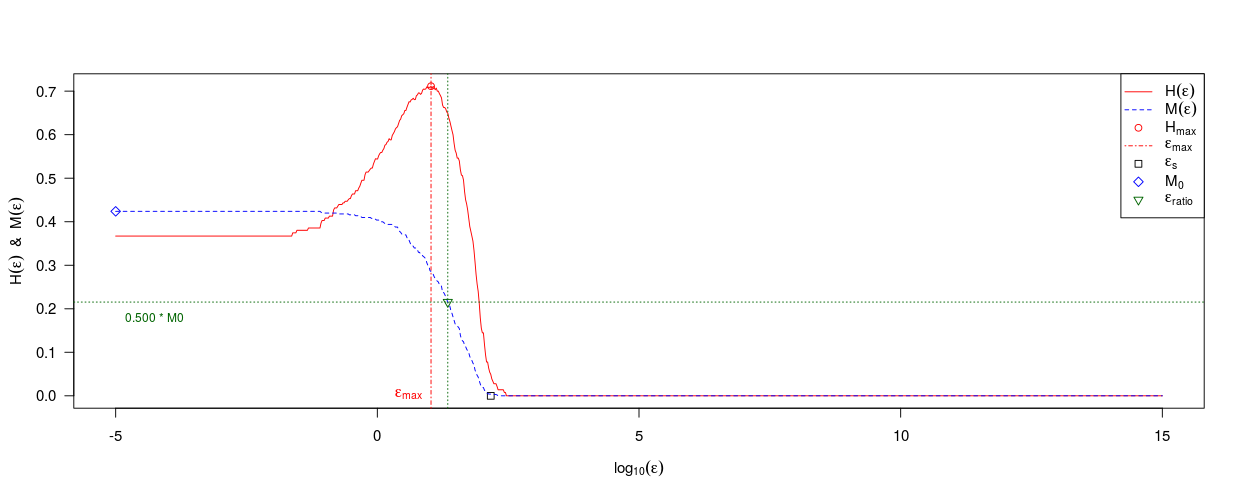}
        \\
 \hline

          \multirow{1}{*}{$f_{23}$}  & \includegraphics[width=3.5cm]{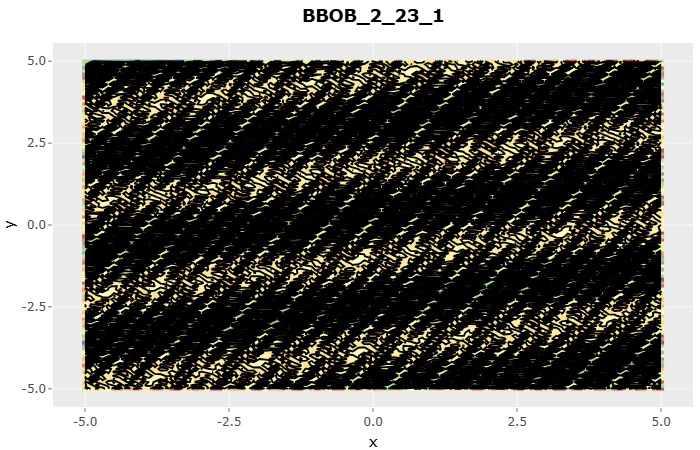}
        & \includegraphics[width=3.5cm]{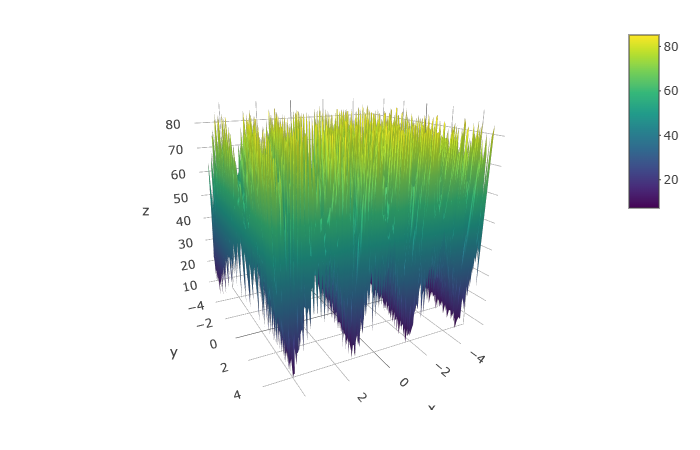} 
        & \includegraphics[width=3.5cm]{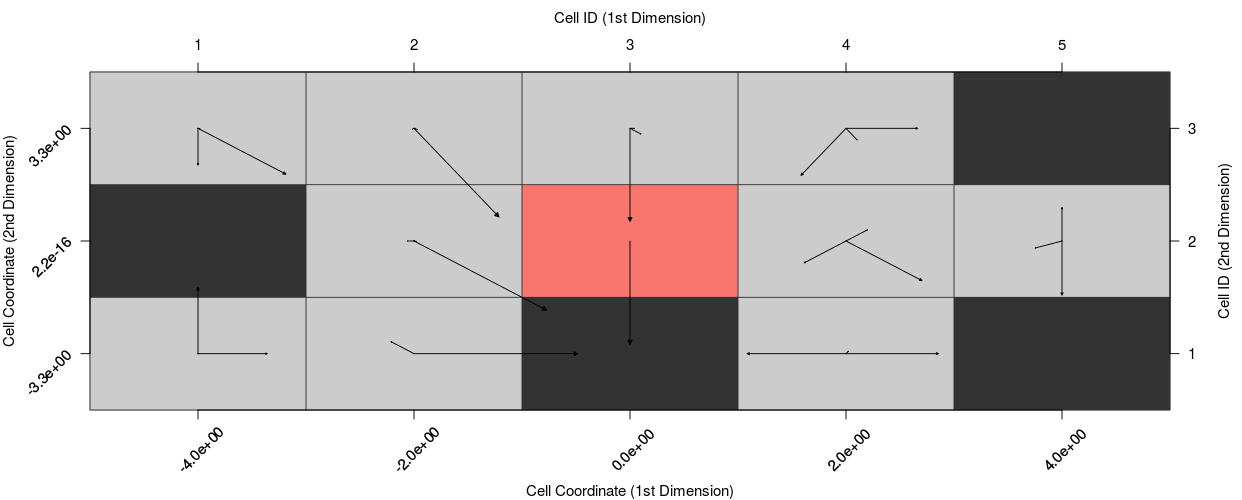}
        & \includegraphics[width=3.5cm]{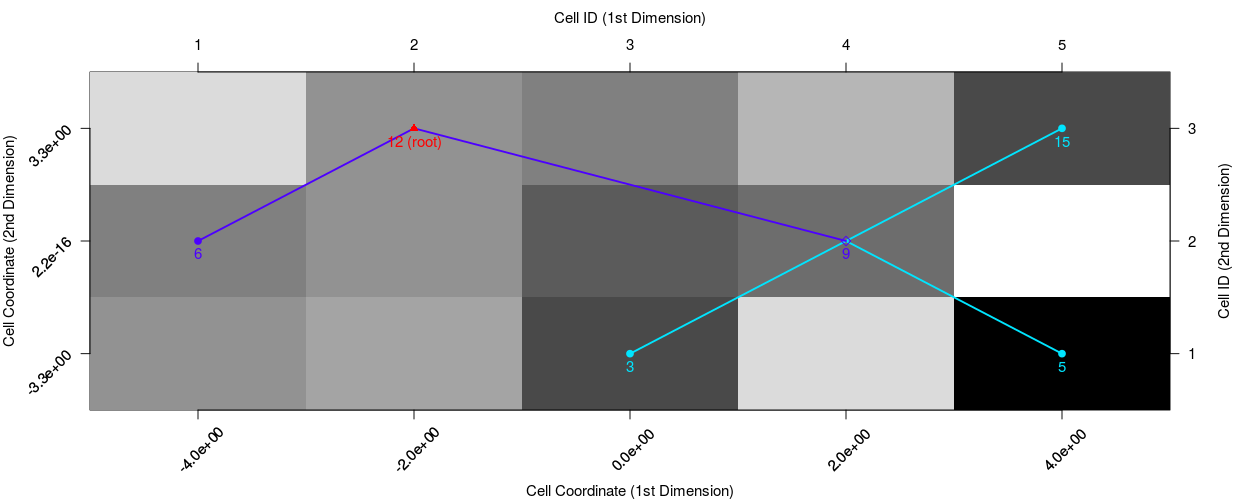}
        & \includegraphics[width=4.5cm]{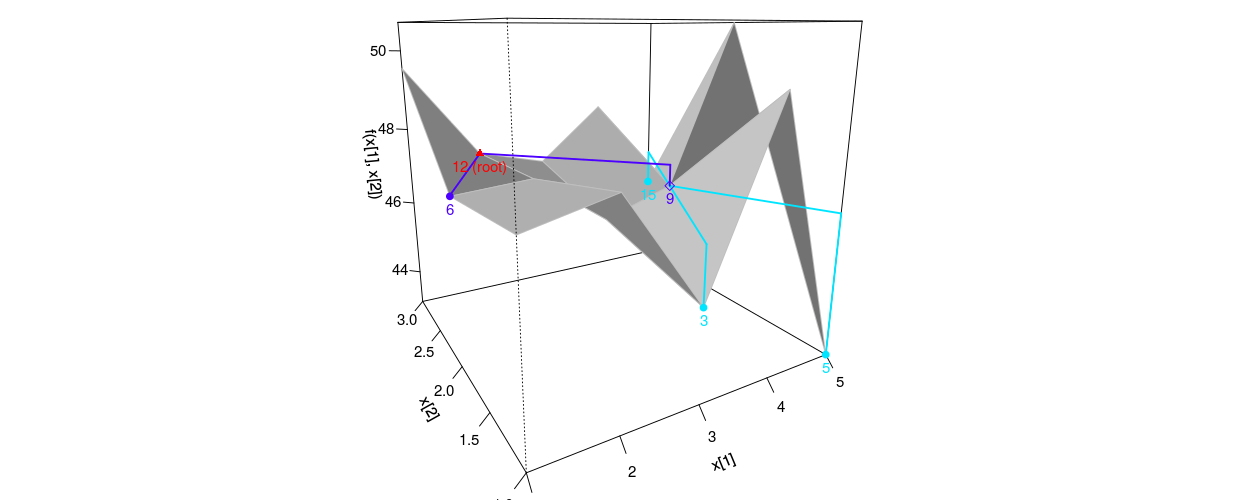}
        & \includegraphics[width=3.5cm]{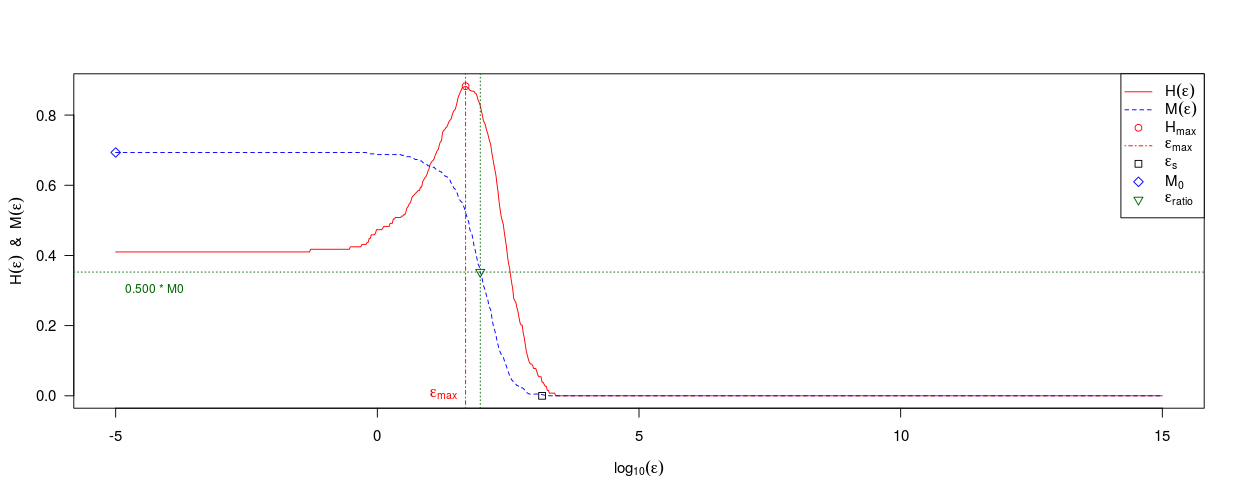}
        \\
 \hline
        \multirow{1}{*}{$f_{24}$}  & \includegraphics[width=3.5cm]{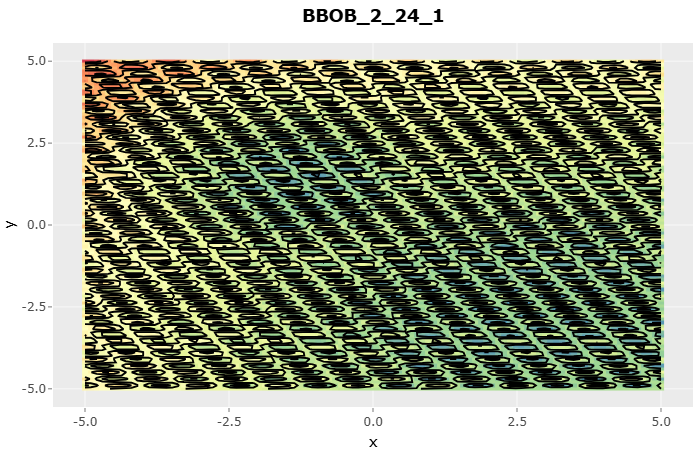}
        & \includegraphics[width=3.5cm]{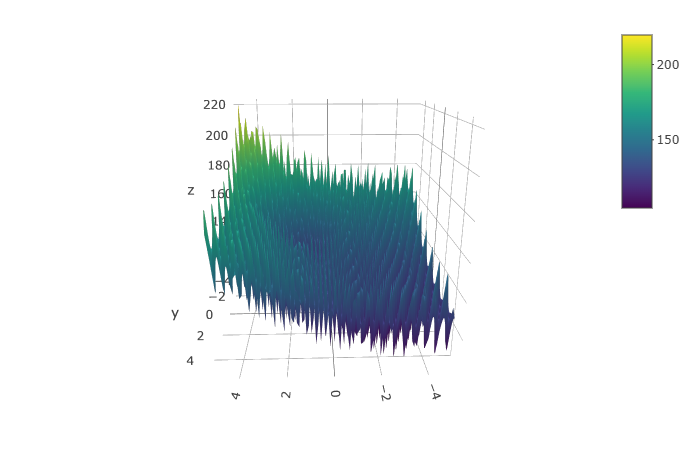} 
        & \includegraphics[width=3.5cm]{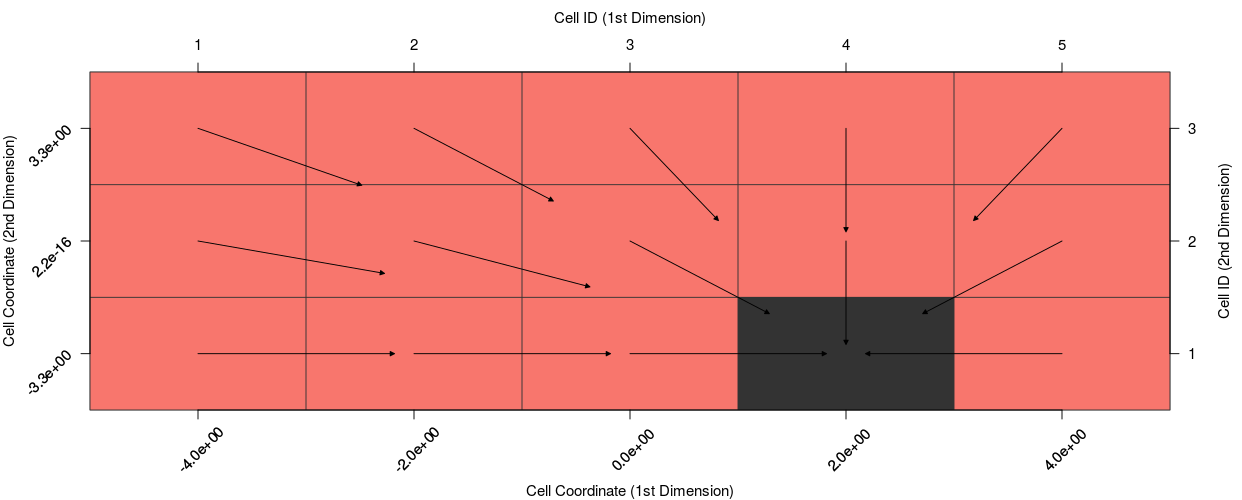}
        & \includegraphics[width=3.5cm]{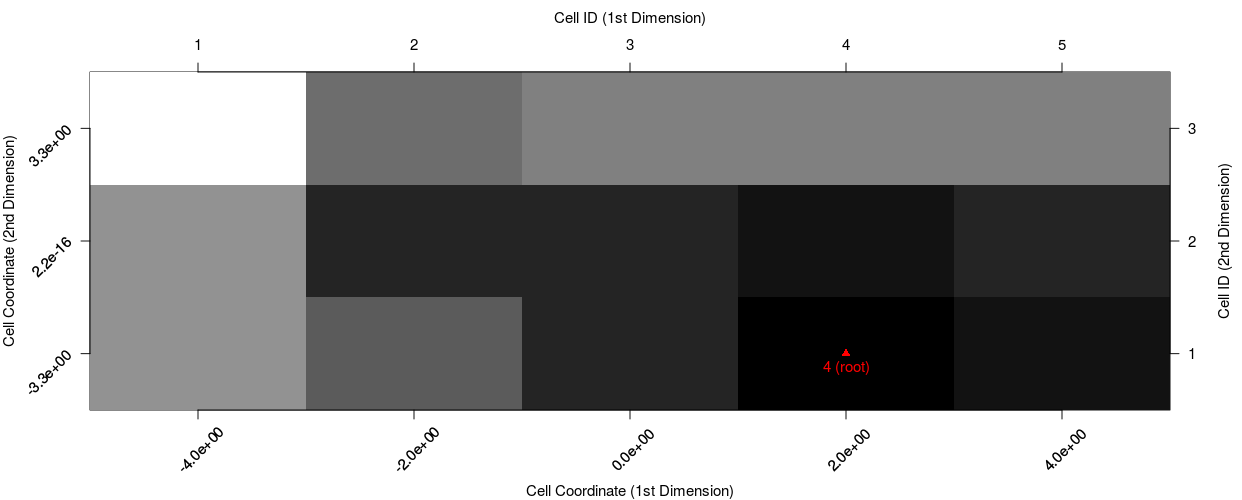}
        & \includegraphics[width=4.5cm]{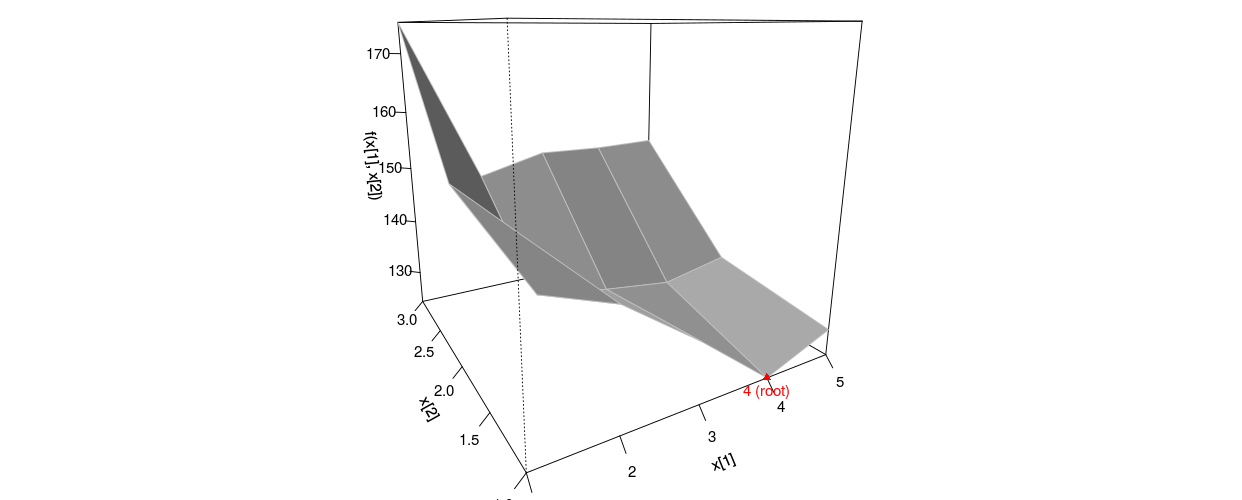}
        & \includegraphics[width=3.5cm]{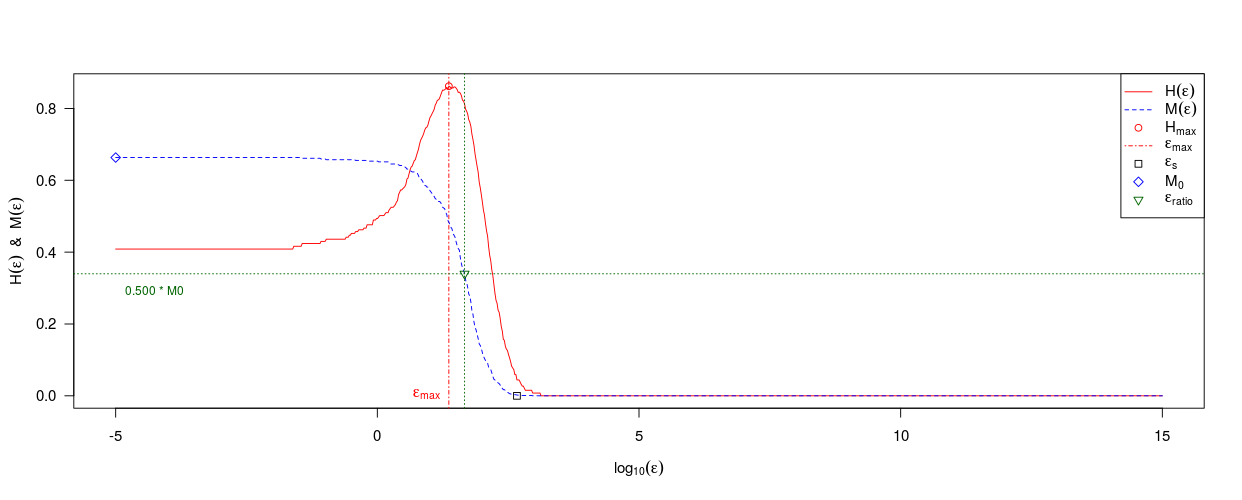}
        \\
 \hline

 \end{tabular}
\end{table*}
\end{landscape}

\begin{table*}[h!]
\centering
 \caption{Information Content visualization of all BBOB functions using iid = 1 on d=5.}
\label{Tab ELA-4}
    \centering
   \scriptsize \begin{tabular}{c c c }

    \multirow{2}{*}{}  \includegraphics[,width=6cm]{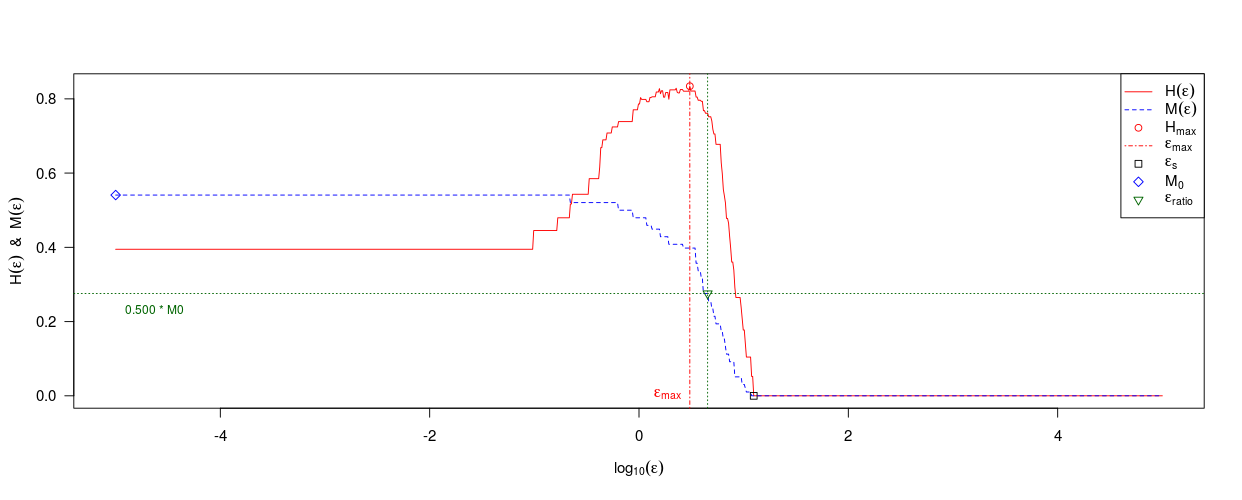}
     & \includegraphics[width=6cm]{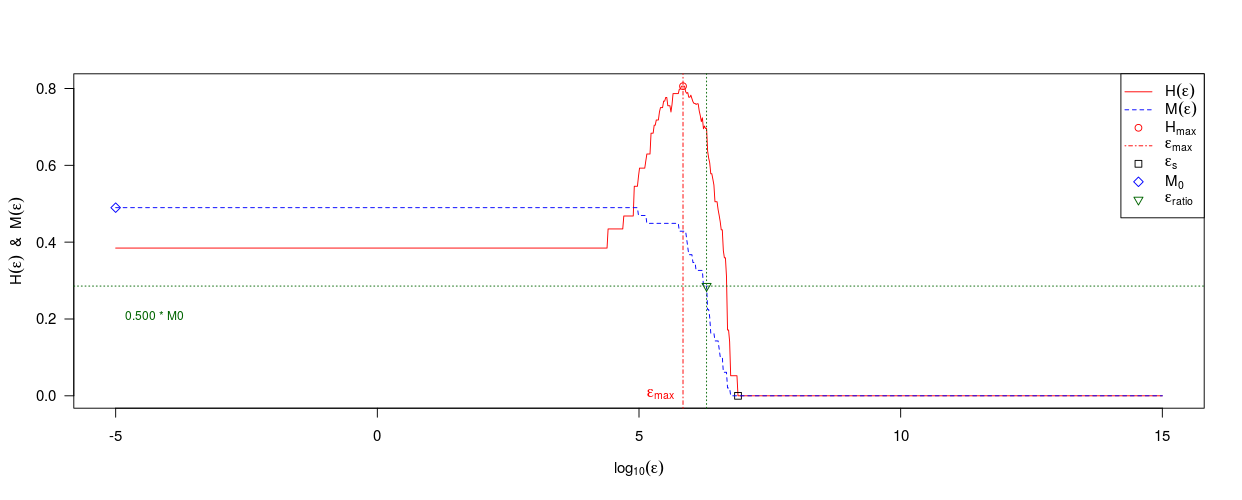} 
     & \includegraphics[width=6cm]{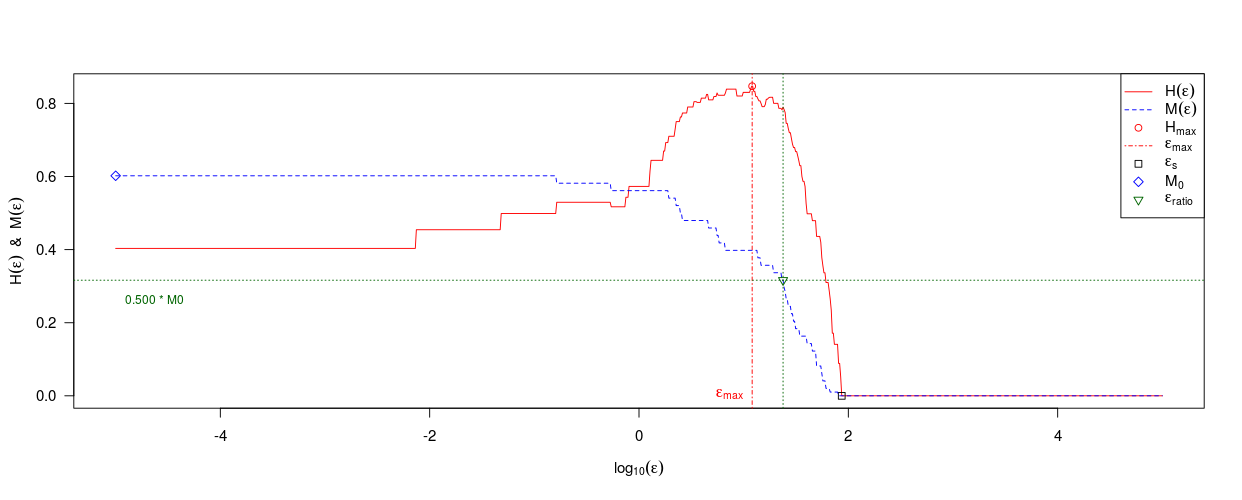}
    \\
    {$f_1$} & {$f_2$} & {$f_3$}
    \\

    \multirow{2}{*}{}  \includegraphics[,width=6cm]{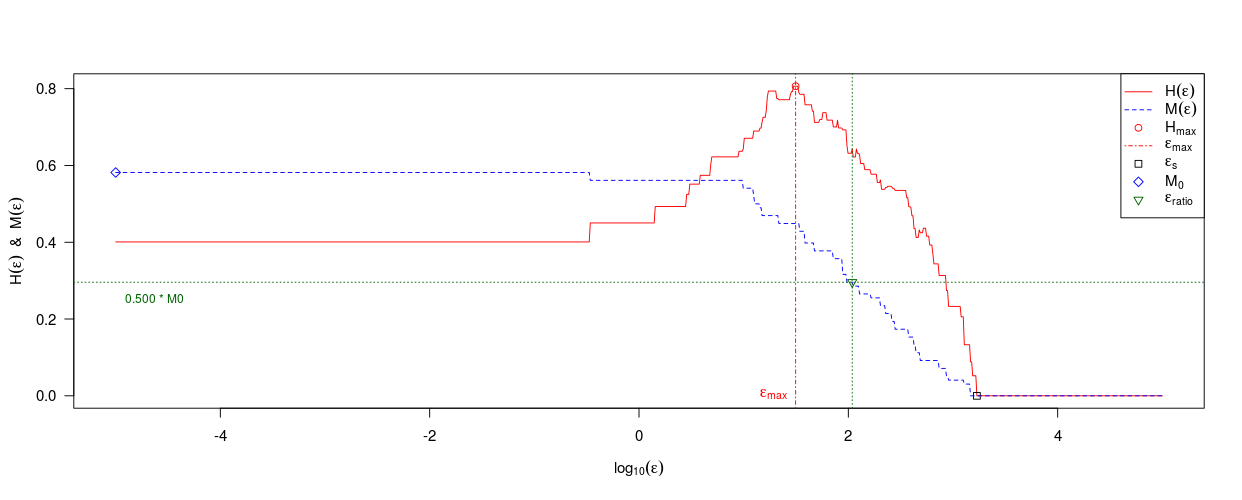}
     & \includegraphics[width=6cm]{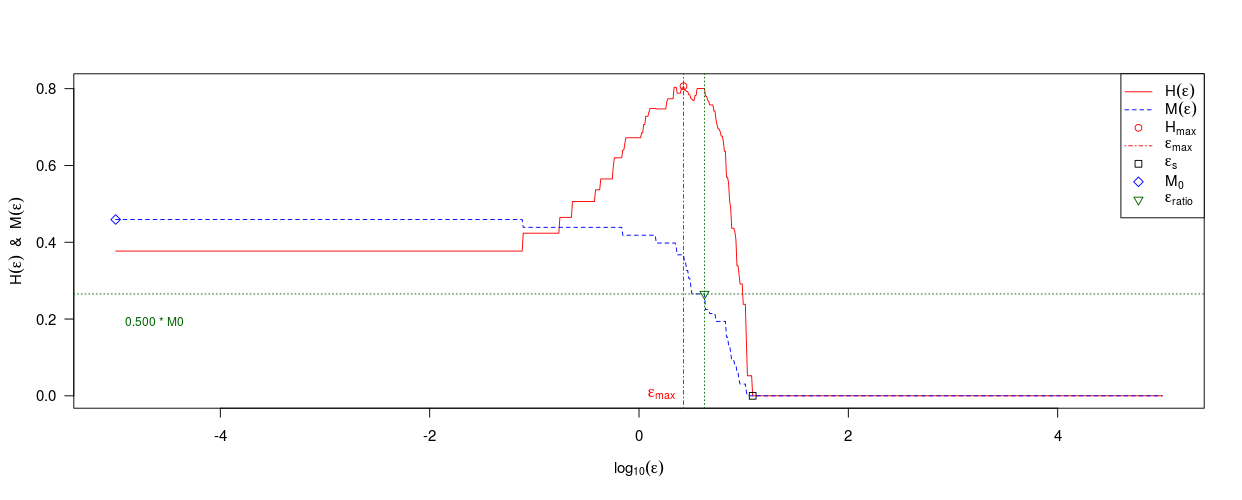} 
     & \includegraphics[width=6cm]{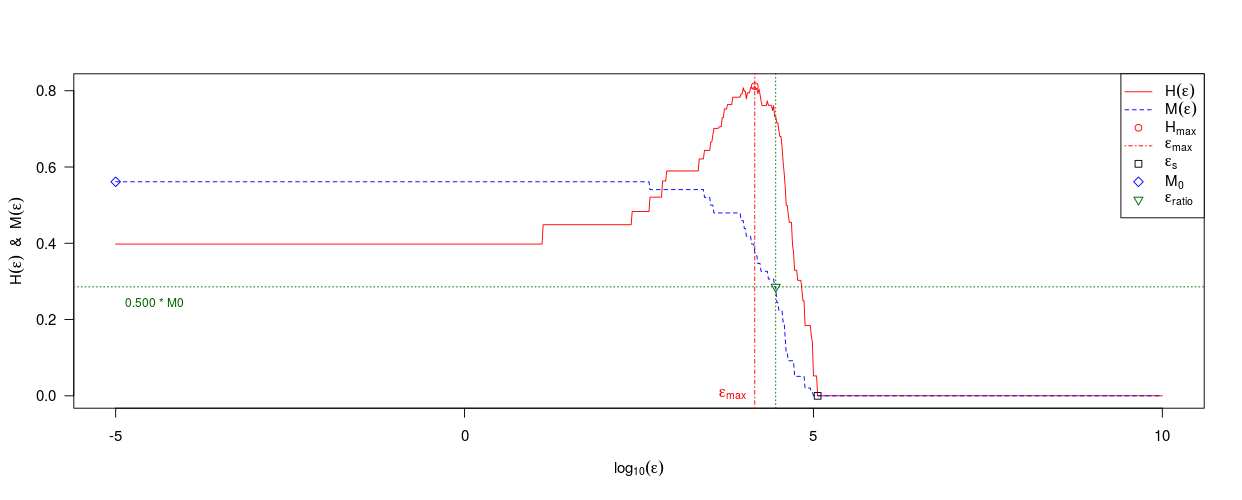}
    \\
    {$f_4$} & {$f_5$} & {$f_6$}
    \\

    \multirow{2}{*}{}  \includegraphics[,width=6cm]{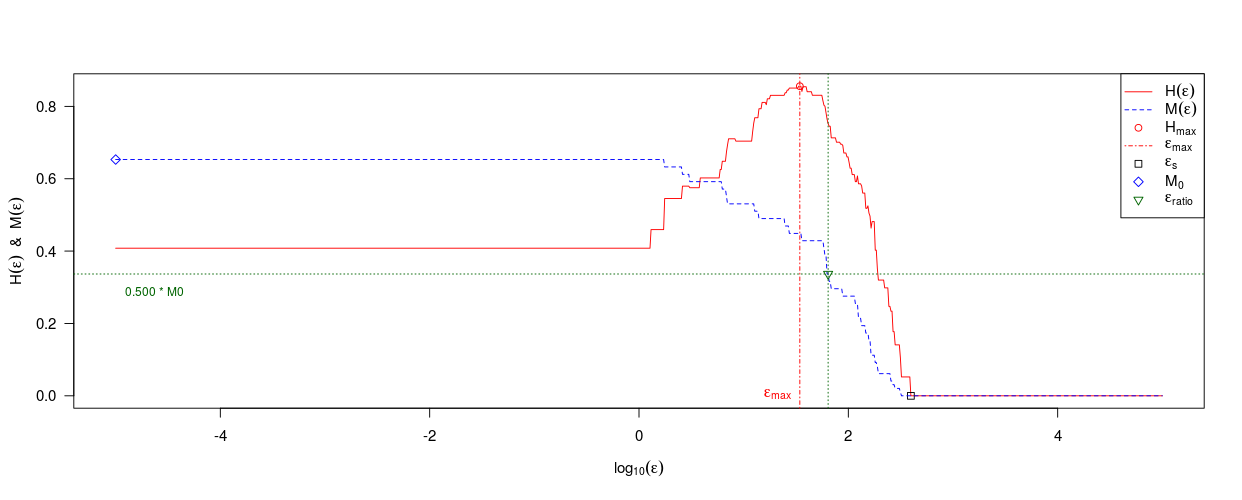}
     & \includegraphics[width=6cm]{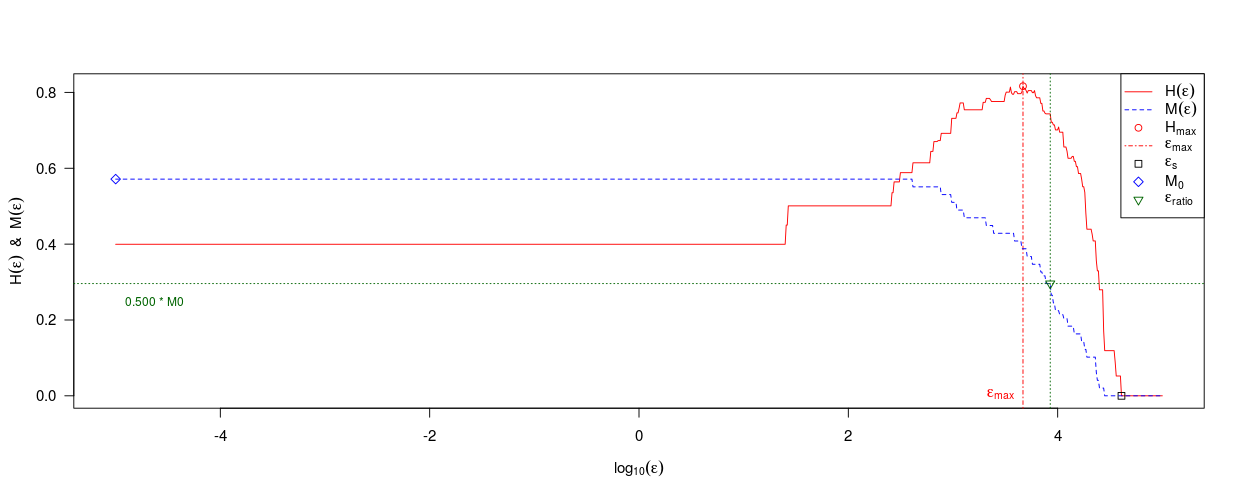} 
     & \includegraphics[width=6cm]{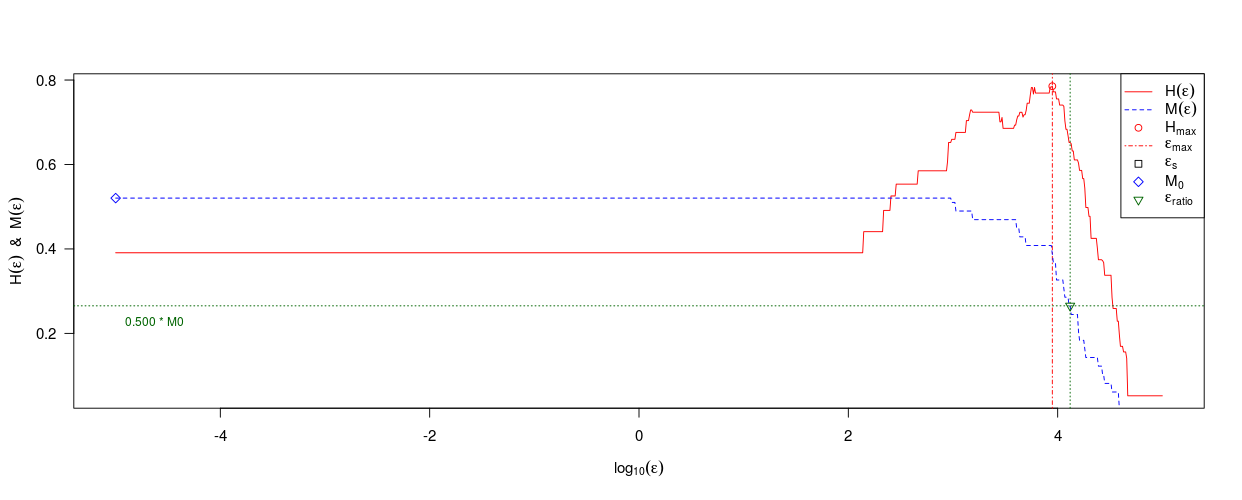}
    \\
    {$f_7$} & {$f_8$} & {$f_9$}
    \\

    \multirow{2}{*}{}  \includegraphics[,width=6cm]{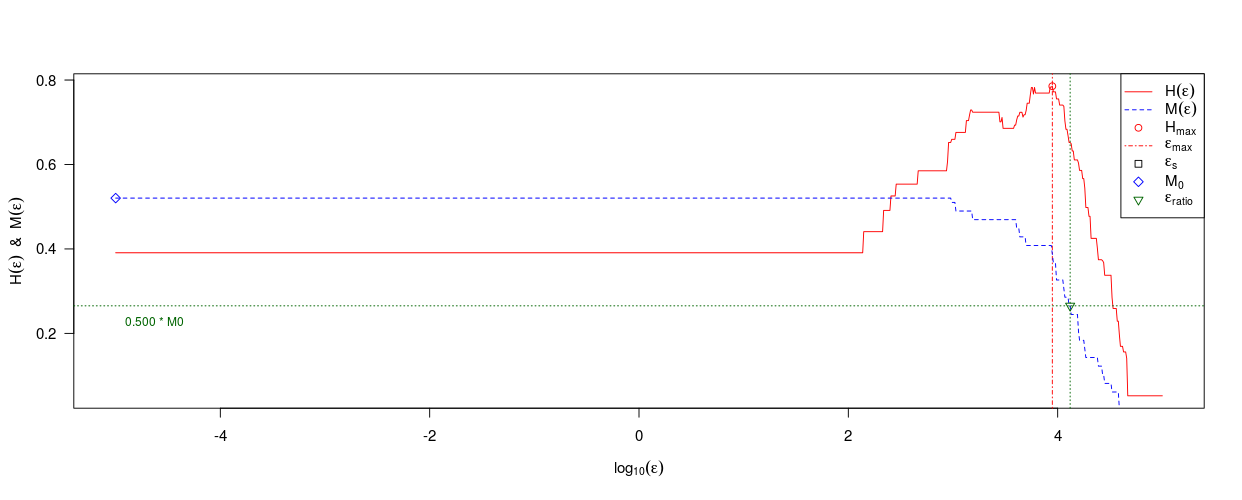}
     & \includegraphics[width=6cm]{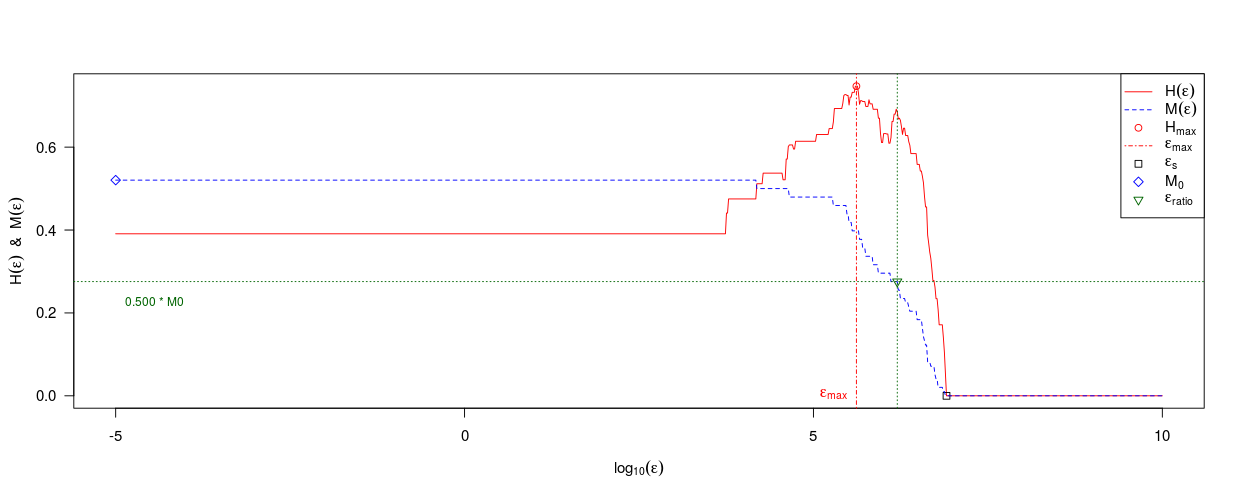} 
     & \includegraphics[width=6cm]{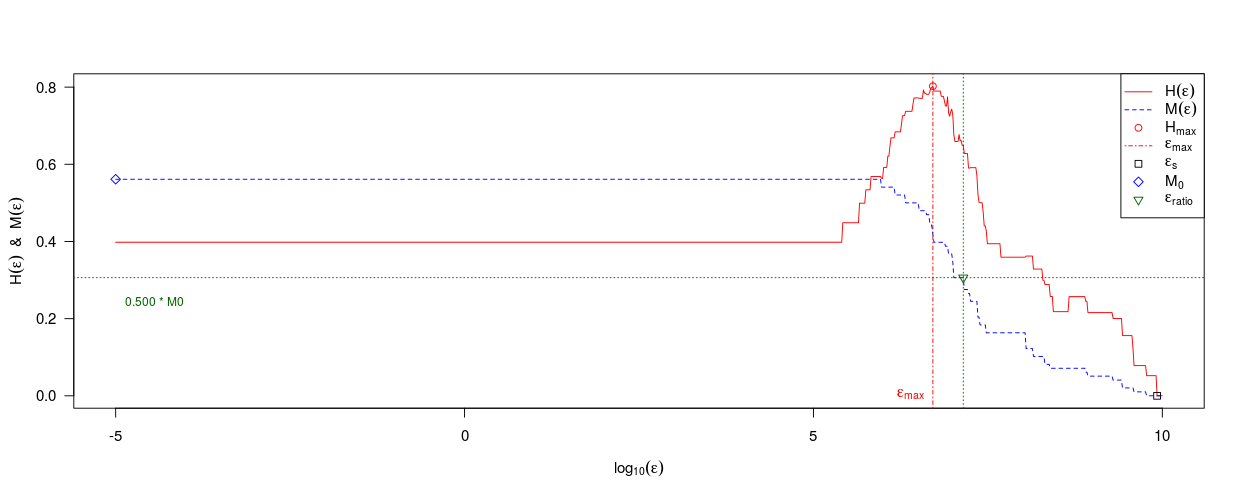}
    \\
    {$f_{10}$} & {$f_{11}$} & {$f_{12}$}
    \\

    \multirow{2}{*}{}  \includegraphics[,width=6cm]{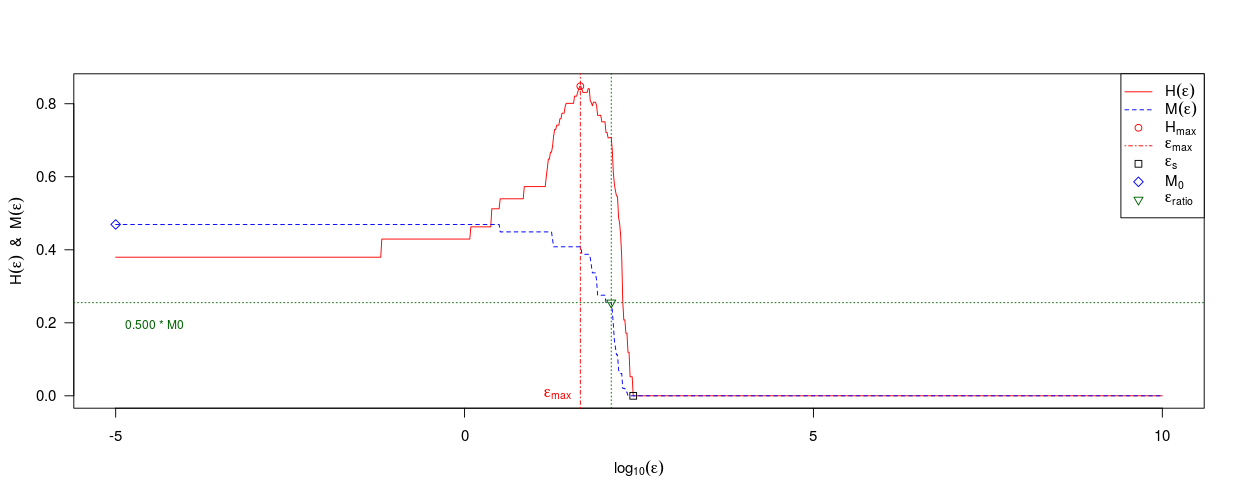}
     & \includegraphics[width=6cm]{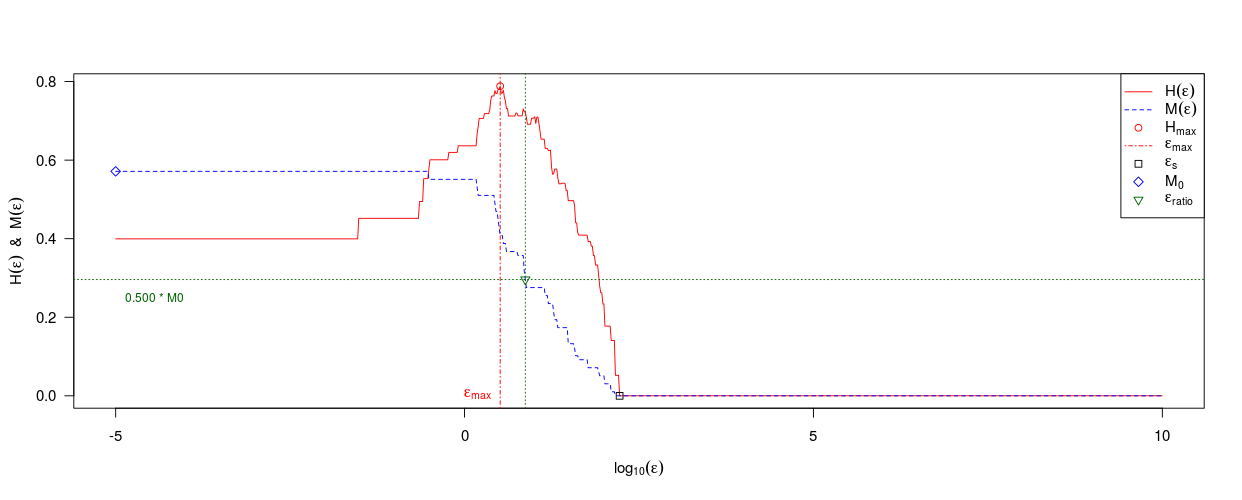} 
     & \includegraphics[width=6cm]{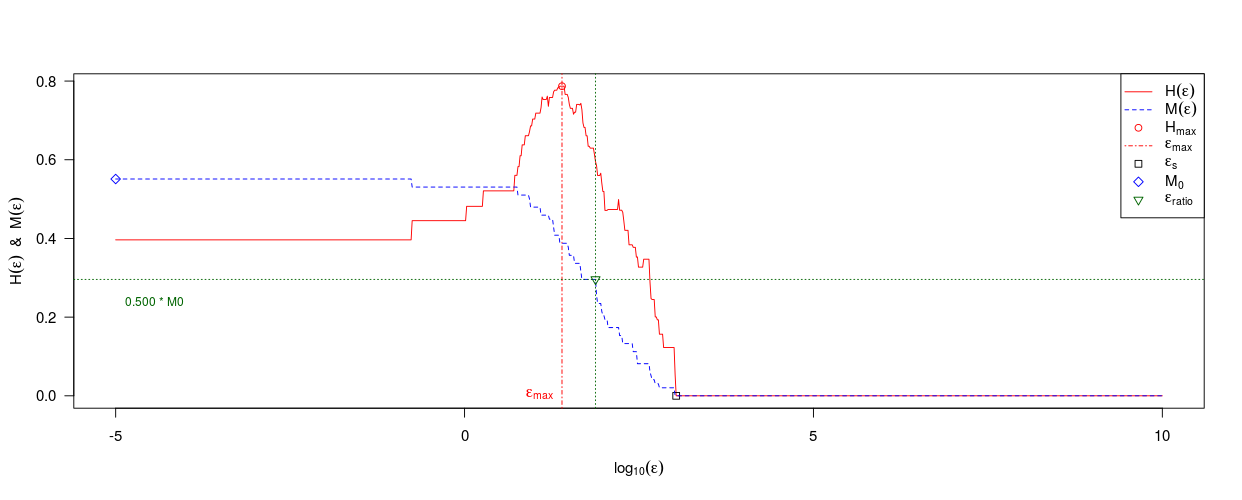}
    \\
    {$f_{13}$} & {$f_{14}$} & {$f_{15}$}
    \\

     \multirow{2}{*}{}  \includegraphics[,width=6cm]{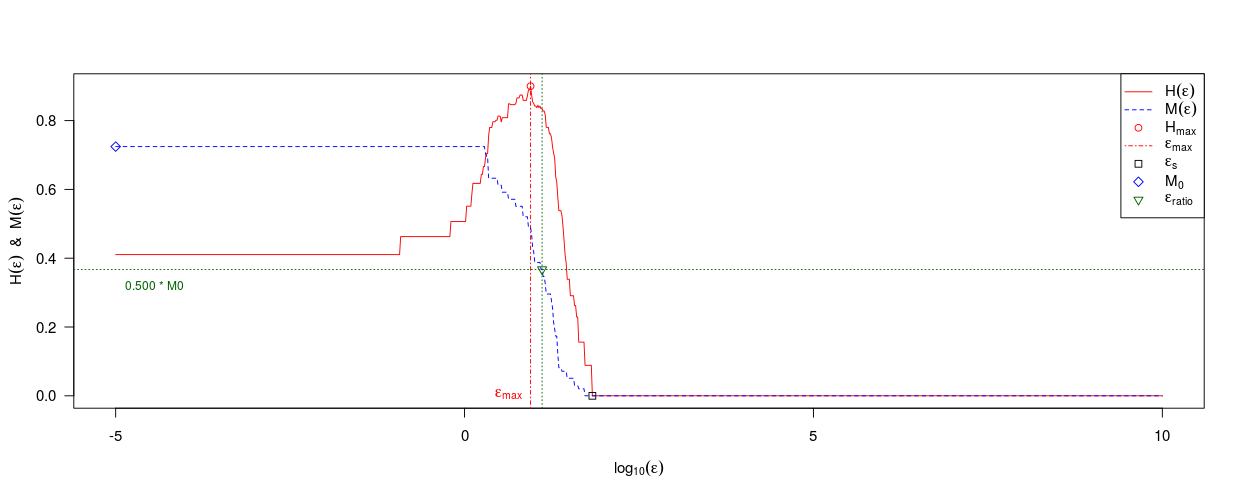}
     & \includegraphics[width=6cm]{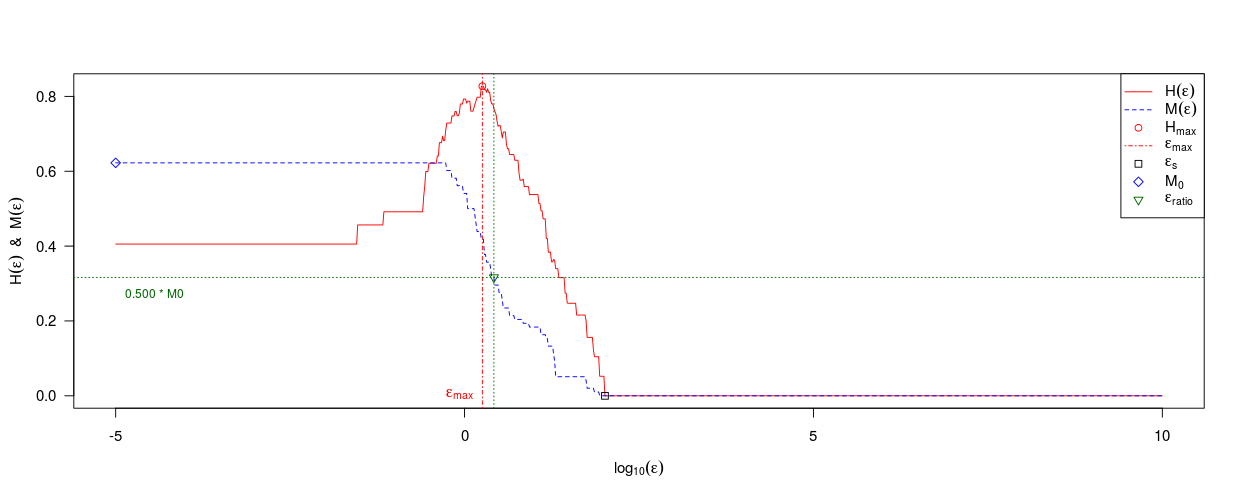} 
     & \includegraphics[width=6cm]{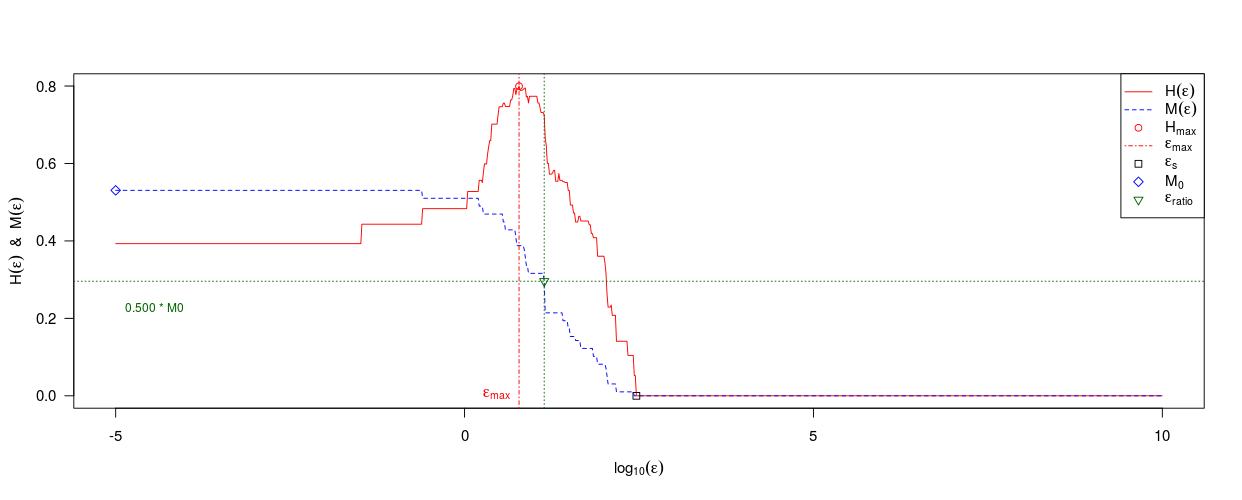}
    \\
    {$f_{16}$} & {$f_{17}$} & {$f_{18}$}
    \\
 
    \multirow{2}{*}{}  \includegraphics[,width=6cm]{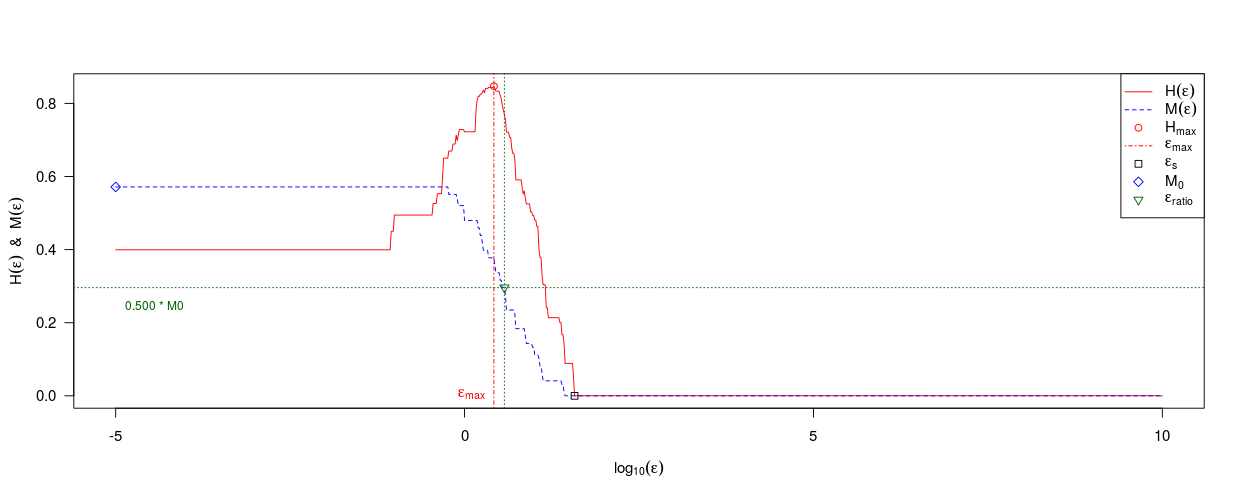}
     & \includegraphics[width=6cm]{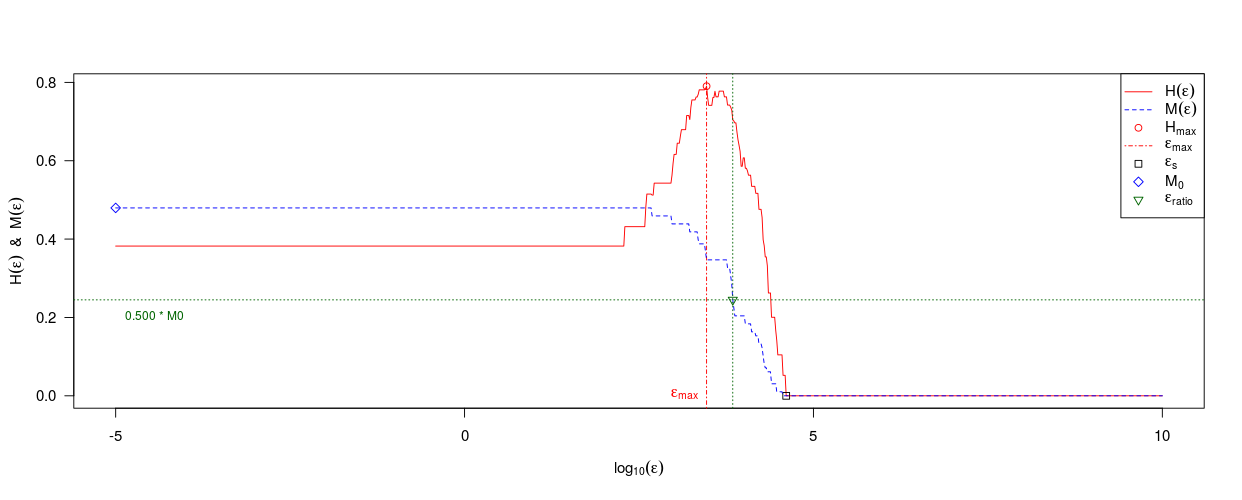} 
     & \includegraphics[width=6cm]{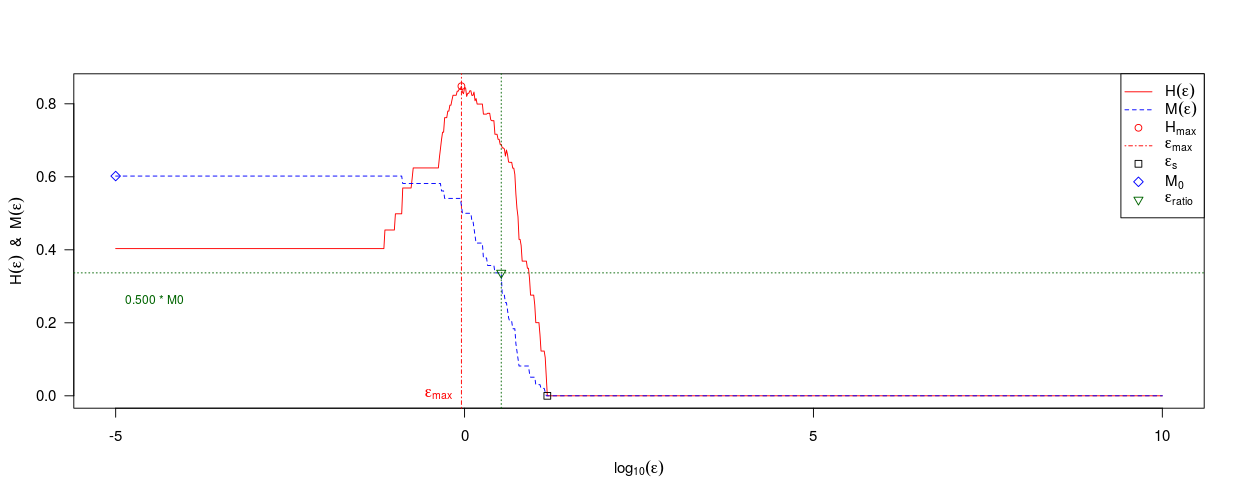}
    \\
    {$f_{19}$} & {$f_{20}$} & {$f_{21}$}
    \\
 
     \multirow{2}{*}{}  \includegraphics[,width=6cm]{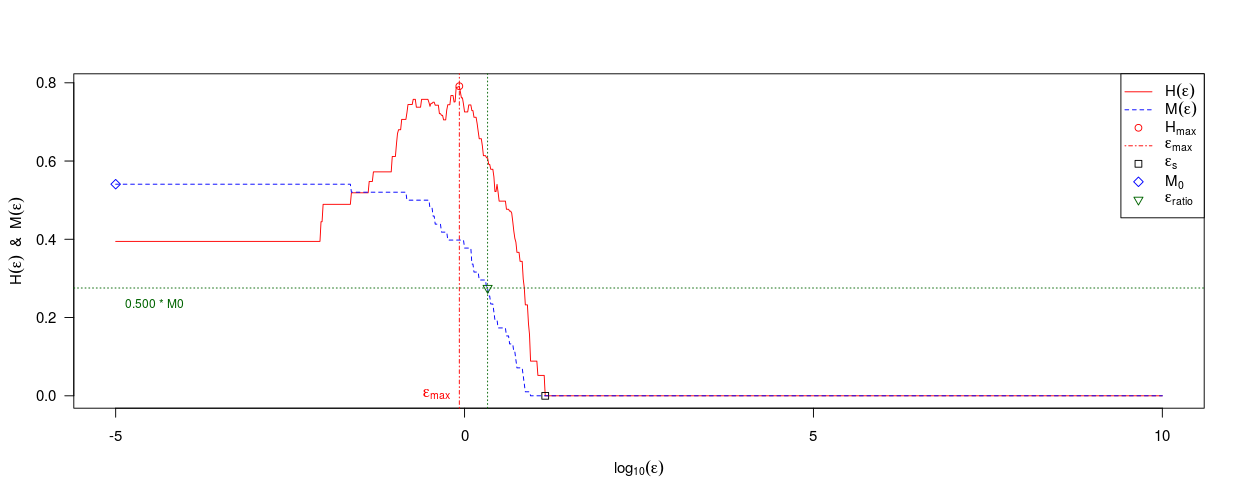}
     & \includegraphics[width=6cm]{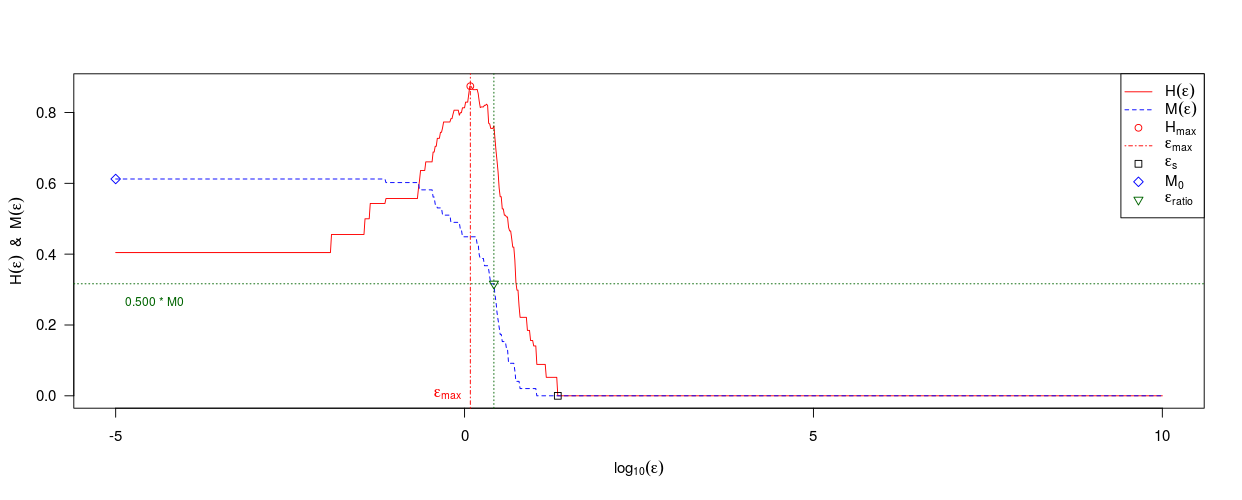} 
     & \includegraphics[width=6cm]{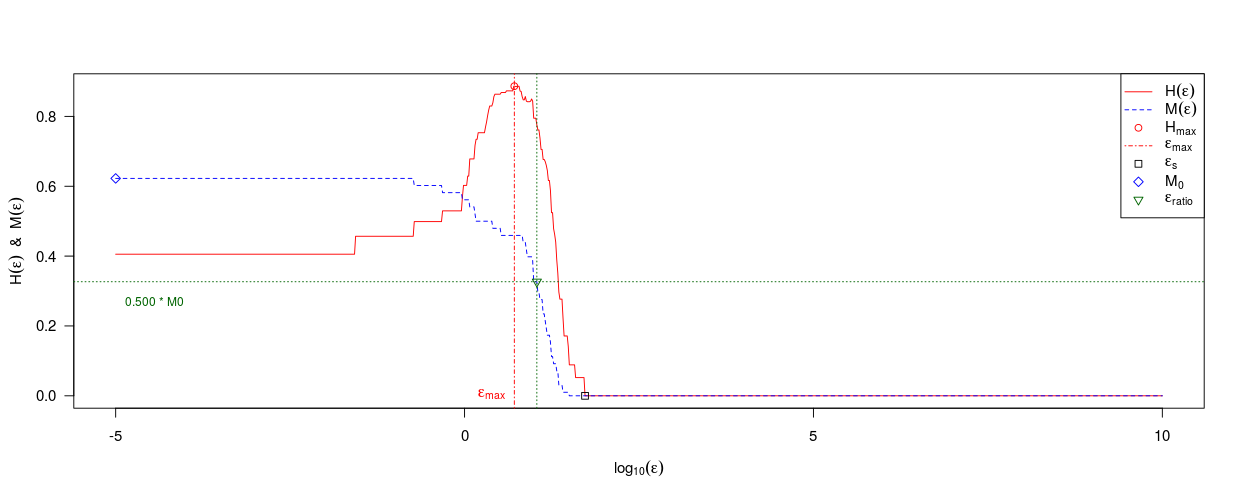}
    \\
    {$f_{22}$} & {$f_{23}$} & {$f_{24}$}
    \\

 \end{tabular}
\end{table*}

\begin{table*}[htp]
\centering
\caption{Analysis of ELA features BBOB functions on d=5.}
\label{Tab E01}
\resizebox{1\textwidth}{!}{
\begin{tabular}{c c c c c c c c c c c}
\hline
\multirow{2}{*}{\textbf{Function}} & \textbf{iid} & \textbf{ela\_meta.lin} & \textbf{ela\_distr} & \textbf{ela\_distr.num-} & \textbf{nbc.nn\_nb.cor} & \textbf{ic.h\_max} & \textbf{ic.eps\_s} & \textbf{ic.eps\_max} & \textbf{ic.eps\_ratio} & \textbf{ic.m0} \\
& & \textbf{\_simple.adj\_r2} & \textbf{.skewness} & \textbf{ber\_of\_peaks} & & & & & & \\
\hline
\multirow{5}{*}{$f_1$} & 1 & $6.08 \times 10^{-1}$ & $5.36 \times 10^{-1}$ & 2 & $6.48 \times 10^{-1}$ & $8.18 \times 10^{-1}$ & $-9.76 \times 10^{-1}$ & $1.60 \times 10^{-2}$ & $-1.50 \times 10^{0}$ & $4.90 \times 10^{-1}$ \\
& 2 & $8.60 \times 10^{-1}$ & $4.01 \times 10^{-1}$ & 1 & $6.65 \times 10^{-1}$ & $8.18 \times 10^{-1}$ & $-1.08 \times 10^{0}$ & $1.75 \times 10^{-2}$ & $-1.56 \times 10^{0}$ & $4.78 \times 10^{-1}$ \\
& 3 & $8.32 \times 10^{-1}$ & $4.22 \times 10^{-1}$ & 1 & $6.48 \times 10^{-1}$ & $8.10 \times 10^{-1}$ & $-1.06 \times 10^{0}$ & $1.67 \times 10^{-2}$ & $-1.58 \times 10^{0}$ & $5.02 \times 10^{-1}$ \\
& 4 & $8.09 \times 10^{-1}$ & $4.12 \times 10^{-1}$ & 2 & $6.54 \times 10^{-1}$ & $8.12 \times 10^{-1}$ & $-1.10 \times 10^{0}$ & $1.39 \times 10^{-2}$ & $-1.64 \times 10^{0}$ & $4.72 \times 10^{-1}$ \\
& 5 & $7.91 \times 10^{-1}$ & $4.60 \times 10^{-1}$ & 1 & $6.47 \times 10^{-1}$ & $8.10 \times 10^{-1}$ & $-1.08 \times 10^{0}$ & $1.52 \times 10^{-2}$ & $-1.58 \times 10^{0}$ & $4.89 \times 10^{-1}$ \\
\hline
\multirow{5}{*}{$f_2$} & 1 & $6.63 \times 10^{-1}$ & $1.03 \times 10^{0}$ & 2 & $5.63 \times 10^{-1}$ & $7.78 \times 10^{-1}$ & $-7.36 \times 10^{-1}$ & $1.60 \times 10^{-2}$ & $-1.44 \times 10^{0}$ & $5.01 \times 10^{-1}$ \\
& 2 & $1.80 \times 10^{-2}$ & $6.45 \times 10^{-1}$ & 2 & $5.40 \times 10^{-1}$ & $7.84 \times 10^{-1}$ & $-6.36 \times 10^{-1}$ & $2.31 \times 10^{-2}$ & $-1.22 \times 10^{0}$ & $5.19 \times 10^{-1}$ \\
& 3 & $8.32 \times 10^{-1}$ & $4.22 \times 10^{-1}$ & 1 & $6.48 \times 10^{-1}$ & $8.10 \times 10^{-1}$ & $-1.06 \times 10^{0}$ & $7.29 \times 10^{-3}$ & $-1.74 \times 10^{0}$ & $5.23 \times 10^{-1}$ \\
& 4 & $8.70 \times 10^{-1}$ & $8.53 \times 10^{-1}$ & 2 & $5.79 \times 10^{-1}$ & $7.15 \times 10^{-1}$ & $-7.76 \times 10^{-1}$ & $6.96 \times 10^{-3}$ & $-1.66 \times 10^{0}$ & $5.09 \times 10^{-1}$ \\
& 5 & $5.87 \times 10^{-3}$ & $6.37 \times 10^{-1}$ & 2 & $5.09 \times 10^{-1}$ & $7.88 \times 10^{-1}$ & $-5.96 \times 10^{-1}$ & $2.53 \times 10^{-2}$ & $-1.18 \times 10^{0}$ & $5.15 \times 10^{-1}$ \\
\hline
\multirow{5}{*}{$f_3$} & 1 & $5.83 \times 10^{-1}$ & $6.29 \times 10^{-1}$ & 1 & $6.13 \times 10^{-1}$ & $8.20 \times 10^{-1}$ & $-8.96 \times 10^{-1}$ & $1.92 \times 10^{-2}$ & $-1.46 \times 10^{0}$ & $5.28 \times 10^{-1}$ \\
& 2 & $5.87 \times 10^{-1}$ & $1.96 \times 10^{0}$ & 2 & $6.23 \times 10^{-1}$ & $7.36 \times 10^{-1}$ & $-5.96 \times 10^{-1}$ & $3.18 \times 10^{-3}$ & $-1.98 \times 10^{0}$ & $4.93 \times 10^{-1}$ \\
& 3 & $6.97 \times 10^{-1}$ & $1.44 \times 10^{0}$ & 2 & $6.23 \times 10^{-1}$ & $6.84 \times 10^{-1}$ & $-6.36 \times 10^{-1}$ & $2.90 \times 10^{-3}$ & $-1.90 \times 10^{0}$ & $4.87 \times 10^{-1}$ \\
& 4 & $8.55 \times 10^{-1}$ & $5.49 \times 10^{-1}$ & 1 & $6.35 \times 10^{-1}$ & $8.00 \times 10^{-1}$ & $-9.56 \times 10^{-1}$ & $1.83 \times 10^{-2}$ & $-1.58 \times 10^{0}$ & $4.95 \times 10^{-1}$ \\
& 5 & $6.01 \times 10^{-1}$ & $1.84 \times 10^{0}$ & 2 & $6.21 \times 10^{-1}$ & $7.24 \times 10^{-1}$ & $-6.56 \times 10^{-1}$ & $2.77 \times 10^{-3}$ & $-1.94 \times 10^{0}$ & $5.04 \times 10^{-1}$ \\
\hline
\multirow{5}{*}{$f_4$} & 1 & $3.12 \times 10^{-1}$ & $2.42 \times 10^{0}$ & 3 & $6.10 \times 10^{-1}$ & $7.55 \times 10^{-1}$ & $-5.16 \times 10^{-1}$ & $4.01 \times 10^{-3}$ & $-1.96 \times 10^{0}$ & $5.02 \times 10^{-1}$ \\
& 2 & $3.52 \times 10^{-1}$ & $1.97 \times 10^{0}$ & 4 & $6.37 \times 10^{-1}$ & $7.39 \times 10^{-1}$ & $-5.96 \times 10^{-1}$ & $4.19 \times 10^{-3}$ & $-1.84 \times 10^{0}$ & $5.00 \times 10^{-1}$ \\
& 3 & $5.88 \times 10^{-1}$ & $1.59 \times 10^{0}$ & 2 & $6.13 \times 10^{-1}$ & $7.23 \times 10^{-1}$ & $-6.56 \times 10^{-1}$ & $5.28 \times 10^{-3}$ & $-1.70 \times 10^{0}$ & $4.91 \times 10^{-1}$ \\
& 4 & $2.96 \times 10^{-1}$ & $1.75 \times 10^{0}$ & 4 & $6.16 \times 10^{-1}$ & $7.82 \times 10^{-1}$ & $-6.16 \times 10^{-1}$ & $6.65 \times 10^{-3}$ & $-1.80 \times 10^{0}$ & $4.99 \times 10^{-1}$ \\
& 5 & $2.63 \times 10^{-1}$ & $2.93 \times 10^{0}$ & 4 & $6.25 \times 10^{-1}$ & $7.89 \times 10^{-1}$ & $-5.16 \times 10^{-1}$ & $2.90 \times 10^{-3}$ & $-2.14 \times 10^{0}$ & $5.33 \times 10^{-1}$ \\
\hline
\multirow{5}{*}{$f_5$} & 1 & $1.00 \times 10^{0}$ & $-1.12 \times 10^{-1}$ & 1 & $6.15 \times 10^{-1}$ & $8.08 \times 10^{-1}$ & $-1.22 \times 10^{0}$ & $1.21 \times 10^{-2}$ & $-1.66 \times 10^{0}$ & $4.73 \times 10^{-1}$ \\
& 2 & $1.00 \times 10^{0}$ & $1.68 \times 10^{-2}$ & 2 & $6.14 \times 10^{-1}$ & $8.03 \times 10^{-1}$ & $-1.20 \times 10^{0}$ & $1.39 \times 10^{-2}$ & $-1.66 \times 10^{0}$ & $4.81 \times 10^{-1}$ \\
& 3 & $1.00 \times 10^{0}$ & $4.49 \times 10^{-2}$ & 2 & $6.27 \times 10^{-1}$ & $8.05 \times 10^{-1}$ & $-1.20 \times 10^{0}$ & $1.39 \times 10^{-2}$ & $-1.66 \times 10^{0}$ & $4.79 \times 10^{-1}$ \\
& 4 & $1.00 \times 10^{0}$ & $-9.57 \times 10^{-2}$ & 1 & $5.78 \times 10^{-1}$ & $8.04 \times 10^{-1}$ & $-1.20 \times 10^{0}$ & $1.46 \times 10^{-2}$ & $-1.66 \times 10^{0}$ & $4.65 \times 10^{-1}$ \\
& 5 & $1.00 \times 10^{0}$ & $1.04 \times 10^{-1}$ & 1 & $6.14 \times 10^{-1}$ & $8.05 \times 10^{-1}$ & $-1.20 \times 10^{0}$ & $1.27 \times 10^{-2}$ & $-1.66 \times 10^{0}$ & $4.64 \times 10^{-1}$ \\
\hline
\multirow{5}{*}{$f_6$} & 1 & $6.13 \times 10^{-1}$ & $9.78 \times 10^{-1}$ & 4 & $6.42 \times 10^{-1}$ & $7.82 \times 10^{-1}$ & $-9.36 \times 10^{-1}$ & $9.62 \times 10^{-3}$ & $-1.70 \times 10^{0}$ & $4.99 \times 10^{-1}$ \\
& 2 & $7.13 \times 10^{-1}$ & $1.71 \times 10^{0}$ & 3 & $6.01 \times 10^{-1}$ & $7.26 \times 10^{-1}$ & $-9.56 \times 10^{-1}$ & $5.53 \times 10^{-3}$ & $-1.86 \times 10^{0}$ & $4.75 \times 10^{-1}$ \\
& 3 & $7.36 \times 10^{-1}$ & $8.27 \times 10^{-1}$ & 4 & $6.04 \times 10^{-1}$ & $7.96 \times 10^{-1}$ & $-9.36 \times 10^{-1}$ & $1.75 \times 10^{-2}$ & $-1.60 \times 10^{0}$ & $4.77 \times 10^{-1}$ \\
& 4 & $8.95 \times 10^{-1}$ & $7.61 \times 10^{-1}$ & 2 & $6.40 \times 10^{-1}$ & $7.90 \times 10^{-1}$ & $-1.04 \times 10^{0}$ & $1.05 \times 10^{-2}$ & $-1.70 \times 10^{0}$ & $4.69 \times 10^{-1}$ \\
& 5 & $9.07 \times 10^{-1}$ & $7.15 \times 10^{-1}$ & 3 & $6.19 \times 10^{-1}$ & $7.51 \times 10^{-1}$ & $-9.36 \times 10^{-1}$ & $1.46 \times 10^{-2}$ & $-1.62 \times 10^{0}$ & $4.46 \times 10^{-1}$ \\
\hline
\multirow{5}{*}{$f_7$} & 1 & $3.80 \times 10^{-2}$ & $2.03 \times 10^{0}$ & 4 & $6.20 \times 10^{-1}$ & $8.01 \times 10^{-1}$ & $-9.56 \times 10^{-1}$ & $1.10 \times 10^{-2}$ & $-1.72 \times 10^{0}$ & $5.09 \times 10^{-1}$ \\
& 2 & $7.80 \times 10^{-1}$ & $1.01 \times 10^{0}$ & 1 & $6.24 \times 10^{-1}$ & $7.79 \times 10^{-1}$ & $-1.02 \times 10^{0}$ & $1.39 \times 10^{-2}$ & $-1.68 \times 10^{0}$ & $4.79 \times 10^{-1}$ \\
& 3 & $7.87 \times 10^{-1}$ & $1.58 \times 10^{0}$ & 3 & $6.15 \times 10^{-1}$ & $7.50 \times 10^{-1}$ & $-9.96 \times 10^{-1}$ & $8.00 \times 10^{-3}$ & $-1.78 \times 10^{0}$ & $4.74 \times 10^{-1}$ \\
& 4 & $3.84 \times 10^{-1}$ & $1.09 \times 10^{0}$ & 2 & $6.18 \times 10^{-1}$ & $7.93 \times 10^{-1}$ & $-9.16 \times 10^{-1}$ & $1.67 \times 10^{-2}$ & $-1.60 \times 10^{0}$ & $5.33 \times 10^{-1}$ \\
& 5 & $2.17 \times 10^{-1}$ & $1.45 \times 10^{0}$ & 3 & $6.11 \times 10^{-1}$ & $7.77 \times 10^{-1}$ & $-9.96 \times 10^{-1}$ & $1.21 \times 10^{-2}$ & $-1.72 \times 10^{0}$ & $4.88 \times 10^{-1}$ \\
\hline
\multirow{5}{*}{$f_8$} & 1 & $3.88 \times 10^{-1}$ & $9.23 \times 10^{-1}$ & 2 & $6.18 \times 10^{-1}$ & $7.81 \times 10^{-1}$ & $-7.36 \times 10^{-1}$ & $1.27 \times 10^{-2}$ & $-1.40 \times 10^{0}$ & $4.99 \times 10^{-1}$ \\
& 2 & $6.55 \times 10^{-1}$ & $1.20 \times 10^{0}$ & 1 & $6.43 \times 10^{-1}$ & $7.49 \times 10^{-1}$ & $-7.56 \times 10^{-1}$ & $1.16 \times 10^{-2}$ & $-1.62 \times 10^{0}$ & $4.98 \times 10^{-1}$ \\
& 3 & $5.99 \times 10^{-1}$ & $9.97 \times 10^{-1}$ & 1 & $6.34 \times 10^{-1}$ & $7.57 \times 10^{-1}$ & $-7.96 \times 10^{-1}$ & $1.52 \times 10^{-2}$ & $-1.56 \times 10^{0}$ & $5.15 \times 10^{-1}$ \\
& 4 & $5.88 \times 10^{-1}$ & $1.14 \times 10^{0}$ & 2 & $6.37 \times 10^{-1}$ & $7.47 \times 10^{-1}$ & $-7.76 \times 10^{-1}$ & $1.60 \times 10^{-2}$ & $-1.54 \times 10^{0}$ & $4.97 \times 10^{-1}$ \\
& 5 & $6.19 \times 10^{-1}$ & $9.76 \times 10^{-1}$ & 1 & $6.38 \times 10^{-1}$ & $7.54 \times 10^{-1}$ & $-8.16 \times 10^{-1}$ & $1.46 \times 10^{-2}$ & $-1.48 \times 10^{0}$ & $5.05 \times 10^{-1}$ \\
\hline
\multirow{5}{*}{$f_9$} & 1 & $5.29 \times 10^{-2}$ & $2.06 \times 10^{0}$ & 3 & $6.20 \times 10^{-1}$ & $7.39 \times 10^{-1}$ & $-7.56 \times 10^{-1}$ & $1.10 \times 10^{-2}$ & $-1.64 \times 10^{0}$ & $4.69 \times 10^{-1}$ \\
& 2 & $6.07 \times 10^{-2}$ & $1.91 \times 10^{0}$ & 3 & $6.34 \times 10^{-1}$ & $7.62 \times 10^{-1}$ & $-7.56 \times 10^{-1}$ & $1.21 \times 10^{-2}$ & $-1.54 \times 10^{0}$ & $5.30 \times 10^{-1}$ \\
& 3 & $5.96 \times 10^{-2}$ & $2.02 \times 10^{0}$ & 2 & $6.25 \times 10^{-1}$ & $7.62 \times 10^{-1}$ & $-7.96 \times 10^{-1}$ & $1.10 \times 10^{-2}$ & $-1.62 \times 10^{0}$ & $4.97 \times 10^{-1}$ \\
& 4 & $7.44 \times 10^{-2}$ & $3.06 \times 10^{0}$ & 4 & $6.30 \times 10^{-1}$ & $7.70 \times 10^{-1}$ & $-9.96 \times 10^{-1}$ & $6.65 \times 10^{-3}$ & $-1.86 \times 10^{0}$ & $5.17 \times 10^{-1}$ \\
& 5 & $5.85 \times 10^{-2}$ & $2.13 \times 10^{0}$ & 4 & $6.40 \times 10^{-1}$ & $7.62 \times 10^{-1}$ & $-8.56 \times 10^{-1}$ & $8.77 \times 10^{-3}$ & $-1.62 \times 10^{0}$ & $4.97 \times 10^{-1}$ \\
\hline
\multirow{5}{*}{$f_{10}$} & 1 & $6.31 \times 10^{-1}$ & $1.66 \times 10^{0}$ & 1 & $5.51 \times 10^{-1}$ & $7.50 \times 10^{-1}$ & $-9.16 \times 10^{-1}$ & $7.29 \times 10^{-3}$ & $-1.78 \times 10^{0}$ & $4.95 \times 10^{-1}$ \\
& 2 & $6.43 \times 10^{-1}$ & $1.65 \times 10^{0}$ & 3 & $5.29 \times 10^{-1}$ & $7.36 \times 10^{-1}$ & $-9.76 \times 10^{-1}$ & $8.00 \times 10^{-3}$ & $-1.86 \times 10^{0}$ & $4.69 \times 10^{-1}$ \\
& 3 & $7.87 \times 10^{-4}$ & $1.20 \times 10^{0}$ & 3 & $5.25 \times 10^{-1}$ & $7.87 \times 10^{-1}$ & $-7.96 \times 10^{-1}$ & $1.52 \times 10^{-2}$ & $-1.50 \times 10^{0}$ & $5.17 \times 10^{-1}$ \\
& 4 & $7.83 \times 10^{-1}$ & $1.86 \times 10^{0}$ & 3 & $6.31 \times 10^{-1}$ & $7.59 \times 10^{-1}$ & $-1.04 \times 10^{0}$ & $5.04 \times 10^{-3}$ & $-1.96 \times 10^{0}$ & $4.96 \times 10^{-1}$ \\
& 5 & $7.36 \times 10^{-1}$ & $2.06 \times 10^{0}$ & 3 & $5.42 \times 10^{-1}$ & $7.28 \times 10^{-1}$ & $-1.02 \times 10^{0}$ & $5.28 \times 10^{-3}$ & $-1.98 \times 10^{0}$ & $4.89 \times 10^{-1}$ \\
\hline
\multirow{5}{*}{$f_{11}$} & 1 & $6.20 \times 10^{-1}$ & $1.60 \times 10^{0}$ & 2 & $5.07 \times 10^{-1}$ & $7.23 \times 10^{-1}$ & $-8.16 \times 10^{-1}$ & $7.29 \times 10^{-3}$ & $-1.70 \times 10^{0}$ & $4.82 \times 10^{-1}$ \\
& 2 & $3.63 \times 10^{-1}$ & $2.11 \times 10^{0}$ & 3 & $5.07 \times 10^{-1}$ & $7.59 \times 10^{-1}$ & $-9.56 \times 10^{-1}$ & $8.00 \times 10^{-3}$ & $-1.80 \times 10^{0}$ & $4.99 \times 10^{-1}$ \\
& 3 & $6.81 \times 10^{-1}$ & $1.65 \times 10^{0}$ & 3 & $5.50 \times 10^{-1}$ & $7.40 \times 10^{-1}$ & $-9.56 \times 10^{-1}$ & $9.62 \times 10^{-3}$ & $-1.84 \times 10^{0}$ & $5.07 \times 10^{-1}$ \\
& 4 & $1.90 \times 10^{-3}$ & $1.73 \times 10^{0}$ & 4 & $4.72 \times 10^{-1}$ & $7.52 \times 10^{-1}$ & $-8.56 \times 10^{-1}$ & $1.21 \times 10^{-2}$ & $-1.68 \times 10^{0}$ & $5.05 \times 10^{-1}$ \\
& 5 & $3.20 \times 10^{-2}$ & $1.52 \times 10^{0}$ & 3 & $4.73 \times 10^{-1}$ & $7.64 \times 10^{-1}$ & $-8.56 \times 10^{-1}$ & $1.21 \times 10^{-2}$ & $-1.70 \times 10^{0}$ & $5.20 \times 10^{-1}$ \\
\hline
\multirow{5}{*}{$f_{12}$} & 1 & $1.06 \times 10^{-1}$ & $1.94 \times 10^{1}$ & 4 & $6.49 \times 10^{-1}$ & $7.47 \times 10^{-1}$ & $-1.40 \times 10^{0}$ & $3.02 \times 10^{-5}$ & $-4.14 \times 10^{0}$ & $4.93 \times 10^{-1}$ \\
& 2 & $4.46 \times 10^{-1}$ & $5.60 \times 10^{0}$ & 4 & $6.54 \times 10^{-1}$ & $6.75 \times 10^{-1}$ & $-1.10 \times 10^{0}$ & $5.78 \times 10^{-4}$ & $-2.62 \times 10^{0}$ & $4.92 \times 10^{-1}$ \\
& 3 & $3.44 \times 10^{-1}$ & $5.73 \times 10^{0}$ & 7 & $6.43 \times 10^{-1}$ & $6.61 \times 10^{-1}$ & $-8.96 \times 10^{-1}$ & $1.45 \times 10^{-4}$ & $-2.84 \times 10^{0}$ & $4.67 \times 10^{-1}$ \\
& 4 & $2.38 \times 10^{-1}$ & $6.51 \times 10^{0}$ & 6 & $6.23 \times 10^{-1}$ & $6.88 \times 10^{-1}$ & $-1.06 \times 10^{0}$ & $2.10 \times 10^{-4}$ & $-2.90 \times 10^{0}$ & $4.87 \times 10^{-1}$ \\
& 5 & $1.60 \times 10^{-1}$ & $1.01 \times 10^{1}$ & 7 & $6.44 \times 10^{-1}$ & $7.35 \times 10^{-1}$ & $-1.04 \times 10^{0}$ & $1.66 \times 10^{-4}$ & $-3.30 \times 10^{0}$ & $5.09 \times 10^{-1}$ \\
\hline
\multirow{5}{*}{$f_{13}$} & 1 & $7.44 \times 10^{-1}$ & $3.28 \times 10^{-1}$ & 1 & $6.26 \times 10^{-1}$ & $8.03 \times 10^{-1}$ & $-1.10 \times 10^{0}$ & $1.75 \times 10^{-2}$ & $-1.56 \times 10^{0}$ & $4.71 \times 10^{-1}$ \\
& 2 & $8.34 \times 10^{-1}$ & $-4.79 \times 10^{-2}$ & 1 & $6.49 \times 10^{-1}$ & $8.18 \times 10^{-1}$ & $-1.16 \times 10^{0}$ & $1.60 \times 10^{-2}$ & $-1.62 \times 10^{0}$ & $4.83 \times 10^{-1}$ \\
& 3 & $9.60 \times 10^{-1}$ & $1.97 \times 10^{-2}$ & 1 & $6.63 \times 10^{-1}$ & $8.03 \times 10^{-1}$ & $-1.16 \times 10^{0}$ & $1.92 \times 10^{-2}$ & $-1.60 \times 10^{0}$ & $4.60 \times 10^{-1}$ \\
& 4 & $5.53 \times 10^{-1}$ & $4.65 \times 10^{-1}$ & 1 & $6.24 \times 10^{-1}$ & $8.09 \times 10^{-1}$ & $-1.06 \times 10^{0}$ & $1.83 \times 10^{-2}$ & $-1.56 \times 10^{0}$ & $4.78 \times 10^{-1}$ \\
& 5 & $2.67 \times 10^{-1}$ & $3.61 \times 10^{-1}$ & 2 & $6.30 \times 10^{-1}$ & $8.20 \times 10^{-1}$ & $-9.96 \times 10^{-1}$ & $2.01 \times 10^{-2}$ & $-1.52 \times 10^{0}$ & $4.93 \times 10^{-1}$ \\
\hline

\end{tabular}}
\end{table*}

\begin{table}[H]
\centering
\caption{Analysis of ELA features BBOB functions on $d=5$.}
\label{Tab E02}
\resizebox{1\textwidth}{!}{
\begin{tabular}{ c c c c c c c c c c c}
\hline
\multirow{2}{*}{\textbf{Function}} & \textbf{iid} & \textbf{ela\_meta.lin} & \textbf{ela\_distr} & \textbf{ela\_distr.num-} & \textbf{nbc.nn\_nb.cor} & \textbf{ic.h\_max} & \textbf{ic.eps\_s} & \textbf{ic.eps\_max} & \textbf{ic.eps\_ratio} & \textbf{ic.m0} \\
& & \textbf{\_simple.adj\_r2} & \textbf{.skewness} & \textbf{ber\_of\_peaks} & & & & & & \\
\hline

\multirow{5}{*}{$f_{14}$} & 1 & $6.42 \times 10^{-1}$ & $2.74 \times 10^{0}$ & 5 & $6.40 \times 10^{-1}$ & $7.08 \times 10^{-1}$ & $-9.76 \times 10^{-1}$ & $2.53 \times 10^{-3}$ & $-2.20 \times 10^{0}$ & $4.85 \times 10^{-1}$ \\
& 2 & $6.47 \times 10^{-1}$ & $2.52 \times 10^{0}$ & 3 & $6.50 \times 10^{-1}$ & $7.15 \times 10^{-1}$ & $-9.56 \times 10^{-1}$ & $1.92 \times 10^{-3}$ & $-2.18 \times 10^{0}$ & $4.75 \times 10^{-1}$ \\
& 3 & $4.60 \times 10^{-1}$ & $2.58 \times 10^{0}$ & 4 & $6.28 \times 10^{-1}$ & $7.49 \times 10^{-1}$ & $-9.36 \times 10^{-1}$ & $4.19 \times 10^{-3}$ & $-1.94 \times 10^{0}$ & $4.94 \times 10^{-1}$ \\
& 4 & $2.98 \times 10^{-2}$ & $2.89 \times 10^{0}$ & 4 & $6.34 \times 10^{-1}$ & $7.45 \times 10^{-1}$ & $-8.56 \times 10^{-1}$ & $3.65 \times 10^{-3}$ & $-1.86 \times 10^{0}$ & $4.97 \times 10^{-1}$ \\
& 5 & $4.20 \times 10^{-1}$ & $3.32 \times 10^{0}$ & 4 & $6.38 \times 10^{-1}$ & $7.69 \times 10^{-1}$ & $-9.76 \times 10^{-1}$ & $4.01 \times 10^{-3}$ & $-2.02 \times 10^{0}$ & $4.97 \times 10^{-1}$ \\
\hline
\multirow{5}{*}{$f_{15}$} & 1 & $5.56 \times 10^{-1}$ & $2.38 \times 10^{0}$ & 3 & $6.44 \times 10^{-1}$ & $7.52 \times 10^{-1}$ & $-8.36 \times 10^{-1}$ & $3.33 \times 10^{-3}$ & $-2.02 \times 10^{0}$ & $4.89 \times 10^{-1}$ \\
& 2 & $5.04 \times 10^{-1}$ & $1.40 \times 10^{0}$ & 1 & $6.07 \times 10^{-1}$ & $8.13 \times 10^{-1}$ & $-9.56 \times 10^{-1}$ & $1.16 \times 10^{-2}$ & $-1.70 \times 10^{0}$ & $4.93 \times 10^{-1}$ \\
& 3 & $3.70 \times 10^{-1}$ & $2.05 \times 10^{0}$ & 5 & $6.09 \times 10^{-1}$ & $8.21 \times 10^{-1}$ & $-8.96 \times 10^{-1}$ & $7.29 \times 10^{-3}$ & $-1.84 \times 10^{0}$ & $5.22 \times 10^{-1}$ \\
& 4 & $4.28 \times 10^{-1}$ & $3.56 \times 10^{0}$ & 5 & $6.10 \times 10^{-1}$ & $7.67 \times 10^{-1}$ & $-8.16 \times 10^{-1}$ & $3.49 \times 10^{-3}$ & $-2.10 \times 10^{0}$ & $5.06 \times 10^{-1}$ \\
& 5 & $3.16 \times 10^{-1}$ & $2.14 \times 10^{0}$ & 3 & $5.83 \times 10^{-1}$ & $7.92 \times 10^{-1}$ & $-1.10 \times 10^{0}$ & $6.06 \times 10^{-3}$ & $-1.90 \times 10^{0}$ & $5.26 \times 10^{-1}$ \\
\hline
\multirow{5}{*}{$f_{16}$} & 1 & $3.72 \times 10^{-3}$ & $1.12 \times 10^{0}$ & 2 & $3.37 \times 10^{-1}$ & $8.75 \times 10^{-1}$ & $-4.35 \times 10^{-1}$ & $3.50 \times 10^{-2}$ & $-1.18 \times 10^{0}$ & $6.51 \times 10^{-1}$ \\
& 2 & $5.17 \times 10^{-3}$ & $9.31 \times 10^{-1}$ & 1 & $3.96 \times 10^{-1}$ & $8.57 \times 10^{-1}$ & $-4.75 \times 10^{-1}$ & $4.01 \times 10^{-2}$ & $-1.12 \times 10^{0}$ & $6.39 \times 10^{-1}$ \\
& 3 & $-5.42 \times 10^{-4}$ & $1.01 \times 10^{0}$ & 2 & $4.04 \times 10^{-1}$ & $8.58 \times 10^{-1}$ & $-4.35 \times 10^{-1}$ & $3.50 \times 10^{-2}$ & $-1.18 \times 10^{0}$ & $6.68 \times 10^{-1}$ \\
& 4 & $-4.41 \times 10^{-4}$ & $8.85 \times 10^{-1}$ & 2 & $3.91 \times 10^{-1}$ & $8.70 \times 10^{-1}$ & $-4.15 \times 10^{-1}$ & $3.83 \times 10^{-2}$ & $-1.14 \times 10^{0}$ & $6.78 \times 10^{-1}$ \\
& 5 & $-3.38 \times 10^{-3}$ & $1.15 \times 10^{0}$ & 2 & $3.58 \times 10^{-1}$ & $8.60 \times 10^{-1}$ & $-4.35 \times 10^{-1}$ & $3.19 \times 10^{-2}$ & $-1.20 \times 10^{0}$ & $6.59 \times 10^{-1}$ \\
\hline
\multirow{5}{*}{$f_{17}$} & 1 & $5.31 \times 10^{-1}$ & $2.78 \times 10^{0}$ & 4 & $5.24 \times 10^{-1}$ & $7.70 \times 10^{-1}$ & $-6.96 \times 10^{-1}$ & $3.33 \times 10^{-3}$ & $-2.14 \times 10^{0}$ & $5.51 \times 10^{-1}$ \\
& 2 & $2.69 \times 10^{-1}$ & $4.64 \times 10^{0}$ & 5 & $5.11 \times 10^{-1}$ & $8.34 \times 10^{-1}$ & $-7.96 \times 10^{-1}$ & $5.53 \times 10^{-3}$ & $-2.04 \times 10^{0}$ & $5.95 \times 10^{-1}$ \\
& 3 & $3.17 \times 10^{-1}$ & $4.90 \times 10^{0}$ & 7 & $5.07 \times 10^{-1}$ & $8.30 \times 10^{-1}$ & $-9.16 \times 10^{-1}$ & $3.49 \times 10^{-3}$ & $-2.12 \times 10^{0}$ & $5.85 \times 10^{-1}$ \\
& 4 & $5.13 \times 10^{-1}$ & $2.33 \times 10^{0}$ & 5 & $5.43 \times 10^{-1}$ & $7.74 \times 10^{-1}$ & $-7.56 \times 10^{-1}$ & $6.96 \times 10^{-3}$ & $-1.92 \times 10^{0}$ & $5.65 \times 10^{-1}$ \\
& 5 & $3.30 \times 10^{-1}$ & $2.12 \times 10^{0}$ & 4 & $4.94 \times 10^{-1}$ & $8.64 \times 10^{-1}$ & $-7.56 \times 10^{-1}$ & $1.60 \times 10^{-2}$ & $-1.56 \times 10^{0}$ & $6.22 \times 10^{-1}$ \\
\hline
\multirow{5}{*}{$f_{18}$} & 1 & $4.70 \times 10^{-1}$ & $3.59 \times 10^{0}$ & 4 & $5.31 \times 10^{-1}$ & $7.97 \times 10^{-1}$ & $-7.56 \times 10^{-1}$ & $3.04 \times 10^{-3}$ & $-2.20 \times 10^{0}$ & $6.03 \times 10^{-1}$ \\
& 2 & $2.80 \times 10^{-1}$ & $3.62 \times 10^{0}$ & 6 & $5.24 \times 10^{-1}$ & $8.31 \times 10^{-1}$ & $-8.16 \times 10^{-1}$ & $5.53 \times 10^{-3}$ & $-2.00 \times 10^{0}$ & $5.99 \times 10^{-1}$ \\
& 3 & $3.43 \times 10^{-1}$ & $4.68 \times 10^{0}$ & 6 & $4.92 \times 10^{-1}$ & $8.13 \times 10^{-1}$ & $-8.16 \times 10^{-1}$ & $3.65 \times 10^{-3}$ & $-2.12 \times 10^{0}$ & $6.19 \times 10^{-1}$ \\
& 4 & $4.10 \times 10^{-1}$ & $3.73 \times 10^{0}$ & 5 & $5.52 \times 10^{-1}$ & $7.81 \times 10^{-1}$ & $-9.36 \times 10^{-1}$ & $3.65 \times 10^{-3}$ & $-2.12 \times 10^{0}$ & $5.65 \times 10^{-1}$ \\
& 5 & $3.39 \times 10^{-1}$ & $1.47 \times 10^{0}$ & 3 & $5.13 \times 10^{-1}$ & $8.56 \times 10^{-1}$ & $-6.96 \times 10^{-1}$ & $2.11 \times 10^{-2}$ & $-1.44 \times 10^{0}$ & $6.22 \times 10^{-1}$ \\
\hline
\multirow{5}{*}{$f_{19}$} & 1 & $7.35 \times 10^{-2}$ & $2.02 \times 10^{0}$ & 3 & $5.67 \times 10^{-1}$ & $7.97 \times 10^{-1}$ & $-8.16 \times 10^{-1}$ & $1.16 \times 10^{-2}$ & $-1.66 \times 10^{0}$ & $5.62 \times 10^{-1}$ \\
& 2 & $5.20 \times 10^{-2}$ & $2.12 \times 10^{0}$ & 3 & $5.61 \times 10^{-1}$ & $8.01 \times 10^{-1}$ & $-7.76 \times 10^{-1}$ & $1.05 \times 10^{-2}$ & $-1.64 \times 10^{0}$ & $5.09 \times 10^{-1}$ \\
& 3 & $5.09 \times 10^{-2}$ & $1.70 \times 10^{0}$ & 4 & $5.50 \times 10^{-1}$ & $7.93 \times 10^{-1}$ & $-7.96 \times 10^{-1}$ & $1.21 \times 10^{-2}$ & $-1.58 \times 10^{0}$ & $5.28 \times 10^{-1}$ \\
& 4 & $7.16 \times 10^{-2}$ & $1.93 \times 10^{0}$ & 3 & $5.56 \times 10^{-1}$ & $7.98 \times 10^{-1}$ & $-7.76 \times 10^{-1}$ & $1.01 \times 10^{-2}$ & $-1.72 \times 10^{0}$ & $5.61 \times 10^{-1}$ \\
& 5 & $2.26 \times 10^{-2}$ & $3.04 \times 10^{0}$ & 5 & $5.03 \times 10^{-1}$ & $7.96 \times 10^{-1}$ & $-8.96 \times 10^{-1}$ & $8.00 \times 10^{-3}$ & $-1.74 \times 10^{0}$ & $5.23 \times 10^{-1}$ \\
\hline
\multirow{5}{*}{$f_{20}$} & 1 & $6.56 \times 10^{-1}$ & $1.40 \times 10^{0}$ & 1 & $6.29 \times 10^{-1}$ & $7.74 \times 10^{-1}$ & $-8.76 \times 10^{-1}$ & $1.10 \times 10^{-2}$ & $-1.66 \times 10^{0}$ & $5.17 \times 10^{-1}$ \\
& 2 & $6.69 \times 10^{-1}$ & $1.42 \times 10^{0}$ & 2 & $6.24 \times 10^{-1}$ & $7.78 \times 10^{-1}$ & $-8.96 \times 10^{-1}$ & $1.16 \times 10^{-2}$ & $-1.72 \times 10^{0}$ & $4.97 \times 10^{-1}$ \\
& 3 & $6.49 \times 10^{-1}$ & $1.47 \times 10^{0}$ & 2 & $6.22 \times 10^{-1}$ & $7.67 \times 10^{-1}$ & $-9.16 \times 10^{-1}$ & $9.18 \times 10^{-3}$ & $-1.70 \times 10^{0}$ & $4.99 \times 10^{-1}$ \\
& 4 & $6.68 \times 10^{-1}$ & $1.36 \times 10^{0}$ & 1 & $6.06 \times 10^{-1}$ & $7.79 \times 10^{-1}$ & $-8.76 \times 10^{-1}$ & $1.16 \times 10^{-2}$ & $-1.68 \times 10^{0}$ & $5.01 \times 10^{-1}$ \\
& 5 & $6.57 \times 10^{-1}$ & $1.50 \times 10^{0}$ & 2 & $6.25 \times 10^{-1}$ & $7.79 \times 10^{-1}$ & $-8.96 \times 10^{-1}$ & $9.18 \times 10^{-3}$ & $-1.72 \times 10^{0}$ & $5.30 \times 10^{-1}$ \\
\hline
\multirow{5}{*}{$f_{21}$} & 1 & $1.51 \times 10^{-2}$ & $-1.28 \times 10^{0}$ & 2 & $4.13 \times 10^{-1}$ & $8.21 \times 10^{-1}$ & $-5.16 \times 10^{-1}$ & $2.65 \times 10^{-2}$ & $-1.28 \times 10^{0}$ & $5.95 \times 10^{-1}$ \\
& 2 & $1.25 \times 10^{-2}$ & $-1.15 \times 10^{0}$ & 1 & $4.39 \times 10^{-1}$ & $8.25 \times 10^{-1}$ & $-4.95 \times 10^{-1}$ & $2.42 \times 10^{-2}$ & $-1.26 \times 10^{0}$ & $5.90 \times 10^{-1}$ \\
& 3 & $2.67 \times 10^{-2}$ & $-1.16 \times 10^{0}$ & 1 & $4.15 \times 10^{-1}$ & $8.26 \times 10^{-1}$ & $-4.75 \times 10^{-1}$ & $2.91 \times 10^{-2}$ & $-1.20 \times 10^{0}$ & $5.89 \times 10^{-1}$ \\
& 4 & $9.02 \times 10^{-3}$ & $-1.25 \times 10^{0}$ & 1 & $3.69 \times 10^{-1}$ & $8.06 \times 10^{-1}$ & $-5.16 \times 10^{-1}$ & $2.91 \times 10^{-2}$ & $-1.22 \times 10^{0}$ & $5.70 \times 10^{-1}$ \\
& 5 & $4.01 \times 10^{-2}$ & $-1.12 \times 10^{0}$ & 2 & $4.93 \times 10^{-1}$ & $8.11 \times 10^{-1}$ & $-4.95 \times 10^{-1}$ & $2.31 \times 10^{-2}$ & $-1.22 \times 10^{0}$ & $6.03 \times 10^{-1}$ \\
\hline
\multirow{5}{*}{$f_{22}$} & 1 & $6.65 \times 10^{-2}$ & $-2.71 \times 10^{0}$ & 5 & $4.70 \times 10^{-1}$ & $7.92 \times 10^{-1}$ & $-4.75 \times 10^{-1}$ & $1.10 \times 10^{-2}$ & $-1.54 \times 10^{0}$ & $5.83 \times 10^{-1}$ \\
& 2 & $4.71 \times 10^{-2}$ & $-2.23 \times 10^{0}$ & 3 & $4.84 \times 10^{-1}$ & $7.94 \times 10^{-1}$ & $-4.95 \times 10^{-1}$ & $1.33 \times 10^{-2}$ & $-1.52 \times 10^{0}$ & $5.66 \times 10^{-1}$ \\
& 3 & $5.01 \times 10^{-2}$ & $-2.20 \times 10^{0}$ & 3 & $4.58 \times 10^{-1}$ & $7.93 \times 10^{-1}$ & $-4.95 \times 10^{-1}$ & $1.46 \times 10^{-2}$ & $-1.52 \times 10^{0}$ & $5.50 \times 10^{-1}$ \\
& 4 & $4.30 \times 10^{-2}$ & $-2.06 \times 10^{0}$ & 2 & $4.59 \times 10^{-1}$ & $7.81 \times 10^{-1}$ & $-4.55 \times 10^{-1}$ & $1.27 \times 10^{-2}$ & $-1.54 \times 10^{0}$ & $5.53 \times 10^{-1}$ \\
& 5 & $1.07 \times 10^{-1}$ & $-2.34 \times 10^{0}$ & 3 & $5.12 \times 10^{-1}$ & $7.85 \times 10^{-1}$ & $-5.56 \times 10^{-1}$ & $1.16 \times 10^{-2}$ & $-1.58 \times 10^{0}$ & $5.64 \times 10^{-1}$ \\
\hline
\multirow{5}{*}{$f_{23}$} & 1 & $9.37 \times 10^{-3}$ & $4.08 \times 10^{-1}$ & 1 & $3.88 \times 10^{-1}$ & $8.75 \times 10^{-1}$ & $-2.55 \times 10^{-1}$ & $5.54 \times 10^{-2}$ & $-9.96 \times 10^{-1}$ & $6.51 \times 10^{-1}$ \\
& 2 & $-3.33 \times 10^{-4}$ & $3.10 \times 10^{-1}$ & 1 & $3.99 \times 10^{-1}$ & $8.73 \times 10^{-1}$ & $-3.15 \times 10^{-1}$ & $5.29 \times 10^{-2}$ & $-9.96 \times 10^{-1}$ & $6.76 \times 10^{-1}$ \\
& 3 & $4.65 \times 10^{-3}$ & $4.10 \times 10^{-1}$ & 1 & $3.60 \times 10^{-1}$ & $8.64 \times 10^{-1}$ & $-2.95 \times 10^{-1}$ & $6.67 \times 10^{-2}$ & $-9.96 \times 10^{-1}$ & $6.61 \times 10^{-1}$ \\
& 4 & $1.19 \times 10^{-4}$ & $3.93 \times 10^{-1}$ & 1 & $4.02 \times 10^{-1}$ & $8.77 \times 10^{-1}$ & $-3.15 \times 10^{-1}$ & $5.54 \times 10^{-2}$ & $-9.96 \times 10^{-1}$ & $6.78 \times 10^{-1}$ \\
& 5 & $1.53 \times 10^{-4}$ & $3.46 \times 10^{-1}$ & 1 & $4.30 \times 10^{-1}$ & $8.82 \times 10^{-1}$ & $-3.35 \times 10^{-1}$ & $5.29 \times 10^{-2}$ & $-9.96 \times 10^{-1}$ & $6.45 \times 10^{-1}$ \\
\hline
\multirow{5}{*}{$f_{24}$} & 1 & $2.46 \times 10^{-1}$ & $3.36 \times 10^{-1}$ & 1 & $5.15 \times 10^{-1}$ & $8.52 \times 10^{-1}$ & $-7.36 \times 10^{-1}$ & $2.78 \times 10^{-2}$ & $-1.36 \times 10^{0}$ & $5.77 \times 10^{-1}$ \\
& 2 & $2.91 \times 10^{-1}$ & $4.08 \times 10^{-1}$ & 2 & $5.44 \times 10^{-1}$ & $8.59 \times 10^{-1}$ & $-8.16 \times 10^{-1}$ & $2.01 \times 10^{-2}$ & $-1.44 \times 10^{0}$ & $5.76 \times 10^{-1}$ \\
& 3 & $2.51 \times 10^{-1}$ & $4.72 \times 10^{-1}$ & 2 & $5.37 \times 10^{-1}$ & $8.60 \times 10^{-1}$ & $-7.96 \times 10^{-1}$ & $1.92 \times 10^{-2}$ & $-1.42 \times 10^{0}$ & $5.72 \times 10^{-1}$ \\
& 4 & $2.29 \times 10^{-1}$ & $3.40 \times 10^{-1}$ & 1 & $5.25 \times 10^{-1}$ & $8.48 \times 10^{-1}$ & $-7.16 \times 10^{-1}$ & $3.04 \times 10^{-2}$ & $-1.32 \times 10^{0}$ & $5.92 \times 10^{-1}$ \\
& 5 & $2.54 \times 10^{-1}$ & $4.29 \times 10^{-1}$ & 2 & $5.22 \times 10^{-1}$ & $8.46 \times 10^{-1}$ & $-7.76 \times 10^{-1}$ & $2.20 \times 10^{-2}$ & $-1.36 \times 10^{0}$ & $5.66 \times 10^{-1}$ \\
\hline
\end{tabular}}
\end{table}
\section{Algorithmic Explainablity of PSO}\label{sec:3}

In this section, a explainability frame work is proposed for PSO. Mathematically, the following two equations govern the PSO framework:
\begin{equation}\label{eq1}
  v_i (t+1)=wv_i (t)+c_1r_1(p_{best,i}-x_i (t))+c_2r_2(g_{best}-x_i (t)),     
\end{equation}
\begin{equation}\label{eq2}   
x_i (t+1)=x_i (t)+v_i (t+1).    
\end{equation}
The movement dynamics of the PSO algorithm is governed by these two fundamental equations (\ref{eq1} and \ref{eq2}). Eq.~(\ref{eq1}) determines velocity, while Eq.~(\ref{eq2}) controls position of the particles (candidate solutions). In these formulations, $v_i (t)$ represents the velocity vector and $x_i (t)$ denotes the position vector of the $i^{th}$  particle in iteration $t$.


Apart from these parameter values, variations in communication topologies significantly influence interactions and information sharing among the particles, The neighbourhood topology in PSO is generally constructed with bidirectional connections, meaning particles maintain mutual influence if particle $j$ belongs to the neighbourhood particle $i$, then $i$ is reciprocally included in $j$'s neighbourhood. In this structure, the particles share information with the neighbouring particles with which they are connected and adapt their trajectories to the best ``local solution" $(pbest)$  found in their local network. The algorithm could theoretically support a wide variety of the topologies, but in practice most of the work has been on a few that have been adopted.
patterns.


\subsection{Swarm Behaviour Analysis on Population Topologies}
In this subsection some of the significant communication topologies in PSO are presented, the \textbf{Star topology} uses a single particle which has a direct communication with all other particles in the swarm, but only with the best solution discovered by a single particle of the population. This architecture results in a complete information network in which all particles are updated by the global optimal position  \cite{miranda2008stochastic, ni2013new}. The topology’s design ensures immediate propagation of the best-known solution throughout the whole swarm was used, which resulted in good directionality of the swarm's guiding force. The structural composition naturally ensures the spread of elite solutions and also the uniformity of influence among all the individuals in the swarm. This is the most information-rich network pattern that can be used in swarm intelligence systems. In the \textbf{Ring topology}, the particles are arranged in a ring topology, where every particle only has $k$ nearest neighbors. In this decentralized architecture, there are two modes of operation: one static with fixed neighbourhood relationships and another dynamic with relationships which can vary during the search process. The distance metric used in particle interactions are Minkowski $p$-norm distance measures; $p$ = 1 (Manhattan distance $L_1$), and $p$ = 2 (Euclidean distance $L_2$) are used for determining the distance measure between swarm members. The limited connectivity pattern leads to the presence of small information channels within the population which as a natural consequence, reduces the extent of solution homogenization throughout the population \cite{liu2016topology}. In the \textbf{Von Neumann topology}, the connection between particles is based on relationships between Delannoy numbers in a multidimensional grid. Each pattern corresponds to a set of combinatorial rules which prescribe the size of the neighborhood of particles depending on the specified interaction range $r$ and the dimensionality of the problem. The resulting lattice structure creates regular, spatially-distributed connections between particles, establishing a systematic information flow pattern across the search space. This geometric arrangement provides distinct advantages in maintaining organized particle interactions while preserving structured exploration capabilities \cite{von1935complete, lynn2018population}.

A PSO explainer model based on the IOHxplainer ~\cite{van2025explainable} environment. IOHxplainer provides a powerful framework for capturing, visualising and quantifying the dynamics of algorithms, by associating algorithm performance with features of exploratory landscape analysis (ELA) and other descriptive algorithm features. This directly follows our previous work on the dynamic visualization of STNs, enabling real-time interactive tracking of the temporal solution trajectories and discovering important transition points during optimization runs. The visualization layer of our current PSO system is informed by that prior experience in designing intuitive visual analytics for dynamic systems. We hope to identify interpretable features of swarm search behaviour, parameter sensitivities, and variations in performance due to landscape. This would provide improved transparency in the understanding of PSO and would also help to select a suitable algorithm, to tune the parameters and to design an adaptive strategy for complex optimization problems.

\subsubsection{IOHxplainer}\label{subsec:1}
IOHxplainer is a benchmarking platform for comparing performance of iterative optimization heuristics (IOHs), including evolutionary algorithms and swarm-based algorithms~\cite{van2025explainable}. It uses explainable artificial intelligence (XAI) to automatically create informative visualizations and statistical analysis, making the assessment of millions of algorithm configurations of large-scale empirical studies much more efficient. Through this one, researchers can learn the interaction of components of an algorithm and their contribution to the performance of the algorithm in different problems more effectively areas. Many different algorithm setups are tested over a variety of benchmark functions, and the performance within the system is tested and recorded for use, such as at any time. IOHxplainer begins by defining the space of configurations for the algorithm that users wish to explore. This involves establishing a maximum or minimum value for continuous parameters and maximum or minimum values for integer parameters, and defining any relationships between these parameters on condition. The platform supports these data are generated by both a thorough grid search and random sampling, allowing for easier exploration and analysis of potentially very large configuration spaces. IOHxplainer offers a unique benefit for the study of modular algorithms that offer many configuration options, but can also be used to analyze conventional algorithms as well as their hyperparameter spaces. The configuration space is defined in a flexible manner (see also the definition of other approaches, such as SMAC~\cite{garcia2023explainable, lindauer2019boah}), which allows flexibility in regards to different optimization scenarios.


\begin{figure*}[h]
    \centering
    \includegraphics[width=\linewidth]{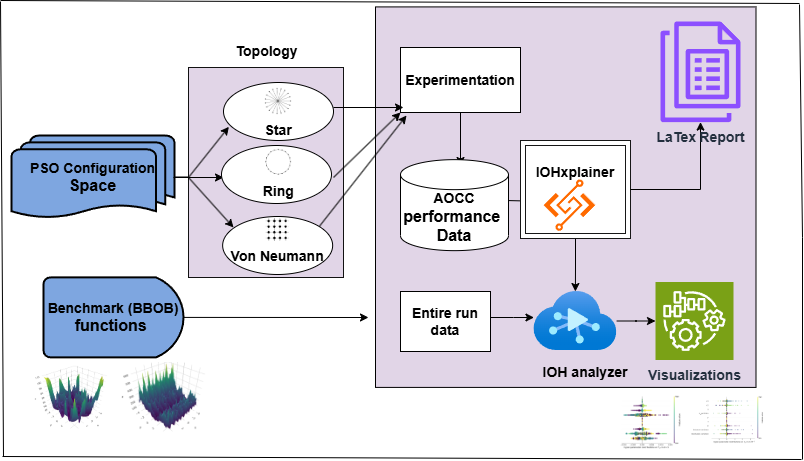}

      \caption{A Framework for Automated and Explainable Analysis of PSO Configurations using IOHxplainer .} \label{f2}
\end{figure*}

For this research, we selected the IOHxplainer framework which was designed and tested for evolutionary algorithms like modCMA \cite{kostovska2022importance} and modDE~\cite{van2024explainable}. We have restructured this framework to model and explain the action of PSO algorithm and in particular, the dynamics of PSO algorithm coupled with different communication topology. Figure \ref{f2} illustrates in detail a representation of this framework.


The structured, three-stage experimental pipeline for the development of the IOHxplainer to integrate PSO is outlined. The development of the IOHxplainer to integrate PSO followed a structured, three-stage experimental pipeline. Firstly the basic PSO parameters were set and fixed to create a baseline configuration space of PSO parameters, as described in Table \ref{tab2}. Secondly, this base algorithm was implemented with three different communication topologies: Star, Ring, and Von Neumann, and each was extensively tested against this wide range of BBOB test functions with five different stopping criteria, using the number of function evaluations as the basis for the test results. Finally, the experimental data generation phase or AOCC performance data, which is the creation of data based on the configured PSO algorithms for all BBOB functions, was performed with many precautions to yield rich, time-series datasets. In this process a number of repetitions were to be carried out independently for each algorithm-topology-function combination to obtain statistical significance, and a set number of function calls was to be used while maintaining a fixed budget of function calls. Crucially, a complete log of the optimization trajectory was generated during each run, recording the best-so-far fitness value at a relatively high frequency, typically after each evaluation, giving a complete record of the optimization process over time. This resulted in a comprehensive ``entire run data" output, a standardized collection of files containing the full history of each experiment, which was then perfectly structured for subsequent ingestion and analysis by the IOHprofiler tools. Algorithm \ref{algo:1} provides the details of setup framework of integration of IOHxplainer with PSO.

\begin{algorithm}[h]
\caption{Integrated PSO--IOHxplainer Framework}\label{algo:1}
\begin{algorithmic}[1]
\State \textbf{Import:} \texttt{pyswarms}, \texttt{ConfigurationSpace}, \texttt{Star}, \texttt{Ring}, \texttt{VonNeumann}, \texttt{explainer}, \texttt{seaborn}
\State \textbf{Define configuration space:} \texttt{confSpace}: $\{c_1, c_2, w, n\_{\text{particles}}, k, p, r\}$
with discrete parameter values
\State \textbf{Initialize topology:} $\text{Topology()}$
\State \textbf{Define function} \texttt{run\_pso(func, config, budget, dim)}:
\Statex \hspace{1.5em} Extract hyperparameters from \texttt{config}
\Statex \hspace{1.5em} Initialize PSO via \texttt{GeneralOptimizerPSO}
\Statex \hspace{1.5em} \Return \texttt{optimizer.optimize(func, iters=budget)}
\State Initialize: \texttt{explainer}(\texttt{run\_pso}, confSpace, \texttt{algname="PSO"})
\State Set experimental parameters: \text{dimension}, \; \text{func}, \; \text{iids}=[1,2,3,4,5], \; 
\State Configure sampling method: \texttt{"grid"}, budget, \texttt{seed=1}
\State Execute: \texttt{explainer.run(parallel=False, checkpoint="data.csv")}
\State Save results: \texttt{explainer.save\_results("results.pkl")}
\State Generate statistics: \texttt{df = explainer.performance\_stats()}
\State Explain results: \texttt{explainer.explain(partial\_dependence=False, best\_config=True)}
\end{algorithmic}
\end{algorithm}

The improvements are designed to make the impacts of topology configurations more apparent on the algorithm's performance, using a set of XAI techniques. This allows information-rich visualisations and data visualisations in the form of statistics to be derived from detailed empirical analysis, so that potentially countless PSO algorithmic architectures can be evaluated. Also, hyper-parameter dependencies can be defined and taken into account, which results in a more fine-grained approach to the configuration of PSO algorithm and the considered topologies. We use the well-known BBOB noiseless functions~\cite{hansen2009real} as a part of the COCO framework~\cite{hansen2022anytime} to set up a particular PSO configuration and test it against 24 diverse noiseless functions with a different number of parameters and a different number of evaluations.

At the end of the experimental runs that are run concurrently, the framework records the fine details of performance (as per the set budget, set target, or flexible time performance measures). The `explainer' part of the framework then uses sophisticated XAI methods, including the \texttt{TreeSHAP} technique ~\cite{lundberg2020local} to compute the SHAP values for the different parts of the PSO algorithm and hyperparameter of its topology ~\cite{lundberg2017unified}. These values indicate the marginal impact of each element on the performance metrics across different runs. The SHAP values are carefully computed and displayed for every function in the benchmark suite, albeit for each function the number of computations is limited. This visualization, often presented as a \textit{Beeswarm }, \textit{Bar }, and \textit{Decision} plots provide PSO designers and practitioners with a clear view of how specific hyperparameters influence performance under varied conditions and across different topologies. This data, once gathered, is processed either in combination with previous datasets or as a stand-alone analysis to extract SHAP values. The details of experiments and results are compiled in subsequent sections and related resources are provided at GitHub link: \textcolor{blue}{\url{https://github.com/GitNitin02/ioh_pso}}.


\subsection{STN-based topologies network}
\begin{figure*}[h]
    \centering
    \includegraphics[width=\linewidth]{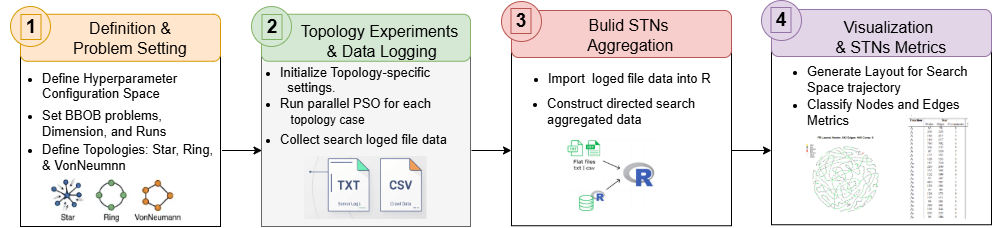}

      \caption{Standardized Parallel working Flow chart of the PSO STNs Pipeline.} \label{flow:stn}
\end{figure*}

Structural characteristics that are often not enough from network metrics can be well perceived using visualization. For the visual representation of the axes, throughout this study, we will use the conventional node – edge diagram, where the nodes are represented by points in the Euclidean space (2d or 3d), and the edges are represented on straight or curved lines connecting the nodes themselves. Moreover, for directed networks, arrowheads are added to make it clear that the links are directed. A directed graph, search trajectory network (\(\mathcal{G}(N,E)\)) \cite{ochoa2021search,gupta2026interpretability}  consists of a node of $N$ and an edge of $E$. Once the data logs of a predetermined number of runs of a specific algorithm-problem instance pair are collected, a post processing step gives a summary of all the representative solutions and transitions to create a single network object. Another mapping is needed to provide a correspondence between each representative solution and the spatial coordinates and the objective value for the solution. We generated all graph visualizations in this paper using the igraph package \cite{csardi2006igraph} within the R programming environment. For node positioning, we examined several force-directed layout algorithms, specifically the Fruchterman–Reingold ~\cite{gajdovs2016parallel} and Kamada–Kawai  ~\cite{sarpkamada} methods. To systematically realize this encoding, a standardized parallel working flowchart of the PSO-STNs pipeline is constructed in figure \ref{flow:stn}


The STN framework encodes progressive search behaviours into an interpretable topological structure through a discretization-based transformation. The transformation initiates with a partitioning of the search space into uniform cells, where each dimension is divided into intervals of length \(10^{-\lambda}\), with \(\lambda\) denoting the solution precision that governs the coarseness of the grid. Under the assumption of identical bounds across all dimensions, \(\lambda\) is scaled appropriately to adapt the granularity to both the problem's dimensionality and its domain extent:
\begin{equation}
\lambda = 2 - \lfloor \log_{10} (d \cdot (U - L)) \rfloor,
\end{equation}
where \(d\) denotes the dimensionality, and \(U\) and \(L\) are the upper and lower bounds of the search space, respectively.  At the same time, the search algorithm keeps a historical database \(\mathcal{H}\) which can store all
Calculated fitness of agents and fitnesses at each epoch and provide them as a data source for network synthesis. Formally, this archive is defined as:

\begin{equation}
\mathcal{H} = \{ A_t, F(A_t) \mid \forall t = 1, \dots, T_{\max} \},
\end{equation}
where \(A_t\) represents the set of agents at epoch \(t\), \(F(\cdot)\) is the fitness evaluation function, and \(T_{\max}\) denotes the total number of iterations executed. This archive is subsequently transformed via the mapping:
\begin{equation}
\Psi_{\lambda} : \mathcal{H} \longrightarrow (\mathcal{G},\; \mathcal{Q}),
\end{equation}
which yields a directed graph \(\mathcal{G}\) alongside its associated metric profile \(\mathcal{Q}\) for subsequent analysis. We systematically integrate STN analysis into the PSO model through a multi-stage pipeline, as formalized in Algorithm \ref{algo: stn}.

\begin{algorithm}[h]\footnotesize 
\caption{STN Pipeline for PSO with Communication Topologies}
\label{algo: stn}
\begin{algorithmic}[1]
\State \textbf{Import:} \texttt{ioh}, \texttt{numpy}, \texttt{pandas}, \texttt{pyswarms}, \texttt{concurrent.futures} (Python); \texttt{igraph}, \texttt{plyr} (R)
\State \textbf{Define:} $\mathcal{C} = \{(c_1, c_2, w, n, \theta)\}$, $d$, Runs=$R$, $E=100d$, BBOB $\mathcal{F}$
\State \textbf{Define:} Topologies $\mathcal{T} = \{\text{Star}, \text{Ring}, \text{VonNeumann}\}$ with $\Theta_{\text{Star}}=\varnothing$, $\Theta_{\text{Ring}}=\{(k,p)\}$, $\Theta_{\text{VonNeumann}}=\{(r,p)\}$
\State \textbf{Initialize:} $\text{map\_to\_node}(x, \lambda) = \text{round}(x / 10^{-\lambda}) \cdot 10^{-\lambda}$, $\lambda = 2 - \lfloor \log_{10}(d \cdot (U - L)) \rfloor$

\State \textbf{Stage 1 - Parallel PSO Runs:} \textsc{PSO\_STN\_Logging}($\mathcal{F}, d, R, \mathcal{C}, \mathcal{T}$):
\ForAll{$\tau \in \mathcal{T}$ \textbf{and} $f \in \mathcal{F}$}
    \State Compute $\lambda = 2 - \lfloor \log_{10}(d \cdot (U - L)) \rfloor$
    \ForAll{$(c_1, c_2, w, n, \theta) \in \mathcal{C} \times \Theta_\tau$}
        \For{$r = 1$ to $R$}
            \State Run PSO in parallel with topology $\tau$ and params $(c_1, c_2, w, n, \theta)$
            \While{$done < E$}
                \State Evolve for $chunk = \min(d, E - done)$ iterations
                \State Log discretized node $\text{map\_to\_node}(x_{best}, \lambda)$ and fitness
                \State $done \gets done + chunk$
            \EndWhile
        \EndFor
    \EndFor
\EndFor

\State \textbf{Stage 2 - Aggregation:} \textsc{Aggregate\_STNs}(results):
\ForAll{config results}
    \State Aggregate nodes and edges across $R$ runs
    \State Save STN file: \texttt{PSO\_\{$\tau$\}\_F\{fid\}\_d\{d\}\_C1\{$c_1$\}\_C2\{$c_2$\}\_w\{$w$\}\_n\{$n$\}\_params\{$\theta$\}STN.txt}
    \State Format: \textsc{Run Fitness1 Solution1 Fitness2 Solution2}
\EndFor

\State \textbf{Stage 3 - R Construction:} \textsc{Build\_STNs}() and \textsc{Decorate\_STN}($G$):
\ForAll{trace files}
    \State Build directed graph $G(V,E,W)$ from nodes and weighted edges
    \State Classify nodes as \{start, end, best, medium\} and edges as \{improving, equal, worsening\}
    \State Save as \texttt{.RData}
\EndFor

\State \textbf{Stage 4 - Visualization:} \textsc{Visualize\_STNs}($\mathcal{G}$):
\ForAll{$G \in \mathcal{G}$}
    \State Decorate $G$ with colors/sizes and generate Fruchterman-Reingold \& Kamada-Kawai layouts
    \State Export PDF with legend
\EndFor

\State \textbf{Execute:} Stage 1 $\rightarrow$ Stage 2 $\rightarrow$ Stage 3 $\rightarrow$ Stage 4
\end{algorithmic}
\end{algorithm}

\section{Experimental Setup and Results }\label{sec:4}
In this section, an experimental design testing PSO and its topological variants, with special concern about algorithmic explainability using our analytical framework (section \ref{sec:3}) is presented. The comprehensive BBOB suite of 24 functions, each with their own mathematical characteristics and optimization problem is used in our evaluation. The benchmark set comprises five classes: separable functions ($f_1$-$f_5$); functions with low-moderate conditioning ($f_6$ - $f_9$); highly conditioned, unimodal functions ($f_{10}$ - $f_{14}$); structured multimodal functions ($f_{15}$ - $f_{19}$); and weakly structured multimodal functions ($f_{20}$ - $f_{24}$). Within these groups, we further distinguish between unimodal ($f_1$, $f_2$, $f_5$ - $f_{14}$), multimodal ($f_3$, $f_4$, $f_{15}$, $f_{16}$, $f_{20}$, $f_{23}$, $f_{24}$), and highly multimodal ($f_{17}$ - $f_{19}$, $f_{21}$, $f_{22}$) functions.  This carefully selected collection allows us to fully evaluate PSO's performance across a range of problem types, from simple convex landscapes, to more complex rugged optimization surfaces. We have integrated these variety of benchmark functions, together with our explainability framework, we are able to offer both quantitative measures of performance and qualitative understandings of how these functions operate. Various optimization problems are addressed for different PSO configurations. By this experimental design, we can systematically investigate the advantages and disadvantages of each of these different topologies within the entire range of possibilities for optimization difficulty.

The two problem dimensions (2d and 5d), with all functions for each PSO configuration are considered in our experimental investigation. Three topologies of swarms are used: star, ring and Von Neumann, and the number of iterations is set to be 100 and 500. PSO parameters for both dimensional spaces are described in the baseline table (Table ~\ref{tab2}) for each topology. The methodology used in achieving statistical reliability is done by running 5 independent runs for the  five instances of each benchmark function.

\subsection{Performance Metrics}
We employ an anytime performance measure to evaluate the generated algorithms efficiently on a complete set of benchmark functions. This is important because it captures the full convergence profile of an algorithm, and therefore rewards those that find good solutions quickly and reliably, rather than those that converge to a decent solution by the final budget. Specifically, it quantifies performance over the complete budget, instead of only looking at the final objective function value. We use the normalized Area Over the Convergence Curve (AOCC) (Eq.~(\ref{eq3})), as introduced in ~\cite{van2025explainable}. The AOCC is calculated as:

\begin{table}[h]
    \centering
    \caption{PSO module and their configurable hyperparameter for each topologies for d=2 \& 5.}\label{tab2}
    \begin{tabular}{l  c c}
        \hline
        \textbf{Hyperparameter} & \textbf{Shorthand} & \textbf{Domain} \\
        \hline
        Cognitive coefficient & $c_1$ & \{0.3, 0.5, 0.7, 0.9\} \\
        Social coefficient & $c_2$ & \{0.2, 0.4, 0.6, 0.7\} \\
        Inertia weight & $w$ & \{0.9, 0.5, 0.7\} \\
        Number of particles & $n$ & \{50, 100, 150\} \\
        Nearest $k$ neighbors & $k$ & \{1, 2, 3\} \\
        Minkowski $p$-norm & $p$ & \{1, 2\} \\
        Delannoy numbers & $r$ & \{1, 2\} \\
        \hline
    \end{tabular}
\end{table}

\begin{equation}\label{eq3}
    AOCC(\bar y) = \frac{1}{B} 
\sum_{i=1}^{B}
 \left[1-  {(\min(\max((y_i),lb),ub)-lb)}/{(ub-lb)}\right] 
\end{equation} 

The AOCC metric is derived using the series of best found function values ($\bar y$) for $B$ (evaluation budget) with the lower bound $lb$ and upper bound $ub$ of the function value. The measure is inspired by the empirical cumulative distribution function (ECDF) describing probability distribution of optimization outcomes, \cite{lopez2024using} which serves as a building block for this performance measure. In particular, AOCC is the area under the curve of the ECDF over an infinite number of target values between the minimum and the maximum.

All function values were logarithmically transformed before being calculated with the AOCC, keeping values between the range $[-5, 5]$ for $2d$ and $5d$ problems to follow to standard BBOB benchmarking procedures. The experimental setup consisted of two computing platforms: two 2d evaluations were carried out on an Intel i7 with 32GB RAM and 8 cores, while 5d experiments were implemented on a more powerful Intel Xeon W-1270P workstation with 96GB RAM and 8 cores, to account for the higher computational requirements of higher-dimensional optimization.

\section{Hyperparameter Impact Analysis}\label{sec:5}
We examined $1,728$ PSO configurations across  twenty four BBOB functions, focusing on the impact of different communication topologies and hyperparameters on performance. Using SHAP-based plots for 2d \& 5d (Table~\ref{tab2}), we analyse the contribution parameters such as n\_particles, $c_1$, $c_2$, $w$, $p$, and $k$. The parameters of the PSO algorithm were tuned by using the tested and recommended values reported in the literature and the recommended range. Based on the analysis of \cite{985692} that shows ($c_1 +c_2 <4$) is a criterion that most commonly satisfies for the stability of a particle velocity, the cognitive ($c_1$) and social ($c_2$) coefficients were chosen from the intervals that satisfy this criterion for effective convergence and particle velocity without diverging. The inertia weight ($w$) as \{0.9, 0.5, 0.7\} adopted the standard set of inertia weight values in \cite{shi1998modified} where a high value ($w$ = 0.9) keeps the momentum to continue exploring the solution space globally and lower values ($w$ = 0.5, 0.7) allow for local exploitation. The swarm sizes ($n$) of \{50, 100, 150\} were chosen, since they are representative of the range of sizes examined in the literature as investigated in \cite{blackwell2007particle}. Finally, the parameters for neighborhood topology ($k,p, r$) were limited to low integer values to model fundamental local interaction structures—such as the Ring and Von Neumann topologies identified as high-performing by \cite{kennedy2002population}—without inducing excessive computational complexity. Respective SHAP values show the influence of each parameter on the AOCC, and are depicted as violet or yellow dots. The difference in the two colours is clearly defined for all topologies and parameters, e.g. n\_particles, $c_1$, $c_2$, $p$, and, $k$, etc. and others. In particular, positive SHAP values are shown as yellow dots and negative SHAP values are shown as violet dots.

\subsection{Experimental findings and Discussion for d=2}
\begin{figure*}[h]
    \centering
    \begin{minipage}{0.40\textwidth}
        \includegraphics[width=\linewidth]{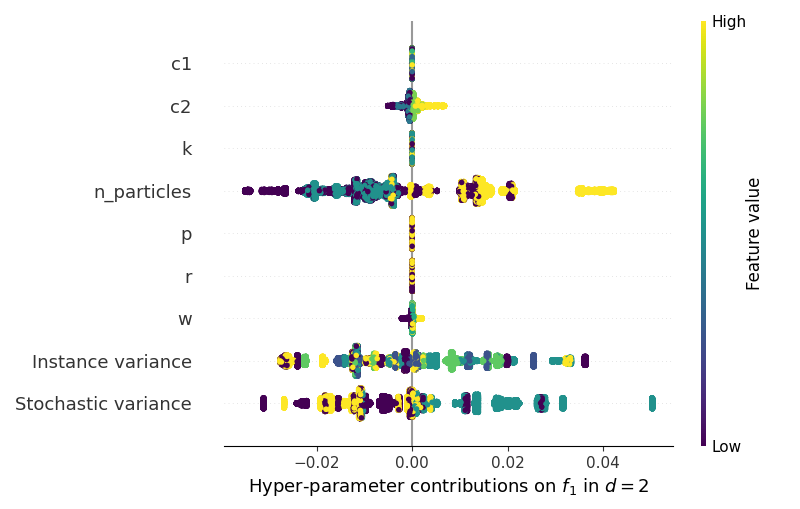}
    \end{minipage}
    \hfill
    \begin{minipage}{0.29\textwidth}
        \includegraphics[width=\linewidth]{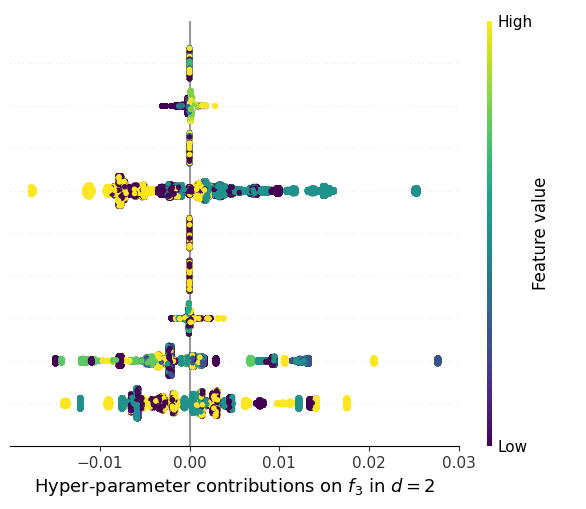}
    \end{minipage}
    \hfill
    \begin{minipage}{0.29\textwidth}
        \includegraphics[width=\linewidth]{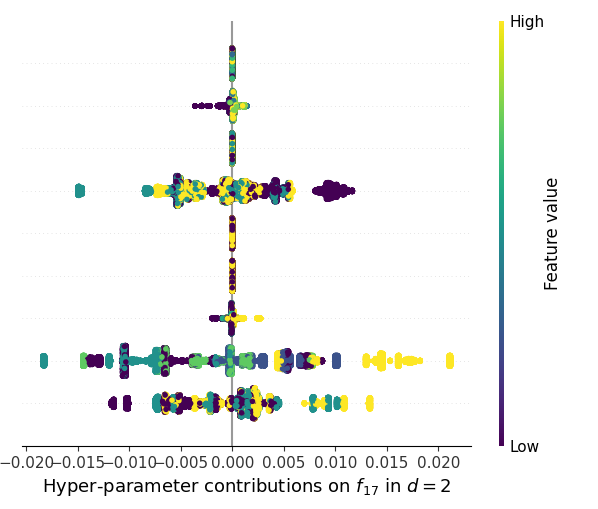}
    \end{minipage}
    \caption{The Impact of Hyperparameters on Performance for Benchmark Function in a 2d PSO with Star Topology} \label{fig2S}
\end{figure*}

\begin{figure*}[h]
    \centering
    \begin{minipage}{0.40\textwidth}
        \includegraphics[width=\linewidth]{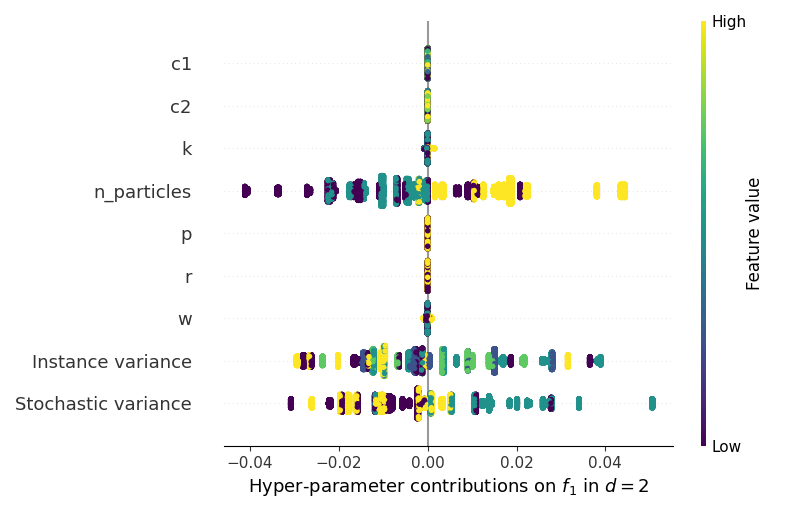}
    \end{minipage}
    \hfill
    \begin{minipage}{0.29\textwidth}
        \includegraphics[width=\linewidth]{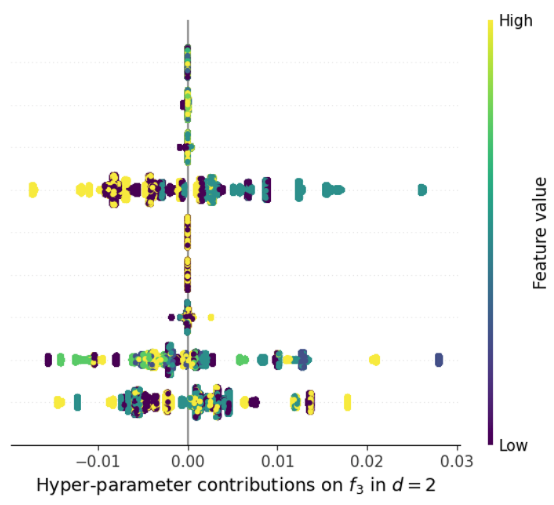}
    \end{minipage}
    \hfill
    \begin{minipage}{0.29\textwidth}
        \includegraphics[width=\linewidth]{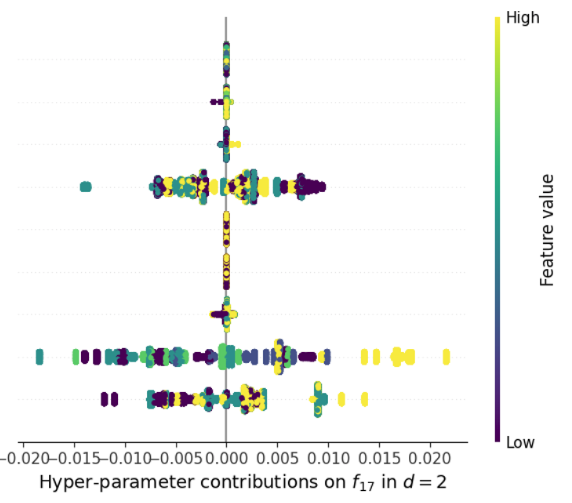}
    \end{minipage}
    \caption{The Impact of Hyperparameters on Performance for Benchmark Function in a 2d PSO with Ring Topology} \label{fig2R}
\end{figure*}

\begin{figure*}[h]
    \centering
    \begin{minipage}{0.40\textwidth}
        \includegraphics[width=\linewidth]{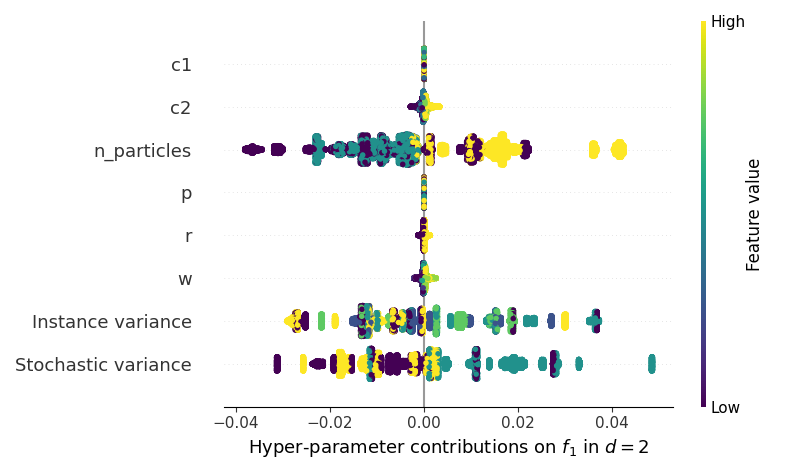}
    \end{minipage}
    \hfill 
    \begin{minipage}{0.29\textwidth}
        \includegraphics[width=\linewidth]{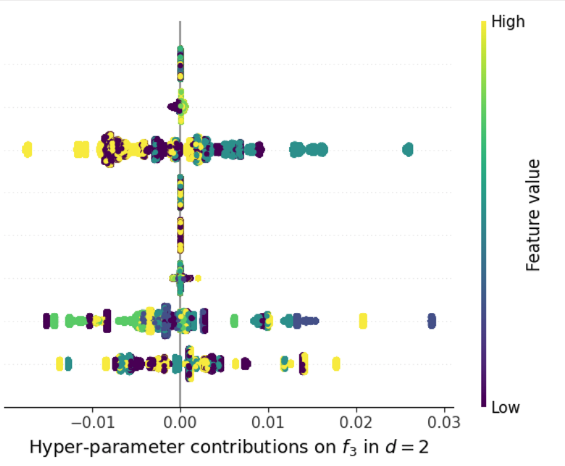}
    \end{minipage}
    \hfill
    \begin{minipage}{0.29\textwidth}
        \includegraphics[width=\linewidth]{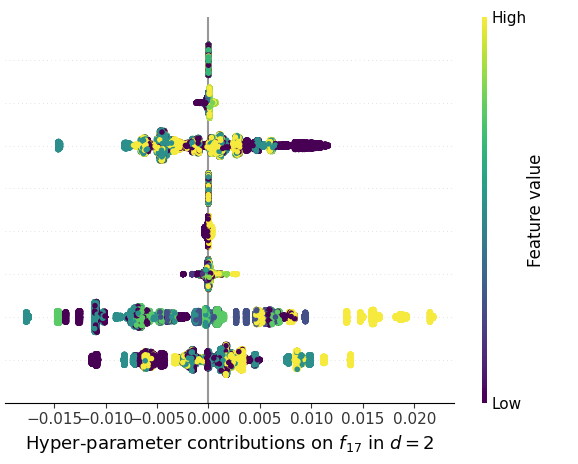}
    \end{minipage}
  \caption{The Impact of Hyperparameters on Performance for Benchmark Function in a 2d PSO with Von Neumann Topology} \label{fig2V}
\end{figure*}

For each of the three architectures: Star, Ring, and Von Neumann, the distinct topological impact is shown in Figures \ref{fig2S}-\ref{fig2V}, on functions $f_1$, $f_3$, and $f_{17}$, respectively. When we observe that the impact of hyperparameter function $f_1$ (uni-model) changes depend on all the three considered topology in our study i.e., Star, Ring, or Von Neumann. In the star configuration, the role of social acceleration coefficient ($c_2$) and  the inertia weight ($w$) is significant. When $c_2$ values are higher, we see better performance because of a strong centralized effect. The influences of $c_1, c_2$, and $w$ are less pronounced for the Ring topology, associated with slower information flow rates, with no single parameter dominating. In the Von Neumann arrangement, we can observe a wider variety of influences, especially from $c_2$ and $w$, which indicates that this grid-like structure helps with more adaptable convergence. Overall, the differences we see due to instance variance and randomness are quite small, meaning the effects of the hyper-parameters are consistent across various problem situations and are resilient against randomness in the PSO process. This points to a stable and trustworthy optimization method.


The function $f_3$ (multi-modal) performs well and consistently with all three topologies and only a slight sensitivity to the tuning of the hyperparameters. The Star topology has almost all contributions of the parameters centered around zero, which indicates that the performance was relatively stable across the different configurations. The same applies for the Ring topology, leading to a flat distribution of contributions and a further confirmation of its robustness towards tuning. Also, the Von Neumann topology has low variability; no one parameter is much of a factor in changing results. Furthermore, in all topologies the variance is kept low in both instances and stochastic variance, and clearly showing that the performance of $f_3$ is stable and consistent, and that it will work well regardless of the randomness or different problem instances. It is a good reference for PSO tuning for 2d optimization.


In contrast, the highly multi-modal function $f_{17}$ is very sensitive to the choice of hyperparameters for all topologies. Strong influences on $c_1, c_2$, and $w$ suggest that cognitive and social learning factors are important to convergence in the Star topology. In the Ring topology, the contributions are moderately but perceptibly lower, pronounced by $c_2$ and $w$, indicating that constraints on information diffusion have only moderate effects on sensitivity. The Von Neumann topology falls somewhere in the middle where $c_2, w$, and n\_particles have clear impacts and which involves the effect of structure and the size of the swarm on the exploration. Although sensitive, the variances are relatively small in all instances (instance variance, stochastic variance) which implies that the effects of the hyperparameters are the same for various runs and instance types. Even for this more complex one, this guarantees reliable and reproducible tuning results.


\subsection{Experimental findings and Discussion for d=5}
\begin{figure*}[h!]
    \centering
    \begin{minipage}{0.40\textwidth}
        \includegraphics[width=\linewidth]{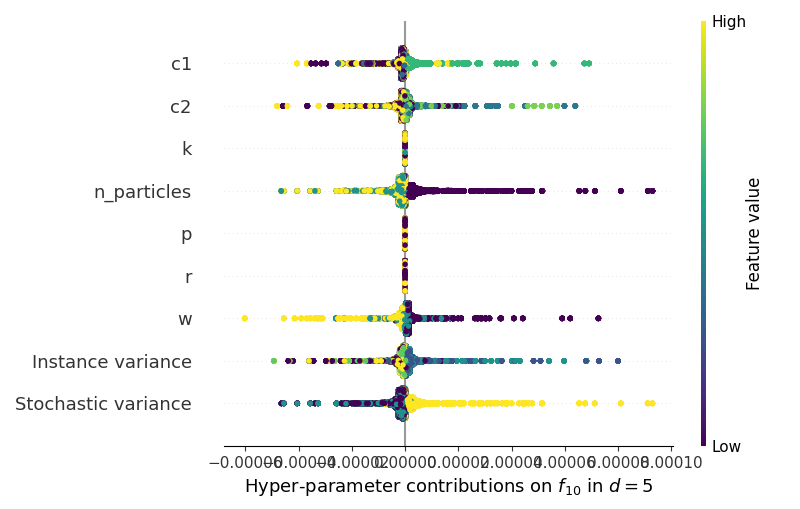}
    \end{minipage}
    \hfill 
    \begin{minipage}{0.29\textwidth}
        \includegraphics[width=\linewidth]{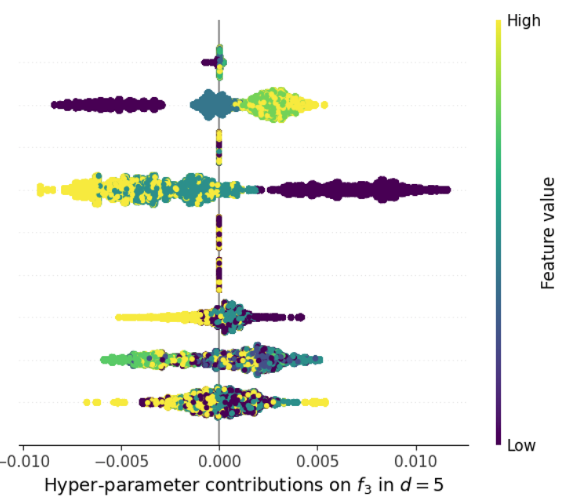}
    \end{minipage}
    \hfill
    \begin{minipage}{0.29\textwidth}
        \includegraphics[width=\linewidth]{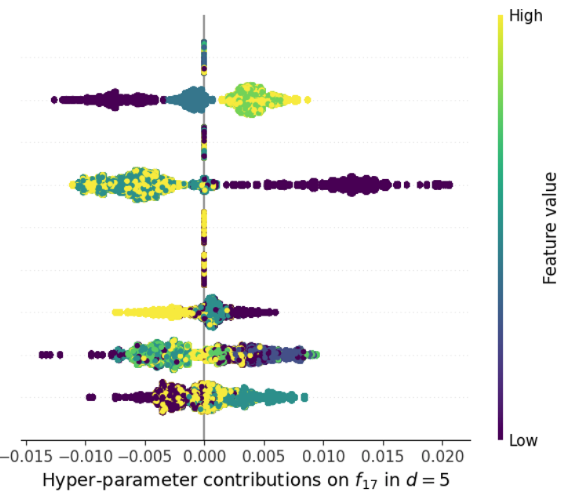}
    \end{minipage}
   \caption{The Impact of Hyperparameters on Performance for Benchmark Function in a 5d PSO with Star Topology} \label{fig5S}
\end{figure*}

\begin{figure*}[h!]
    \centering
    \begin{minipage}{0.40\textwidth}
        \includegraphics[width=\linewidth]{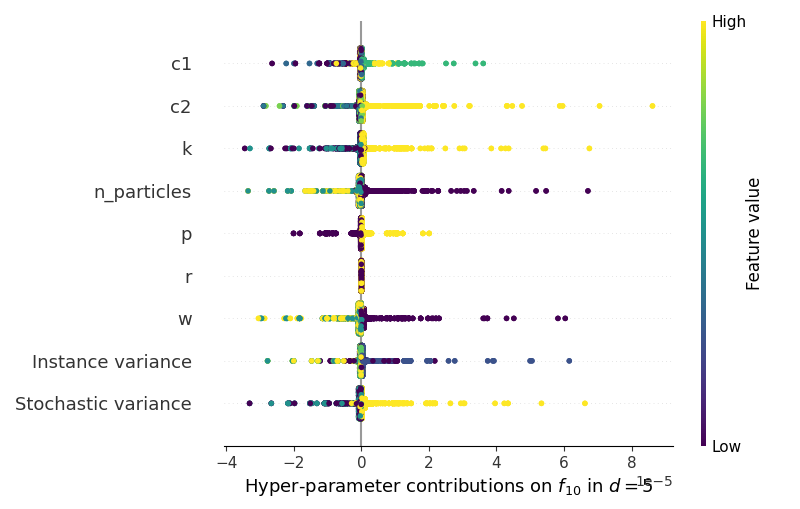}
    \end{minipage}
    \hfill 
    \begin{minipage}{0.29\textwidth}
        \includegraphics[width=\linewidth]{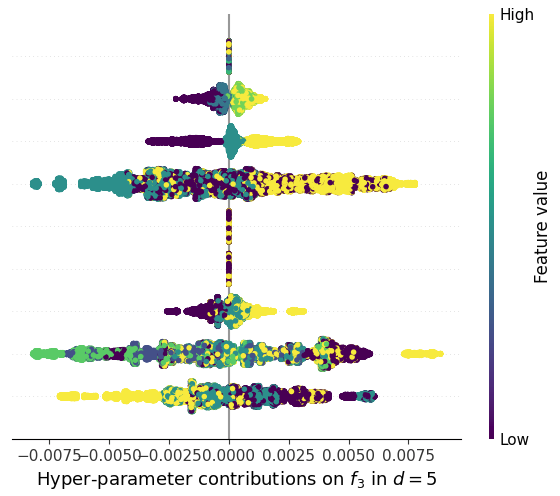}
    \end{minipage}
    \hfill
    \begin{minipage}{0.29\textwidth}
        \includegraphics[width=\linewidth]{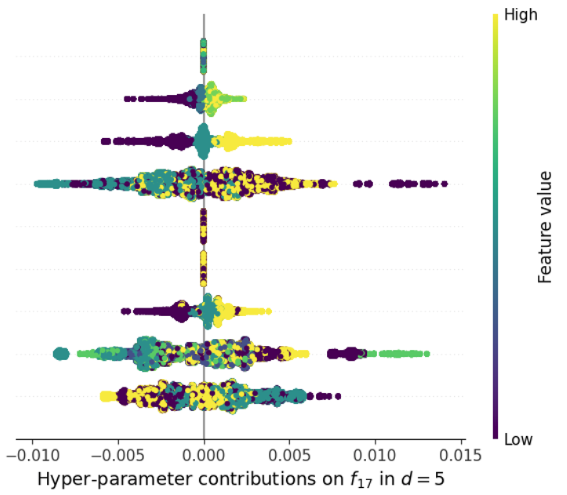}
    \end{minipage}
   \caption{The Impact of Hyperparameters on Performance for Benchmark Function in a 5d PSO with Ring Topology} \label{fig5R}
\end{figure*}

\begin{figure*}[h!]
    \centering
    \begin{minipage}{0.40\textwidth}
        \includegraphics[width=\linewidth]{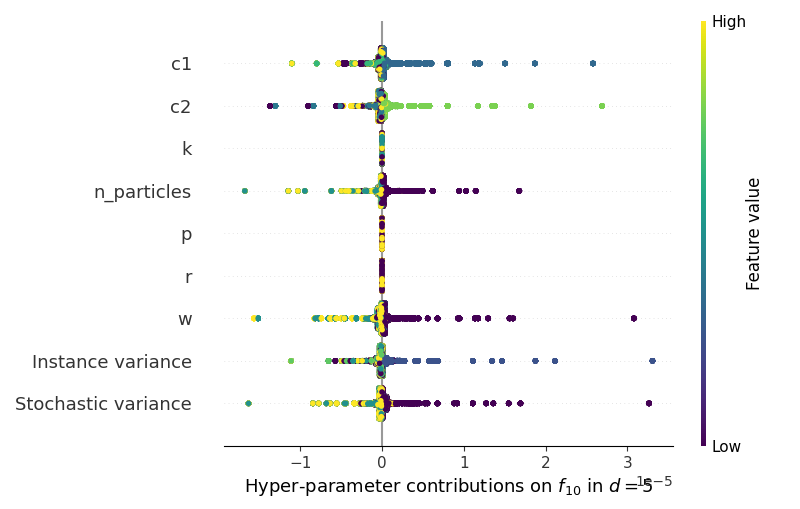}
    \end{minipage}
    \hfill
    \begin{minipage}{0.29\textwidth}
        \includegraphics[width=\linewidth]{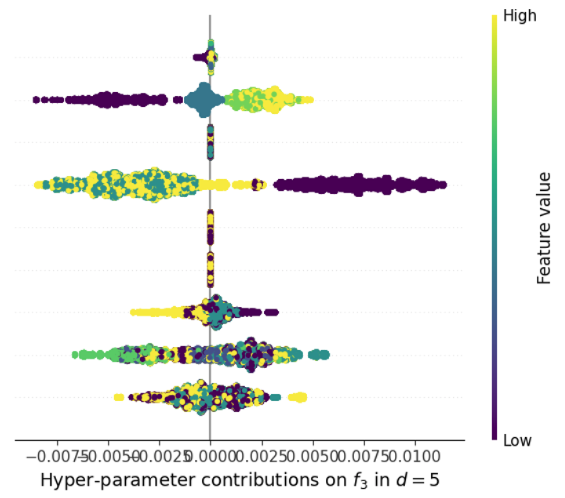}
    \end{minipage}
    \hfill
    \begin{minipage}{0.29\textwidth}
        \includegraphics[width=\linewidth]{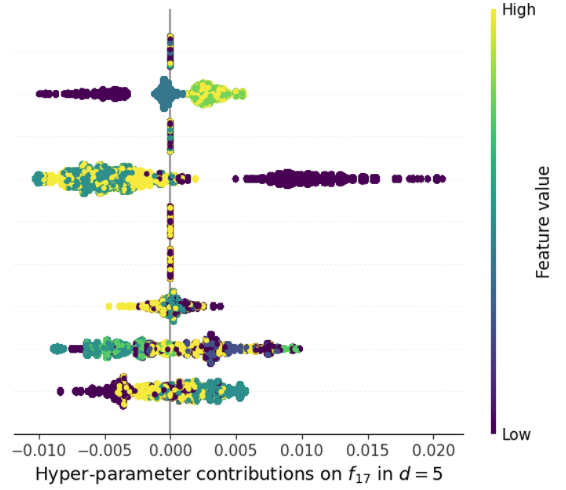}
    \end{minipage}
   \caption{The Impact of Hyperparameters on Performance for Benchmark Function in a 5d PSO with Von Neumann Topology} \label{fig5V}
\end{figure*}

Likewise, from figures \ref{fig5S} -\ref{fig5V},  the functions $f_{10}$, $f_3$, and $f_{17}$ present significant differences in the contributions of their hyperparameters in all three topologies with $c_2$ playing the most significant role. The contributions are relatively evenly split across the three nodes in the Star topology, though show more variability in $c_1, c_2,$ and $ k,$, indicating that there is greater sensitivity to exploration and neighborhood size. The Ring topology, however, takes this away. variance decreases towards $c_2$, with $k$ becomes a less important factor, indicating a greater dependence on local information. The Von Neumann topology also shows that $c_2$ and $r$ are important with a smaller spread and less stochastic variance, showing greater consistency in performance. $c_2$ and $w$ are still the primary drivers for all topologies, but the Von Neumann structure is the most robust followed by Star and then Ring.


Likewise, parameter affects of all topologies are more consistent and distinct for the functions $f_3$ and $f_{17}$. Star and Von Neumann have periodic contributions from $w$, $c_2$ and $c_1$ which will ensure convergence reliably. Ring shows slightly more dispersion but still maintains stable influence, reflecting the easier landscape where local communication suffices. The n\_particles parameter (number of particles in the swarm) shows negative contributions (violet in all SHAP plots) on all three topologies (Star, Ring, Von Neumann), showing that: As a swarm size grows, more data points will be needed to achieve the same optimization results. The remaining graphs and plots for each function have been added to the repository and supplemental file, giving a complete view of the behavior and outputs of each function. The supplementary file is available at \textcolor{blue}{\url{https://github.com/GitNitin02/ioh_pso/blob/main/Supplementary.pdf}}.

\subsection{Quantitative Analysis via STN Metrics}
To systematically analyze and compare the search behavior of PSO with different communication topologies, we employ a set of quantitative metrics derived from the constructed Search Trajectory Networks (STNs). These metrics capture distinct aspects of the search dynamics, ranging from the structural complexity of the trajectories to the efficiency of exploration and the degree of convergence.  The basic STNcomponents–nodes, edges, and connected components–as outlined by Ochoa et al. \cite{ochoa2021search} are used for the calculation of the metrics. The basic elements of STNs are nodes, edges, and connected components. The number of nodes visited by the swarm ($|N|$) gives an idea of the extent of the exploration performed, high numbers favour and low numbers favour the swarm converging quickly. The number of directed transitions between successive states is noted by the edges ($|E|$), which represents the overall search activity. A sub-network is disconnected from the rest of the STN if it has no nodes in common with the other sub-networks, and the number of such sub-networks is called the number of connected components ($C$). A single component ($C$=1) means a continuous, connected multiple components ($C$ > 1) show that the search trajectory fragments into distinct episodes of exploration, suggesting that the particles have spread to other parts of the search space without any re-connections. We present three derived measures beyond mere counts that extend explainability.

\begin{figure}[h]
    \centering
    \begin{minipage}{0.30\textwidth}
     \includegraphics[width=\linewidth]{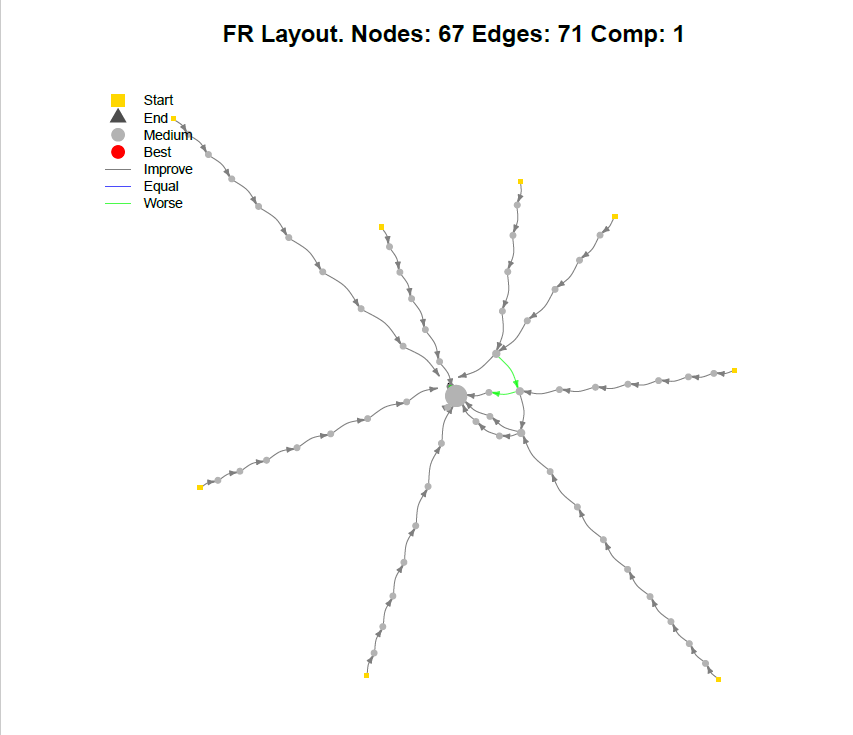}
       \subcaption{$f_1$ (unimodal)}
    \end{minipage}
    \hfill
    \begin{minipage}{0.30\textwidth}
        \includegraphics[width=\linewidth]{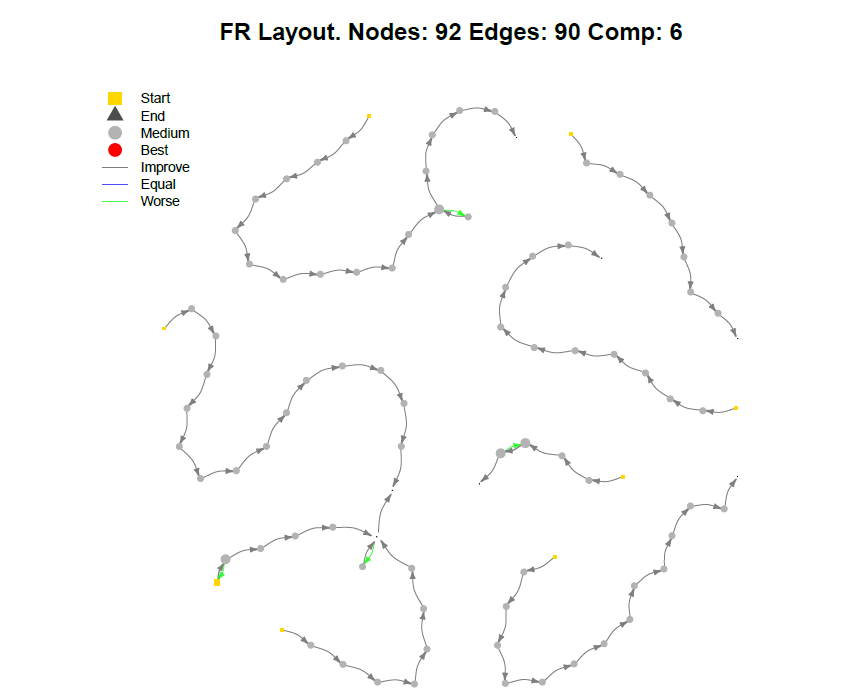}
        \subcaption{ $f_{3}$ (multimodal) }
    \end{minipage}
     \hfill
    \begin{minipage}{0.30\textwidth}
        \includegraphics[width=\linewidth]{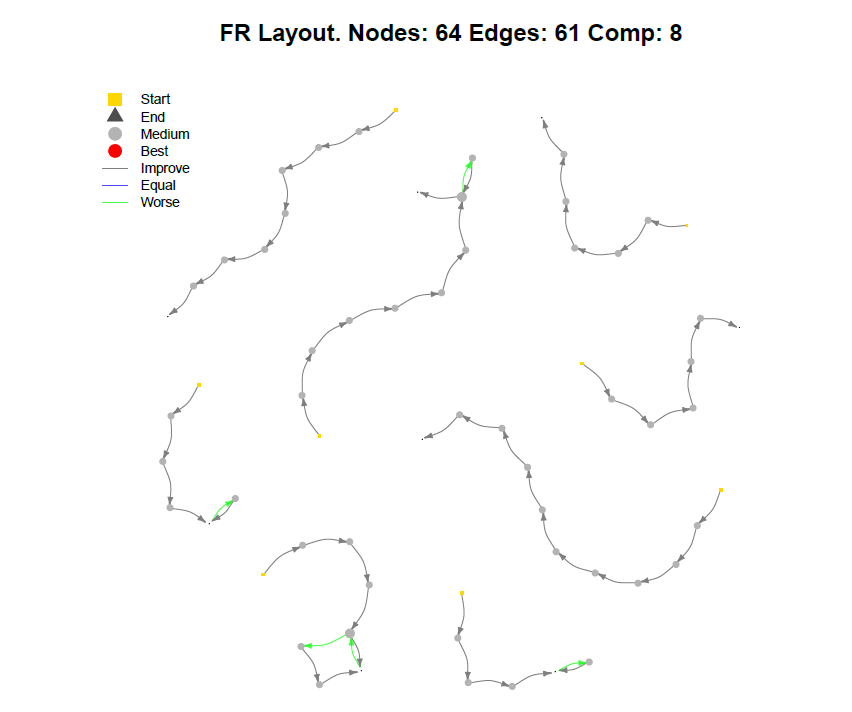}
        \subcaption{ $f_{17}$ (highly multimodal) }
    \end{minipage}
    \hfill
     \caption{FR layout for Benchmark Functions using config ($C_1$=0.5, $C_2$=0.7, $w$=0.9, $n$=50) in a 5 d star topology  } \label{fig: FR:star}
\end{figure}
\begin{figure}[h]
    \centering
    \begin{minipage}{0.30\textwidth}
     \includegraphics[width=\linewidth]{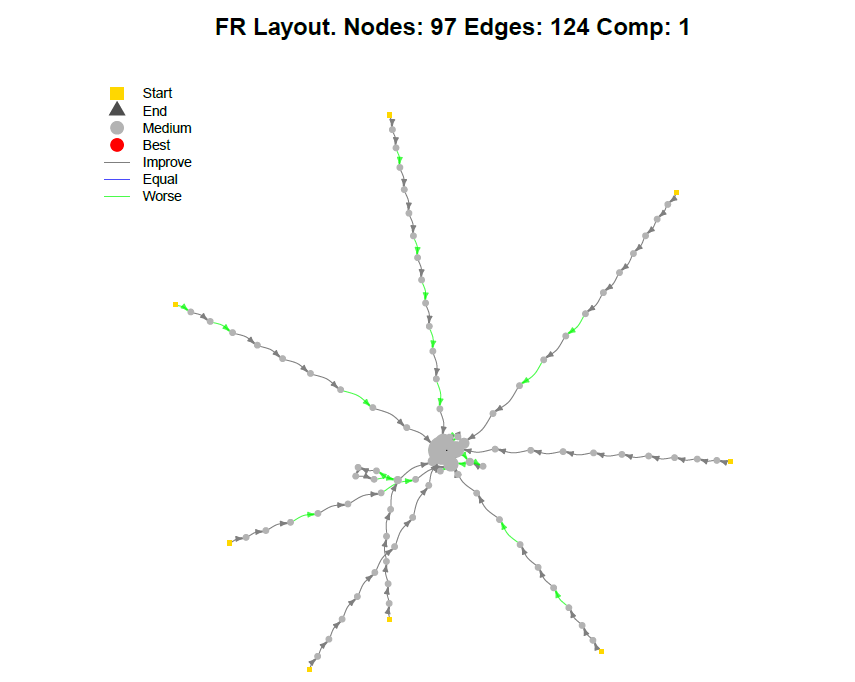}
       \subcaption{$f_1$ (unimodal)}
    \end{minipage}
    \hfill
    \begin{minipage}{0.30\textwidth}
        \includegraphics[width=\linewidth]{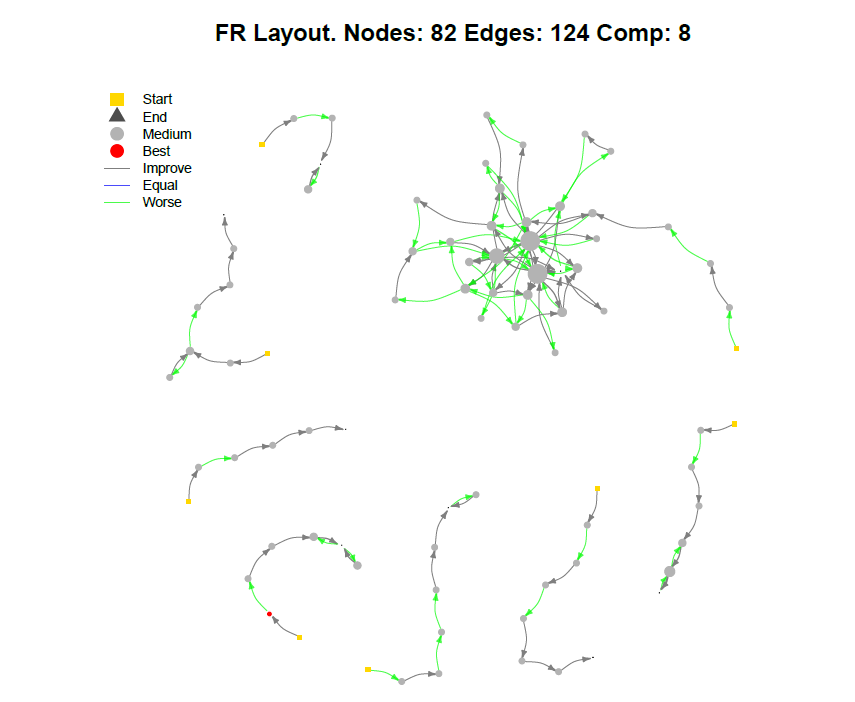}
        \subcaption{$f_{3}$ (multimodal) }
    \end{minipage}
     \hfill
    \begin{minipage}{0.30\textwidth}
        \includegraphics[width=\linewidth]{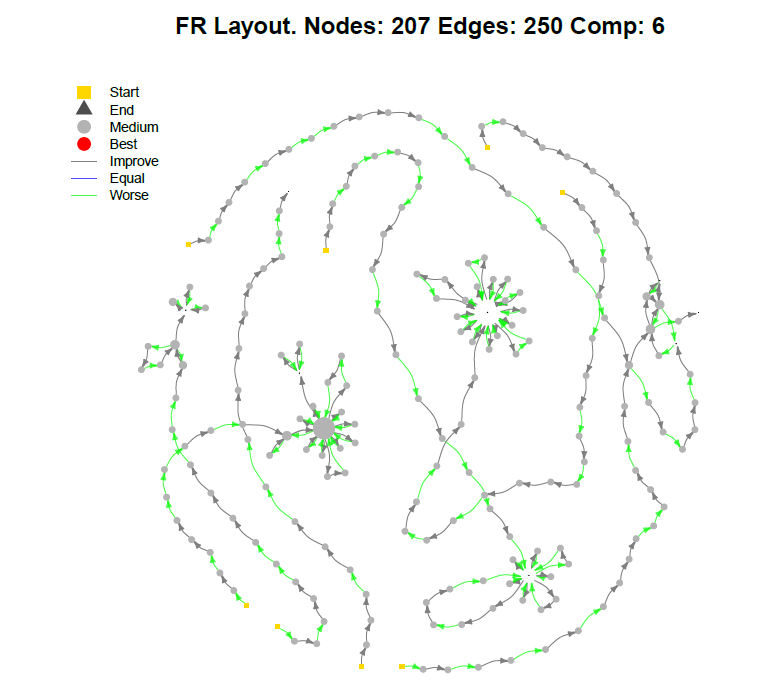}
        \subcaption{$f_{17}$ (highly multimodal) }
    \end{minipage}
    \hfill
     \caption{FR layout for Benchmark Functions using config ($C_1$=0.5, $C_2$=0.7, $w$=0.9, $n$=50, $k$=1, and $p$=1) in a 5 d Ring topology  } \label{fig: FR:ring}
\end{figure}
\begin{figure}[h]
    \centering
    \begin{minipage}{0.30\textwidth}
     \includegraphics[width=\linewidth]{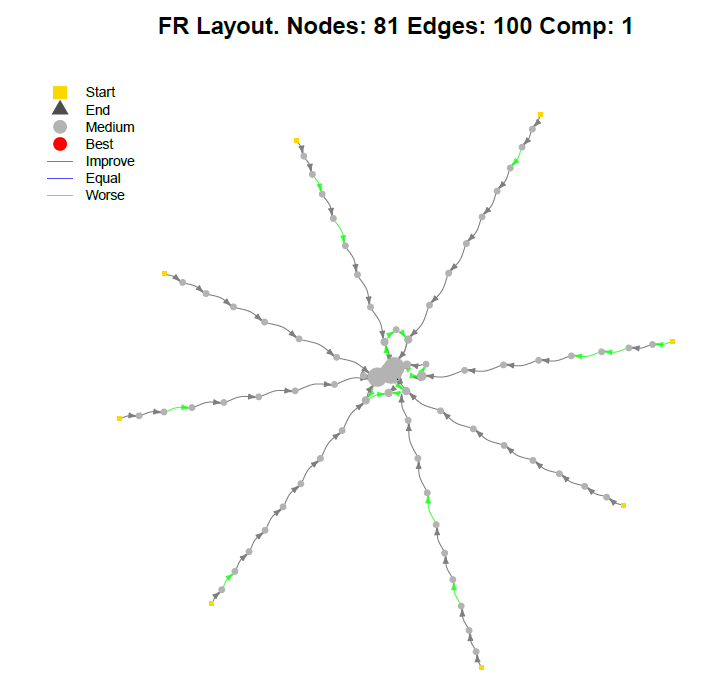}
       \subcaption{$f_1$ (unimodal)}
    \end{minipage}
    \hfill
    \begin{minipage}{0.30\textwidth}
        \includegraphics[width=\linewidth]{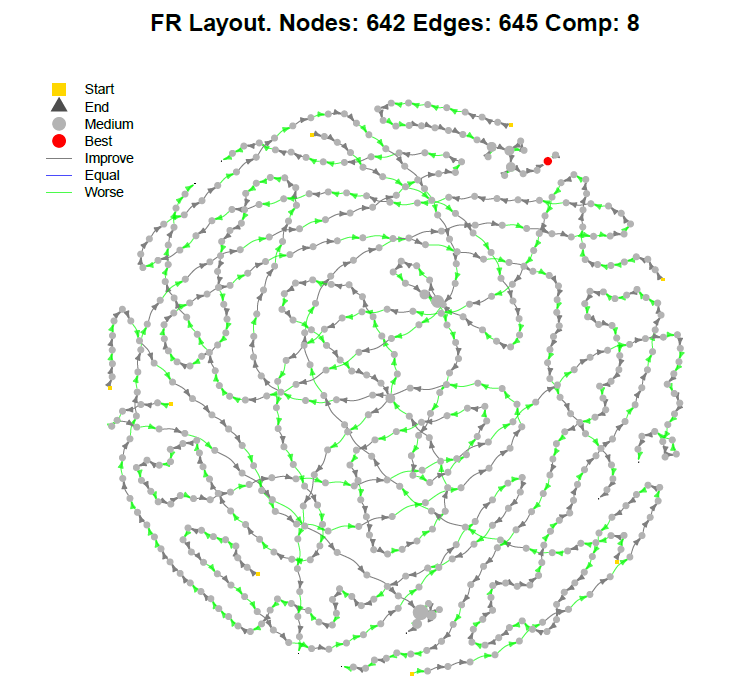}
        \subcaption{$f_{3}$ (multimodal) }
    \end{minipage}
     \hfill
    \begin{minipage}{0.30\textwidth}
        \includegraphics[width=\linewidth]{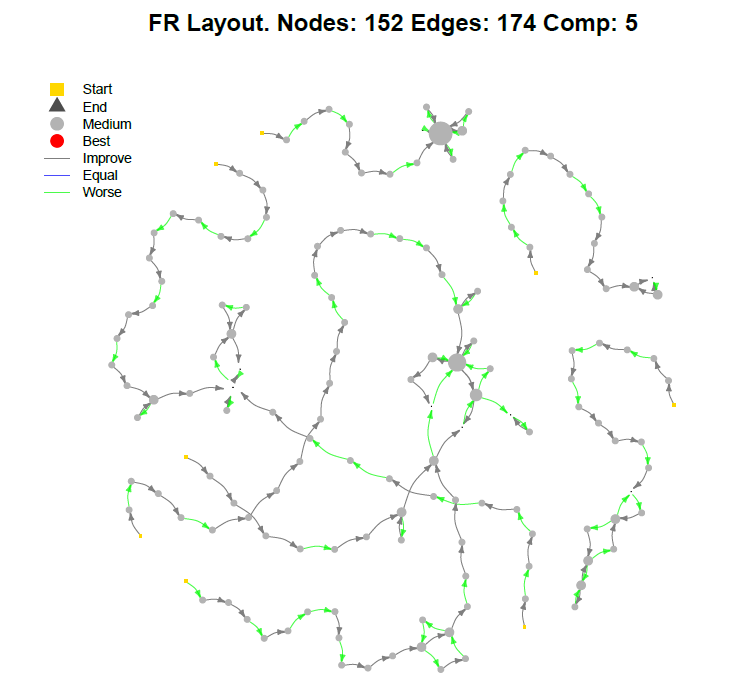}
        \subcaption{$f_{17}$ (highly multimodal) }
    \end{minipage}
    \hfill
     \caption{FR layout for Benchmark Functions using config ($C_1$=0.5, $C_2$=0.7, $w$=0.9, $n$=50, $p$=1, and $r$=2) in a 5 d VonNeumann topology  } \label{fig: FR:von}
\end{figure}

\begin{enumerate}[label=\roman*.]
    \item  \textbf{Connectivity Density ($\delta$):}  This measure gives a uniform measure of the network's connectivity/complexity, regardless of size. It is used in a similar way as density in the conventional graph theory. A higher value means that, on average, each state in the search trajectory leads to a higher value. Increased number of transitions, indicating increased complexity in search pattern, which may be revisiting or cyclic. This can be useful to help identify algorithms that produce linear and efficient paths versus meandering or repetitive paths.. Connectivity Density is defined as:
    \begin{equation} 
        \delta = {|E|} / {|N|}
    \end{equation}
    \item  \textbf{Fragmentation Score (F):} This metric combines information about both the number of nodes ($|N|$) and the number of disconnected sub-networks ($C$). This metric quantifies the total ``\textit{exploration effort}” expended across disconnected search episodes. A large fragmentation score means that the search process consists of numerous isolated search episodes. It can be very informative for the study of population based algorithms such as PSO whose particles end up at several different points (local optima) that form several disconnected components.
    \begin{equation}
        F= |N|* C
    \end{equation}
  Combines two correlated but separable parameters of search behaviour (breadth of exploration and degree of fragmentation) in a single overall figure for easier comparison of different algorithms or parameter settings.

    \item \textbf{Search Efficiency ($ \eta $):} This indicator is based on the principle of quantifying the “cost of exploration”. The number of states ($|N|$) or search episodes ($C$) found by the algorithm per edge (or transition) is quantified in the network. The smaller the value of $ \eta $ the more efficient the algorithm will be and so the fewer transitions it will take to get to a certain level of exploration. Search Efficiency is the ratio of Fragmentation Score divided by number of edges:
    
    \begin{equation}
        \eta = (|N|* C) / |E|
    \end{equation}
 
\end{enumerate}
Table \ref{Tab: Stn} presents the complete set of metrics-including nodes ($N$), edges ($E$), connected components ($C$), connectivity density ($\delta$), fragmentation score (F), and search efficiency ($\eta$)-for the Star, Ring, and Von Neumann topologies across all 24 BBOB benchmark functions. To complement these quantitative measures, Figures [\ref{fig: FR:star}-\ref{fig: FR:von}] visualize the STN structures using the FR layout with color-coded node classifications.  Table \ref{Tab:Stn_Summary} provides a summary of the range, average and relevant interpretable behavior insights that can be extracted from each topology and Table \ref{Tab:cross} makes it easy to compare the behavioral signatures side by side. The analysis of the combined networks shows three different patterns: (1) The Star topology, has a better convergence performance on simple functions but gets heavily fragmented on complex multimodal functions; (2) The Ring topology, performs with an exactly fixed number of 8 connected components for all functions, without reaching any consensus, and does not show a higher performance in terms of number of nodes in the connected component with the highest density of nodes, but rather in terms of the largest number of nodes in the connected component with the lowest density of nodes; (3) The Von Neumann topology generalizes by having the lowest density of nodes in the largest connected component, which indicates near linear exploration, but has the highest number of nodes in the smallest connected component, and is the worst performing in terms of search efficiency, even though it has a good performance in terms of number of connected components. The results show that the effect of communication topology on the fundamental understanding of trade-off between exploration and exploitation in PSO can be comprehended using the STN framework in a powerful way that is also easily interpreted.


\begin{table*}[t]
\centering
\caption{Summary of STN metrics—namely, nodes, edges, compositional structure, connectivity density, fragmentation score, and search efficiency—for PSO with Star ($C_1=0.5$, $C_2=0.7$, $w=0.9$, $n=50$), Ring ($C_1=0.5$, $C_2=0.7$, $w=0.9$, $n=50$, $k=1$, $p=1$), and Von Neumann ($C_1=0.5$, $C_2=0.7$, $w=0.9$, $n=50$, $p=1$, $r=2$) topologies across the BBOB benchmark functions}
\label{Tab: Stn}
\resizebox{\textwidth}{!}{%
\begin{tabular}{c|cccccc|cccccc|cccccc}
\hline
\multirow{2}{*}{\textbf{Function}} & \multicolumn{6}{c|}{\textbf{Star}} & \multicolumn{6}{c}{\textbf{Ring}} & \multicolumn{6}{c}{\textbf{VonNeumann}} \\
\cline{2-19}
 & $N$ & $E$ & $C$ & $\delta$ & F & $\eta$ & $N$ & $E$ & $C$ & $\delta$ & F & $\eta$ & $N$ & $E$ & $C$ & $\delta$ & F & $\eta$ \\ \hline
$f_1$  & 65 & 78 & 1 & 1.200 & 65 & 0.833 & 92 & 118 & 8 & 1.283 & 736 & 6.237 & 71 & 90 & 1 & 1.268 & 71 & 0.789 \\
$f_2$  & 203 & 225 & 2 & 1.108 & 406 & 1.804 & 79 & 81 & 8 & 1.025 & 632 & 7.802 & 378 & 437 & 1 & 1.156 & 378 & 0.865 \\
$f_3$  & 168 & 217 & 5 & 1.292 & 840 & 3.871 & 68 & 79 & 8 & 1.162 & 544 & 6.886 & 715 & 714 & 7 & 0.999 & 5005 & 7.010 \\
$f_4$  & 116 & 117 & 6 & 1.009 & 696 & 5.949 & 125 & 180 & 8 & 1.440 & 1000 & 5.556 & 155 & 167 & 2 & 1.077 & 310 & 1.856 \\
$f_5$  & 780 & 792 & 8 & 1.015 & 6240 & 7.879 & 404 & 687 & 8 & 1.701 & 3232 & 4.705 & 800 & 792 & 8 & 0.990 & 6400 & 8.081 \\
$f_6$  & 201 & 237 & 5 & 1.179 & 1005 & 4.241 & 94 & 145 & 8 & 1.543 & 752 & 5.186 & 270 & 285 & 3 & 1.056 & 810 & 2.842 \\
$f_7$  & 97 & 110 & 7 & 1.134 & 679 & 6.173 & 70 & 83 & 8 & 1.186 & 560 & 6.747 & 137 & 157 & 6 & 1.146 & 822 & 5.236 \\
$f_8$  & 133 & 153 & 3 & 1.150 & 399 & 2.608 & 68 & 65 & 8 & 0.956 & 544 & 8.369 & 188 & 222 & 2 & 1.181 & 376 & 1.694 \\
$f_9$  & 124 & 153 & 3 & 1.234 & 372 & 2.431 & 83 & 120 & 8 & 1.446 & 664 & 5.533 & 135 & 179 & 1 & 1.326 & 135 & 0.754 \\
$f_{10}$ & 162 & 210 & 8 & 1.296 & 1296 & 6.171 & 98 & 129 & 8 & 1.316 & 784 & 6.078 & 571 & 573 & 8 & 1.004 & 4568 & 7.972 \\
$f_{11}$ & 220 & 240 & 8 & 1.091 & 1760 & 7.333 & 81 & 100 & 8 & 1.235 & 648 & 6.480 & 800 & 792 & 8 & 0.990 & 6400 & 8.081 \\
$f_{12}$ & 237 & 355 & 8 & 1.498 & 1896 & 5.341 & 181 & 284 & 8 & 1.569 & 1448 & 5.099 & 475 & 473 & 8 & 0.996 & 3800 & 8.034 \\
$f_{13}$ & 152 & 199 & 8 & 1.309 & 1216 & 6.111 & 120 & 160 & 8 & 1.333 & 960 & 6.000 & 409 & 432 & 7 & 1.056 & 2863 & 6.627 \\
$f_{14}$ & 92 & 107 & 2 & 1.163 & 184 & 1.720 & 96 & 147 & 8 & 1.531 & 768 & 5.224 & 95 & 133 & 1 & 1.400 & 95 & 0.714 \\
$f_{15}$ & 410 & 502 & 8 & 1.224 & 3280 & 6.534 & 109 & 189 & 8 & 1.734 & 872 & 4.614 & 800 & 792 & 8 & 0.990 & 6400 & 8.081 \\
$f_{16}$ & 139 & 194 & 7 & 1.396 & 973 & 5.015 & 74 & 87 & 8 & 1.176 & 592 & 6.805 & 715 & 707 & 8 & 0.989 & 5720 & 8.091 \\
$f_{17}$ & 97 & 99 & 8 & 1.021 & 776 & 7.838 & 80 & 113 & 8 & 1.413 & 640 & 5.664 & 205 & 230 & 2 & 1.122 & 410 & 1.783 \\
$f_{18}$ & 128 & 173 & 8 & 1.352 & 1024 & 5.919 & 101 & 155 & 8 & 1.535 & 808 & 5.213 & 447 & 469 & 2 & 1.049 & 894 & 1.906 \\
$f_{19}$ & 107 & 111 & 8 & 1.037 & 856 & 7.712 & 97 & 120 & 8 & 1.237 & 776 & 6.467 & 704 & 713 & 8 & 1.013 & 5632 & 7.899 \\
$f_{20}$ & 98 & 101 & 8 & 1.031 & 784 & 7.762 & 74 & 74 & 8 & 1.000 & 592 & 8.000 & 136 & 143 & 6 & 1.051 & 816 & 5.706 \\
$f_{21}$ & 249 & 341 & 6 & 1.369 & 1494 & 4.381 & 99 & 137 & 8 & 1.384 & 792 & 5.781 & 132 & 139 & 5 & 1.053 & 660 & 4.748 \\
$f_{22}$ & 195 & 244 & 4 & 1.251 & 780 & 3.197 & 128 & 214 & 8 & 1.672 & 1024 & 4.785 & 225 & 255 & 2 & 1.133 & 450 & 1.765 \\
$f_{23}$ & 202 & 225 & 8 & 1.114 & 1616 & 7.182 & 67 & 75 & 8 & 1.119 & 536 & 7.147 & 800 & 792 & 8 & 0.990 & 6400 & 8.081 \\
$f_{24}$ & 96 & 106 & 8 & 1.104 & 768 & 7.245 & 73 & 80 & 8 & 1.096 & 584 & 7.300 & 742 & 737 & 8 & 0.993 & 5936 & 8.054 \\
\hline
\end{tabular}}
\end{table*}

\begin{table*}[t]
\centering
\caption{Summary of STN metrics and derived features for PSO with Star, Ring, and Von Neumann topologies across the BBOB benchmark functions}
\label{Tab:Stn_Summary}
\resizebox{\textwidth}{!}{%
\begin{tabular}{l|cccccc}
\hline
\textbf{Topology} & \textbf{Metric} & \textbf{Range} & \textbf{Average} & \textbf{Interpretability Insight} \\
\hline
\multirow{6}{*}{\textbf{Star}} 
 & Nodes & 65 -- 780 & $\sim$184 & High variability; scales with problem complexity \\
 & Edges & 78 -- 792 & $\sim$228 & Moderate connectivity \\
 & Components & 1 -- 8 & $\sim$5.9 & Fragmentation increases with problem difficulty \\
 & Connectivity Density ($\delta$) & 1.009 -- 1.498 & $\sim$1.19 & Moderate; indicates some revisiting behavior \\
 & Fragmentation Score ($F$) & 65 -- 6,240 & $\sim$1,095 & Very high on complex functions ($f_5$, $f_{15}$) \\
 & Search Efficiency ($\eta$) & 0.833 -- 7.879 & $\sim$4.59 & Efficient on simple ($f_1$: 0.833); inefficient on complex ($f_5$: 7.879) \\
\hline
\multirow{6}{*}{\textbf{Ring}} 
 & Nodes & 67 -- 404 & $\sim$120 & Consistently lower than Star; limited exploration \\
 & Edges & 65 -- 687 & $\sim$166 & Moderate but with high-density cases \\
 & Components & 8 (always) & 8 & Never converges; always 8 disconnected components \\
 & Connectivity Density ($\delta$) & 0.956 -- 1.734 & $\sim$1.36 & Highest average; most revisiting/cyclic behavior \\
 & Fragmentation Score ($F$) & 536 -- 3,232 & $\sim$960 & Consistently high ($8 \times N$) due to fixed components \\
 & Search Efficiency ($\eta$) & 4.614 -- 8.369 & $\sim$6.12 & Consistently inefficient; never below 4.6 \\
\hline
\multirow{6}{*}{\textbf{Von Neumann}} 
 & Nodes & 71 -- 800 & $\sim$392 & Highest overall; often maxes out at 800 \\
 & Edges & 90 -- 792 & $\sim$405 & Scales with nodes; near-linear for complex functions \\
 & Components & 1 -- 8 & $\sim$4.8 & Mixed; good convergence on simple functions \\
 & Connectivity Density ($\delta$) & 0.989 -- 1.400 & $\sim$1.07 & Lowest average; near-linear trajectories \\
 & Fragmentation Score ($F$) & 71 -- 6,400 & $\sim$1,881 & Highest overall; extreme on complex functions \\
 & Search Efficiency ($\eta$) & 0.714 -- 8.091 & $\sim$4.52 & Very efficient on simple ($\eta<1$); poor on complex ($\eta>7$) \\
\hline
\end{tabular}}
\end{table*}


\begin{table*}[t]
\centering
\caption{Comparative behavioral signatures of PSO communication topologies based on STN metrics}
\label{Tab:cross}
\resizebox{\textwidth}{!}{%
\begin{tabular}{l|ccc}
\hline
\textbf{Aspect} & \textbf{Star} & \textbf{Ring} & \textbf{Von Neumann} \\ \hline
Convergence on Simple Functions & \ding{51}\ Excellent ($C=1$, $\eta<2$) &  Moderate converges ($C=8$) & \ding{51}\ Excellent ($C=1$, $\eta<1$) \\
Exploration Breadth & Moderate ($N \approx 184$) & Limited ($N \approx 120$) & Extensive ($N \approx 392$) \\
Fragmentation & Increases with complexity & Always high (8 components) & Increases with complexity \\
Connectivity Density & Moderate ($\delta \approx 1.19$) & Highest ($\delta \approx 1.36$) & Lowest ($\delta \approx 1.07$) \\
Search Efficiency & Variable (0.83--7.88) & Consistently poor (4.61--8.37) & Bimodal (0.71--8.09) \\
Landscape Sensitivity & High & Low & High \\
\hline
\end{tabular}}
\end{table*}

\subsection{PSO configuration selection}
This analysis distinguishes between specialized and general hyperparameter configurations using four key metrics. The ``Single Best Mean" (sbm) and ``Single Best Std" (sbs) quantify the performance of a best configuration optimized for a single function; the mean represents its average performance score, while the standard deviation captures the variability of that performance across different algorithm instances and random seeds. Conversely, the ``Avg-Best Mean" (abm) and ``Avg-Best Std" (abs) describe the best general-purpose configuration, which is selected for its robust performance across all benchmark functions. Here, the abm indicates its overall average performance, and the abs measures the consistency of that performance when applied to the diverse set of functions. In essence, the ``Single Best" metrics reflect specialized, peak performance on a specific function, whereas the ``Avg-Best" metrics reflect the reliability and generalizability of a configuration across a broader problem domain.

Table~\ref{Tab: d=2} compares the \textit{Star}, \textit{Ring}, and \textit{Von Neumann} topologies across all 24 BBOB functions at dimension $d=2$. For most functions, the three topologies produce nearly identical results. For example, on $f_2$, all three topologies achieve $Sbm \approx 0.976$--$0.977$ with $R^2_{\text{train}} \approx 0.98$, showing that topology structure barely affects performance on smoother, less deceptive functions. Likewise, at low dimension on $f_1$, Star, Ring and Von Neumann all report $Sbm \approx 0.750$ with similar values of $Sbs \approx 0.946$, indicating similar behavior in this case. For more multi-modal functions, however, the picture changes slightly: on $f_8$, the $Sbm$ is slightly larger in Ring ($\approx 0.858$) than in Star ($ \approx 0.854$) and on $f_{21}$, again, the $Sbm$ value is slightly larger in Ring ($\approx 0.769$ ) than in Star ($\approx 0.763$) and Von Neumann ($\approx 0.763$). The small improvements, usually ranging from $0.004$--$0.006$ in $Sbm$, are due to Ring's limited neighborhood structure which results in slower information dissemination within the swarm and helps to prevent over-early convergence on rugged terrain. The specialist-vs generalist trade-off is seen across almost all the functions, with $Abm$ consistently lower than $Sbm$, and $R^2_{\text{train}}$ values consistently high (often > $0.95$--$0.98$) for the majority of function dimensions, including $f_5$ ($R^2 \approx 0.996$ for Star), showing a strong and reliable model fit at this dimension. In general, Table ~\ref{Tab: d=2} reveals that there is little difference numerically among the choices of topology for $d=2$, and that Ring and Von Neumann improve a few thousandths of $Sbm$ on isolated multimodal functions.


Table~\ref{Tab: d=5} repeats the comparison at $d=5$, where the numerical gaps between topologies widen considerably. On $f_1$, Ring reaches $Sbm \approx 0.863$, clearly ahead of Star's $Sbm \approx 0.780$ and Von Neumann's $\approx 0.812$ --- a gap of roughly $0.05$--$0.08$, far larger than anything observed at $d=2$.  Again on $f_3$, Ring has the highest score, this time at $Sbm \approx 0.967$ while Star is at $\approx 0.950$ and Von Neumann is at $\approx 0.952$, thus highlighting the advantage of restricted neighborhood topologies—this advantage increases with the size of the search space. Furthermore, the specialist-to-generalist drop, as seen in the $Abm$ dropping to around $0.846$--$0.862$ even with the best topologies, is more evident here; for example, this drop that can be seen around $f_1$, is a much larger gap than what can typically be seen at $d=2$, where $Abm$ drops by sub-$0.01$ on the best topologies, demonstrating that a single generalist configuration becomes harder to optimize across the entire suite as dimensionality increases. The $R^2_{\text{train}}$ values also become far less consistent -- $f_2$ shows an extremely low $R^2 \approx 0.28$--$0.41$ across all three topologies despite near-perfect $Sbm \approx 0.997$--$0.998$, meaning that although absolute performance is excellent, the fitted model struggles to explain the variance, likely due to a flat or plateau-like fitness landscape.  Functions such as $f_{10}$ and $f_{13}$ show similarly depressed $R^2$ (in the $0.3$--$0.65$ range) despite strong Sbm/Abm scores.  Combined, the results here show that at $d=5$ the choice of topology is much more critical than it is at $d=2$ : Ring and Von Neumann both achieve several percentage point improvements in Sbm on some functions, while the reliability of the resulting performance model ($R^2_{\text{train}}$) is less stable throughout the benchmark suite.  The optimal parameter configurations summarized in Table \ref{T3} provide Single best Config for the specialized roles of different population topologies.



.

To summarise, PSO topology should be determined based on the problem complexity. The Star topology is used in simple, unimodal problem instances because it takes a short time to converge; the Ring topology is employed in rugged and multi-modal problem instances where diversity is a key consideration and is important. Balanced topology is best for deceptive or highly complex functions: Von Neumann topology. However, in applications with unknown characteristics of the problems, adaptive or hybrid approaches may be more suitable.

In the topology comparison in the PSO topology, $R^2$ (R-squared) \cite{miles2005r} can be used as an additional figure to assess the extent to which the performance of each topology is consistent with an ideal optimization path or expected convergence behaviour. For instance, in the case of unimodal functions, the convergence is fast and consistent, following a predictable trend, in the topology of the star, if this was the case, its $R^2$ would be very high, meaning the performance would closely follow the simplicity of the problem. In the multi-modal functions, however, poor diversity maintenance in the Star topology may result in a lower $R^2$ value, possibly because convergence occurs too early, while the Ring and Von Neumann topologies may show higher $R^2$ values due to their ability to maintain diversity more effectively, reflecting a more stable alignment with the problem complexity.

 \subsubsection{Time Complexity Analysis}
Differences in the topology of the swarm have a more pronounced effect on the computational efficiency of PSO's as the dimension of the problem is increased. In a 2-dimensional problem, our experiments show a distinct difference in execution times: star topology is the most efficient (7.2 hours), Von Neumann (6.56 hours) and Ring (9.35 hours) are less efficient than star topology. This trend continues and extends in the 5-dimensional problems, where Von Neumann is still 56.87 hours, Star's time is 61.17 hours and Ring's time is 75.52 hours. However, the performance difference is maintained across dimensions, may be because Von Neumann's communication pattern is structured (but decentralized), thus it seems to scale better because of the balanced information flow which does not cause bottlenecks. By contrast, updates to the Ring topology seem to get worse with the increase in the number of dimensions, as the time increase seems to be compounding.The former (Ring topology) have sequential neighbor based updates, while they seem to take a long time to increase with the number of dimensions. The Star topology falls somewhere between the Ring and Von Neumann topologies, in that it has a global topology of communication that is more scalable than Ring, but still more expensive than Von Neumann. The results confirm that the topology selection is increasingly important as the dimension of a problem to optimize increases, since the time difference increases significantly in higher-dimensional problems. The results given can be used to inform the design of algorithms, especially where a combination of solution quality and computing resources plays an important role. Also the computation required for this experiment is very high: even for the problem size of 20 dimensions, 20 times for 30 repetitions, takes about 5 hours. The experimental results have been sent to the given GitHub link for 20-dimensional problems. The need to parallelize evaluations and deal with the massive processing power stress is evident from this heavy load. Unfortunately, the computational cost for going to higher dimensions or a larger sample of test functions is currently prohibitive, and our experimental analysis is thus limited.

\begin{table*}[htp]\centering
 \caption{Performance of Star, Ring and Von Neumann Topologies over all BBOB functions on d=2.}
\label{Tab: d=2}
    \centering
   \scriptsize \begin{tabular}{ccccccccc}
    \hline
      \textbf{Function}  & \textbf{Topologies} &  \textbf{sbm}  & \textbf{sbs} & \textbf{abm} & \textbf{abs} & \textbf{all mean} & \textbf{all std}  & \textbf{R2 train}\\ \hline
      
        \multirow{3}{*}{$f_1$} & Star & 0.750 &	0.946&	0.777&	0.974&	0.766&	0.958& 0.976\\ 
        ~ & Ring & 0.750 & 0.946 & 0.777 & 0.974 & 0.768 & 0.957
          & \textbf{\color{red}{0.983}} \\ 
        ~ & Von Neumann & 0.750 & 0.946 & 0.777 & 0.974 &	0.767 &	0.958  &9.80
  \\ \hline

        \multirow{3}{*}{$f_2$} & Star & \textcolor{red}{0.977}&	0.977	&0.983&	0.981&	0.981&	0.978
 & 0.986
\\ 
        ~ & Ring &	0.976 &	0.973&	0.983&	0.981 &	0.982 &	0.978
 & 0.987\\ 
        ~ & Von Neumann & 0.976 & 0.973 & 0.983 & 0.981 & 0.982&	0.978 & \textbf{\color{red}{0.988}}
 \\ \hline

        \multirow{3}{*}{$f_3$} & Star &0.899&	0.972	&0.899&	0.972&	0.907&	0.979
 & 0.966
\\ 
        ~ & Ring &	0.899 &	0.972 &	0.899 &	0.972 &	0.907 &	0.979
        &\textbf{\color{red}{0.977}} 
   \\ 
        ~ & Von Neumann & 0.899 & 0.972 & 0.899 & 0.972 & 0.907 &	0.979 & 0.969
  \\ \hline

        \multirow{3}{*}{$f_4$} & Star	&	\textcolor{red}{0.906}	&0.970&	0.922&	0.985&	0.913&	0.978
& 0.987\\ 
        ~ & Ring &	0.905 &	0.969&	0.922&	0.985&	0.914& 0.977
 &\textbf{\color{red}{0.993}} 
  \\ 
        ~ & Von Neumann & 0.901 & 0.969 & 0.922 & 0.985 &	0.913&	0.978 & 0.983
  \\ \hline

        \multirow{3}{*}{$f_5$} & Star&	0.845&	0.973	&0.845&	0.973&	0.863&	0.977
  &\textbf{\color{red}{ 0.996}} 
\\ 
        ~ & Ring &	0.845 &	0.973& 0.845 &	0.973 &	0.862 &	0.977
 & 0.994
   \\ 
        ~ & Von Neumann & 0.845 &	0.973&	0.845&	0.973	&0.863&	0.977   & 0.990
  \\ \hline

        \multirow{3}{*}{$f_6$} & Star&	 0.859&	0.965&	0.862&	0.977&	0.867&	0.972
   & 0.979\\ 
        ~ & Ring &	\textcolor{red}{0.862} &	0.977 &	0.862 &	0.977 &	0.869 &	0.971
 &\textbf{\color{red}{ 0.994}} 
   \\ 
        ~ & Von Neumann & 0.856& 0.965 & 0.862 & 0.977 & 0.868 & 0.971 & 0.978
  \\ \hline

        \multirow{3}{*}{$f_7$} & Star & 0.808 &	0.968 &	0.822 &	0.957 &	0.816 &	0.962
 & 0.964
\\ 
        ~ & Ring &	0.809 &	0.954 &	0.822 &	0.957 &	0.817 &	0.962
 & 0.980
   \\ 
        ~ & Von Neumann & 0.809& 0.954 & 0.822 & 0.957 &	0.817 &	0.962 & \textbf{\color{red}{ 0.982}}
  \\ \hline

        \multirow{3}{*}{$f_8$} & Star&	0.854& 0.957	& 0.867& 0.942&	0.866& 0.952
  & 0.969
\\ 
        ~ & Ring &	\textcolor{red}{0.858} &	0.957 &	0.867 &	0.942 &	0.866 &	0.952
     &\textbf{\color{red}{ 0.984}} 
  \\ 
        ~ & Von Neumann & 0.856	&0.962 & 0.867 & 0.942 &	0.866&	0.952  & 0.977
  \\ \hline

        \multirow{3}{*}{$f_9$} & Star&	0.837& 0.962&	0.837& 0.962& 0.857& 0.953 & 0.954 \\ 
        ~ & Ring &	0.837 &	0.962 &	0.837 &	0.962 &	0.858 &	0.953
 &\textbf{\color{red}{ 0.988}} 
   \\ 
        ~ & Von Neumann & 0.837 & 0.962 & 0.837 & 0.962 & 0.857&	0.951 & 0.963
  \\ \hline

        \multirow{3}{*}{$f_{10}$} & Star& 0.976&	0.970	&0.976&	0.970&	0.984&	0.981
       & 0.961 \\ 
        ~ & Ring &	0.976 &	0.970 &	0.976 &	0.970 &	0.985 &	0.982
 & \textbf{\color{red}{ 0.982}}
   \\ 
        ~ & Von Neumann & 0.976& 0.970 & 0.976 & 0.970 &	0.984 &	0.982 &  0.954
  \\ \hline

        \multirow{3}{*}{$f_{11}$} & Star &0.940&	0.958	& 0.940& 0.958&	0.940&	0.960 &  0.980
\\ 
        ~ & Ring &	0.940 &	0.958 &	0.940&	0.958 &	0.940 &	0.960
  & 0.980
   \\ 
        ~ & Von Neumann & 0.941& 0.958 & 0.940 & 0.958 &	0.940 &	0.960 & 0.977
  \\ \hline

        \multirow{3}{*}{$f_{12}$} & Star & 0.983&	0.982&	0.985&	0.987&	0.986&	0.987
 & 0.943

\\ 
        ~ & Ring &	\textcolor{red}{0.984} &	0.987 &	0.986 &	0.987 &	0.987 &	0.987
  &\textbf{\color{red}{ 0.991}} 
   \\ 
        ~ & Von Neumann & 0.982& 0.985 & 0.982 & 0.985 &	0.986 &	0.987 & 0.958

  \\ \hline

        \multirow{3}{*}{$f_{13}$} & Star & 0.927&	0.976&	0.932&	0.977&	0.937&	0.973
       &\textbf{\color{red}{ 0.972}} 

\\ 
        ~ & Ring  &	0.928 &	0.975 &	0.931 &	0.974 &	0.938 &	0.973
 & 0.958 \\ 
        ~ & Von Neumann & 0.928& 0.975 & 0.934 & 0.976 &	0.938&	0.972 &  0.970
  \\ \hline

        \multirow{3}{*}{$f_{14}$} & Star  &0.749&	0.966&	0.753&	0.966 &	0.760 &	0.966
 & 0.975 \\ 
        ~ & Ring &	\textcolor{red}{0.757} &	0.963 &	0.762 &	0.963 &	0.762 &	0.966
 &\textbf{\color{red}{ 0.985}}
  \\ 
        ~ & Von Neumann & 0.751& 0.967&	0.754& 0.963 & 0.761 &	0.966 &  0.966
  \\ \hline

        \multirow{3}{*}{$f_{15}$} & Star & 0.891 &	0.970&	0.895&	0.967&	0.899&	0.976
     & 0.982 \\ 
        ~ & Ring  &	\textcolor{red}{0.893} &	0.969 &	0.896 &	0.966 &	0.900 &	0.976  & \textbf{\color{red}{ 0.985}} \\ 
        ~ & Von Neumann &  0.892 & 0.969 &	0.894 &	0.968 &	0.900 &	0.976  & 0.984 \\ \hline

        \multirow{3}{*}{$f_{16}$} & Star & 0.832 &	0.945 &	0.852 &	 0.968 & 0.842 & 0.961
 &\textbf{\color{red}{ 0.981}} 

\\ 
        ~ & Ring  &	0.832 &	0.945 &	0.853 &	0.968 &	0.843 &	0.962
  & 0.980
   \\ 
        ~ & Von Neumann & 0.832& 0.945 & 0.846 & 0.963 &0.843	&0.961 &   0.974
  \\ \hline

        \multirow{3}{*}{$f_{17}$} & Star &  0.826	& 0.987&	0.826&	0.987&	0.832&	0.983
 & 0.962\\ 
        ~ & Ring  &	\textcolor{red}{0.827} &	0.984 &	0.829 &	0.985 &	0.833 &	0.983
  &\textbf{\color{red}{ 0.982}} 
   \\ 
        ~ & Von Neumann & 0.825& 0.981 &0.826&	0.980&	0.833&	0.982  &   0.953
  \\ \hline

        \multirow{3}{*}{$f_{18}$} & Star & 0.875 &	0.978& 	0.875&	0.984&	0.889&	0.977
  & 0.968
\\ 
        ~ & Ring &	\textcolor{red}{0.877} &	0.981 &	0.879 &	0.979 &	0.889 &	0.977
 & 0.968
   \\ 
        ~ & Von Neumann & 0.873 & 0.982& 0.879&	0.981&	0.889&	0.977  &   0.969
  \\ \hline

        \multirow{3}{*}{$f_{19}$} & Star & 0.755 &	0.941 &	0.766 &	0.944 &	0.779 &	0.953 & 0.956 \\
        
        ~ & Ring &	\textcolor{red}{0.763} &	0.946 &	0.769 &	0.940 &	0.781 &	0.953
       & \textbf{\color{red}{ 0.982}} 
   \\ 
        ~ & Von Neumann & 0.755& 0.943&	0.763 &	0.947 &	0.780&	0.953   &  0.972
  \\ \hline

        \multirow{3}{*}{$f_{20}$} & Star & 0.852 &	0.982 &	0.859 &	0.977 &	0.856 &	0.980 & 0.995 \\ 
        ~ & Ring &	0.852 &	0.982&	0.859 &	0.975 &	0.856 &	0.980
& \textbf{\color{red}{ 0.998}}
   \\ 
        ~ & Von Neumann & 0.849 & 0.976& 0.850 & 0.971&	0.855 &	0.980   & 0.992
  \\ \hline

         \multirow{3}{*}{$f_{21}$} & Star & 0.763& 0.934 & 0.778	& 0.960 &	0.779 &	0.952 & 0.962
\\ 
        ~ & Ring &	\textcolor{red}{0.769} &	0.942 &	0.779 &	0.961& 0.781 &	0.955
  & \textbf{\color{red}{ 0.989}}
   \\ 
        ~ & Von Neumann &  0.763 &	0.934 &	0.779 &	0.965 &	0.781 &	0.954 &  0.957
  \\ \hline

         \multirow{3}{*}{$f_{22}$} & Star & 0.763 & 0.921 & 0.763	& 0.921 &	0.780 &	0.942 & 0.965
\\ 
        ~ & Ring &	\textcolor{red}{0.774} &	0.934&	0.781 &	0.940 &	0.781 &	0.940
  & \textbf{\color{red}{ 0.972}} 
   \\ 
        ~ & Von Neumann & 0.773& 0.916&	0.778 &	0.951 &	0.781 &	0.942 & 0.964
 \\ \hline

         \multirow{3}{*}{$f_{23}$} & Star &	0.851 &	0.978 &	0.861	& 0.974& 0.859 & 0.979  &\textbf{\color{red}{ 0.978}}
\\ 
        ~ & Ring &	0.851 &	0.978 &	0.860 &	0.975 &	0.858 &	0.979
 & 9.76 \\ 
        ~ & Von Neumann & 0.851&	0.978&	0.856 &	0.978 &	0.858 & 0.979 & 0.966 
  \\ \hline

         \multirow{3}{*}{$f_{24}$} & Star &	0.877 &	0.984&	0.886	&0.987&	0.882&	0.882 & 0.952
\\ 
        ~ & Ring 	& 0.877 &	0.984 &	0.886 &	0.987 &	0.883 &	0.987 &\textbf{\color{red}{ 0.966}} 
   \\ 
        ~ & Von Neumann & 0.877& 0.984 &	0.884 &	0.989 &	0.882 &	0.987  &  0.930
  \\ \hline

   \end{tabular}
\end{table*}

\begin{table*}[htp]
\centering
 \caption{Performance of Star, Ring and Von Neumann Topologies over all BBOB functions on d=5.}
\label{Tab: d=5}
    \centering
   \scriptsize \begin{tabular}{ccccccccc}
    \hline
      \textbf{Function}  & \textbf{Topologies} &  \textbf{sbm}  & \textbf{sbs} & \textbf{abm} & \textbf{abs} & \textbf{all mean} & \textbf{all std}  & \textbf{R2 train}\\ \hline
      
        \multirow{3}{*}{$f_1$} & Star & 0.780 &	0.968&	0.780&	0.968&	0.846&	0.967&	\textcolor{red}{0.932} \\ 
        ~ & Ring & \textcolor{red}{0.863} &	0.980 &	0.863 &	0.980 &	0.882 &	0.978 &	0.895 \\ 
        ~ & Von Neumann & 0.812 & 0.971 & 0.812 & 0.971 &	0.862 &	0.972 &	0.917
\\ \hline

        \multirow{3}{*}{$f_2$} & Star&	0.997&	0.996&	0.997&	0.996&	0.998&	1.000&	0.283 \\ 
        ~ & Ring & 0.998 &	1.000 &	0.998 &	1.000 &	0.998 &	1.000 & 0.327	\\ 
        ~ & Von Neumann & 0.998 & 1.000 & 0.998 & 1.000 &	0.998 &	1.000 &	\textcolor{red}{0.409} \\ \hline

        \multirow{3}{*}{$f_3$} & Star&	0.950 &	0.992 &	0.950&	0.992 &	0.964&	0.990 &	0.793
\\ 
        ~ & Ring &	\textcolor{red}{0.967} &	0.988 &	0.967 &	0.988 &	0.976 &	0.990 &	0.800
\\ 
        ~ & Von Neumann & 0.952 & 0.990 &	0.954 &	0.992 &	0.968 &	0.990 &	\textcolor{red}{0.814}
\\ \hline

        \multirow{3}{*}{$f_4$} & Star	& 0.960 & 0.990 & 0.960 &	0.990 &	0.974 &	0.990 &	0.800
\\ 
        ~ & Ring & \textcolor{red}{0.979} &	0.994 &	0.980 &	0.995 &	0.988 &	0.991 &	0.840 \\ 
        ~ & Von Neumann  & 0.966 &	0.990 &	0.968 &	0.991 &	0.978 &	0.990 &	\textcolor{red}{0.848}
\\ \hline

        \multirow{3}{*}{$f_5$} & Star&	0.929 &	0.989 &	0.929&	0.989 &	0.938&	0.990&	0.885 \\ 
        ~ & Ring &	\textcolor{red}{0.935} &	0.990 &	0.938 &	0.992 & 0.942 &	0.990 &	\textcolor{red}{0.927} \\ 
        ~ & Von Neumann & 0.932 & 0.992 &	0.933 &	0.992 &	0.940 &	0.990 &	0.923
\\ \hline

        \multirow{3}{*}{$f_6$} & Star &	0.926 &	0.963 &	0.926 &	0.963&	0.959 &	0.972 &	0.884\\ 
        ~ & Ring &	\textcolor{red}{0.957} &	0.970 &	0.957 &	0.967 &	0.978 &	0.977 &	0.881\\ 
        ~ & Von Neumann & 0.934 & 0.971 & 0.943 & 0.972 &	0.968 &	0.974 &	\textcolor{red}{0.907}
\\ \hline

        \multirow{3}{*}{$f_7$} & Star&	0.870 &	0.971 &	0.870 &	0.971 &	0.902 &	0.970 &	0.881 \\ 
        ~ & Ring &	\textcolor{red}{0.902} &	0.973 &	0.908 &	0.978 &	0.929 &	0.971 &\textcolor{red}{	0.912} \\ 
        ~ & Von Neumann  & 0.874 &	0.982 &	0.874 &	0.982 &	0.909 &	0.971 &	0.899
        \\ \hline

        \multirow{3}{*}{$f_8$} & Star&	0.948 &	0.977 &	0.948 &	0.977 &	0.987 &	0.983 &	0.817
\\ 
        ~ & Ring &	\textcolor{red}{0.989} &	0.981 &	0.989 &	0.981 &	0.997 &	0.993 &	\textcolor{red}{0.863}
\\ 
        ~ & Von Neumann & 0.969	& 0.986 &	0.970 &	0.986 &	0.992 &	0.988 &	0.848
\\ \hline

        \multirow{3}{*}{$f_9$} & Star&	0.952 &	0.984 &	0.952 &	0.984 &	0.985 &	0.983 &	0.841
\\ 
        ~ & Ring &	\textcolor{red}{0.993} &	0.988 &	0.993 &	0.988 &	0.997 &	0.995 &	\textcolor{red}{0.919}
\\ 
        ~ & Von Neumann & 0.953 & 0.989 & 0.955 &	0.987 &	0.988&	0.985 &	0.865
\\ \hline

        \multirow{3}{*}{$f_{10}$} & Star&	0.998 &	1.000 &	0.998 &	1.000 &	0.998 &	1.000 &	0.349
\\ 
        ~ & Ring &	0.998 &	0.999 &	0.998 &	1.000 &	0.998 &	1.000 &	0.152
\\ 
        ~ & Von Neumann  & 0.998 &	1.000 &	0.998 &	1.000 &	0.998 &	1.000 &	\textcolor{red}{0.413}
\\ \hline

        \multirow{3}{*}{$f_{11}$} & Star &	0.961 &	0.976 &	0.961 &	0.976 &	0.982 &	0.982 &	\textcolor{red}{0.658}
\\ 
        ~ & Ring &	\textcolor{red}{0.965} &	0.979 &	0.965 &	0.979 &	0.985 &	0.984 &0.654	
\\ 
        ~ & Von Neumann  & 0.961 &	0.976 &	0.963 &	0.980 &	0.981 &	0.982 &	0.604
\\ \hline

        \multirow{3}{*}{$f_{12}$} & Star & 0.998 & 1.000 & 0.998 &	1.000 &	0.998 &	1.000 &	0.734
\\ 
        ~ & Ring & 0.998 & 1.000 &	0.998 & 1.000 &	0.998 &	1.000 &0.734
\\ 
        ~ & Von Neumann & 0.998 & 1.000 &	0.998 &	1.000 &	0.998 &	1.000	& \textcolor{red}{0.649}
\\ \hline


        \multirow{3}{*}{$f_{13}$} & Star & 0.988 & 0.990 & 0.988 &	0.990 &	0.996 &	0.996 &	0.681
\\ 
        ~ & Ring  &	\textcolor{red}{0.996} &	0.992 &	0.998 &	1.000 &	0.998 &	0.999 & \textcolor{red}{0.711}
\\ 
        ~ & Von Neumann & 0.990 & 0.993 & 0.990 & 0.993 &	0.997 &	0.997 &	0.702
\\ \hline

        \multirow{3}{*}{$f_{14}$} & Star  &	0.766 &	0.966 &	0.766 &	0.966 &	0.817 &	0.973 &	0.882
\\ 
        ~ & Ring &	\textcolor{red}{0.832} &	0.976 &	0.832 &	0.976 &	0.846 &	0.976 &	0.890
        \\ 
        ~ & Von Neumann  & 0.791 &	0.976 &	0.791 &	0.977 &	0.828 &	0.976 &	\textcolor{red}{0.912}
\\ \hline

        \multirow{3}{*}{$f_{15}$} & Star & 0.949 & 0.990 & 0.949	& 0.990 &	0.965 &	0.990 &	0.761
\\ 
        ~ & Ring 	&\textcolor{red}{0.966} & 0.990 &	0.966 &	0.990 &	0.976&	0.991 &	0.729
\\ 
        ~ & Von Neumann  & 0.956 &	0.993 &	0.957 &	0.991 &	0.969 &	0.990 & \textcolor{red}{0.776}
\\ \hline

        \multirow{3}{*}{$f_{16}$} & Star & 0.888 & 0.986 & 0.888 &	0.986 &	0.898 &	0.986 &	0.685
\\ 
        ~ & Ring  &	\textcolor{red}{0.891} &	0.987 &	0.892 &	0.986 &	0.898 &	0.985 &\textcolor{red}{0.756}
\\ 
        ~ & Von Neumann  & 0.886 &	0.989 &	0.890 &	0.987 &	0.896 &	0.985 &	0.746
\\ \hline

        \multirow{3}{*}{$f_{17}$} & Star &	0.822 &	0.980 &	0.822 &	0.980 &	0.847 &	0.984 &	0.845 \\ 
        ~ & Ring  &	 \textcolor{red}{0.849} &	0.986 &	0.851 &	0.986 &	0.863 &	0.989 &	0.783
\\ 
        ~ & Von Neumann & 0.832 & 0.982 & 0.833 & 0.982 &	0.853 &	0.986 & \textcolor{red}{0.854}
\\ \hline

        \multirow{3}{*}{$f_{18}$} & Star & 0.878 & 0.980 & 0.878 &	0.980 &	0.903 &	0.985 &	0.843 \\ 
        ~ & Ring &	\textcolor{red}{0.906} &	0.983 &	0.910 &	0.985 &	0.917 &	0.986 &	0.812
\\ 
        ~ & Von Neumann & 0.884 & 0.982 & 0.886 & 0.982 &	0.908 &	0.985 & \textcolor{red}{0.862}
\\ \hline

        \multirow{3}{*}{$f_{19}$} & Star & 0.840 & 0.986 & 0.840 &	0.986 &	0.859 &	0.987 &	0.657 \\ 
        ~ & Ring &	\textcolor{red}{0.859} &	0.981 &	0.859 &	0.989 &	0.871 &	0.990 &	0.572
\\ 
        ~ & Von Neumann & 0.840 & 0.984 & 0.844 & 0.987 &	0.863 &	0.987 &	\textcolor{red}{0.677}
\\ \hline

        \multirow{3}{*}{$f_{20}$} & Star & 0.859 & 0.988 & 0.859 &	0.988 &	0.894 &	0.969 & \textcolor{red}{0.938}
\\ 
        ~ & Ring &	\textcolor{red}{0.897} &	0.974 &	0.897 &	0.974 &	0.941 &	0.949 &	0.891
\\ 
        ~ & Von Neumann & 0.865&	0.986 &	0.866 &	0.987 &	0.905 &	0.962 &	0.928
\\ \hline

         \multirow{3}{*}{$f_{21}$} & Star & 0.859 & 0.955 & 0.867 &	0.980 &	0.875 &	0.974 &	0.825 \\ 
        ~ & Ring &	\textcolor{red}{0.870} &	0.974 &	0.870 &	0.974 &	0.884 &	0.980 &0.784
\\ 
        ~ & Von Neumann & 0.845 & 0.956 & 0.849 & 0.957 &	0.873 &	0.972 &	0.825
\\ \hline

         \multirow{3}{*}{$f_{22}$} & Star & 0.837 & 0.921 & 0.837 &	0.933 &	0.883 &	0.961 &	0.809
\\ 
        ~ & Ring &	\textcolor{red}{0.879} &	0.951 &	0.884 &	0.969 &	0.909 &	0.974 &	0.784
\\ 
        ~ & Von Neumann & 0.862 & 0.940 & 0.862 & 0.943 &	0.892 &	0.969 &	\textcolor{red}{0.823}
\\ \hline

         \multirow{3}{*}{$f_{23}$} & Star & 0.837 & 0.985 & 0.845 &	0.985 &	0.843 &	0.987 &	0.719 \\ 
        ~ & Ring &	0.837 &	0.983 &	0.839 &	0.988 &	0.843 &	0.988 &	0.733
 \\ 
        ~ & Von Neumann & \textcolor{red}{0.838} & 0.989 & 0.847 & 0.989 &	0.843 &	0.988 &	\textcolor{red}{0.762}
\\ \hline

         \multirow{3}{*}{$f_{24}$} & Star & 0.946 & 0.992 & 0.947 &	0.991 &	0.956 &	0.992 &	0.658 \\ 
        ~ & Ring &	\textcolor{red}{0.957} &	0.992 &	0.960 &	0.995 &	0.963 &	0.992 &	\textcolor{red}{0.757}\\ 
        ~ & Von Neumann & 0.948 &	0.994 &	0.950 &	0.995 &	0.957 &	0.992 &	0.749
\\ \hline

\end{tabular}
\end{table*}

\begin{table*}[htp]\centering
 \caption{Single best Config of Star, Ring and Von Neumann Topologies over all BBOB functions on d=5.}
\label{T3}
    \centering
   \scriptsize \begin{tabular}{ccccccccc}
    \hline
      \textbf{Function}  & \textbf{Topologies} &  \textbf{$c_1$}  & \textbf{$c_2$} & \textbf{$k$} & \textbf{$n\_particle$} & \textbf{$p$} & \textbf{$r$}  & \textbf{$w$}\\ \hline
      
        \multirow{3}{*}{$f_1$} & Star & 0.9& 0.7&	1&	50&	1&	1&	0.5

\\ 
        ~ & Ring & 0.3& 0.7&	3&	50&	2&	1&	0.9
   \\ 
        ~ & Von Neumann & 0.9& 0.7&	1&	50&	1&	1&	0.5
  \\ \hline

        \multirow{3}{*}{$f_2$}& Star & 0.7& 0.6&	1&	50&	1&	1&	0.5
\\ 
        ~ & Ring & 0.5& 0.6&	3&	50&	2&	1&	0.7
   \\ 
        ~ & Von Neumann & 0.3& 0.6&	1&	50&	1&	1&	0.5
 \\ \hline

        \multirow{3}{*}{$f_3$} & Star& 0.9& 0.7&	1&	50&	1&	1&	0.5
\\ 
        ~ & Ring & 0.5& 0.7&	3&	50&	1&	1&	0.7
   \\ 
        ~ & Von Neumann & 0.9& 0.7&	1&	50&	1&	1&	0.5
  \\ \hline

        \multirow{3}{*}{$f_4$}&Star & 0.7& 0.7&	1&	50&	1&	1&	0.5
\\ 
        ~ & Ring & 0.5& 0.7&	3&	50&	2&	1&	0.9
  \\ 
        ~ & Von Neumann & 0.3& 0.6&	1&	50&	1&	1&	0.5
  \\ \hline

        \multirow{3}{*}{$f_5$} & Star& 0.7& 0.7&	1&	50&	1&	1&	0.5
\\ 
        ~ & Ring & 0.5& 0.2&	1&	100&	1&	1&	0.9
   \\ 
        ~ & Von Neumann & 0.5& 0.7&	1&	50&	1&	1&	0.5
  \\ \hline

        \multirow{3}{*}{$f_6$} & Star& 0.5& 0.7&	1&	50&	1&	1&	0.5
\\ 
        ~ & Ring & 0.7& 0.7&	3&	50&	2&	1&	0.9
   \\ 
        ~ & Von Neumann & 0.9& 0.6&	1&	50&	1&	1&	0.5
  \\ \hline

        \multirow{3}{*}{$f_7$} & Star & 0.9& 0.6&	1&	50&	1&	1&	0.5
\\ 
        ~ & Ring & 0.3& 0.6&	3&	50&	1&	1&	0.9
   \\ 
        ~ & Von Neumann & 0.7& 0.7&	1&	50&	1&	1&	0.5
  \\ \hline

        \multirow{3}{*}{$f_8$}&Star & 0.9& 0.6&	1&	50&	1&	1&	0.5
\\ 
        ~ & Ring & 0.9& 0.6&	3&	50&	1&	1&	0.7
  \\ 
        ~ & Von Neumann & 0.5& 0.7&	1&	50&	1&	1&	0.5
  \\ \hline

        \multirow{3}{*}{$f_9$} & Star & 0.9& 0.6&	1&	50&	1&	1&	0.5
\\ 
        ~ & Ring & 0.9& 0.7&	3&	50&	2&	1&	0.5
   \\ 
        ~ & Von Neumann & 0.9& 0.6&	1&	50&	1&	1&	0.5
  \\ \hline

        \multirow{3}{*}{$f_{10}$} & Star & 0.7& 0.4&	1&	50&	1&	1&	0.7
\\ 
        ~ & Ring & 0.9& 0.6&	3&	50&	1&	1&	0.9
   \\ 
        ~ & Von Neumann & 0.5& 0.6&	1&	50&	1&	1&	0.5
  \\ \hline

        \multirow{3}{*}{$f_{11}$} & Star & 0.9& 0.4&	1&	50&	1&	1&	0.7
\\ 
        ~ & Ring & 0.9& 0.7&	3&	50&	2&	1&	0.9
   \\ 
        ~ & Von Neumann & 0.7& 0.6&	1&	50&	1&	1&	0.7
  \\ \hline

        \multirow{3}{*}{$f_{12}$} & Star & 0.7& 0.6&	1&	50&	1&	1&	0.5

\\ 
        ~ & Ring & 0.3& 0.6&	1&	50&	1&	1&	0.7
   \\ 
        ~ & Von Neumann & 0.7& 0.6&	1&	50&	1&	1&	0.5
  \\ \hline

        \multirow{3}{*}{$f_{13}$} & Star & 0.7& 0.6&	1&	50&	1&	1&	0.5 

\\ 
        ~ & Ring  & 0.3& 0.7&	3&	50&	2&	1&	0.9
   \\ 
        ~ & Von Neumann & 0.7& 0.7&	1&	50&	1&	1&	0.5
  \\ \hline

        \multirow{3}{*}{$f_{14}$} & Star  & 0.7& 0.7&	1&	50&	1&	1&	0.5

\\ 
        ~ & Ring& 0.5& 0.6&	3&	150&	1&	1&	0.9
  \\ 
        ~ & Von Neumann & 0.5& 0.7&	1&	50&	1&	1&	0.5
  \\ \hline

        \multirow{3}{*}{$f_{15}$} & Star & 0.5& 0.6&	1&	50&	1&	1&	0.5

 \\ 
        ~ & Ring 	& 0.7& 0.6&	3&	50&	1&	1&	0.9
   \\ 
        ~ & Von Neumann& 0.9& 0.6&	1&	50&	1&	1&	0.5
 \\ \hline

        \multirow{3}{*}{$f_{16}$} & Star & 0.5& 0.7&	1&	50&	1&	1&	0.5

\\ 
        ~ & Ring  & 0.9& 0.4&	3&	50&	2&	1&	0.7
   \\ 
        ~ & Von Neumann & 0.9& 0.7&	1&	50&	1&	1&	0.5
  \\ \hline

        \multirow{3}{*}{$f_{17}$} & Star 	& 0.7& 0.6&	1&	50&	1&	1&	0.5

\\ 
        ~ & Ring& 0.3& 0.7&	3&	50&	2&	1&	0.7
   \\ 
        ~ & Von Neumann & 0.9& 0.6&	1&	50&	1&	1&	0.5
  \\ \hline

        \multirow{3}{*}{$f_{18}$} & Star & 0.7& 0.6&	1&	50&	1&	1&	0.5
\\ 
        ~ & Ring & 0.3& 0.7&	3&	50&	2&	1&	0.7
   \\ 
        ~ & Von Neumann & 0.5& 0.7&	1&	50&	1&	1&	0.5
  \\ \hline

        \multirow{3}{*}{$f_{19}$} & Star 	& 0.3& 0.4&	1&	50&	1&	1&	0.5
\\ 
        ~ & Ring & 0.5& 0.6&	3&	50&	1&	1&	0.9
   \\ 
        ~ & Von Neumann & 0.9& 0.6&	1&	50&	1&	1&	0.7
  \\ \hline

        \multirow{3}{*}{$f_{20}$} & Star 	& 0.9& 0.7&	1&	50&	1&	1&	0.5
\\ 
        ~ & Ring & 0.3& 0.7&	3&	50&	2&	1&	0.9
   \\ 
        ~ & Von Neumann & 0.9& 0.7&	1&	50&	1&	1&	0.5
  \\ \hline

         \multirow{3}{*}{$f_{21}$ } & Star  	& 0.3& 0.7&	1&	100&	1&	1&	0.5
\\ 
        ~ & Ring 	& 0.5& 0.4&	3&	50&	1&	1&	0.7
   \\ 
        ~ & Von Neumann & 0.3& 0.7&	1&	50&	1&	1&	0.5
  \\ \hline

         \multirow{3}{*}{$f_{22}$} & Star  	& 0.3& 0.6&	1&	50&	1&	1&	0.5
\\ 
        ~ & Ring & 0.5& 0.7&	3&	50&	1&	1&	0.7 
   \\ 
        ~ & Von Neumann & 0.9& 0.7&	1&	50&	1&	1&	0.5
 \\ \hline

         \multirow{3}{*}{$f_{23}$} & Star & 0.9& 0.7&	1&	50&	1&	1&	0.7
\\ 
        ~ & Ring & 0.7& 0.6&	3&	100&	2&	1&	0.9
   \\ 
        ~ & Von Neumann & 0.9& 0.6&	1&	150&	1&	1&	0.5
  \\ \hline

         \multirow{3}{*}{$f_{24}$} & Star & 0.9& 0.7&	1&	50&	1&	1&	0.5
\\ 
        ~ & Ring 	& 0.7& 0.6&	3&	150&	2&	1&	0.9
   \\ 
        ~ & Von Neumann& 0.5& 0.7&	1&	150&	1&	1&	0.7
  \\ \hline

   \end{tabular}
\end{table*}


 \section{Data-Driven and Interpretable Configuration Learning}\label{sec:6}
The IOHxplainer pipeline processes a large amount of data information on the behaviour and performance of different algorithm configurations for each topology. Many benefits exist to systematically store and re-use this data, but explainable AI (XAI) \cite{ding2022explainability} techniques can be used to obtain key insights from the data. In this section, we feature a few practical examples of how the gathered data can be utilized to improve the choice of algorithm configurations, their setup and their understanding.

\begin{figure}[H]
    \centering
    \includegraphics[scale=0.12]{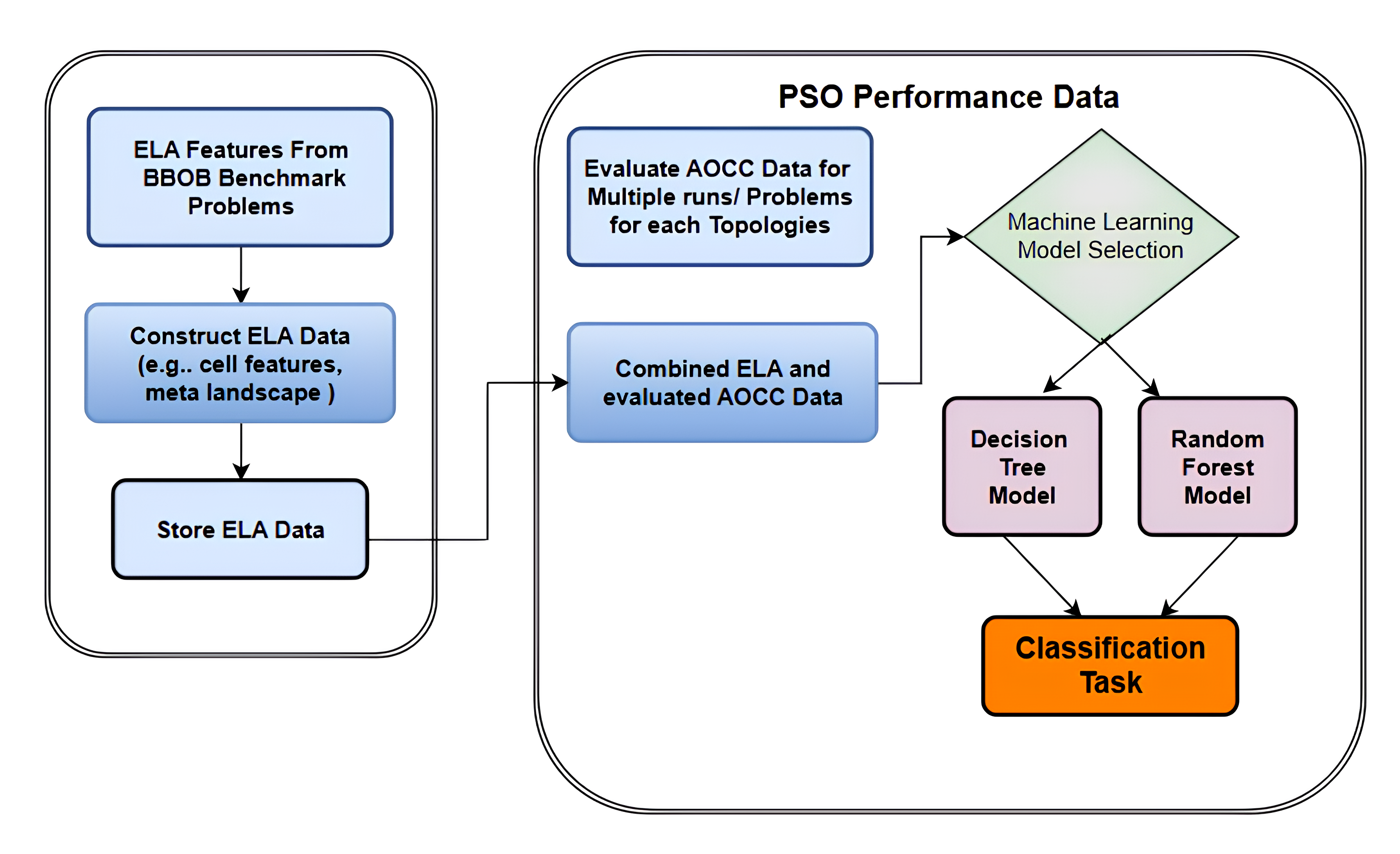}

      \caption{Workflow for classifying PSO AOCC performance using an expanded set of ELA features} \label{flow2}
\end{figure}
The IOHxplainer toolbox has a second benefit: It allows the easy training of machine learning models that are capable of predicting the most appropriate algorithm configuration (choice of modules, hyper parameters) depending on the characteristics of the optimization landscape observed. The workflow for performance classification based on Exploratory Landscape Analysis (ELA) features is shown in \ref{flow2}. The diagram shows the flow of feature extraction, augmentation, model training, model evaluation and prediction of performance of PSO algorithm using AOCC. It is essentially the job of auto-configuration and auto-selection of algorithms, as in the previous studies mentioned in \cite{kerschke2019automated,long2022learning}  and \cite{trajanov2022improving} respectively.


What distinguishes our approach is that we merge the selection of the algorithm and the tuning of its configuration into a single, unified machine learning problem. We achieve this by employing multi-output models that can handle both classification (for discrete choices like module selection) and regression (for continuous settings like hyper-parameters) simultaneously. Specifically, our models are limited to shallow Decision Tree (DT) \cite{song2015decision,wang2025neighborhood} and Random Forest (RF) \cite{pal2005random}. The decision was strategic, representing a trade-off between competing needs for interpretability and predictive performance. We chose the shallow DT for its transparency as a `white-box' model. Its limited depth results in a simple set of decision rules in human-readable form, which directly relate the Exploratory Landscape Analysis (ELA) features to algorithm performance, which is of crucial importance to domain experts. To complement this and to ensure high accuracy we used an RF model. As an ensemble method, RF addresses the potential overfitting of individual trees and robustly captures the complex, non-linear relationships between the wide array of ELA features and the optimal algorithm configuration.

We have collected a large set of features for Exploratory Landscape Analysis (ELA) to demonstrate what we can do. For these features we designed an experiment that sampled 1000 points from each BBOB suite instance. We wanted our machine learning algorithms to find the best performing algorithm configuration from our experiment results for each instance. With this two-pronged modeling approach (RF for strong predictions and DT for clear rule extraction) we train our systems to predict which algorithm configurations will perform well given the ELA features. This approach allows not only to automate the selection and configuration of algorithms but also to have transparent and understandable decision rules due to the interpretability of the used models.
\begin{figure*}[htp]
    \centering
    \begin{minipage}{1\textwidth}
        \includegraphics[width=\linewidth]{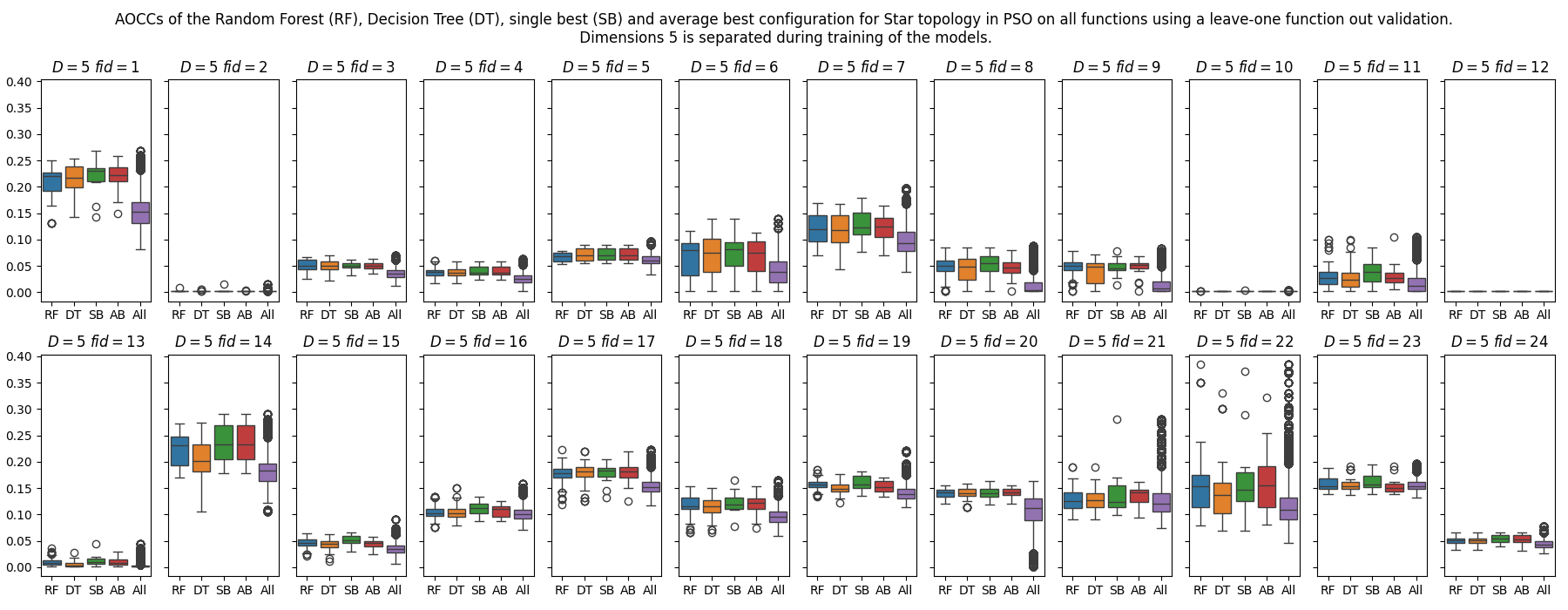}
    \end{minipage}
    \hfill 
   
    \begin{minipage}{1\textwidth}
        \includegraphics[width=\linewidth]{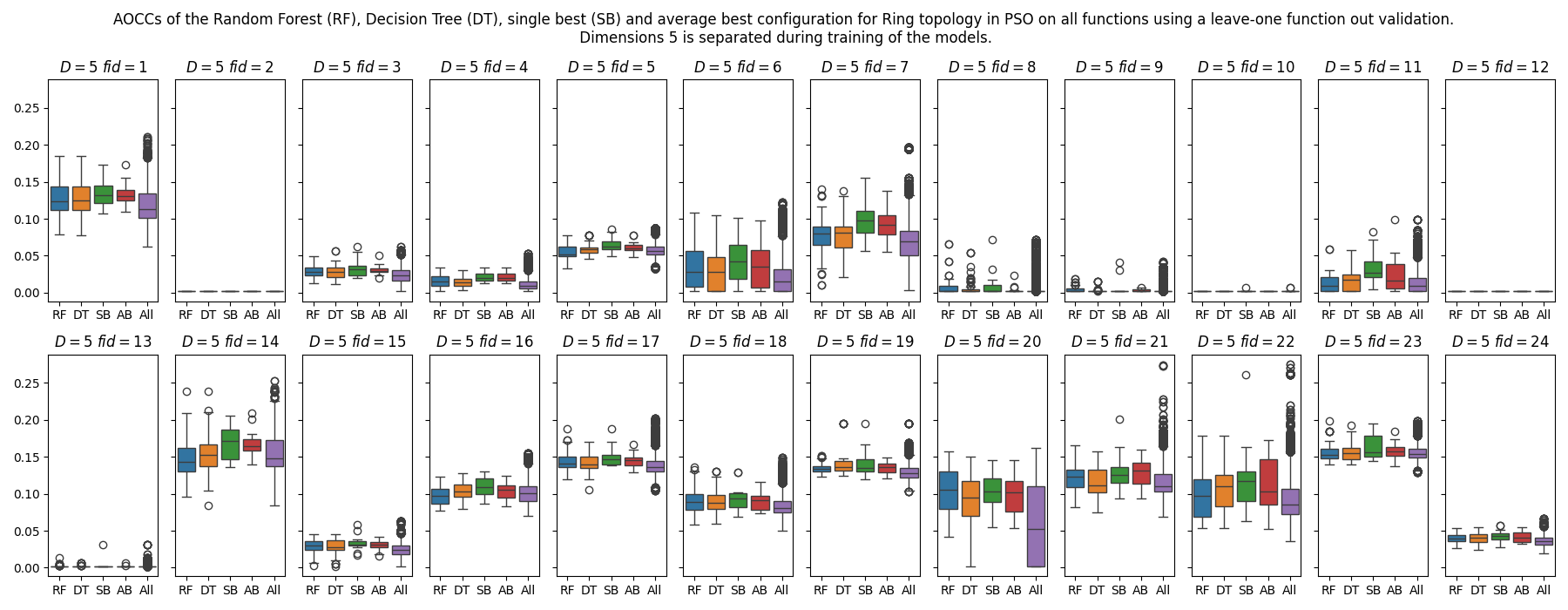}
    \end{minipage}
    \hfill

    \begin{minipage}{1\textwidth}
        \includegraphics[width=\linewidth]{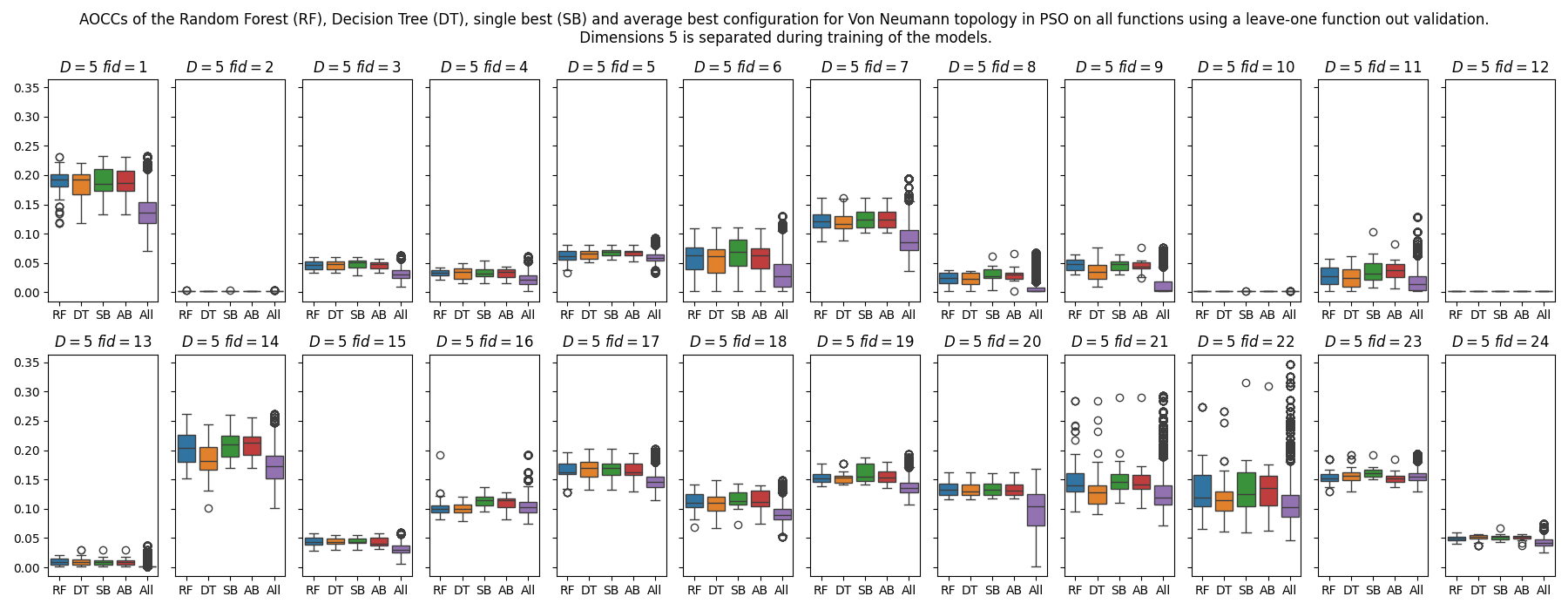}
    \end{minipage}
     \hfill
    
  \caption{AOCCs of RF, DT, SB, and AB configurations for PSO using all topologies on all functions using LoFo. Models are trained excluding dimension 5. } \label{fig Ao1}
\end{figure*}

\begin{figure*}[htp]
    \centering
    \begin{minipage}{1\textwidth}
        \includegraphics[width=\linewidth]{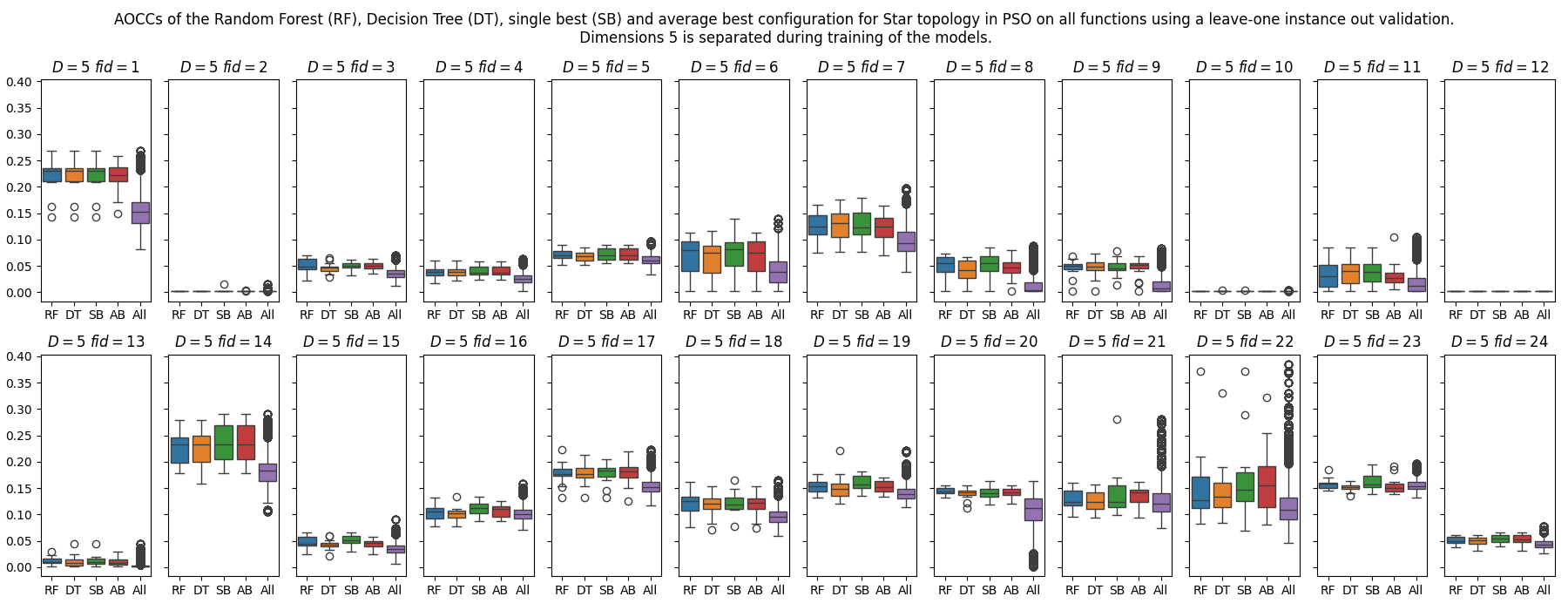 }
    \end{minipage}
    \hfill 
   
    \begin{minipage}{1\textwidth}
        \includegraphics[width=\linewidth]{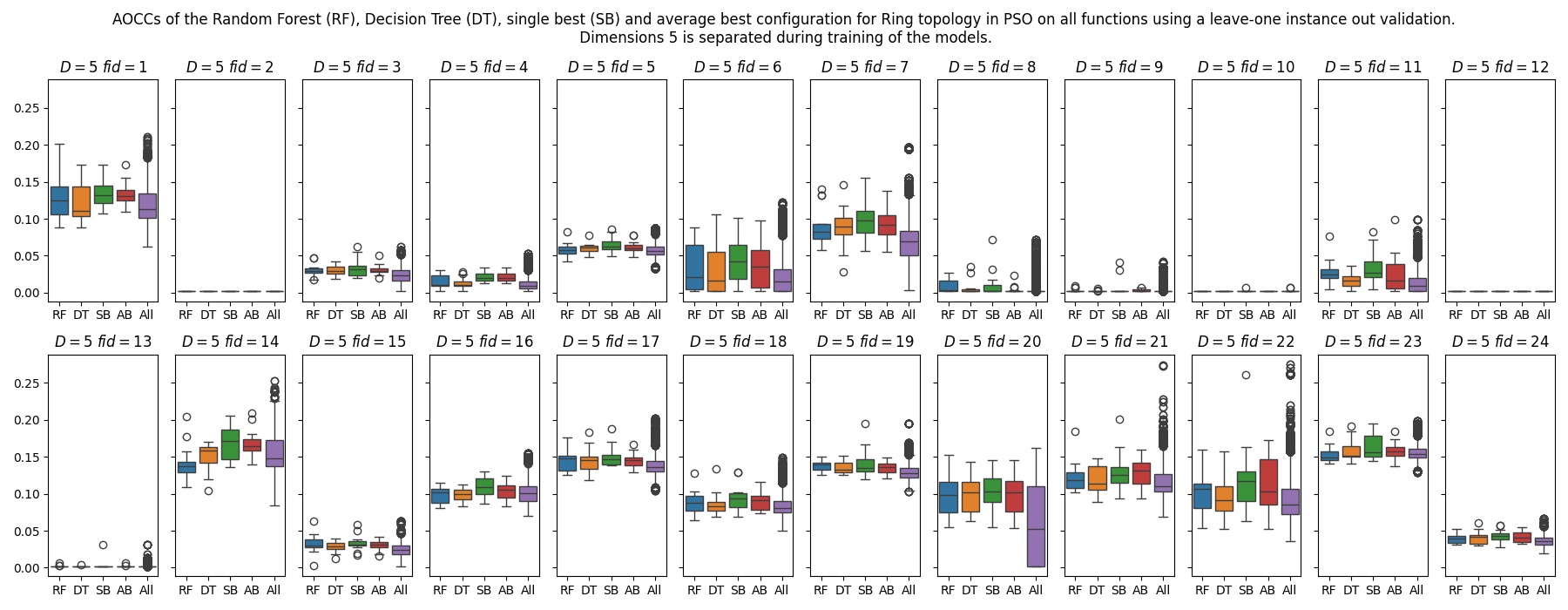}
    \end{minipage}
    \hfill

    \begin{minipage}{1\textwidth}
        \includegraphics[width=\linewidth]{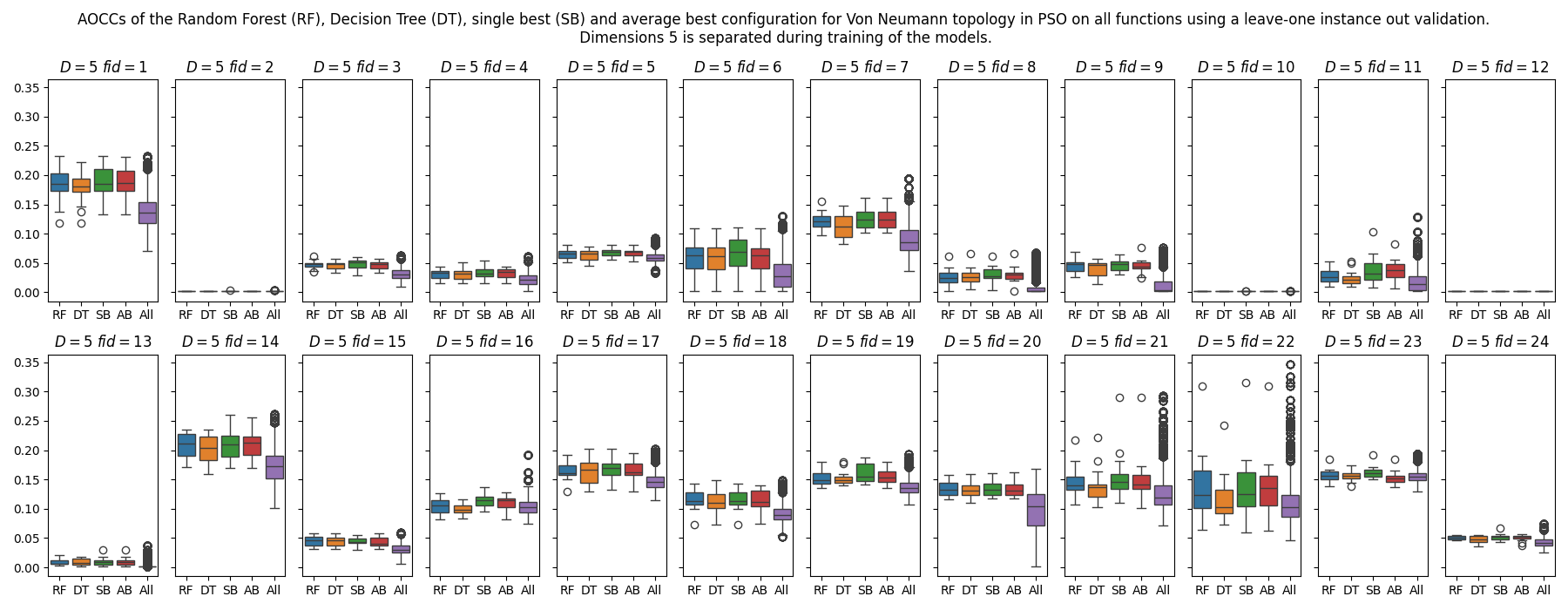}
    \end{minipage}
     \hfill
  \caption{AOCCs of RF, DT, SB, and AB configurations for PSO using all topologies on all functions using LoIo validation. Models are trained excluding dimension 5. } \label{fig Ao2}
\end{figure*}
\begin{figure*}[h!]
    \centering
    \begin{minipage}{0.48\textwidth}
        \includegraphics[width=\linewidth]{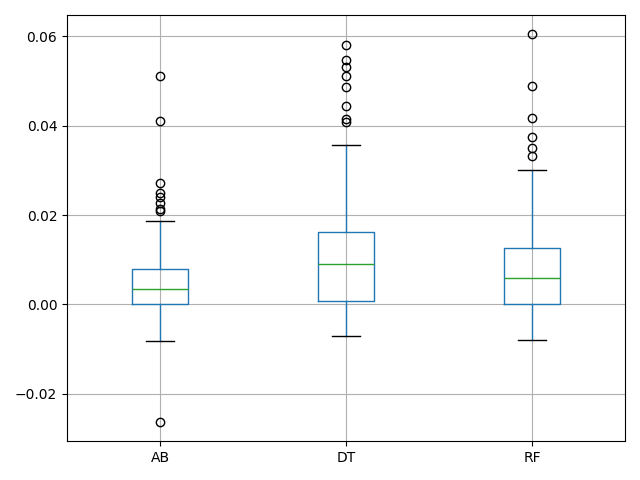}
    \end{minipage}
    \hfill 
    \begin{minipage}{0.48\textwidth}
        \includegraphics[width=\linewidth]{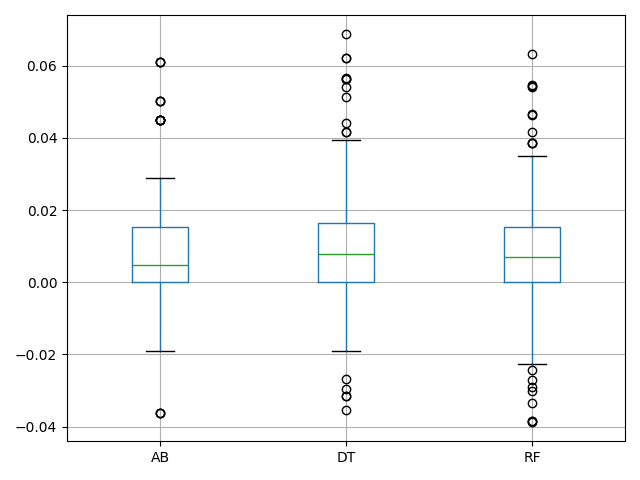}
    \end{minipage}
    \hfill
\caption{AOCC performance loss of PSO using Star topology on BBOB functions (d = 5), comparing RF, shallow DT, and AB configurations against the best single run, using LoFo (left) and LoIo (right) validation. } \label{fig c1}
\end{figure*}
\begin{figure*}[h!]
    \centering
    \begin{minipage}{0.48\textwidth}
        \includegraphics[width=\linewidth]{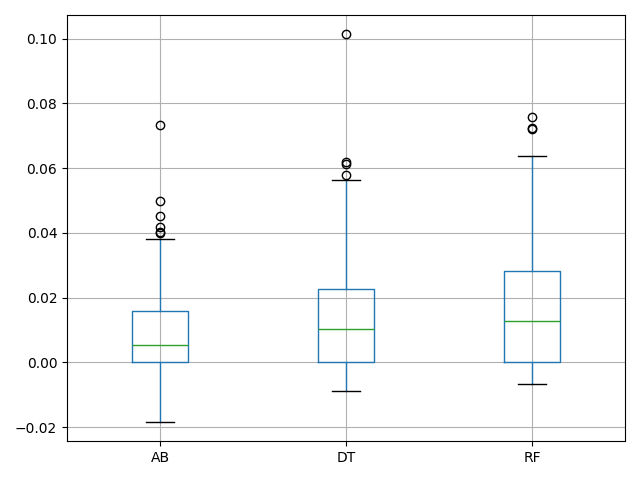}
    \end{minipage}
    \hfill 
    \begin{minipage}{0.48\textwidth}
        \includegraphics[width=\linewidth]{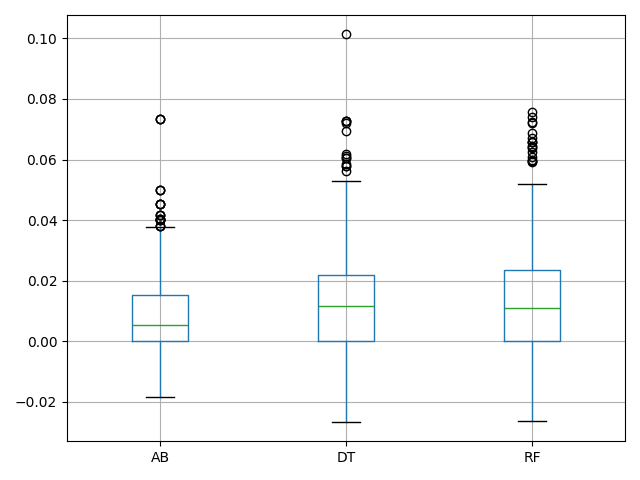}
    \end{minipage}
    \hfill
\caption{AOCC performance loss of PSO using Ring topology on BBOB functions (d = 5), comparing RF, shallow DT, and AB configurations against the best single run, using LoFo (left) and LoIo (right) validation. } \label{fig c2}
\end{figure*}
\begin{figure*}[h!]
    \centering
    \begin{minipage}{0.48\textwidth}
        \includegraphics[width=\linewidth]{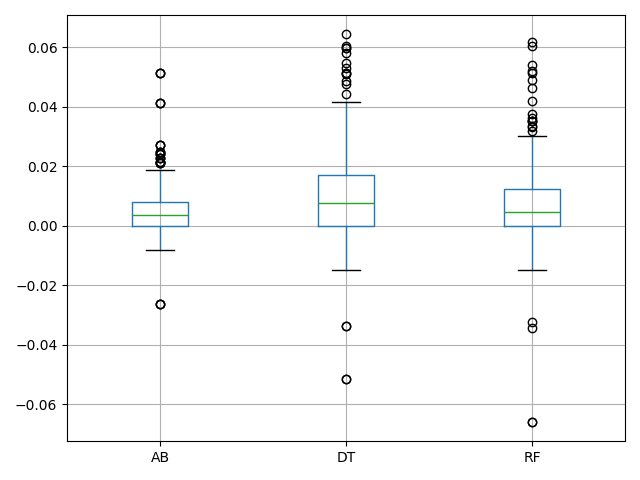}
    \end{minipage}
    \hfill 
    \begin{minipage}{0.48\textwidth}
        \includegraphics[width=\linewidth]{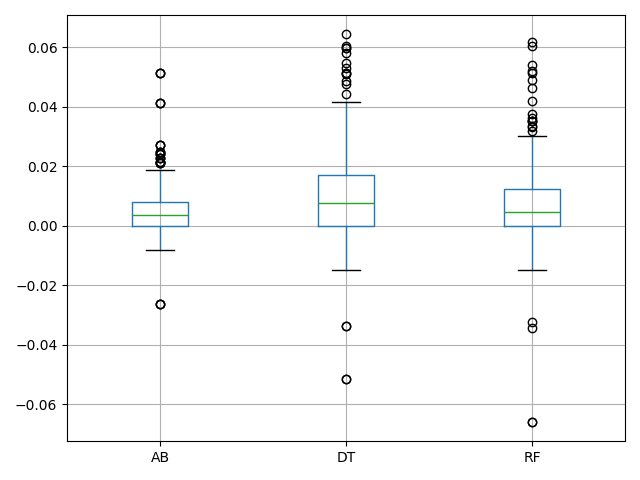}
    \end{minipage}
    \hfill
\caption{AOCC performance loss of PSO using Von Neumann topology on BBOB functions (d = 5), comparing RF, shallow DT, and AB configurations against the best single run, using LoFo (left) and LoIo (right) validation. } \label{fig c3}
\end{figure*}

The topologies and functions are listed and the performance of the AOCC of the Random Forest \cite{pal2005random}, Decision Tree (DT) \cite{song2015decision}, Single Best (SB) and Average Best (AB) configuration is shown in Figures \ref{fig Ao1} and \ref{fig Ao2}. The difference lies with how they're validated: Figure 
 \ref{fig Ao1} tests generalization to unseen functions, and Figure \ref{fig Ao2} tests generalization to unseen problem instances.
In Figures \ref{fig c1}- \ref{fig c3}, the results of each PSO setup strategy are shown with both the leave one function out (LoFo) and the leave one instance out (LoIo) methods for the distance to optimum in the Star, Ring and Von Neumann patterns applied to functions in the BBOB benchmark. In the case of the Star topology, it is evident that the RF model performs much better than the others, reaching the lowest optimum, while AB is performing worse, with the benefit of model-based configurations. The case is a little different on the Ring topology, however, because for all three methods (AB, DT, and RF) the performance is practically the same, and even for the more stable AB approach the performance is not that poor in comparison to the model-based approaches. AB always gets the lowest optimum, and performs better than both DT and RF, which are more variable. Regarding validation methods, LoIo sometimes yields narrower and more reliable distribution of the AOCC for all topologies. This means that it has a better understanding of how the model is going to act for new instances. LoFo, on the other hand, is more difficult to generalize, and displays better differentiation in performance with the methods. Generally, model-based methods appear to be most successful when applied to Star topology, whereas AB works as well as or even better with more organized topologies like Ring and Von Neumann.

\subsection{Explaining PSO via Decision Trees and ELA Features}

\begin{sidewaysfigure}
    \centering
    \includegraphics[width = \linewidth]{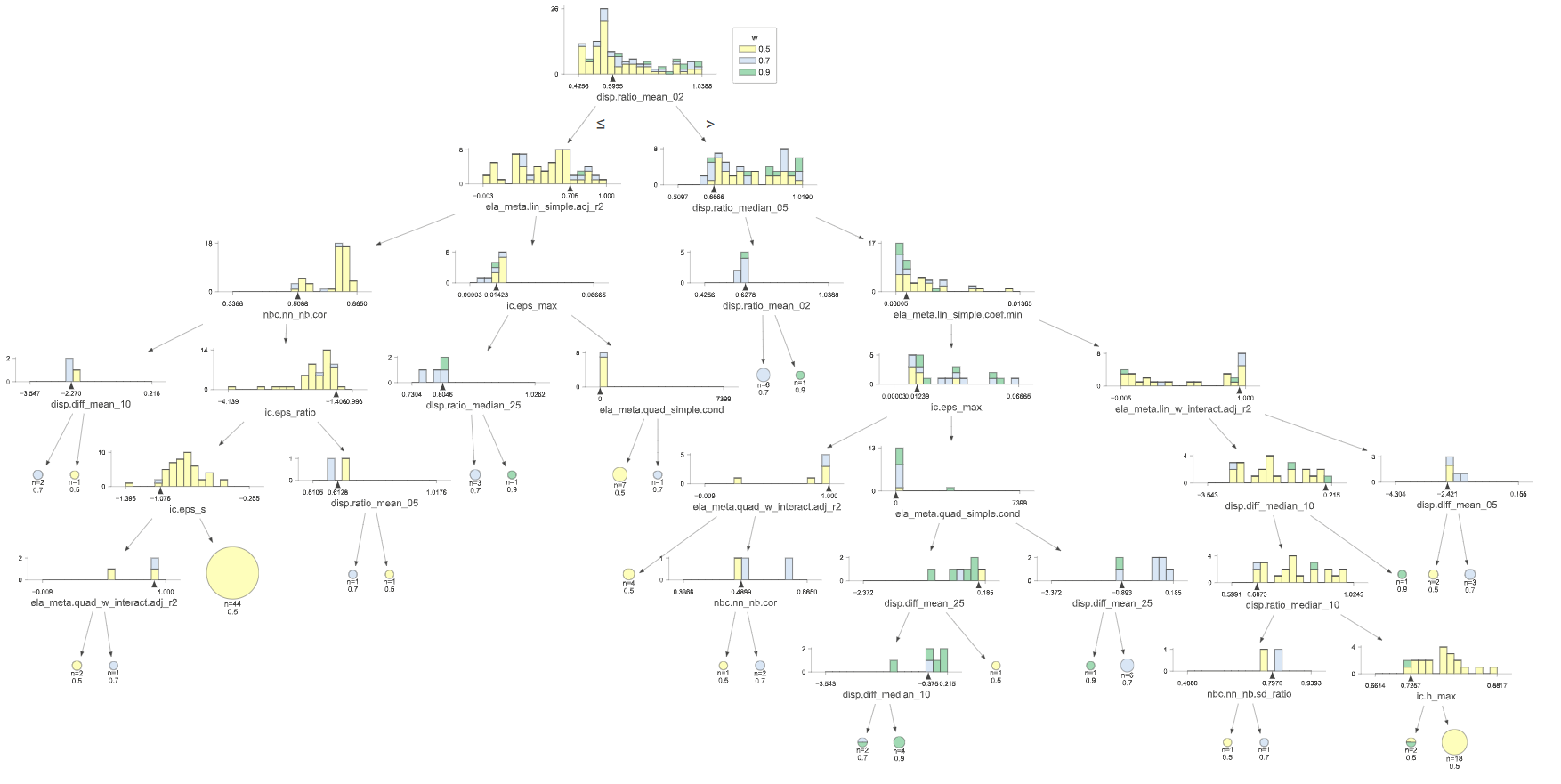}
         \caption{Decision tree (depth=7) for inertia weights ($w$) module in PSO using Star Topology ($d=5$). The nodes show feature splits with value distributions- yellow: false, green: true. The black arrow marks the split threshold.}
    \label{fig:Tr1}
\end{sidewaysfigure}

\begin{sidewaysfigure}
    \centering
   \includegraphics[width=\linewidth]{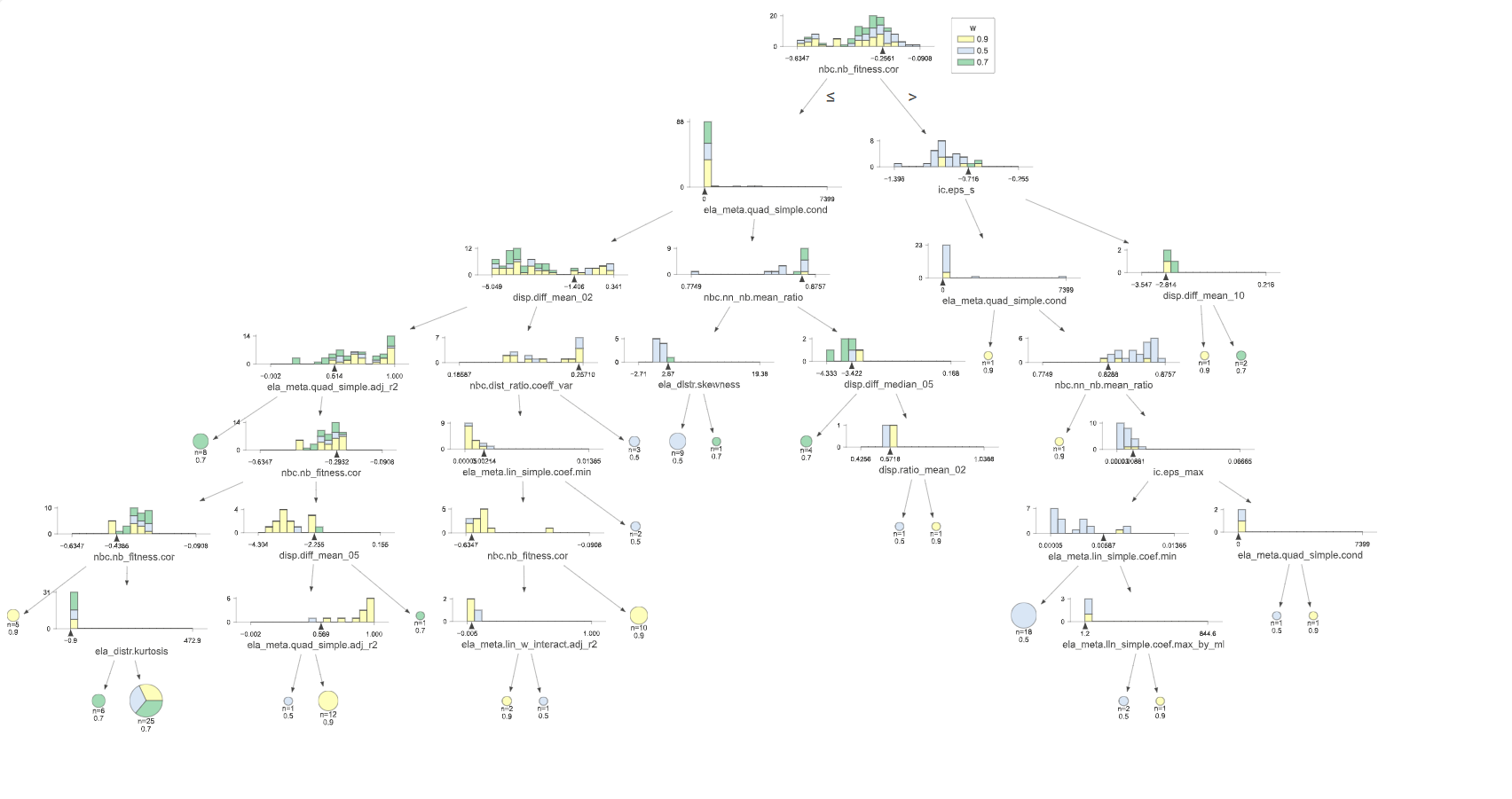}
         \caption{Decision tree (depth=7) for inertia weights ($w$) module in PSO using Ring Topology (d=5). Nodes show feature splits with value distributions—yellow: false, green: true. Black arrow marks split threshold.}\label{fig:Tr2}
\end{sidewaysfigure}

\begin{sidewaysfigure}
\centering
     \includegraphics[width=\linewidth]{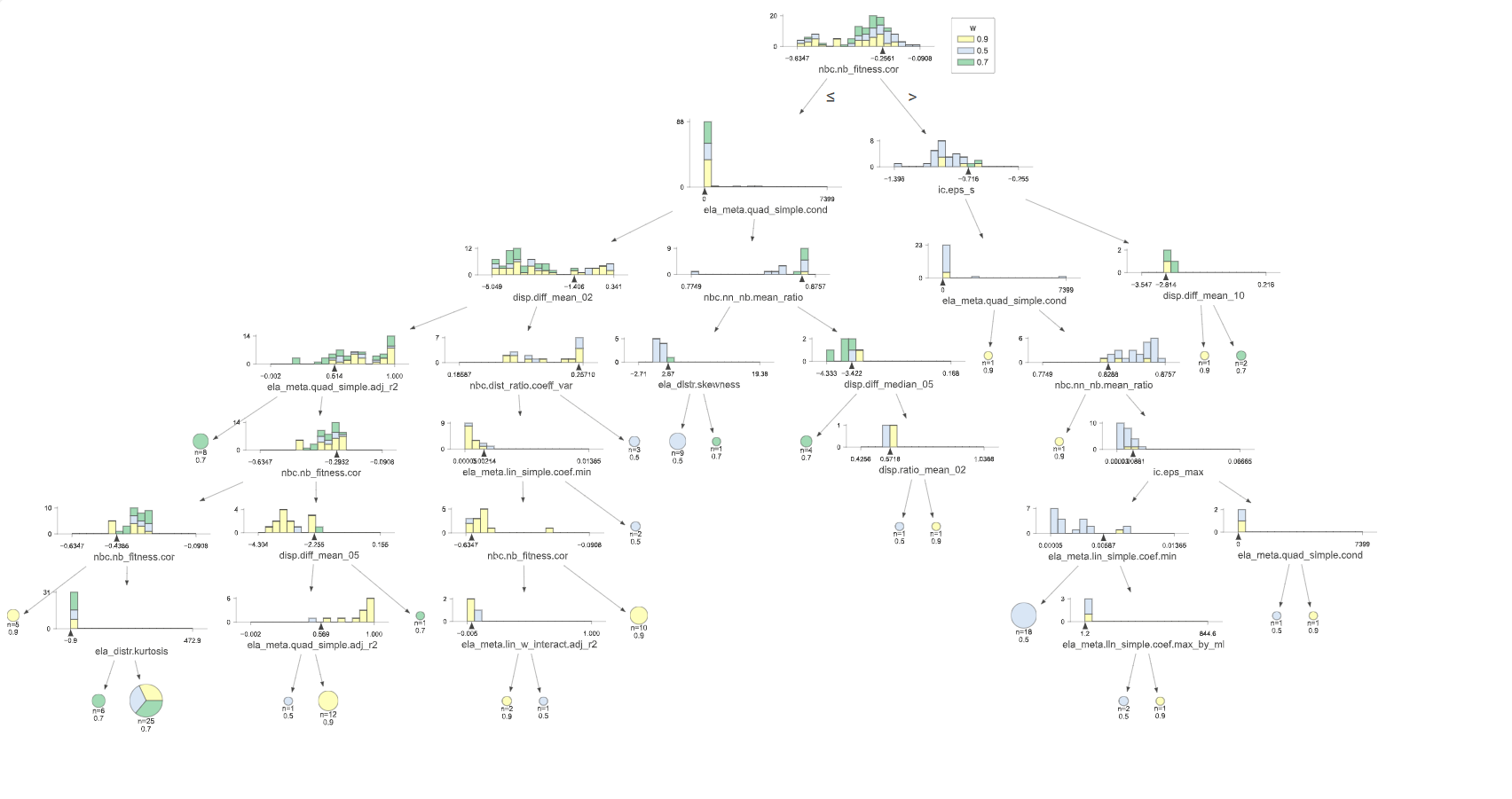}
        \caption{Decision tree (depth=7) for inertia weights ($w$) module in PSO using Von Neumann Topology (d=5). Nodes show feature splits with value distributions—yellow: false, green: true. Black arrow marks split threshold.}\label{fig:Tr3}
\end{sidewaysfigure}

The decision trees for the Ring, Star, and Von Neumann topologies show how different landscape features impact the PSO algorithm's performance at various inertia weights ($w = 0.5, 0.7, 0.9$). For the Ring topology (Figure \ref{fig:Tr1}), the top feature is \texttt{nbc.nb\_fitness.cor}, with \texttt{ela\_meta.quad\_simple.cond} coming in next. These features help categorize how the PSO performs based on different $w$ values. In the Star topology (Figure \ref{fig:Tr2}), \texttt{nbc.nb\_fitness.cor} is also the main feature, but this time, features like \texttt{disp.diff\_mean\_10} and \texttt{ela.distr.skewness} influence the decision-making process. As for the Von Neumann topology (Figure \ref{fig:Tr3}), the main feature here is \texttt{disp.diff.mean\_02}, which is key for predictions, while \texttt{ela\_meta.lin\_simple.coef.max\_by\_m} and \texttt{nbc.nn\_nb.mean\_ratio} lend support. Each decision tree visually illustrates how instances are divided, and the leaf nodes show class distributions, pointing out the best weight settings for each topology.

The Von Neumann layout is one that seems to be both even and informative, as can be seen in the decision tree graphs. It contains important parameters such as \texttt{disp.diff.mean\_02} which aid in ranking the PSO performance for various inertia weights suggesting very good adaptability and more accurate selection of the PSO parameters. This capability enables Von Neumann to adapt seamlessly to different challenges by detecting subtle changes in features, which helps the algorithm determine the optimal settings for different conditions. The Star layout, however, is fairly complicated. Lacks a little bit of a trade-off between speed and strength, as shown by its use of features that result in faster but less evenly distributed class distributions (e.g. \texttt{nbc.nb\_fitness.cor} and \texttt{ela.distr.skewness}). For comparison, the Ring layout is simpler, for the most part with only \texttt{nbc.nb\_fitness.cor}, fewer splits and more uneven leaves. This configuration provides greater consistency in performance, but less flexibility. Overall, Von Neumann appears to be adaptive, and does well in a variety of environments. Star on the other hand, will lead to faster results, but lack some diversity. Ring is a good alternative, but doesn't adjust as gracefully.

\subsection{Limitations}
One major drawback to this experimental approach is the high computational costs for scalability and expense. The methodology is not parameterizable for GPUs, but requires significant CPU resources, which results in a lot of parallel cores needed for calculations. This means that the time complexity is relatively high because of the exhaustive search of hyperparameters space and several iterations of algorithms with different benchmark functions is time-consuming. However, the experiments could not be scaled up to include higher dimensions, further wide-ranging functions or a more comprehensive hyperparameter search, because our current computer power imposes a hard limit on the scale of the experiments.

\section{Conclusion}
\label{sec:7}

This work advances the explainability of the Particle Swarm Optimization (PSO) algorithm through four key contributions. \textit{First}, it highlights the importance of Exploratory Landscape Analysis (ELA) in revealing problem structure, multimodality, and landscape difficulty, demonstrating how these features can guide algorithm selection and configuration. The IOHxplainer framework further enables systematic assessment of hyperparameter effects, particularly elucidating interactions between PSO topologies and landscape characteristics. \textit{Second}, the work presents a comprehensive evaluation of Star, Ring, and Von Neumann topologies. Results indicate that Star topology excels on unimodal or smooth functions with rapid convergence and high stability; Ring topology performs best on multimodal landscapes, balancing exploration and exploitation; and Von Neumann topology offers robust, general-purpose performance. These findings provide actionable insights into topology selection based on landscape properties. \textit{Third}, three novel STN-based metrics—Connectivity Density, Fragmentation Score, and Search Efficiency—are proposed to enhance visual explainability by transparently quantifying topology-specific search organization, network cohesion, and transition effectiveness during PSO optimization. \textit{Finally}, a novel machine learning approach using Random Forest and Decision Tree classifiers on AOCC data demonstrates the feasibility of predicting optimal topology configurations, establishing a data-driven, explainable framework. While focused on PSO, the methodology is extensible to other swarm-based metaheuristics. Future directions include extending the analysis to higher-dimensional problems, integrating additional XAI techniques, and developing adaptive topology selection systems. Overall, this work shifts optimization research from empirical evaluation toward interpretable, explainable algorithm design, with significant implications for both theoretical understanding and practical applications.

\vspace{1cm}
\textbf{Data availability}

All research materials are open to public access and include source code, experimental data and detailed results so as to ensure transparency and repeatability of our research. The full code of the PSO with various topologies (Star, Ring, Von Neumann), benchmark functions and evaluation scripts are available at: \textcolor{blue}{\url{https://github.com/GitNitin02/ioh_pso}}

\vspace{1cm}
\textbf{CRediT authorship contribution statement}

\textbf{Nitin Gupta:} Writing – original draft, Methodology, Experimentation. \textbf{Bapi Dutta:} Review \& editing. \textbf{Anupam Yadav:} Writing –
review \& editing, Supervision.

\vspace{1cm}
\textbf{Declaration of competing interest}

 The authors state that there are no competing interests to declare that might have influenced this research.
 
\vspace{1cm}
\textbf{Acknowledgments}

This work was supported by Dr. B. R. Ambedkar National Institute of Technology, Jalandhar. 

\small
\singlespacing
\bibliographystyle{ACM-Reference-Format}
\bibliography{acmart.bib}

@inproceedings{gupta2025enhancing,
  title={Enhancing explainability and reliable decision-making in particle swarm optimization through communication topologies},
  author={Gupta, Nitin and Bala, Indu and Dutta, Bapi and Mart{\'\i}nez, Luis and Yadav, Anupam},
  booktitle={Proceedings of the Genetic and Evolutionary Computation Conference Companion},
  pages={955--958},
  year={2025}
}

@article{coleto2020artificial,
  title={Artificial bee colony algorithm based on dominance (ABCD) for a hybrid gene selection method},
  author={Coleto-Alcudia, Veredas and Vega-Rodr{\'\i}guez, Miguel A},
  journal={Knowledge-Based Systems},
  volume={205},
  pages={106323},
  year={2020},
  publisher={Elsevier}
}

@article{gonzalez2021addressing,
  title={Addressing topic modeling with a multi-objective optimization approach based on swarm intelligence},
  author={Gonz{\'a}lez-Santos, Carlos and Vega-Rodr{\'\i}guez, Miguel A and P{\'e}rez, Carlos J},
  journal={Knowledge-Based Systems},
  volume={225},
  pages={107113},
  year={2021},
  publisher={Elsevier}
}

@article{liu2023improved,
  title={An improved heuristic mechanism ant colony optimization algorithm for solving path planning},
  author={Liu, Chao and Wu, Lei and Xiao, Wensheng and Li, Guangxin and Xu, Dengpan and Guo, Jingjing and Li, Wentao},
  journal={Knowledge-based systems},
  volume={271},
  pages={110540},
  year={2023},
  publisher={Elsevier}
}

@article{ochoa2021search,
  title={Search trajectory networks: A tool for analysing and visualising the behaviour of metaheuristics},
  author={Ochoa, Gabriela and Malan, Katherine M and Blum, Christian},
  journal={Applied Soft Computing},
  volume={109},
  pages={107492},
  year={2021},
  publisher={Elsevier}
}

@article{yu2025particle,
  title={Particle swarm optimization algorithm based on teaming behavior},
  author={Yu, Yu-Feng and Wang, Ziwei and Chen, Xinjia and Feng, Qiying},
  journal={Knowledge-Based Systems},
  pages={113555},
  year={2025},
  publisher={Elsevier}
}

@article{gupta2026interpretability,
  title={An interpretability framework for Ant Colony Optimization in continuous domains},
  author={Gupta, Nitin and Yadav, Anupam},
  journal={Swarm and Evolutionary Computation},
  volume={107},
  pages={102449},
  year={2026},
 publisher={Elsevier}
}

@article{csardi2006igraph,
  title={The igraph software},
  author={Csardi, Gabor and Nepusz, Tamas},
  journal={Complex syst},
  volume={1695},
  pages={1--9},
  year={2006}
}

@article{gajdovs2016parallel,
  title={A parallel Fruchterman--Reingold algorithm optimized for fast visualization of large graphs and swarms of data},
  author={Gajdo{\v{s}}, Petr and Je{\v{z}}owicz, Tom{\'a}{\v{s}} and Uher, Vojt{\v{e}}ch and Dohn{\'a}lek, Pavel},
  journal={Swarm and evolutionary computation},
  volume={26},
  pages={56--63},
  year={2016},
  publisher={Elsevier}
}

@article{sarpkamada,
  title={Kamada-Kawai Algorithm: A Comprehensive Analysis from Global Graph Layout to Modern Challenges},
  author={SARP, {\"U}mit and DEM{\.I}R, Bilal},
  journal={THEORETICAL AND APPLIED RESEARCH IN NATURAL SCIENCES AND MATHEMATICS},
  pages={63}
}

@article{ding2022explainability,
  title={Explainability of artificial intelligence methods, applications and challenges: A comprehensive survey},
  author={Ding, Weiping and Abdel-Basset, Mohamed and Hawash, Hossam and Ali, Ahmed M},
  journal={Information Sciences},
  volume={615},
  pages={238--292},
  year={2022},
  publisher={Elsevier}
}

@article{liu2016topology,
  title={Topology selection for particle swarm optimization},
  author={Liu, Qunfeng and Wei, Wenhong and Yuan, Huaqiang and Zhan, Zhi-Hui and Li, Yun},
  journal={Information Sciences},
  volume={363},
  pages={154--173},
  year={2016},
  publisher={Elsevier}
}

@article{BARREDOARRIETA202082,
title = {Explainable Artificial Intelligence (XAI): Concepts, taxonomies, opportunities and challenges toward responsible AI},
journal = {Information Fusion},
volume = {58},
pages = {82-115},
year = {2020},
issn = {1566-2535},
author = {Alejandro {Barredo Arrieta} and Natalia Díaz-Rodríguez and Javier {Del Ser} and Adrien Bennetot and Siham Tabik and Alberto Barbado and Salvador Garcia and Sergio Gil-Lopez and Daniel Molina and Richard Benjamins and Raja Chatila and Francisco Herrera}
}

@inproceedings{kennedy1995new,
  title={A new optimizer using particle swarm theory},
  author={Kennedy, James and Eberhart, Russell},
  booktitle={Proceedings of the sixth international symposium on micro machine and human science},
  volume={3943},
  year={1995},
  organization={Nagoya, Japan: IEEE}
}

@article{miranda2008stochastic,
  title={Stochastic star communication topology in evolutionary particle swarms (EPSO)},
  author={Miranda, Vladimiro and Keko, Hrvoje and Junior, Alvaro Jaramillo},
  year={2008}
}

@article{zhang2019enhancing,
  title={Enhancing comprehensive learning particle swarm optimization with local optima topology},
  author={Zhang, Kai and Huang, Qiujun and Zhang, Yimin},
  journal={Information Sciences},
  volume={471},
  pages={1--18},
  year={2019},
  publisher={Elsevier}
}

@article{wang2025neighborhood,
  title={Neighborhood Rough Decision Tree},
  author={Wang, Changzhong and Cui, Xinyu and An, Shuang},
  journal={Information Sciences},
  pages={122266},
  year={2025},
  publisher={Elsevier}
}

@article{ni2013new,
  title={A new logistic dynamic particle swarm optimization algorithm based on random topology},
  author={Ni, Qingjian and Deng, Jianming},
  journal={The Scientific World Journal},
  volume={2013},
  number={1},
  pages={409167},
  year={2013},
  publisher={Wiley Online Library}
}

@article{von1935complete,
  title={On complete topological spaces},
  author={Von Neumann, John},
  journal={Transactions of the American Mathematical Society},
  volume={37},
  number={1},
  pages={1--20},
  year={1935},
  publisher={JSTOR}
}

@article{munoz2015algorithm,
  title={Algorithm selection for black-box continuous optimization problems: A survey on methods and challenges},
  author={Mu{\~n}oz, Mario A and Sun, Yuan and Kirley, Michael and Halgamuge, Saman K},
  journal={Information sciences},
  volume={317},
  pages={224--245},
  year={2015},
  publisher={Elsevier}
}

@article{lynn2018population,
  title={Population topologies for particle swarm optimization and differential evolution},
  author={Lynn, Nandar and Ali, Mostafa Z and Suganthan, Ponnuthurai Nagaratnam},
  journal={Swarm and evolutionary computation},
  volume={39},
  pages={24--35},
  year={2018},
  publisher={Elsevier}
}

@article{van2024explainable,
  title={Explainable benchmarking for iterative optimization heuristics},
  author={van Stein, Niki and Vermetten, Diederick and Kononova, Anna V and B{\"a}ck, Thomas},
  journal={arXiv preprint arXiv:2401.17842},
  year={2024}
}

@article{garcia2023explainable,
  title={Explainable Rules and Heuristics in AI Algorithm Recommendation Approaches—A Systematic Literature Review and Mapping Study},
  author={Garc{\'\i}a-Holgado, A and Vazquez-Ingelmo, Andrea and Garc{\'\i}a-Pe{\~n}alvo, FJ},
  year={2023},
  publisher={Tech Science Press}
}

@article{lindauer2019boah,
  title={Boah: A tool suite for multi-fidelity bayesian optimization \& analysis of hyperparameters},
  author={Lindauer, Marius and Eggensperger, Katharina and Feurer, Matthias and Biedenkapp, Andr{\'e} and Marben, Joshua and M{\"u}ller, Philipp and Hutter, Frank},
  journal={arXiv preprint arXiv:1908.06756},
  year={2019}
}

@phdthesis{hansen2009real,
  title={Real-parameter black-box optimization benchmarking 2009: Noiseless functions definitions},
  author={Hansen, Nikolaus and Finck, Steffen and Ros, Raymond and Auger, Anne},
  year={2009},
  school={INRIA}
}

@article{lundberg2020local,
  title={From local explanations to global understanding with explainable AI for trees},
  author={Lundberg, Scott M and Erion, Gabriel and Chen, Hugh and DeGrave, Alex and Prutkin, Jordan M and Nair, Bala and Katz, Ronit and Himmelfarb, Jonathan and Bansal, Nisha and Lee, Su-In},
  journal={Nature machine intelligence},
  volume={2},
  number={1},
  pages={56--67},
  year={2020},
  publisher={Nature Publishing Group}
}

@article{hansen2022anytime,
  title={Anytime performance assessment in blackbox optimization benchmarking},
  author={Hansen, Nikolaus and Auger, Anne and Brockhoff, Dimo and Tu{\v{s}}ar, Tea},
  journal={IEEE Transactions on Evolutionary Computation},
  volume={26},
  number={6},
  pages={1293--1305},
  year={2022},
  publisher={IEEE}
}

@article{lopez2024using,
  title={Using the empirical attainment function for analyzing single-objective black-box optimization algorithms},
  author={L{\'o}pez-Ib{\'a}{\~n}ez, Manuel and Vermetten, Diederick and Dreo, Johann and Doerr, Carola},
  journal={IEEE Transactions on Evolutionary Computation},
  year={2024},
  publisher={IEEE}
}

@inproceedings{mersmann2011exploratory,
  title={Exploratory landscape analysis},
  author={Mersmann, Olaf and Bischl, Bernd and Trautmann, Heike and Preuss, Mike and Weihs, Claus and Rudolph, G{\"u}nter},
  booktitle={Proceedings of the 13th annual conference on Genetic and evolutionary computation},
  pages={829--836},
  year={2011}
}

@inproceedings{bischl2012algorithm,
  title={Algorithm selection based on exploratory landscape analysis and cost-sensitive learning},
  author={Bischl, Bernd and Mersmann, Olaf and Trautmann, Heike and Preu{\ss}, Mike},
  booktitle={Proceedings of the 14th annual conference on Genetic and evolutionary computation},
  pages={313--320},
  year={2012},
  publisher={ACM}
}

@article{liang2024survey,
  title={A survey of surrogate-assisted evolutionary algorithms for expensive optimization},
  author={Liang, Jing and Lou, Yahang and Yu, Mingyuan and Bi, Ying and Yu, Kunjie},
  journal={Journal of Membrane Computing},
  pages={1--20},
  year={2024},
  publisher={Springer}
}

@article{malan2013survey,
  title={A survey of techniques for characterising fitness landscapes and some possible ways forward},
  author={Malan, Katherine M and Engelbrecht, Andries P},
  journal={Information Sciences},
  volume={241},
  pages={148--163},
  year={2013},
  publisher={Elsevier}
}

@inproceedings{zhang2008sequential,
  title={Sequential particle swarm optimization for visual tracking},
  author={Zhang, Xiaoqin and Hu, Weiming and Maybank, Steve and Li, Xi and Zhu, Mingliang},
  booktitle={2008 IEEE conference on computer vision and pattern recognition},
  pages={1--8},
  year={2008},
  organization={IEEE}
}

@article{hansen2021coco,
  title={COCO: A platform for comparing continuous optimizers in a black-box setting},
  author={Hansen, Nikolaus and Auger, Anne and Ros, Raymond and Mersmann, Olaf and Tu{\v{s}}ar, Tea and Brockhoff, Dimo},
  journal={Optimization Methods and Software},
  volume={36},
  number={1},
  pages={114--144},
  year={2021},
  publisher={Taylor \& Francis}
}

@article{wu2017problem,
  title={Problem definitions and evaluation criteria for the CEC 2017 competition on constrained real-parameter optimization},
  author={Wu, Guohua and Mallipeddi, Rammohan and Suganthan, Ponnuthurai Nagaratnam},
  journal={National University of Defense Technology, Changsha, Hunan, PR China and Kyungpook National University, Daegu, South Korea and Nanyang Technological University, Singapore, Technical Report},
  volume={9},
  pages={2017},
  year={2017},
  publisher={National University of Defense Technology Changsha, China}
}

@misc{rapin2018nevergrad,
  title={Nevergrad-A gradient-free optimization platform},
  author={Rapin, J{\'e}r{\'e}my and Teytaud, Olivier},
  year={2018}
}

@incollection{kerschke2014cell,
  title={Cell mapping techniques for exploratory landscape analysis},
  author={Kerschke, Pascal and Preuss, Mike and Hern{\'a}ndez, Carlos and Sch{\"u}tze, Oliver and Sun, Jian-Qiao and Grimme, Christian and Rudolph, G{\"u}nter and Bischl, Bernd and Trautmann, Heike},
  booktitle={EVOLVE-A Bridge between Probability, Set Oriented Numerics, and Evolutionary Computation V},
  pages={115--131},
  year={2014},
  publisher={Springer}
}

@incollection{kerschke2019comprehensive,
  title={Comprehensive feature-based landscape analysis of continuous and constrained optimization problems using the R-package flacco},
  author={Kerschke, Pascal and Trautmann, Heike},
  booktitle={Applications in statistical computing: from music data analysis to industrial quality improvement},
  pages={93--123},
  year={2019},
  publisher={Springer}
}

@article{kerschke2019automated,
  title={Automated algorithm selection on continuous black-box problems by combining exploratory landscape analysis and machine learning},
  author={Kerschke, Pascal and Trautmann, Heike},
  journal={Evolutionary computation},
  volume={27},
  number={1},
  pages={99--127},
  year={2019},
  publisher={MIT Press One Rogers Street, Cambridge, MA 02142-1209, USA journals-info~…}
}

@inproceedings{long2022learning,
  title={Learning the characteristics of engineering optimization problems with applications in automotive crash},
  author={Long, Fu Xing and van Stein, Bas and Frenzel, Moritz and Krause, Peter and Gitterle, Markus and B{\"a}ck, Thomas},
  booktitle={Proceedings of the Genetic and Evolutionary Computation Conference},
  pages={1227--1236},
  year={2022}
}

@inproceedings{trajanov2022improving,
  title={Improving nevergrad’s algorithm selection wizard ngopt through automated algorithm configuration},
  author={Trajanov, Risto and Nikolikj, Ana and Cenikj, Gjorgjina and Teytaud, Fabien and Videau, Mathurin and Teytaud, Olivier and Eftimov, Tome and L{\'o}pez-Ib{\'a}{\~n}ez, Manuel and Doerr, Carola},
  booktitle={International Conference on Parallel Problem Solving from Nature},
  pages={18--31},
  year={2022},
  organization={Springer}
}

@article{lundberg2017unified,
  title={A unified approach to interpreting model predictions},
  author={Lundberg, Scott M and Lee, Su-In},
  journal={Advances in neural information processing systems},
  volume={30},
  year={2017}
}

@article{shields2016generalization,
  title={The generalization of Latin hypercube sampling},
  author={Shields, Michael D and Zhang, Jiaxin},
  journal={Reliability Engineering \& System Safety},
  volume={148},
  pages={96--108},
  year={2016},
  publisher={Elsevier}
}

@inproceedings{burhenne2011sampling,
  title={Sampling based on Sobol’sequences for Monte Carlo techniques applied to building simulations},
  author={Burhenne, Sebastian and Jacob, Dirk and Henze, Gregor P},
  booktitle={Building Simulation 2011},
  volume={12},
  pages={1816--1823},
  year={2011},
  organization={IBPSA}
}

@incollection{dorigo2018introduction,
  title={An introduction to ant colony optimization},
  author={Dorigo, Marco and Socha, Krzysztof},
  booktitle={Handbook of approximation algorithms and metaheuristics},
  pages={395--408},
  year={2018},
  publisher={Chapman and Hall/CRC}
}

@article{karaboga2007powerful,
  title={A powerful and efficient algorithm for numerical function optimization: artificial bee colony (ABC) algorithm},
  author={Karaboga, Dervis and Basturk, Bahriye},
  journal={Journal of global optimization},
  volume={39},
  number={3},
  pages={459--471},
  year={2007},
  publisher={Springer}
}

@article{hossain2021machine,
  title={Machine learning model optimization with hyper parameter tuning approach},
  author={Hossain, Md Riyad and Timmer, Douglas},
  journal={Glob. J. Comput. Sci. Technol. D Neural Artif. Intell},
  volume={21},
  number={2},
  pages={31},
  year={2021}
}

@article{li2021particle,
  title={Particle swarm optimization algorithm with multiple phases for solving continuous optimization problems},
  author={Li, Jing and Sun, Yifei and Hou, Sicheng},
  journal={Discrete Dynamics in Nature and Society},
  volume={2021},
  number={1},
  pages={8378579},
  year={2021},
  publisher={Wiley Online Library}
}

@ARTICLE{985692,
  author={Clerc, M. and Kennedy, J.},
  journal={IEEE Transactions on Evolutionary Computation}, 
  title={The particle swarm - explosion, stability, and convergence in a multidimensional complex space}, 
  year={2002},
  volume={6},
  number={1},
  pages={58-73},
  doi={10.1109/4235.985692}}

@article{blackwell2007particle,
  title={Particle swarm optimization},
  author={Blackwell, Tim and Kennedy, J and Poli, R},
  journal={Swarm Intelligence},
  volume={1},
  number={1},
  pages={33--57},
  year={2007}
}

@article{yang2025meta,
  title={Meta-Black-Box optimization for evolutionary algorithms: Review and perspective},
  author={Yang, Xu and Wang, Rui and Li, Kaiwen and Ishibuchi, Hisao},
  journal={Swarm and Evolutionary Computation},
  volume={93},
  pages={101838},
  year={2025},
  publisher={Elsevier}
}

@article{van2025explainable,
  title={Explainable benchmarking for iterative optimization heuristics},
  author={van Stein, Niki and Vermetten, Diederick and V. Kononova, Anna and B{\"a}ck, Thomas},
  journal={ACM Transactions on Evolutionary Learning},
  volume={5},
  number={2},
  pages={1--30},
  year={2025},
  publisher={ACM New York, NY}
}

@article{slack2021reliable,
  title={Reliable post hoc explanations: Modeling uncertainty in explainability},
  author={Slack, Dylan and Hilgard, Anna and Singh, Sameer and Lakkaraju, Himabindu},
  journal={Advances in neural information processing systems},
  volume={34},
  pages={9391--9404},
  year={2021}
}

@article{doshi2017towards,
  title={Towards a rigorous science of interpretable machine learning},
  author={Doshi-Velez, Finale and Kim, Been},
  journal={arXiv preprint arXiv:1702.08608},
  year={2017}
}

@article{prager2024pflacco,
  title={Pflacco: Feature-based landscape analysis of continuous and constrained optimization problems in Python},
  author={Prager, Raphael Patrick and Trautmann, Heike},
  journal={Evolutionary Computation},
  volume={32},
  number={3},
  pages={211--216},
  year={2024},
  publisher={MIT Press 255 Main Street, 9th Floor, Cambridge, Massachusetts 02142, USA~…}
}

@inproceedings{kostovska2022importance,
  title={The importance of landscape features for performance prediction of modular CMA-ES variants},
  author={Kostovska, Ana and Vermetten, Diederick and D{\v{z}}eroski, Sa{\v{s}}o and Doerr, Carola and Korosec, Peter and Eftimov, Tome},
  booktitle={Proceedings of the Genetic and Evolutionary Computation Conference},
  pages={648--656},
  year={2022}
}

@article{miles2005r,
  title={R-squared, adjusted R-squared},
  author={Miles, Jeremy},
  journal={Encyclopedia of statistics in behavioral science},
  year={2005},
  publisher={Wiley Online Library}
}

@article{pal2005random,
  title={Random forest classifier for remote sensing classification},
  author={Pal, Mahesh},
  journal={International journal of remote sensing},
  volume={26},
  number={1},
  pages={217--222},
  year={2005},
  publisher={Taylor \& Francis}
}

@article{song2015decision,
  title={Decision tree methods: applications for classification and prediction},
  author={Song, Yan-Yan and Lu, Ying},
  journal={Shanghai archives of psychiatry},
  year={2015},
  publisher={Shanghai Municipal Bureau of Publishing}
}

@inproceedings{shi1998modified,
  title={A modified particle swarm optimizer},
  author={Shi, Yuhui and Eberhart, Russell},
  booktitle={1998 IEEE international conference on evolutionary computation proceedings. IEEE world congress on computational intelligence (Cat. No. 98TH8360)},
  pages={69--73},
  year={1998},
  organization={Ieee}
}

@article{eiben2011parameter,
  title={Parameter tuning for configuring and analyzing evolutionary algorithms},
  author={Eiben, Agoston E and Smit, Selmar K},
  journal={Swarm and evolutionary computation},
  volume={1},
  number={1},
  pages={19--31},
  year={2011},
  publisher={Elsevier}
}

@phdthesis{pedersen2010tuning,
  title={Tuning \& simplifying heuristical optimization},
  author={Pedersen, Magnus Erik Hvass},
  year={2010},
  school={University of Southampton}
}

@article{banks2007review,
  title={A review of particle swarm optimization. Part I: background and development},
  author={Banks, Alec and Vincent, Jonathan and Anyakoha, Chukwudi},
  journal={Natural Computing},
  volume={6},
  number={4},
  pages={467--484},
  year={2007},
  publisher={Springer}
}

@article{bonyadi2017particle,
  title={Particle swarm optimization for single objective continuous space problems: a review},
  author={Bonyadi, Mohammad Reza and Michalewicz, Zbigniew},
  journal={Evolutionary computation},
  volume={25},
  number={1},
  pages={1--54},
  year={2017},
  publisher={MIT Press}
}

@book{bonabeau1999swarm,
  title={Swarm intelligence: from natural to artificial systems},
  author={Bonabeau, Eric and Dorigo, Marco and Theraulaz, Guy},
  number={1},
  year={1999},
  publisher={Oxford university press}
}

@article{bianchi2009survey,
  title={A survey on metaheuristics for stochastic combinatorial optimization},
  author={Bianchi, Leonora and Dorigo, Marco and Gambardella, Luca Maria and Gutjahr, Walter J},
  journal={Natural Computing},
  volume={8},
  number={2},
  pages={239--287},
  year={2009},
  publisher={Springer}
}

@article{engelbrecht2005fundamentals,
  title={Fundamentals of Computational Swarm Intelligence. John Wiley \& Sons, Chichester, UK},
  author={Engelbrecht, AP},
  year={2005}
}

@inproceedings{kennedy2002population,
  title={Population structure and particle swarm performance},
  author={Kennedy, James and Mendes, Rui},
  booktitle={Proceedings of the 2002 Congress on Evolutionary Computation. CEC'02 (Cat. No. 02TH8600)},
  volume={2},
  pages={1671--1676},
  year={2002},
  organization={IEEE}
}
\end{document}